\documentclass[12pt]{article}
\usepackage{natbib}
\usepackage[hidelinks]{hyperref}
\usepackage{fullpage}
\usepackage{graphicx}
\usepackage{caption}
\usepackage{subcaption}
\usepackage{amsmath}
\usepackage{amssymb}
\usepackage{mathtools}
\usepackage{float}
\usepackage{pgfplots}
\pgfplotsset{compat=1.17}

\title{Inference via Sparse Coding in a Hierarchical Vision Model}

\author{Joshua Bowren$^1$, Luis Sanchez-Giraldo$^2$, and Odelia Schwartz$^1$\\
\large{$^1$Department of Computer Science, University of Miami,} \\
\large{Coral Gables, FL, USA} \\
\large{$^2$Department of Electrical and Computer Engineering,}\\
\large{University of Kentucky, Lexington, KY, USA}}
\date{}

\begin{document}

\maketitle

\begin{abstract}
Sparse coding has been incorporated in models of the visual cortex for its computational advantages and connection to biology. But how the level of sparsity contributes to performance on visual tasks is not well understood. In this work, sparse coding has been integrated into an existing hierarchical V2 model \citep{hosoya:jneuro15}, but replacing its independent component analysis (ICA) with an explicit sparse coding in which the degree of sparsity can be controlled. After training, the sparse coding basis functions with a higher degree of sparsity resembled qualitatively different structures, such as curves and corners. The contributions of the models were assessed with image classification tasks, specifically tasks associated with mid-level vision including figure-ground classification, texture classification, and angle prediction between two line stimuli. In addition, the models were assessed in comparison to a texture sensitivity measure that has been reported in V2 \citep{freeman:natneuro13}, and a deleted-region inference task. The results from the experiments show that while sparse coding performed worse than ICA at classifying images, only sparse coding was able to better match the texture sensitivity level of V2 and infer deleted image regions, both by increasing the degree of sparsity in sparse coding. Higher degrees of sparsity allowed for inference over larger deleted image regions. The mechanism that allows for this inference capability in sparse coding is described here.
\end{abstract}

\section{Introduction}
Computational models of the visual cortex have progressed significantly over the past few decades. One approach to modeling cortical neurons, denoted as goal-oriented (or supervised learning), is based on optimizing model goals such as image classification \citep[see e.g., review papers,][]{geisler:arpsych08,yamins:natneuro16,turner:natneuro19}. In recent years, deep neural network models optimized for image classification \citep[e.g.,][]{krizhevsky:nips12, dapello:biorxiv20} have captured neural processing in cortical visual areas \citep{kriegeskorte:arvs15,yamins:natneuro16}, including low and mid level visual cortex \citep[e.g.,][]{cadena:ploscb19,kindel:jov19,pospisil:elife18,laskar:jov20}.

Another approach for modeling cortical visual neurons which is the focus here is denoted as stimulus-oriented (or unsupervised learning). In particular, it has been hypothesized that neurons are matched to the statistical properties of images in the environment \citep[][]{barlow:sencom61, attneave:pr54,simoncelli:arn01} by optimizing statistical constraints such as sparsity or coding efficiency.  For instance, the sparse coding model of \cite{olshausen:nat96}, and models of Independent Component Analysis \citep[ICA;][]{bell:nc95,hyvarinen:nc97}, offered a principled mechanism for the derivation of oriented filters qualitatively similar to simple cells in the primary visual cortex (area V1). Others have proposed methods of deriving models of V1 complex cell responses \citep{hyvarinen:vr01,berkes:jov05,karklin:nat09}, deriving V2 model responses from V1 responses \citep{lee:nips07,coen:jov13,shan:arxiv13,hosoya:jneuro15}, and hierarchical nonlinear models that learn patterns of statistical dependencies \citep{karklin:nc05}. Stimulus-oriented approaches have also been adapted to deep neural networks with success in capturing aspects of the ventral visual cortex \citep{zhuang:nas21}.

In addition to bottom-up approaches of optimizing statistical constraints, stimulus-oriented approaches can be closely tied to top-down generative approaches describing the process by which the signals are generated \citep{rao:pmb02}. For instance, sparse coding can be seen both from the perspective of optimizing sparseness, and as a generative model of images \citep{olshausen:psn06}. Generating and inferring image structure is also an important aspect of vision \citep{yuille:tcs06} in addition to classifying images. While a major emphasis in computer vision has been on image classification, other ideas exist for how inference and other capabilities may be achieved in image models \citep{pei:tis06,zhaoping:ploscb08,goodfellow:nips14,radford:arxiv15,luo:iccv15,svanera:jov21}.

The aim of this work is to investigate how sparse coding can be explicitly integrated into a V2 model \citep{hosoya:jneuro15} in order to introduce an inference mechanism (discussed later), test its performance on an inference task, test its performance on image classification tasks spanning line combinations, figure-ground classification, and texture classification, and to compare the model's texture sensitivity with the texture sensitivity of V2 as reported in the fMRI results of \cite{freeman:natneuro13}. This approach was taken, as opposed to a complete vision model capable of inference like a generative adversarial network, because it allowed for the the principle of a sparse prior to be studied in a model V2 stage when holding lower-level stages (V1 and V1 complex) constant with respect to the effect of the sparse prior.

The work in \cite{hosoya:jneuro15} includes a model of V1 complex cell responses and a dimensionality reduction stage, followed by a version of independent component analysis for overcomplete codes and rectification to form V2-like model neurons. Although ICA results in filter responses with high kurtosis, it does not explicitly optimize for sparsity. In addition, studies comparing ICA and sparse coding in the overcomplete case in a single-layer model have found differences \citep{livezey:jmlr19}. Therefore the purpose of this work was to understand the implications of incorporating an explicit sparse coding at the last stage of the model, for which the sparsity level could also be controlled.

The current understanding of V2 characterizes its receptive fields in terms of its response properties, some of which include cross-orientation suppression \citep{rowekamp:natcom17}, selectivity for angles \citep[e.g.,][]{ito:jneuro04}, selectivity for figure-ground \citep{von:jneuro89,peterhans:jneuro89,zhaoping:neuron05}, and selectivity for texture \citep{freeman:natneuro13,ziemba:nas16,kohler:jneuro16}. However, the receptive fields of V2 neurons are not fully understood, and there is no consensus that any one model best explains V2 units. The model of \cite{hosoya:jneuro15} can capture some properties of V2 neurons, but in this paper, our primary goal was to highlight practical vision capabilities rather than to compare to neural data. The sparse coding model works by finding a dictionary of basis functions such that only a few are needed to reconstruct any given image. Sparse coding was successful for modeling V1, so some continued to perform sparse coding twice to model V2 \citep[e.g.,][]{lee:nips07}. Others have looked at hierarchical nonlinear generative sparse coding models \citep{karklin:nc05}. Here, traditional sparse coding \citep{olshausen:nat96} was performed in the V2-stage of a V2-like model.

The focus of this work was on a hierarchical visual cortical model with sparse coding because sparse coding has various computational advantages \citep{willshaw:nat69,kanerva:nasaames92}, is biologically plausible \citep{field:nc94,olshausen:vn03,olshausen:conb04,rozell:nc08}, and sparse firing has been observed in visual cortical neurons in response to images \citep[e.g., see][]{willmore:jnp11,yoshida:natcom20}, although see also the discussion in the work by \cite{berkes:nips09}. While the sparse coding model of \cite{olshausen:nat96} is not the only method of achieving a sparse neural representation, its underlying generative model provides a coding strategy that models neuron responses as contributions of basis functions that sum to reconstruct the input image rather than linear filter responses to that image. This allows for inference via the mechanism discussed next.

The original approach by \cite{hosoya:jneuro15} performed a variation of ICA for overcomplete codes (here referred to as overcomplete ICA) as its final V2-stage computation in order to derive an overcomplete sparse representation, but sparse coding provides several appealing computational differences. First, while the generative model of sparse coding is linear, its forward transformation is nonlinear (the solution to an optimization problem). By comparison, the forward transform of ICA is linear (multiplication by a filter matrix). Second, unlike ICA, the sparse coding algorithm allows for explicit control of the degree of sparsity. Third, sparse coding explicitly learns a dictionary of basis functions for which each of the model’s responses is interpreted as the contribution of a single basis function to the image.

The degree of sparsity enforced by the L1 regularization coefficient of sparse coding allows the model to focus more on either a faithful reconstruction with low values or structure inference with high values. With low values (low sparsity), many basis functions are available, and the image is reconstructed almost exactly. With high values (high sparsity), only a few basis functions constitute the image reconstruction, and each individual basis function must do more to explain the image (minimize reconstruction error). In practice, reconstruction error increases with fewer basis functions, but more latent information is introduced into the reconstruction. The idea, though counterintuitive, is that higher error may be advantageous. The error allows for missing image information due to events such as occlusion to be discarded in the model’s representation of the image, and the model instead explains the image from an incomplete set of input responses. For this reason, this mechanism is referred to here as the model’s \emph{inductive inference mechanism}.

Both the overcomplete ICA and non-negative sparse coding based models were tested on three classification tasks: a figure-ground detection task, a texture classification task, and an angle discrimination task (see \nameref{sec:methods} for details). Classification was performed by training a linear support vector machine (SVM) on the V2-stage responses generated by both models to give a sense of how linearly-separable the image classes were in the V2-stage representation space. While good performance on these tasks is probably characteristic of a good initial (low-level) vision model, vision models should be expected to perform a large range of functions necessary for understanding the world. One might expect tasks like noise removal, image completion, and content generation to be necessary to compete with the wide range of tasks the human visual system can perform. One such task was explored here: the ability of both models to fill in missing image information when deleted midway through the visual processing pipeline. Good performance on this task would suggest an image understanding beyond an association with labels and provide evidence that the inductive inference mechanism postulated here might benefit other vision models.

The novel contribution of this work is an understanding of the importance of vision tasks such as image classification, image inference, and texture sensitivity, and their implications for model performance. Non-negative sparse coding was found to perform worse on the popular computer vision metric of image classification, but was more closely matched to V2 in terms of texture sensitivity and better inferred deleted image regions in the image inference experiment with the proper value of the regularization coefficient. The results highlight some of the tradeoffs of sparse coding with different sparsity levels for the range of tasks. Also, sparse coding with a larger regularization coefficient (i.e., larger sparsity level) is viewed here as providing an enhancement rather than only a degradation of the model. While reconstruction error becomes higher with a larger regularization coefficient and may seem undesirable, it is proposed here that such a strategy may be useful for vision.

\section{Methods}\label{sec:methods}
This work builds upon the hierarchical unsupervised learning V2 model of \cite{hosoya:jneuro15}. Non-negative sparse coding similar to that of \cite{hoyer:nnsp02} was incorporated in place of overcomplete ICA in order to maintain the non-negative response property of the original model. Also the results were compared to the original overcomplete ICA based model. \cite{hosoya:jneuro15} made ICA overcomplete by increasing the number of independent components in the loss function beyond the number of inputs, then estimating the components with score matching according to \cite{hyvarinen:jmlr05}. The same loss function in \cite{hyvarinen:jmlr05} was minimized here for the original model.

The generative model of sparse coding models images as sparse linear combinations of a set of basis functions given by the matrix $\mathbf{\Phi}$ called a dictionary:
\begin{align}
\mathbf{x} &= \mathbf{\Phi} \mathbf{a} \label{eq:scmodel}
\end{align}

\noindent where the vector $\mathbf{x}$ is the image and the vector $\mathbf{a}$ combines columns of $\mathbf{\Phi}$ and contains mostly zeros (sparse). Since there are a few variations of sparse coding, we define our percise method here. Non-negative sparse coding was performed with scikit-learn \citep[version 0.20.3]{pedregosa:jmlr11} by inferring the basis function matrix $\mathbf{\Phi}$ such that
\begin{align}
\mathbf{\Phi} &= \arg \min_{\mathbf{\Phi}} \Vert \mathbf{X} - \mathbf{\Phi} \mathbf{A} \Vert_\textrm{F}^2 \ \mid \ \Vert \mathbf{\Phi}_i \Vert_2 = 1, \forall i\label{eq:basisobj}
\end{align}

\noindent where $\mathbf{X}$ is a matrix with column input image vectors $\mathbf{x}_i$, $\mathbf{A}$ is a matrix of sparse coefficient column vectors $\mathbf{a}_i$ which are functions of $\mathbf{\Phi}$ and $\mathbf{x}_i$, and the operation $\Vert \ \Vert_F$ is the Forbenius-norm. During each training step, first the sparse coefficient vector $\mathbf{a}_i$ for each input vector $\mathbf{x}_i$ is inferred via LASSO by choosing $\mathbf{a}_i$ such that
\begin{align}
\mathbf{a}_i &= \arg \min_{\mathbf{a}_i} \Vert \mathbf{x}_i - \mathbf{\Phi} \mathbf{a}_i \Vert_2^2 + \lambda \Vert \mathbf{a}_i \Vert_1 \ \mid \ a_{i,j} > 0, \forall i,j.\label{eq:coefobj}
\end{align}

\noindent where the hyperparameter $\lambda$ determines the level of sparsity in the non-negative sparse coding representation. The objective in equation \ref{eq:coefobj} is minimized via coordinate descent with the current value of the basis function matrix $\mathbf{\Phi}$. After inferring all $\mathbf{a}_i$, the basis function matrix is updated with one step of coordinate descent according to its objective in equation \ref{eq:basisobj}. This process was repeated until there was little change in the appearance of the basis function visualizations (see below). This was similar to the method of \cite{olshausen:nat96}, but with a non-negativity constraint.

In the original model, overcomplete ICA was followed by rectification (ReLU) in order to constrain the model V2 responses to be non-negative. The same could be done for sparse coding, but a more natural approach was available via non-negative sparse coding \citep{hoyer:nnsp02}, a method that constrains the responses of sparse coding to be either zero or positive. Non-negative sparse coding is not equivalent to sparse coding with rectification, but constrains the model to find basis functions that combine without inverting contrast (with negative coefficients). The introduction of sparse coding into the overall V2 model also introduces an additional hyperparameter: the L1 regularization coefficient. The L1 regularization coefficient controls the degree of sparsity in the sparse coding responses. Larger values result in fewer active (non-zero) units for reconstructions. Several values for the L1 regularization coefficient in the range of [0.1, 4.0] were explored, including values common for discovering basis functions that are similar to Gabor wavelets in traditional sparse coding as well as much larger values. Values of 0.5 and 4.0 were of particular interest because they maximized the performance on the later classification tasks and forced the model to usually recruit only a few basis functions respectively. The latter approximately maximizes prior information in each individual basis function. This approach is compared to the original ICA model qualitatively with V2 unit visualizations and quantitatively with their performances on several vision tasks. The overcomplete ICA model was less sparse than non-negative sparse coding with a regularization coefficient of 0.5, but we noticed that an approximate ceiling was reached. As the regularization coefficient decreased the classification accuracy increased until a value of about 0.5.

\begin{figure}
	\centering
	\includegraphics[width=\columnwidth]{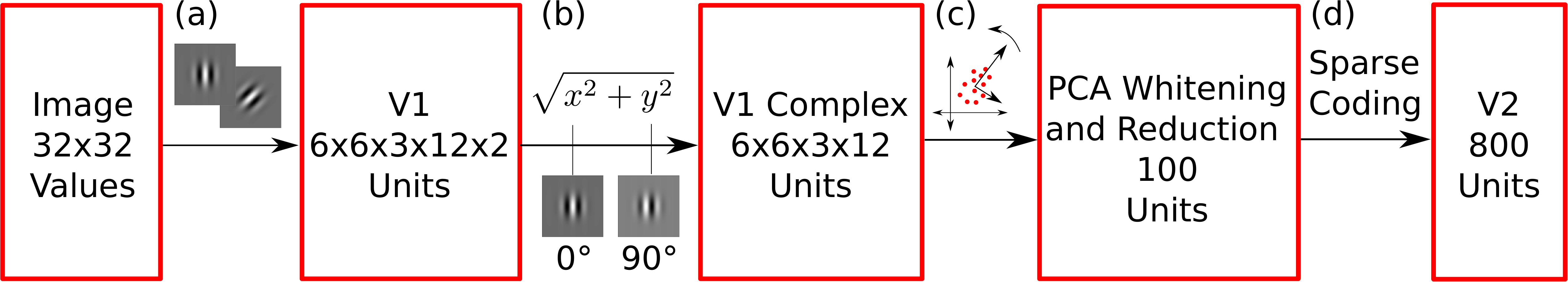}
	\caption{\textbf{Hierarchical V2 Model With Sparse Coding.}  (a) The model begins by computing the responses of the Gabor filters (3 frequencies, 12 orientations, and 2 phases) over the 6x6 central spatial locations. (b) Next, the V1 responses are pooled by taking the square-root of the energy of each pair of filters that are ninety degrees out of phase. (c) The pooling is followed by PCA whitening and reduction down to 100 components. (d) Finally, the representation is expanded with non-negative sparse coding by 8 times. }
	\label{fig:model}
\end{figure}

The models (see figure \ref{fig:model} for an illustration with sparse coding) were trained on 400,000 32x32 image patches from ImageNet ILSVRC12 \citep{russakovsky:ijcv15}. The patches were randomly sampled from images after subtracting the mean and normalizing the variance of the images. Low contrast patches were not included (variance less than 0.32) as was done in \cite{hosoya:jneuro15}. The mean of each patch was also subtracted and its variance normalized. The overcomplete ICA and non-negative sparse coding models were trained for 16 epochs (presentations of the whole training set). The model hyperparameters were matched to that of \cite{hosoya:jneuro15}. The probability density function (the function “G”) of the input under the overcomplete ICA model was the negative log of the hyperbolic cosine function. The model V1 simple cell responses were computed with Gabor filters along the 6x6 center locations of each 32x32 image patch. This is equivalent to 2D convolution of the Gabor filters with the image with a stride of 4 and no padding around the edges of the image. There were 3 frequencies (1.25 cycles, 1.5 cycles, 1.75 cycles), 12 orientations (increments of 15$^\circ$ from 0$^\circ$ to 165$^\circ$), and 2 phases (0$^\circ$ and 90$^\circ$). The filters had a receptive field size of approximately 12x12 pixels. The resulting set of model V1 simple cell responses for the location and parameter choices had a dimension of (6, 6, 3, 12, 2) responses. The model V1 complex cell responses were computed by taking the square-root of the sum of the squares of each quadrature (ninety-degree out of phase) pair of Gabor functions to model phase invariance. The resulting model V1 complex cell responses had a dimension of (6, 6, 3, 12) because the last dimension of the model V1 simple cell responses corresponded to the quadrature pair. Before computing the model V2 responses, the model V1 complex cell responses were pooled with principal component analysis (PCA) by maintaining only the 100 components with the largest eigenvalues. Finally, the V2 responses were computed with overcomplete ICA or non-negative sparse coding with 800 filters or basis functions. The source code for the complete V2 model has been made available at \url{https://notabug.org/jbowren/hv2model}.

The model is illustrated in Figure \ref{fig:model}. The number of components and V2 units matches that of \cite{hosoya:jneuro15}. A second configuration with 11x11 spatial locations and 350 principal components (for the increase in Gabors) was also explored for the inference experiment in order to reconstruct entire patches. The number of V2 units chosen was 2800 to keep the representation 8 times overcomplete.

\begin{figure}
	\centering
	\includegraphics[width=0.6\columnwidth]{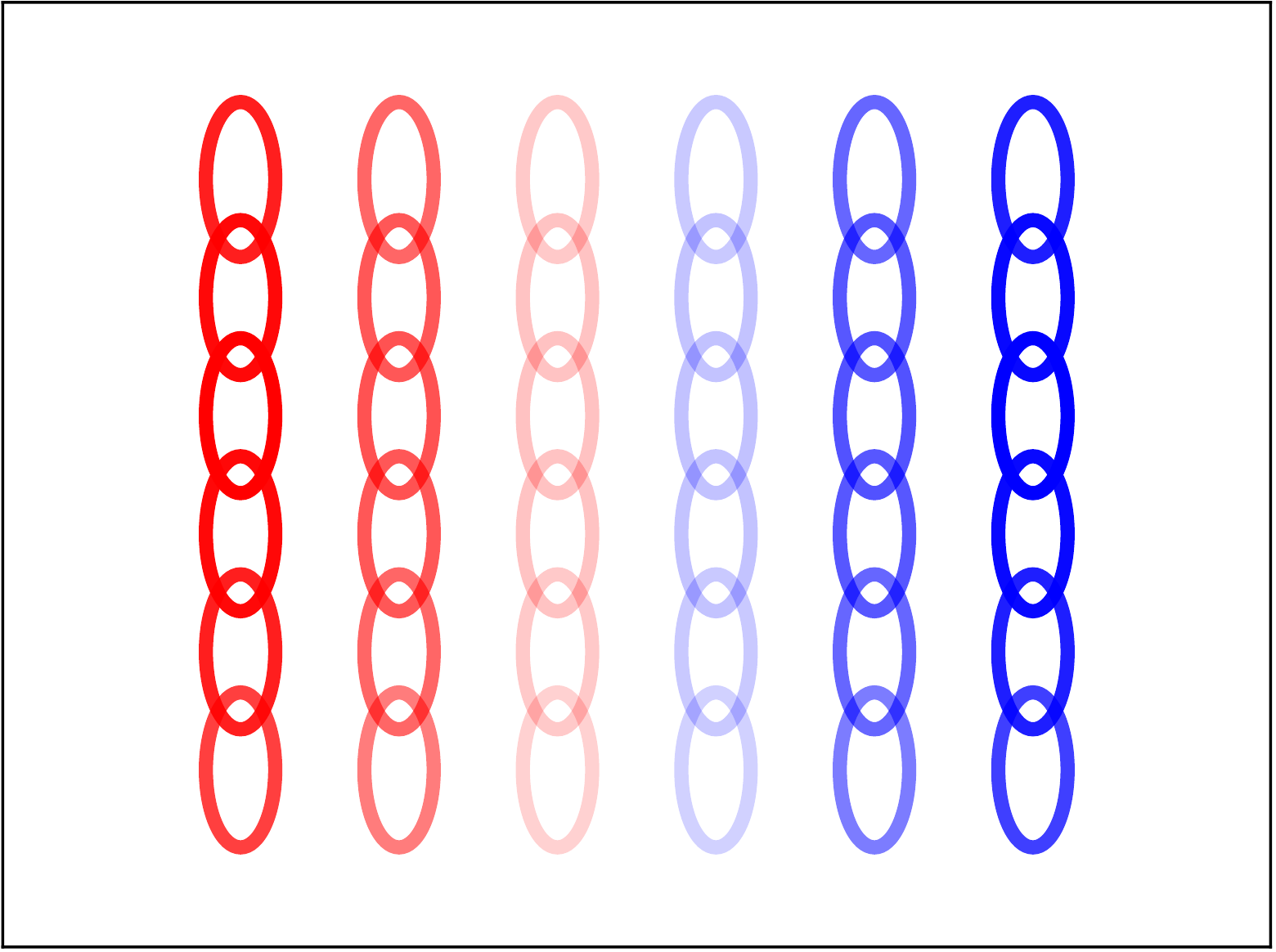}
	\caption{\textbf{V2 Model Unit Visualization.}  There are 6x6 spatial locations, 3 frequencies, and 12 orientations. Opacity reflects the strength of the response and color the sign with red being excitatory and blue inhibitory. Location represents the location of where the Gabor was applied in the image. Orientation represents the orientation of the Gabor. Size represents the frequency of the Gabor (larger size smaller frequency).}
	\label{fig:unit}
\end{figure}

The V2 model neurons were visualized in a similar fashion to that of \cite{hosoya:jneuro15}. First, a 1-of-K representation was inserted into the model as the V2-stage responses where the unit to be visualized is set to 1 and every other unit is set to 0. Next, the model proceeded backward until the corresponding V1 complex responses were obtained. This representation was then plotted in the input 32x32 image space with ovals drawn over the 6x6 center locations in the 32x32 image space. The opacity of the ovals represents the strength of the responses, the color indicates the sign of the response (red for excitatory and blue inhibitory), the size of the ovals reflect the frequency of the Gabors, and the orientation of the ovals represents the orientation of the Gabors. Excitatory (red) Gabors signify the presence of stimuli with the same orientation and frequency (size). Inhibitory (blue) Gabors signify that stimuli with the same orientation and frequency should be absent to maximally excite the unit. An example of a V2 model unit is shown in Figure \ref{fig:unit}. In addition to this visualization, the 6 32x32 image patches that maximally activate each unit are shown in the \nameref{sec:results} to provide insight into the representation of the units.

\begin{figure}
	\centering
	\includegraphics[width=0.38\columnwidth]{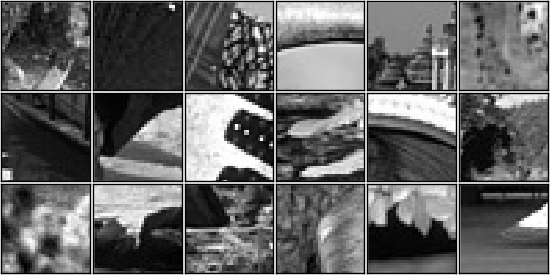}
	\caption{\textbf{32x32 Figure-Ground Image Patches.}  Example 32x32 image patches sampled from the Berkeley Segmentation Database \citep[BSDS300;][]{fowlkes:jov07}. The label for each image patch indicates whether the figure of the image falls primarily on the right side or left side of the image. The labels were determined from the corresponding 32x32 region of the human-drawn contour line map for each image.}
	\label{fig:fg}
\end{figure}

\begin{figure}
	\centering
	\includegraphics[width=0.6\columnwidth]{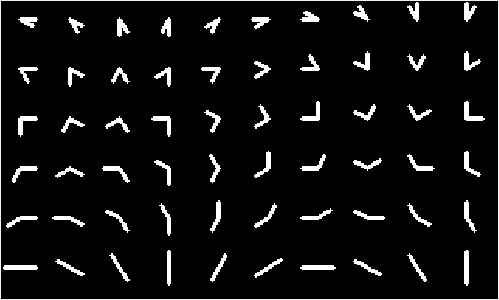}
	\caption{\textbf{32x32 Line Stimuli.} Two lines connected at one end-point with 3x3 center locations, 6 angles, 12 rotations, and 3 lengths. Each line image is a 32x32 image.}
	\label{fig:line}
\end{figure}

The resulting V2 models with non-negative sparse coding and overcomplete ICA were run on three different image datasets. The datasets were selected because they are considered to capture mid-level visual tasks that could be appropriate at the V2 level, namely to perform figure-ground classification, texture classification, and to predict the angle between two lines connected at one end-point. Classification was performed by training a linear SVM (with the SVC classifier of Scikit-Learn \citep{{pedregosa:jmlr11}}) on the model responses of each model configuration. The choice of regularization coefficient could influence the results, so a few values of the regularization coefficient for the linear SVM were tested as well a logistic regression model with the same values of the regularization coefficient (and without a regularization coefficient), but ICA consistently performed the best outside of the error standard deviation bars across all configurations and datasets. We include the classification results for these models and all the values of the regularization coefficient we tested in \nameref{sec:appA}. Since the regularization coefficient did not change the model that performed best, we simply report the result of the SVM classification with a regularization coefficient of 1. For each dataset, 5-fold cross validation was performed and the average accuracy and standard deviation were recorded. For the figure-ground experiment, the images and figure-ground labelings were obtained from the Berkeley Segmentation DataSet \citep[BSDS300;][]{fowlkes:jov07}. Figure and ground refer to regions of images separated by some contour in the image that determines the main region of focus for an observer. The region denoted as the figure is the main region of focus that might grab the foveal attention of an observer while the ground is considered to provide context to the figure. Neural processing is thought to have a mechanism of distinguishing between figure and ground regions \citep[see][for a more in-depth description]{coen:jov13}. \cite{fowlkes:jov07} showed that the regions of images with figure rather than ground were usually smaller and more convex, so these convex regions probably require more advanced features for image classification. A total of 20,000 32x32 image patches (see Figure \ref{fig:fg}) were randomly sampled from the dataset with labels (figure or ground) assigned based on which side of the image (left or right) the human-labeled contour primarily fell. The second experiment included synthetically generated images according to \cite{portilla:ijcv00}. The images were generated to match the low order statistics of different classes of real texture images from the Brodatz dataset \citep{brodatz:dover66}. The models were tested on 30,000 32x32 patches sampled from 15 texture categories; an example of a few texture patch families are shown in the left column of Figure \ref{fig:tex}. The last classification experiment tested the models on line segments joined at one end-point with varying lengths, locations, rotations, and angles between the two lines. There are a total of 3 lengths (10, 15, and 20 pixels), 9 locations (the 3x3 center locations in the image), 12 rotations (0$^\circ$ to 330$^\circ$ with an interval of 30$^\circ$), and 6 angles (30$^\circ$ to 180$^\circ$ with an interval of 30$^\circ$). Examples of the line stimuli are shown in Figure \ref{fig:line}.

\begin{figure}
	\centering
	\begin{subfigure}[h]{\columnwidth}
		{}
		\hfill
		\begin{subfigure}[h]{0.4\columnwidth}
			\includegraphics[width=\columnwidth]{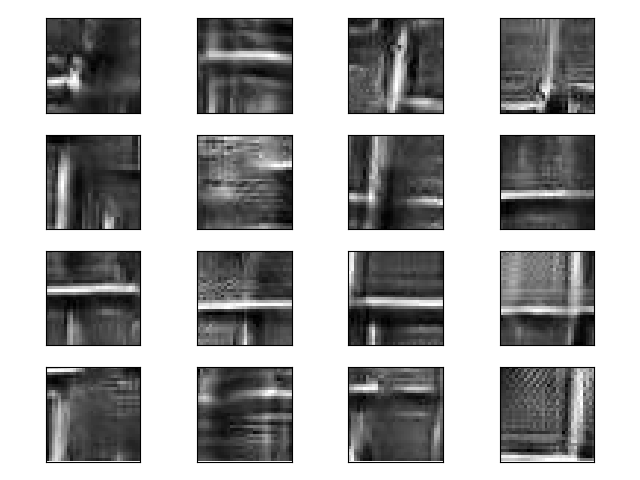}
		\end{subfigure}
		\hfill
		\begin{subfigure}[h]{0.4\columnwidth}
			\includegraphics[width=\columnwidth]{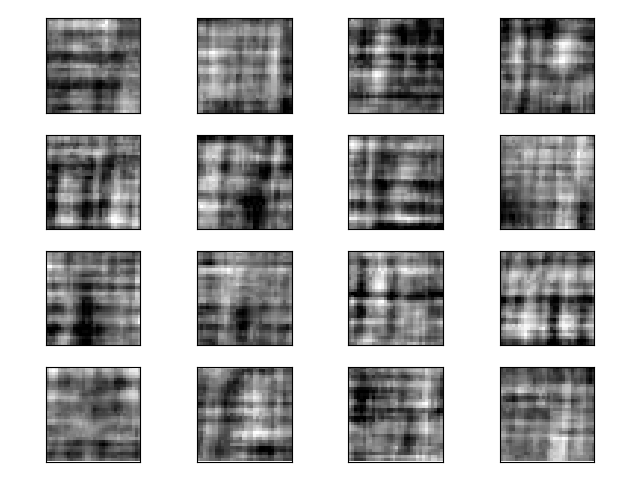}
		\end{subfigure}
		\hfill
		{}
	\end{subfigure} \\
	\vspace{3em}
	\begin{subfigure}[h]{\columnwidth}
		{}
		\hfill
		\begin{subfigure}[h]{0.4\columnwidth}
			\includegraphics[width=\columnwidth]{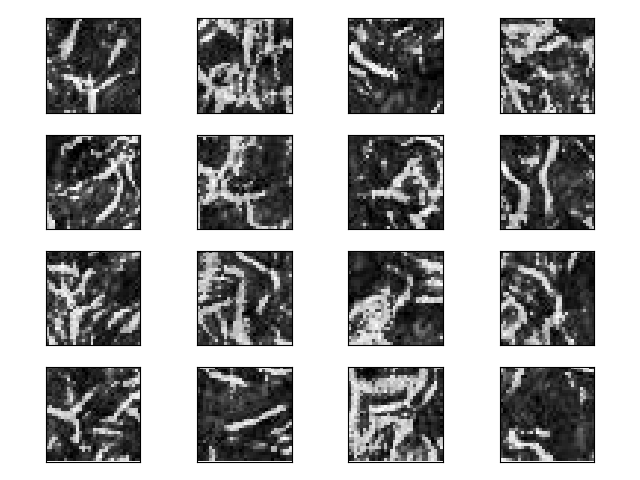}
		\end{subfigure}
		\hfill
		\begin{subfigure}[h]{0.4\columnwidth}
			\includegraphics[width=\columnwidth]{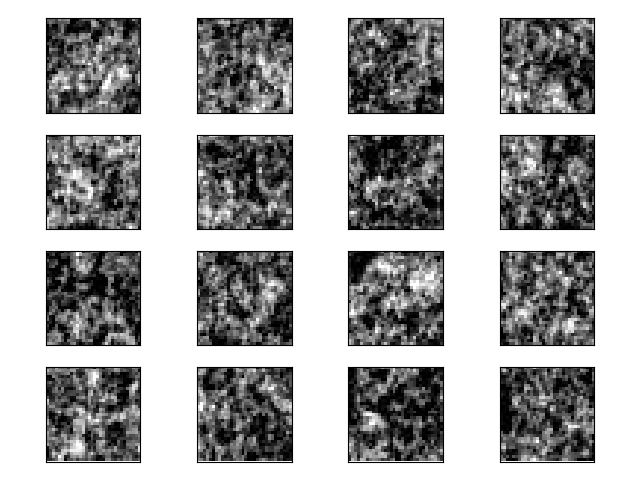}
		\end{subfigure}
		\hfill
		{}
	\end{subfigure} \\
	\vspace{3em}
	\begin{subfigure}[h]{\columnwidth}
		{}
		\hfill
		\begin{subfigure}[h]{0.4\columnwidth}
			\includegraphics[width=\columnwidth]{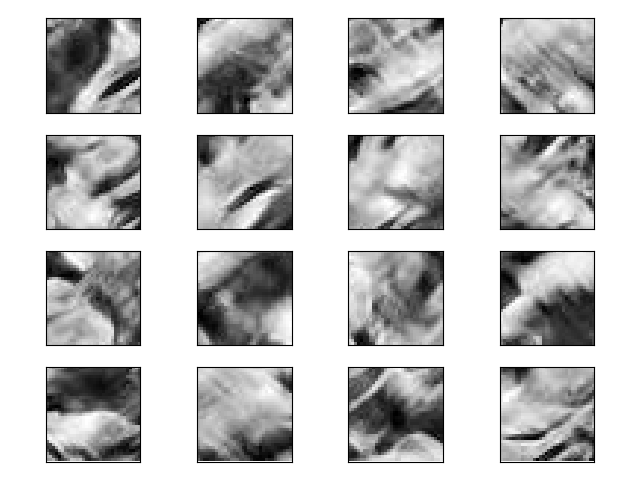}
		\end{subfigure}
		\hfill
		\begin{subfigure}[h]{0.4\columnwidth}
			\includegraphics[width=\columnwidth]{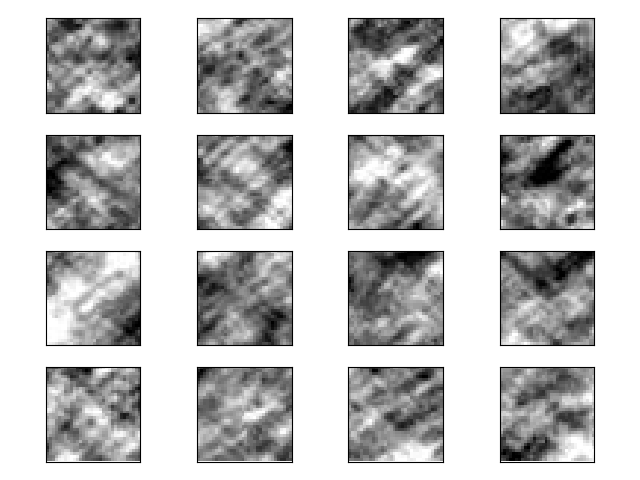}
		\end{subfigure}
		\hfill
		{}
	\end{subfigure}
\end{figure}

\begin{figure}
	\ContinuedFloat
	\centering
	\begin{subfigure}[h]{\columnwidth}
		{}
		\hfill
		\begin{subfigure}[h]{0.4\columnwidth}
			\includegraphics[width=\columnwidth]{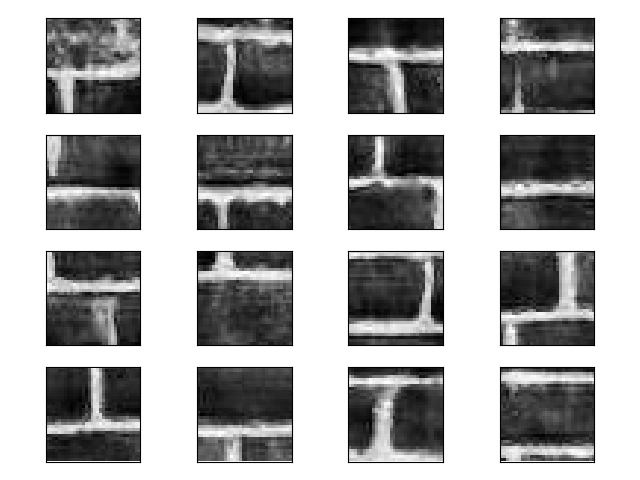}
		\end{subfigure}
		\hfill
		\begin{subfigure}[h]{0.4\columnwidth}
			\includegraphics[width=\columnwidth]{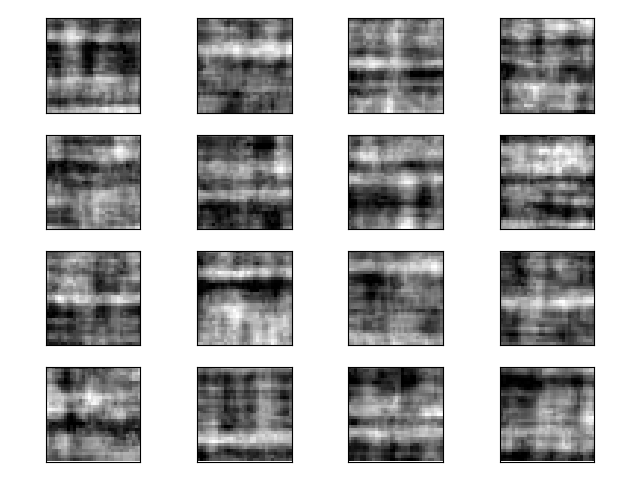}
		\end{subfigure}
		\hfill
		{}
	\end{subfigure}
	\caption{\textbf{32x32 Texture and Noise Patches Extracted from 256x256 Generated Texture and Noise Images.} These 32x32 patches were extracted from 256x256 Brodatz texture images generated with the code made available by \cite{portilla:ijcv00}. The column on the left shows 4x4 grids of 32x32 patches where each grid corresponds to a different texture. The column on the right shows the corresponding spectrally-matched noise versions of the texture patches extracted at the same locations within the spectrally-matched noise versions of the texture images. A complete set of the full size 256x256 texture and noise images are shown in \nameref{sec:appB}.}
	\label{fig:tex}
\end{figure}

Next, the texture sensitivity of the models was measured via the texture modulation index computed with 30,000 32x32 texture patches and 30,000 32x32 spectrally-matched noise versions (see end of paragraph) of the same patches. The texture modulation index is a measure of texture sensitivity in the range of [-1, 1] where 1 indicates the maximal sensitivity for texture and -1 the opposite. The texture modulation index is calculated by taking the difference of the responses of a model (or brain region) to texture stimuli and noise stimuli, then normalizing by the sum of the two. Here, the index was averaged over all of the model neurons. The equation for calculating the modulation index $M$ is given by
\begin{align}
	M = \frac{r_{tex} - r_{noise}}{r_{tex} + r_{noise}}
\end{align}

\noindent where $r_{tex}$ is the response to a texture stimuli and $r_{noise}$ is the response to a spectrally-matched noise version of the texture patch. The 30,000 texture patches were taken from the same textures in the classification experiment. The corresponding 30,000 spectrally-matched noise patches were obtained by taking 32x32 patches at the same locations of the texture patches from spectrally-matched noise versions of the original 256x256 texture stimuli. The spectrally-matched noise versions of the original texture stimuli were generated by first computing the magnitude and phase of a fast-Fourier transform (FFT) of each texture and a corresponding randomly generated Gaussian white noise image. Next, the phase component of the original texture image was replaced with the phase component of the Gaussian white noise image. Finally, the spectrally-matched noise images were obtained by performing the inverse FFT on the new magnitude and phase representation. This ensures that the magnitude of the FFT of the spectrally-matched noise image is the same as the magnitude of the FFT of the synthesized texture image with uniform random phase \citep[see][]{galerne:tip}.

Next the ability of the models to fill in missing information was tested by deleting part of the image representation within the model before non-negative sparse coding or overcomplete ICA, then going backward to reconstruct the image from the models’ responses. The backward computation proceeded by multiplying the V2 responses by the sparse coding dictionary (or ICA mixing matrix for overcomplete ICA) and the inverse PCA whitening matrix to recover the V1 complex responses. Next, the V1 simple responses were calculated from the V1 complex responses and the angles between each pair of responses (in polar coordinates) which were saved in the forward computation; this does not affect the relative reconstruction accuracy of the two models. Finally, the image was reconstructed by convolving the V1 simple responses with the transpose of the original Gabor filters. This is an approximation, but computationally practical to undo the forward Gabor filter transform. A total of 1000 image patches were sampled from ImageNet and fed forward through the model until model V1 complex responses were obtained. Deletion was then performed by setting either a 1x1 or 2x2 region of the V1 complex responses to the minimum value of the responses minus 1. This corresponded to about a 2x2 or 6x6 region respectively in the original image space. Next the responses were filtered with the PCA whitening matrix, then fed through either non-negative sparse coding (with a regularization coefficient of 2.0 or 4.0) or overcomplete independent component analysis. A variety of values of the sparse coding regularization coefficient in the range [0.1, 4.0] were tested, and the pair 2.0 and 4.0 was shown instead of 0.5 and 4.0 because a regularization coefficient of 0.5 was not large enough to significantly change the input representation. Finally, the transformations were undone as detailed above to get the models’ representation in the original image space. The reconstructions were visually inspected side-by-side to the original image patches, and the average mean-squared error (MSE) of the representations and the original image patches was computed for both models and compared.
	
\section{Results}\label{sec:results}

\subsection{Unit Properties}

The visualizations for non-negative sparse coding with regularization coefficients of 0.5 and 4.0 and overcomplete ICA when run with 6x6 spatial locations are shown in Figures \ref{fig:6x6gpsa}, \ref{fig:6x6gpsb}, and \ref{fig:6x6gpsc} respectively. The two models discovered qualitatively different units. The non-negative sparse coding units contain corners, curves, circles, lines, parallel lines, and other structures. The overcomplete ICA units contain iso-oriented excitation with broad, side, cross, and end inhibition units and orientation-convergent excitation with end inhibition units as defined by \cite{hosoya:jneuro15}. Each non-negative sparse coding unit also recruited more excitatory values than inhibitory values while overcomplete ICA had more balance between excitation and inhibition. However, it is important to remind the reader that, unlike overcomplete ICA, in non-negative sparse coding there is no explicit form of forward computation, so positive and negative values do not bear the same meaning. An excitatory value in an overcomplete ICA unit simply means that stimulus was present with the orientation and frequency depicted by the Gabor plot and an inhibitory value the opposite, but the same stimulus can be described by negating the sign of the unit and unit response. By contrast, an excitatory value in a non-negative sparse coding unit means that a stimulus with the given orientation and frequency was useful for reconstructing the input, but the sign of the unit cannot be flipped because the model is non-negative. The results for 11x11 spatial locations with regularization coefficients of 2.0 and 4.0 are shown in Figures \ref{fig:11x11gpsa} and \ref{fig:11x11gpsb}. The units discovered by non-negative sparse coding and overcomplete ICA (figure \ref{fig:11x11gpsc}) with 11x11 spatial locations were similar to that for 6x6 spatial locations.

\begin{figure}
\Large \textbf{(a)} \\
\begin{subfigure}[t]{\linewidth}
	\centering
	\includegraphics[width=0.19\linewidth]{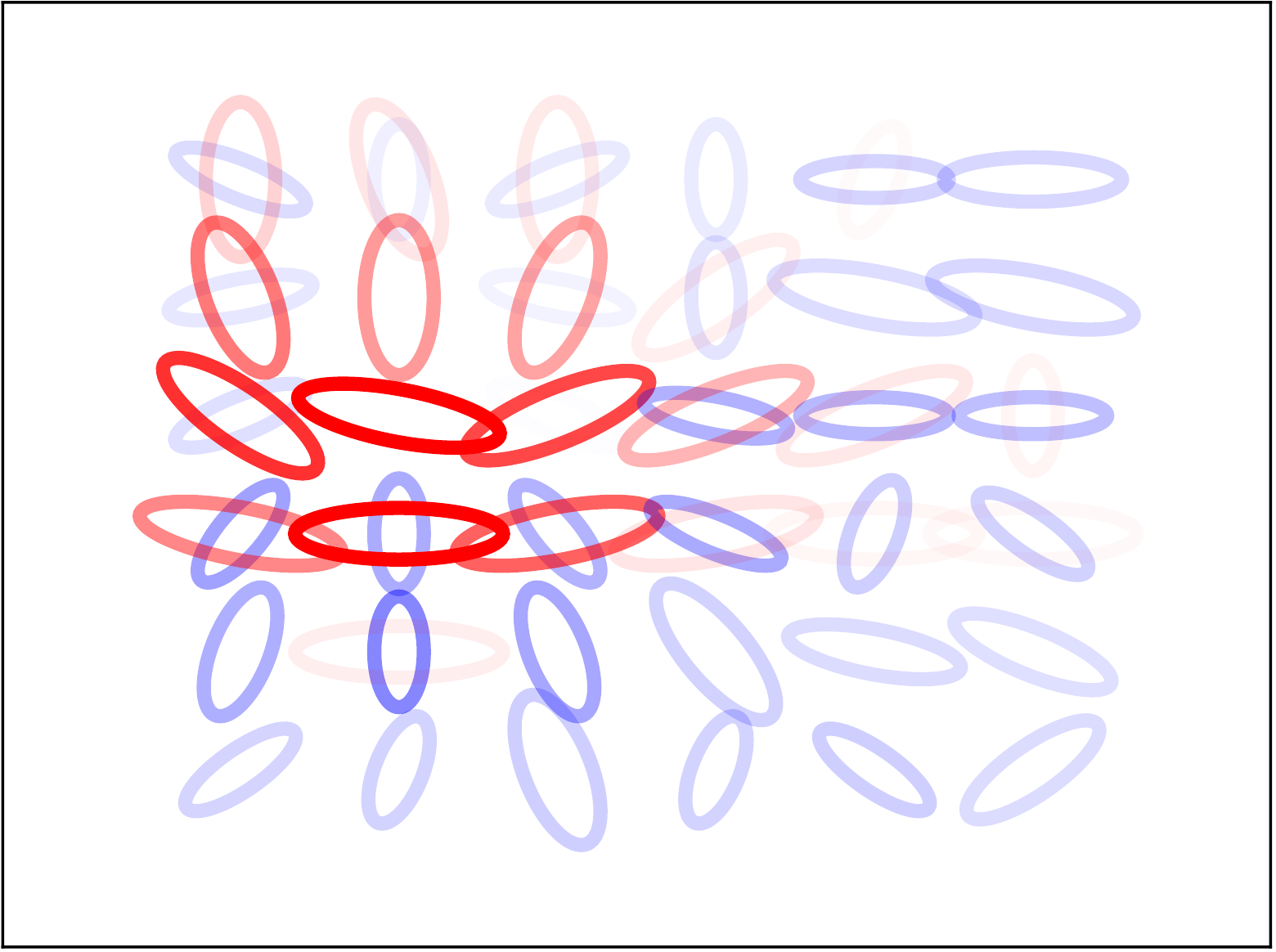}
	\includegraphics[width=0.19\linewidth]{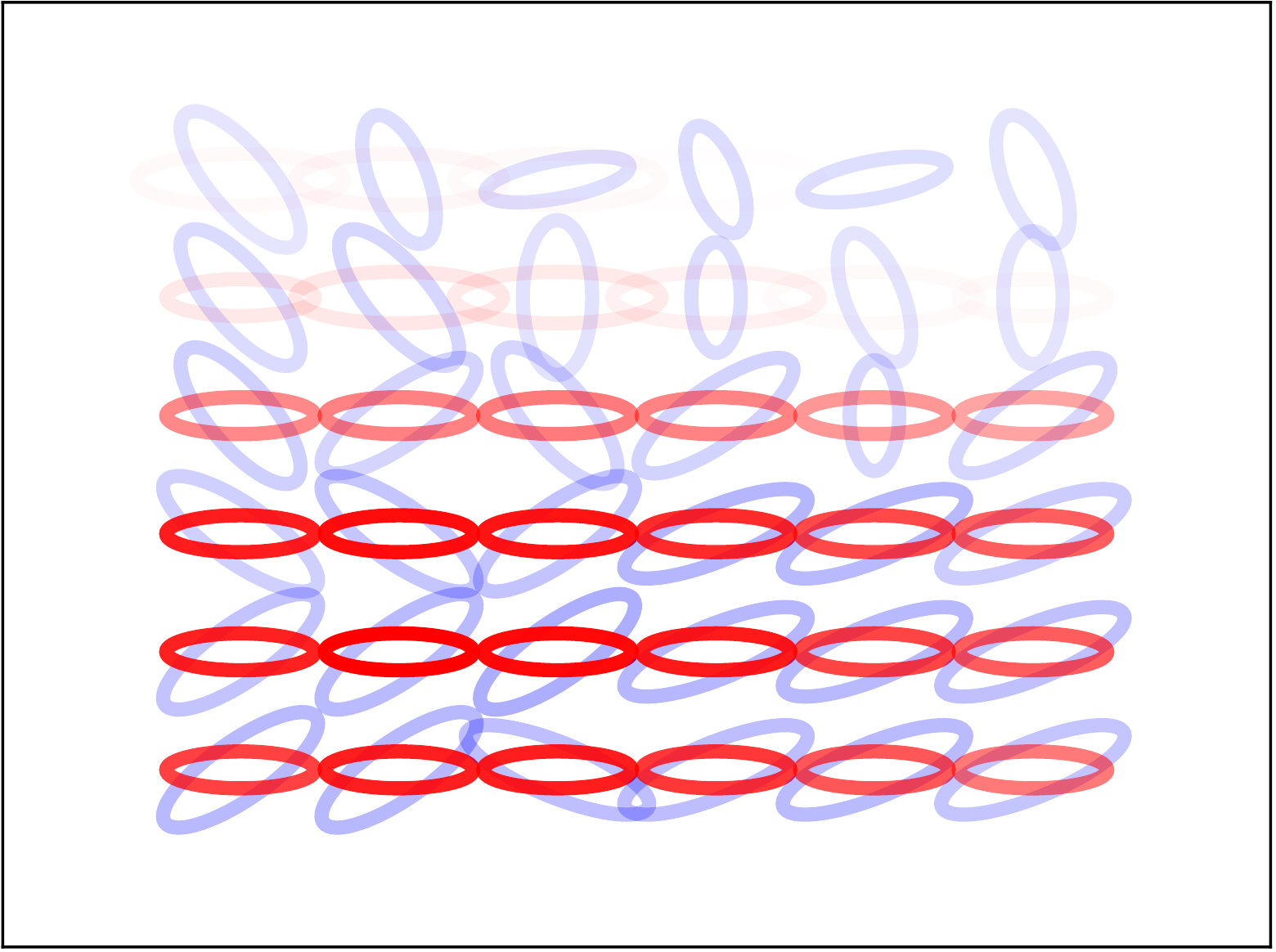}
	\includegraphics[width=0.19\linewidth]{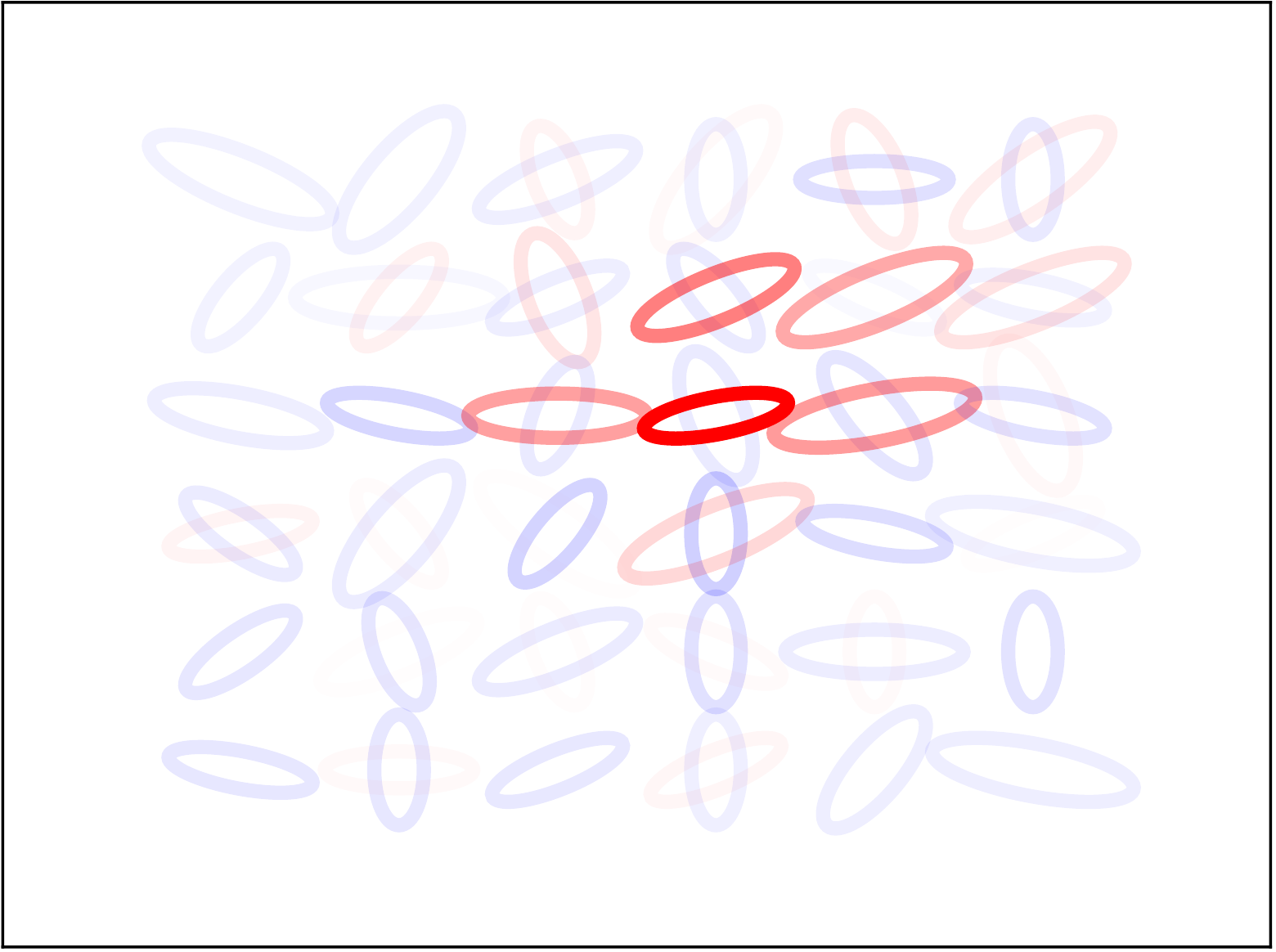}
	\includegraphics[width=0.19\linewidth]{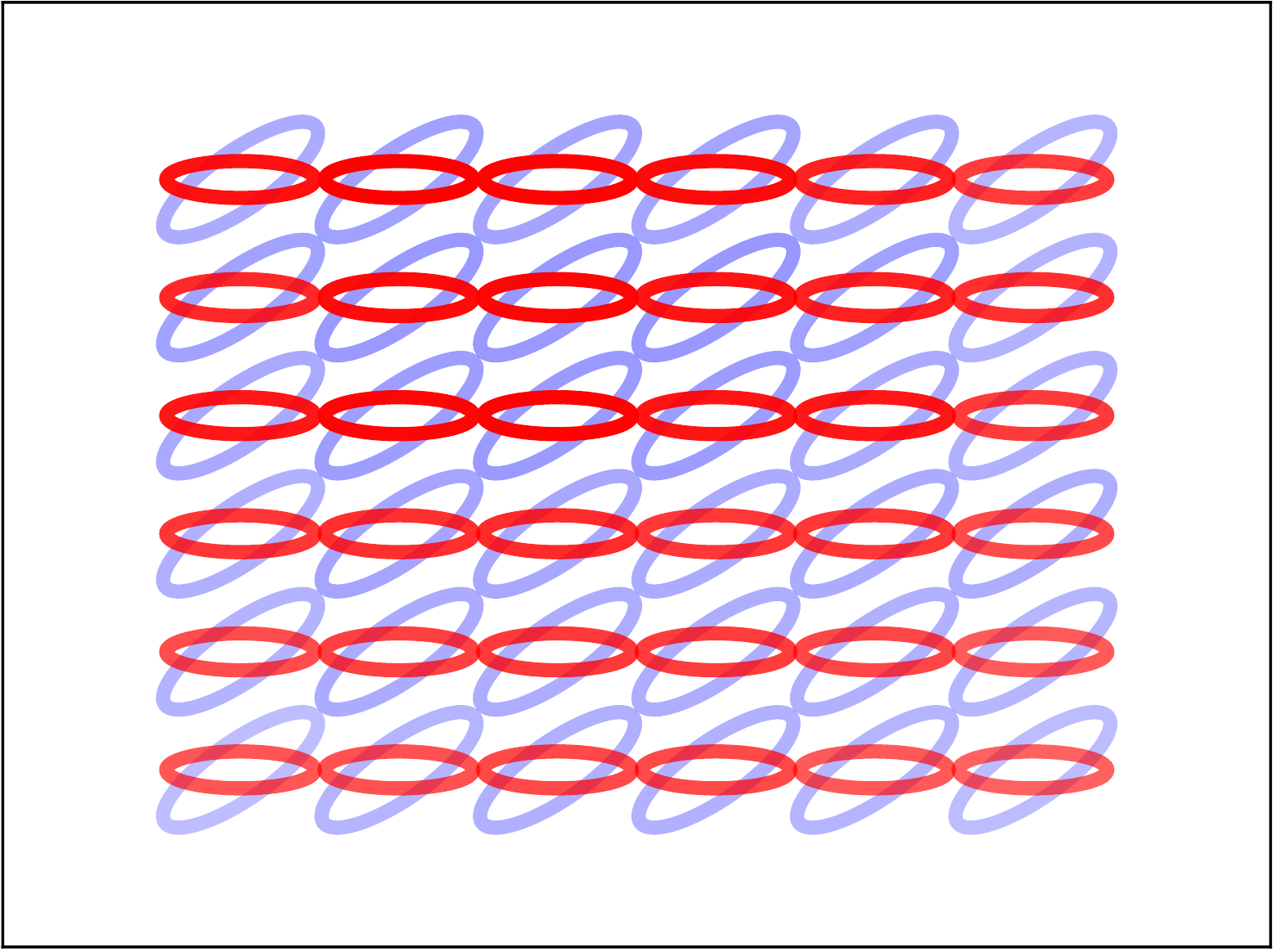}
	\includegraphics[width=0.19\linewidth]{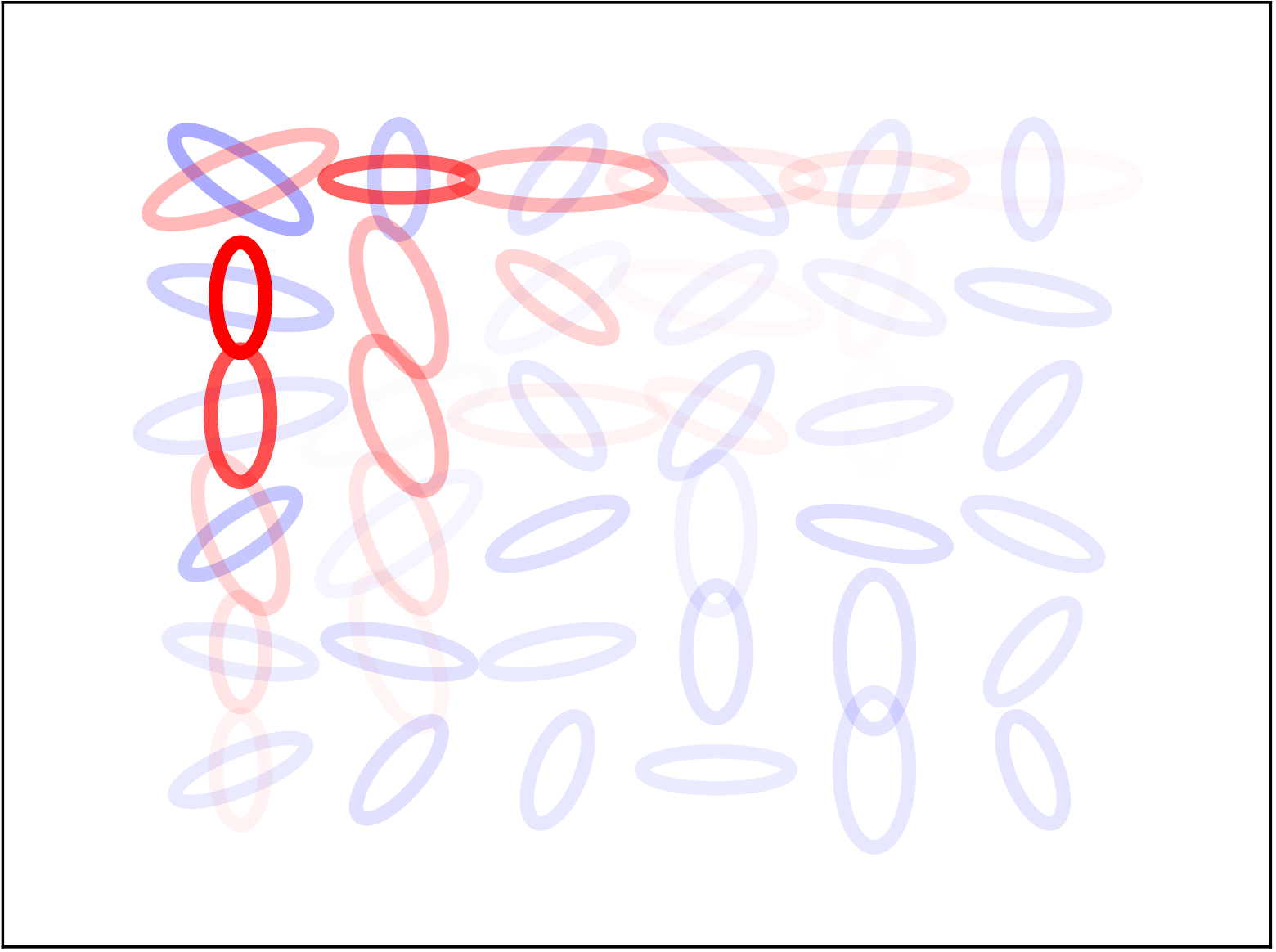}
	\includegraphics[width=0.19\linewidth]{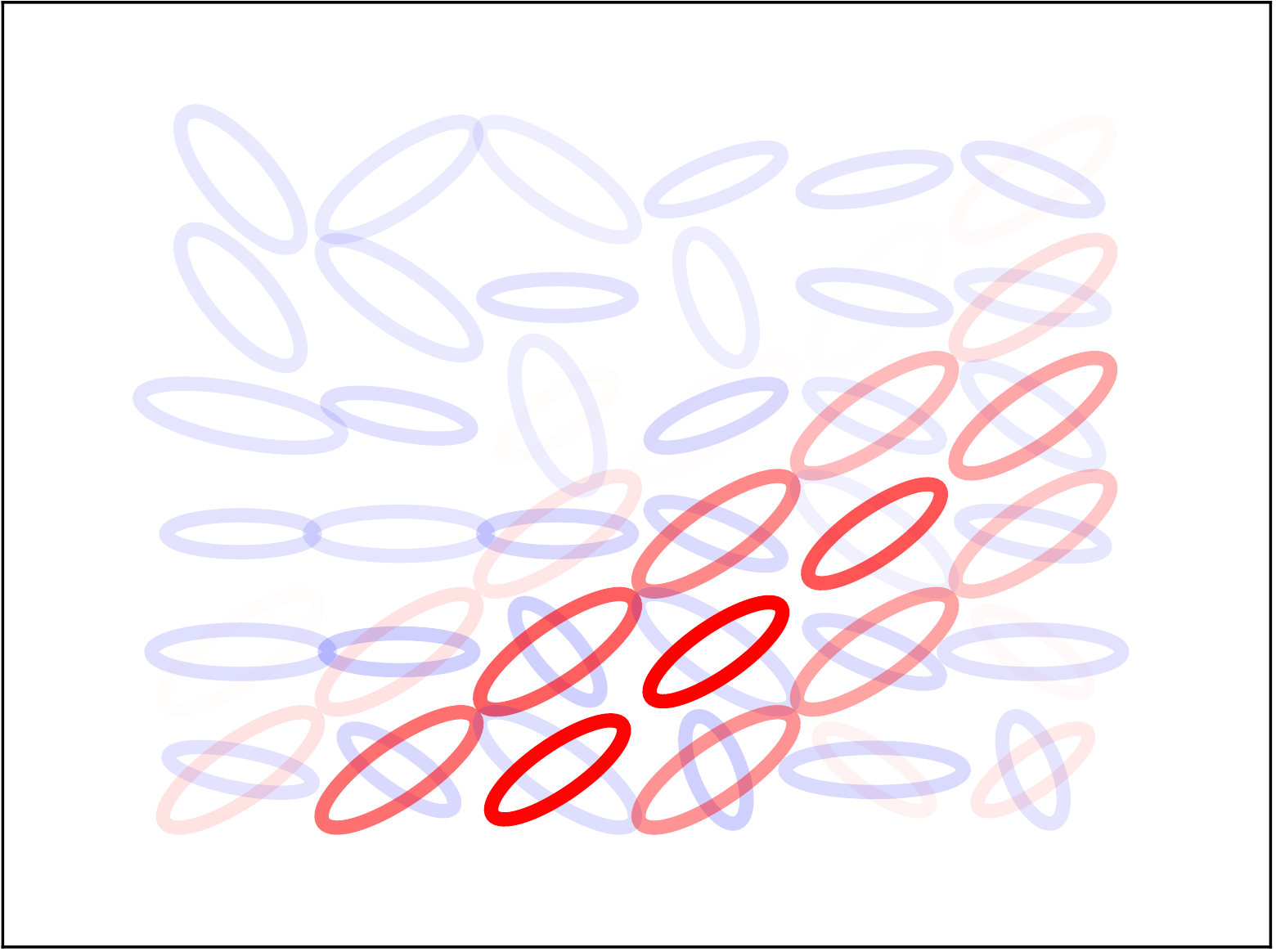}
	\includegraphics[width=0.19\linewidth]{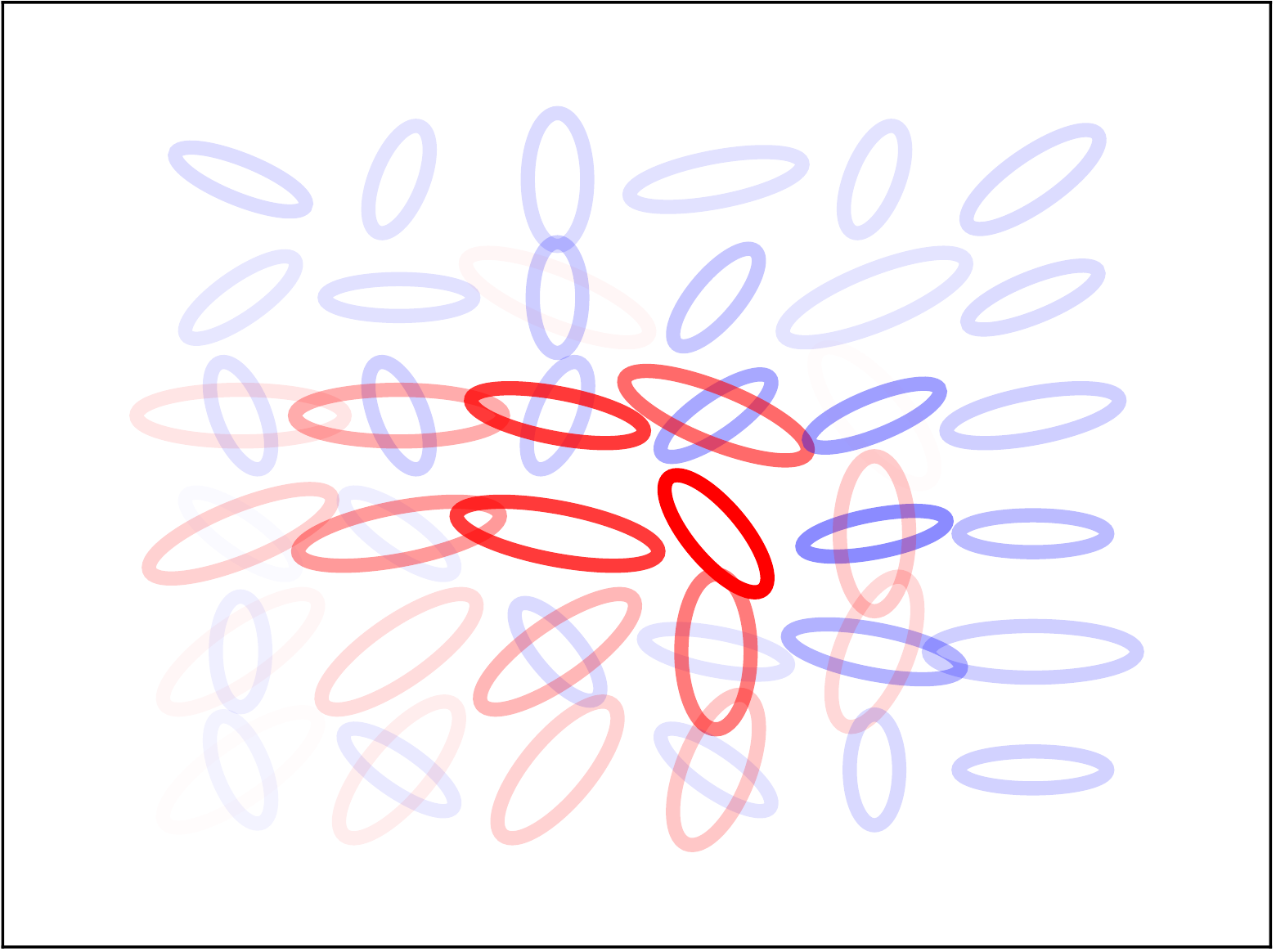}
	\includegraphics[width=0.19\linewidth]{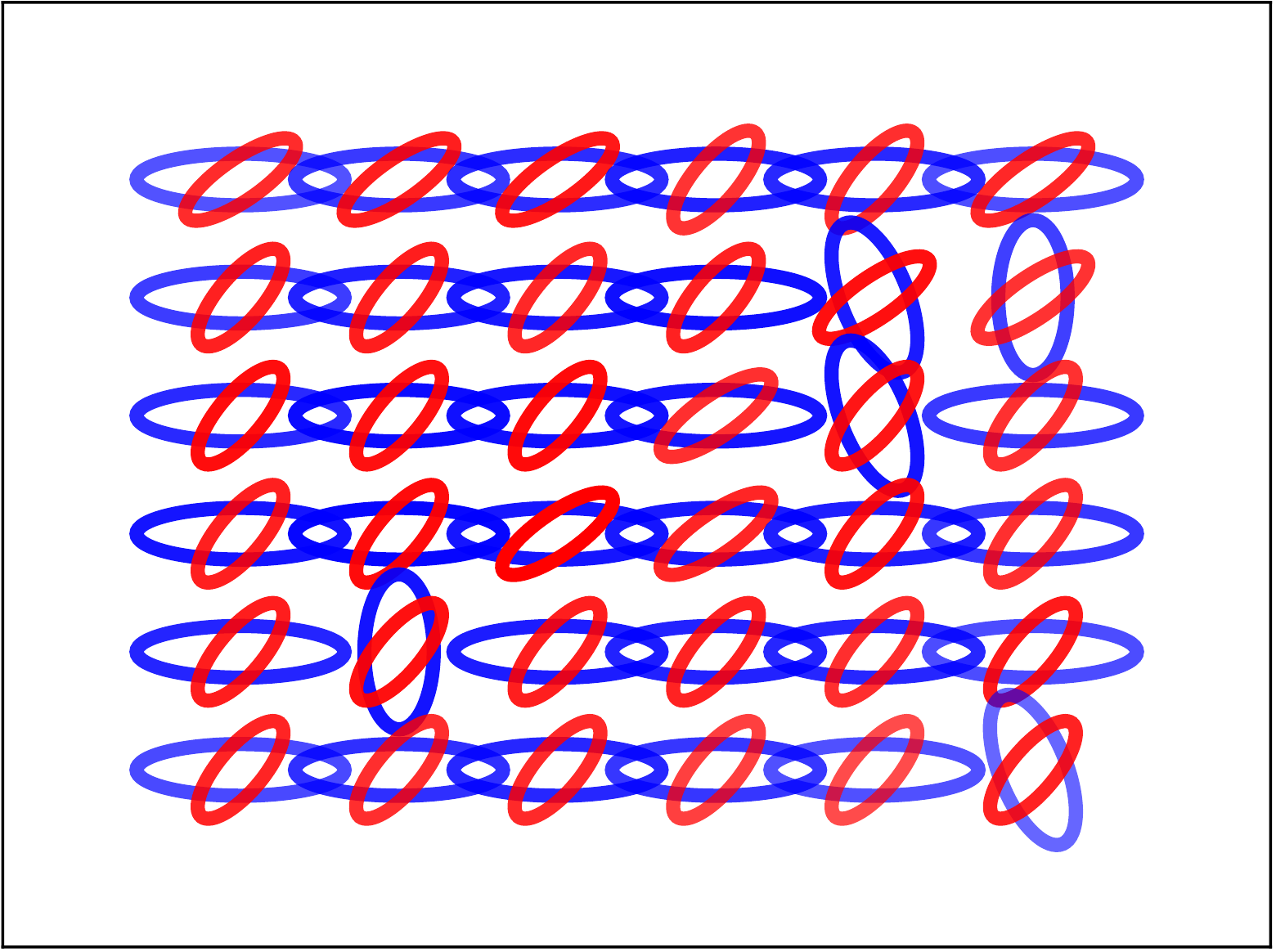}
	\includegraphics[width=0.19\linewidth]{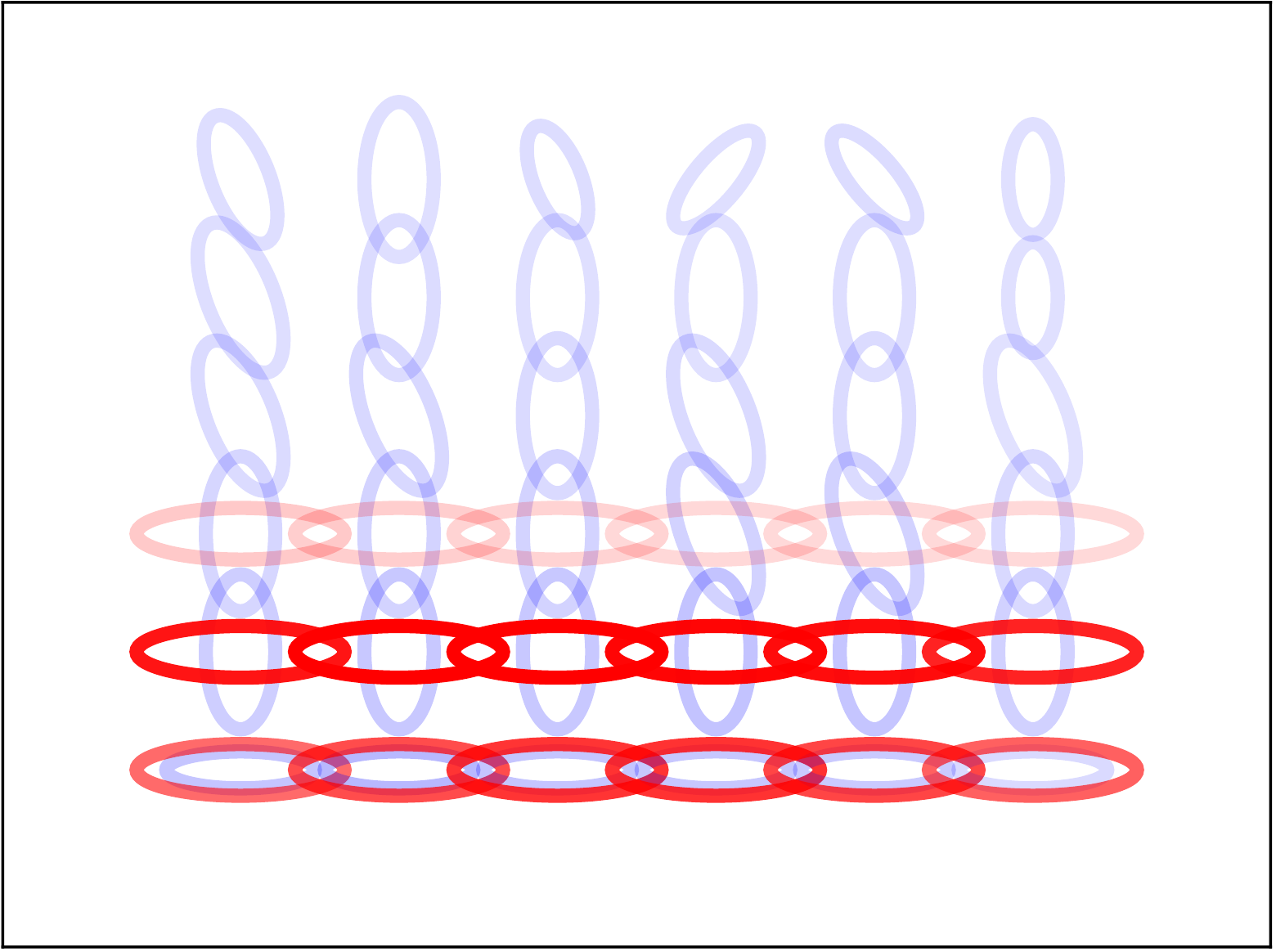}
	\includegraphics[width=0.19\linewidth]{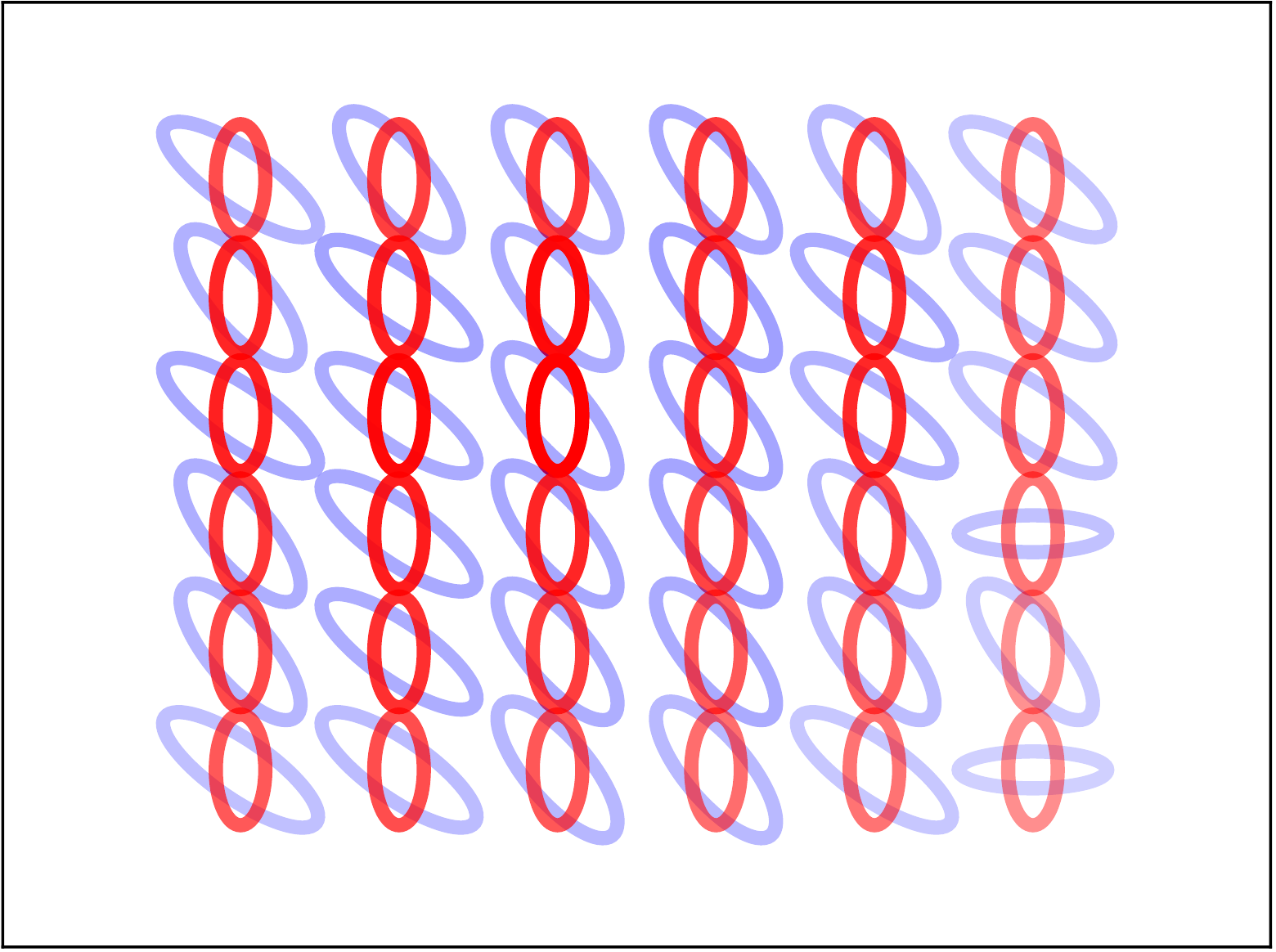}
	\includegraphics[width=0.19\linewidth]{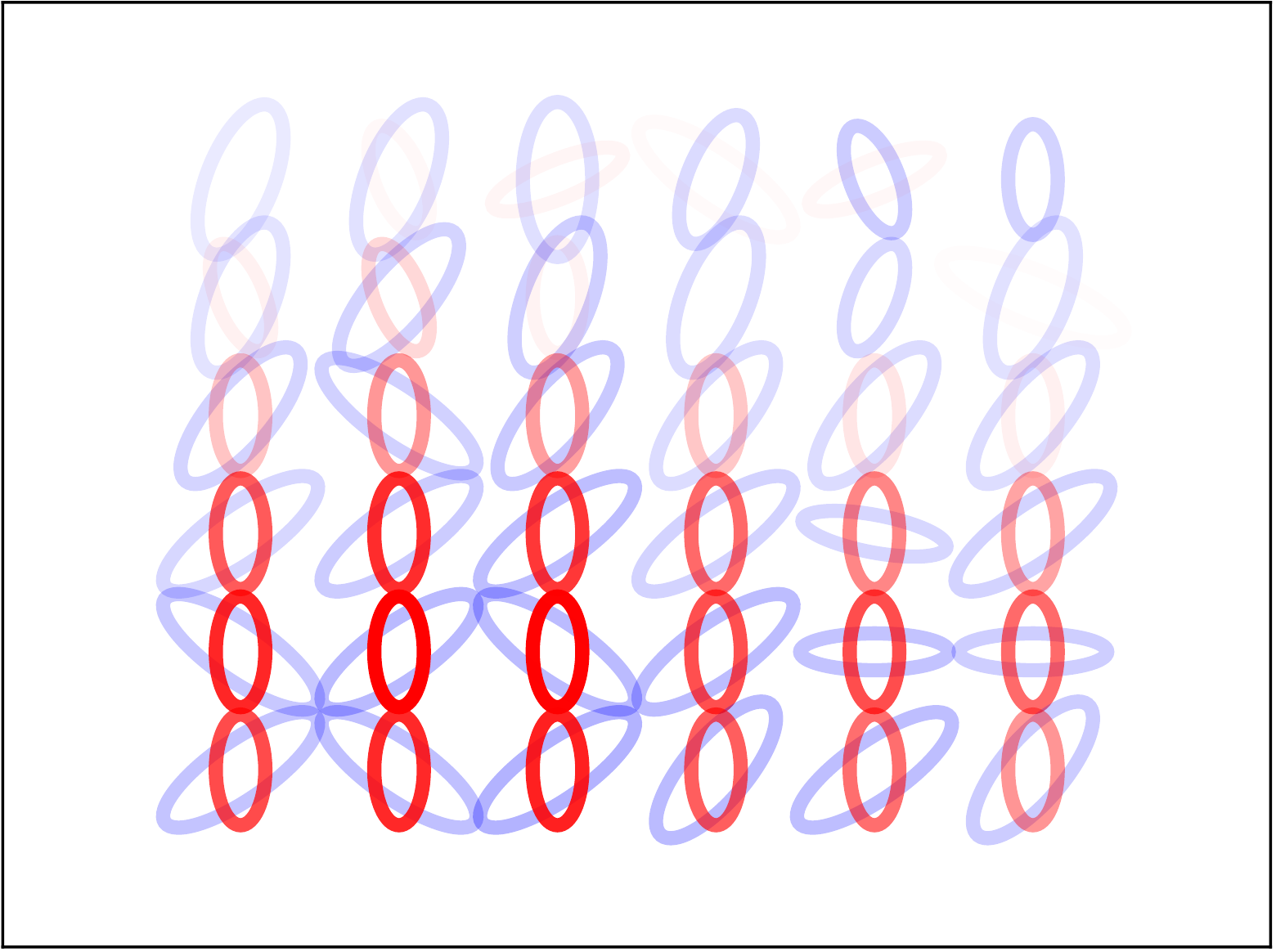}
	\includegraphics[width=0.19\linewidth]{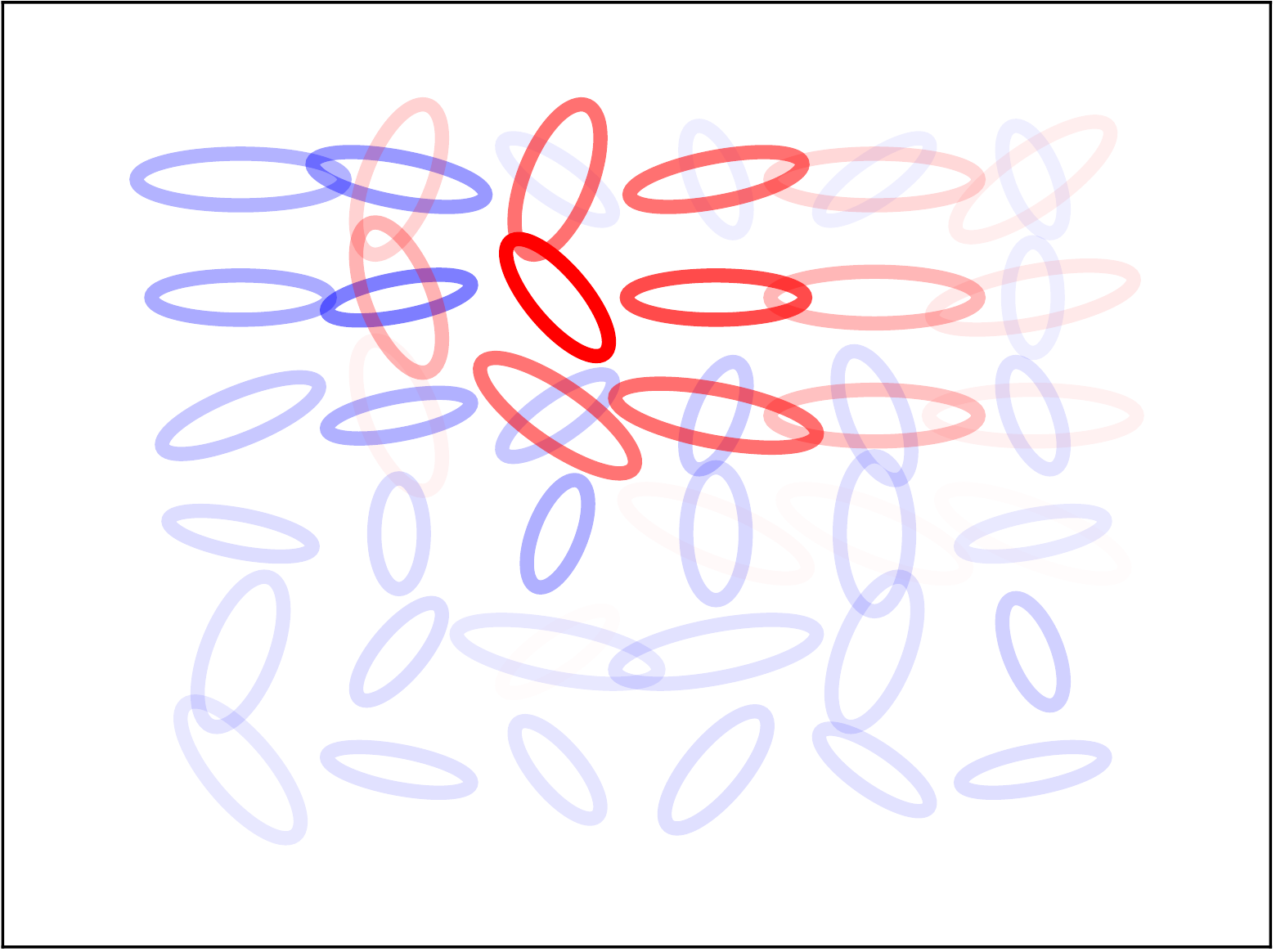}
	\includegraphics[width=0.19\linewidth]{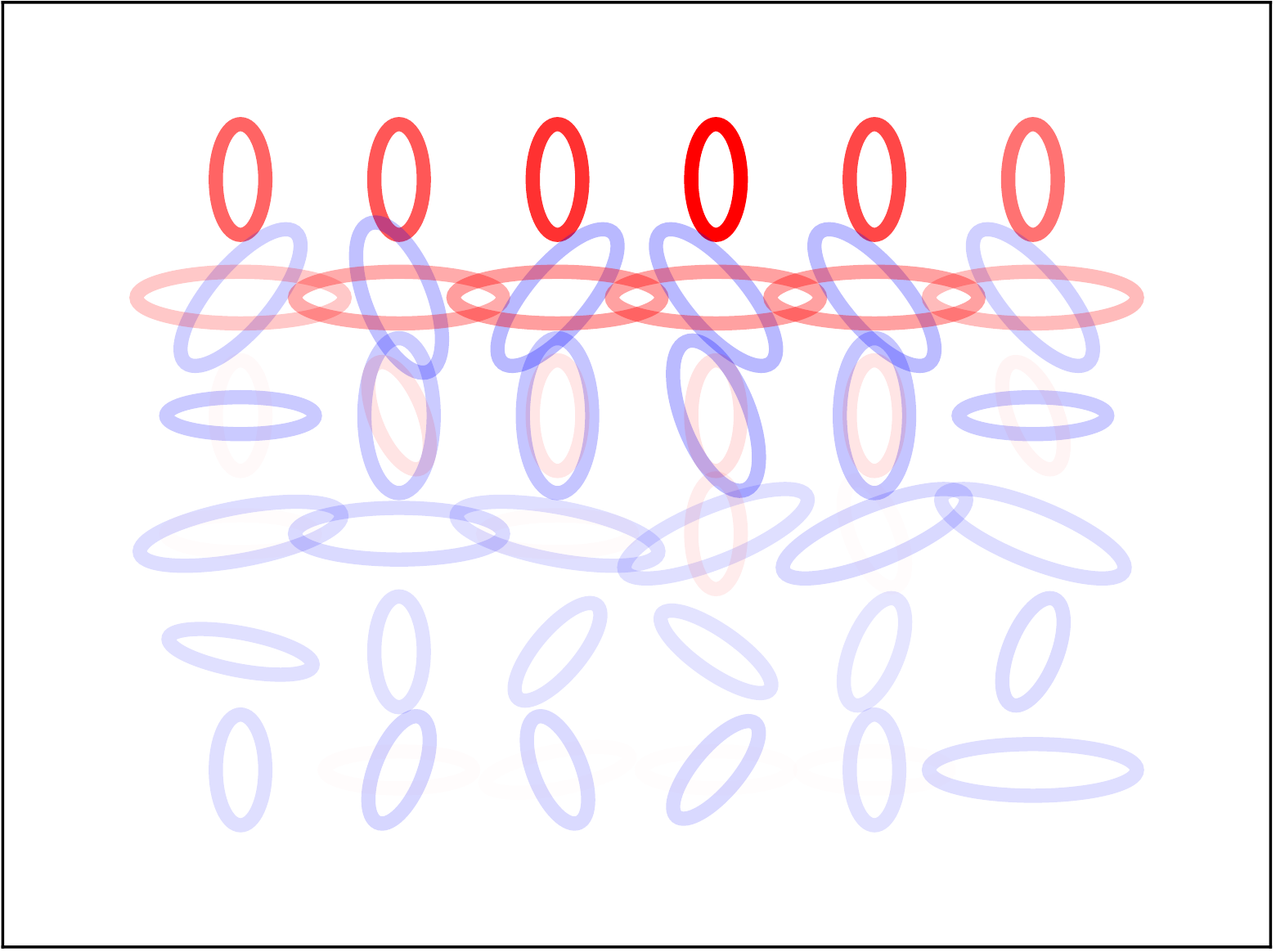}
	\includegraphics[width=0.19\linewidth]{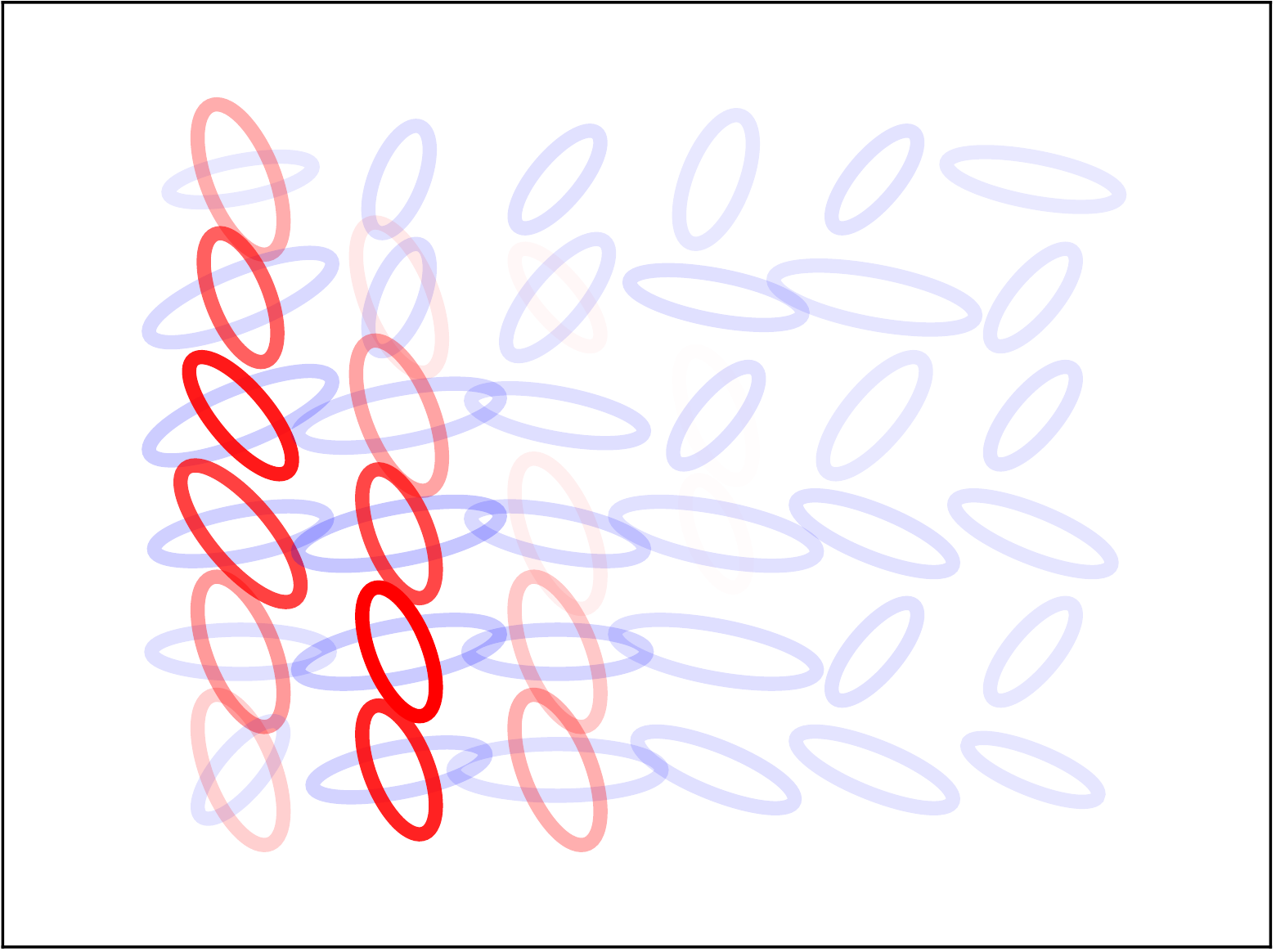}
	\includegraphics[width=0.19\linewidth]{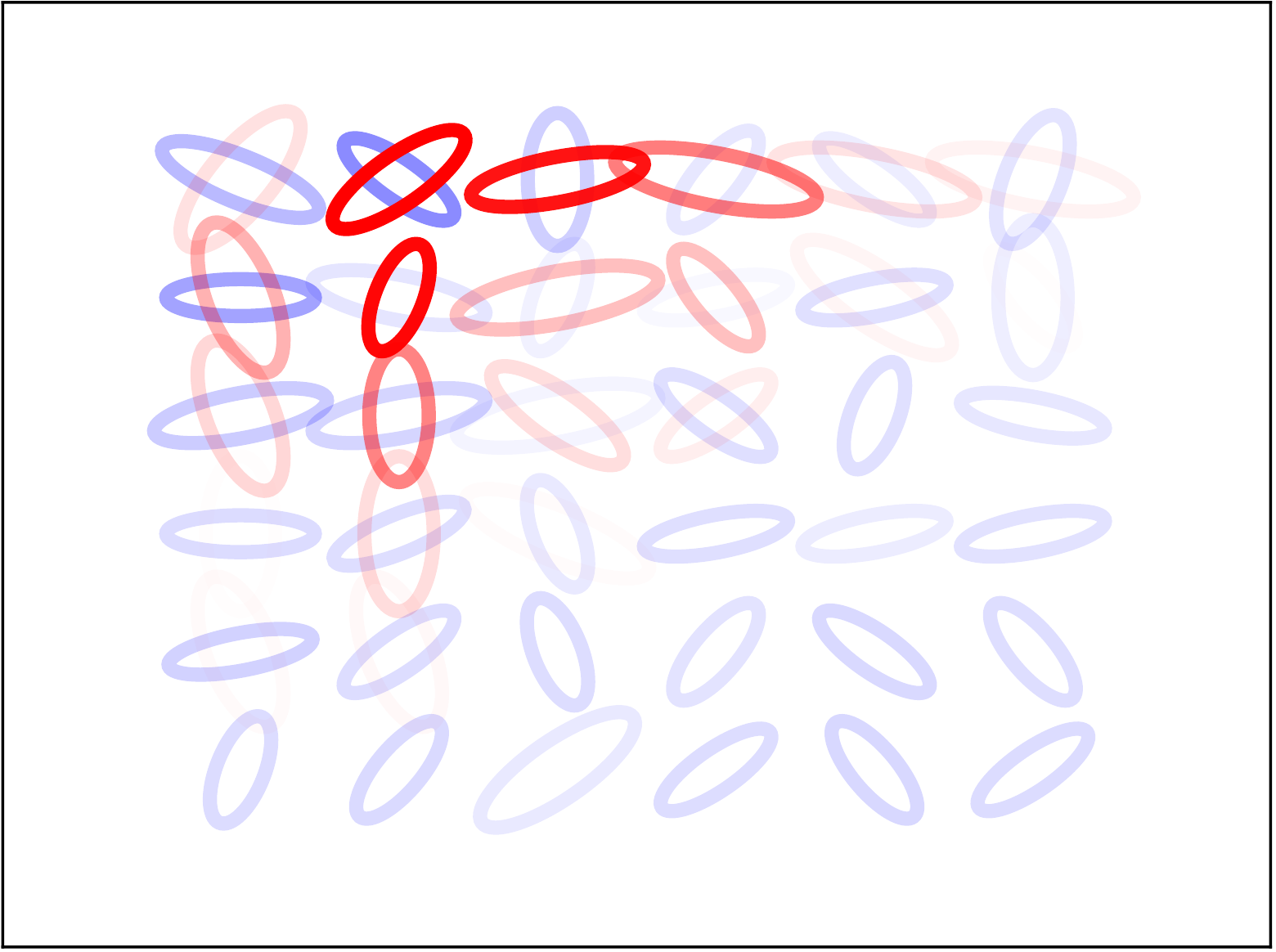}
	\includegraphics[width=0.19\linewidth]{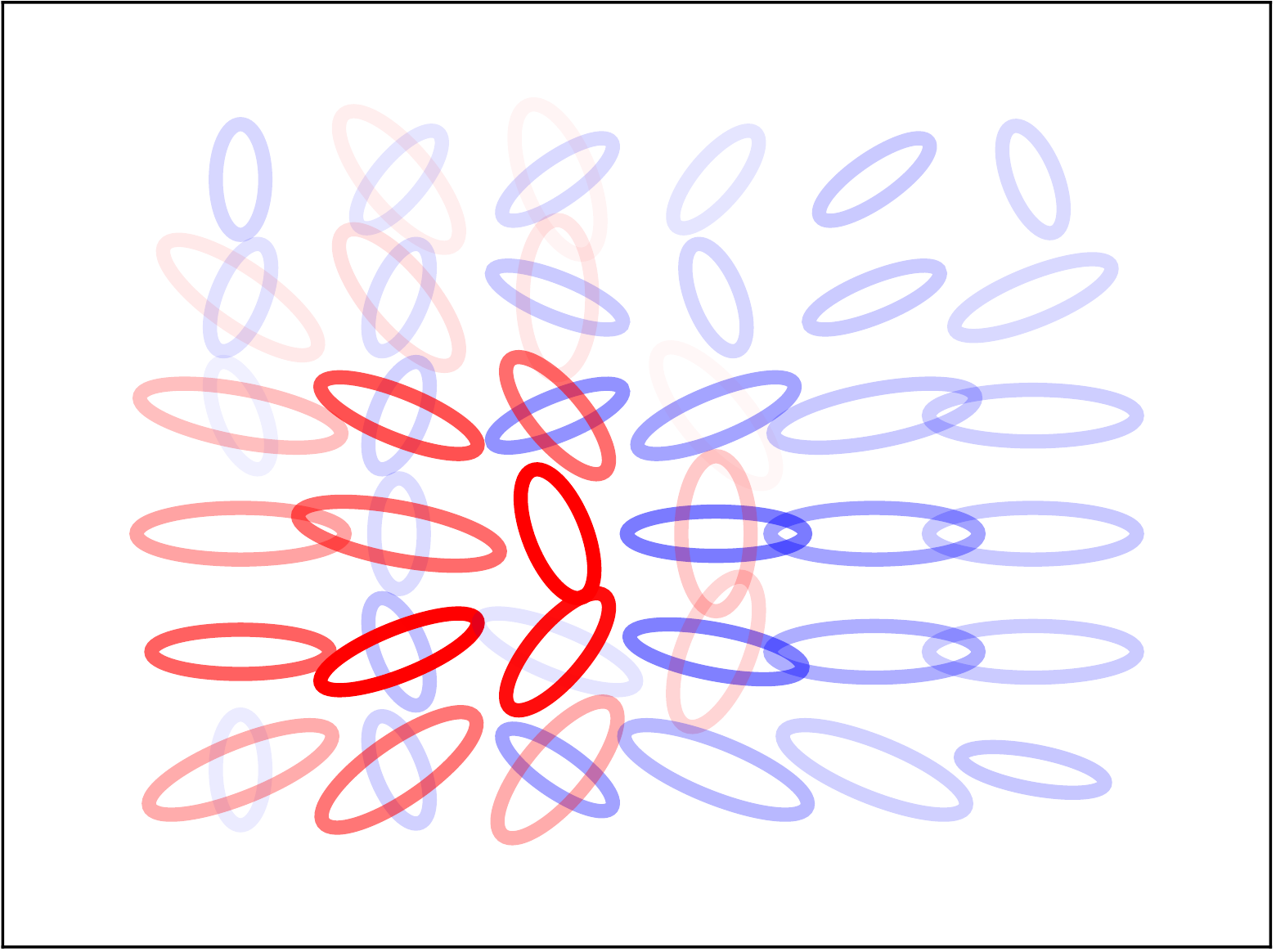}
	\includegraphics[width=0.19\linewidth]{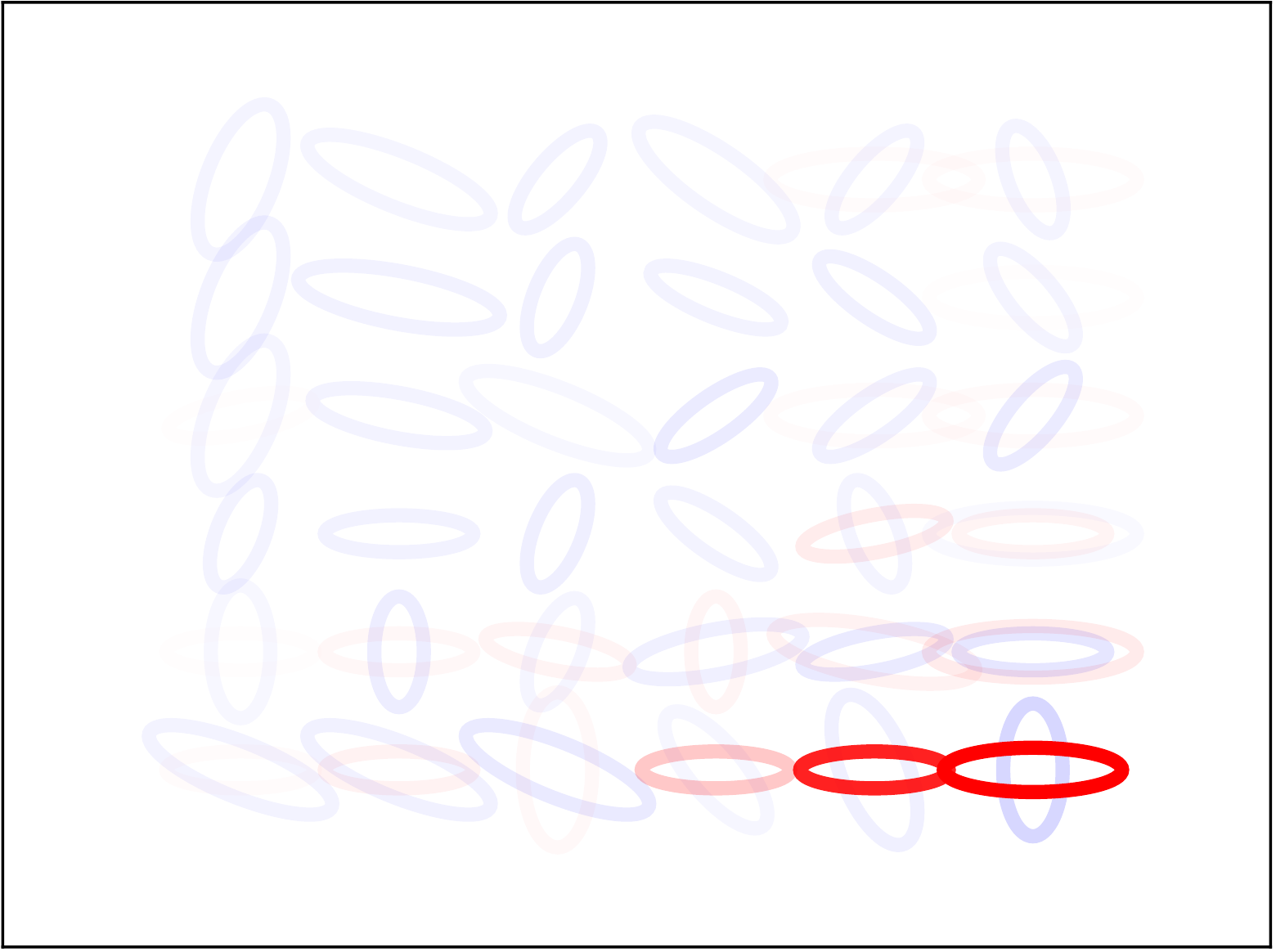}
	\includegraphics[width=0.19\linewidth]{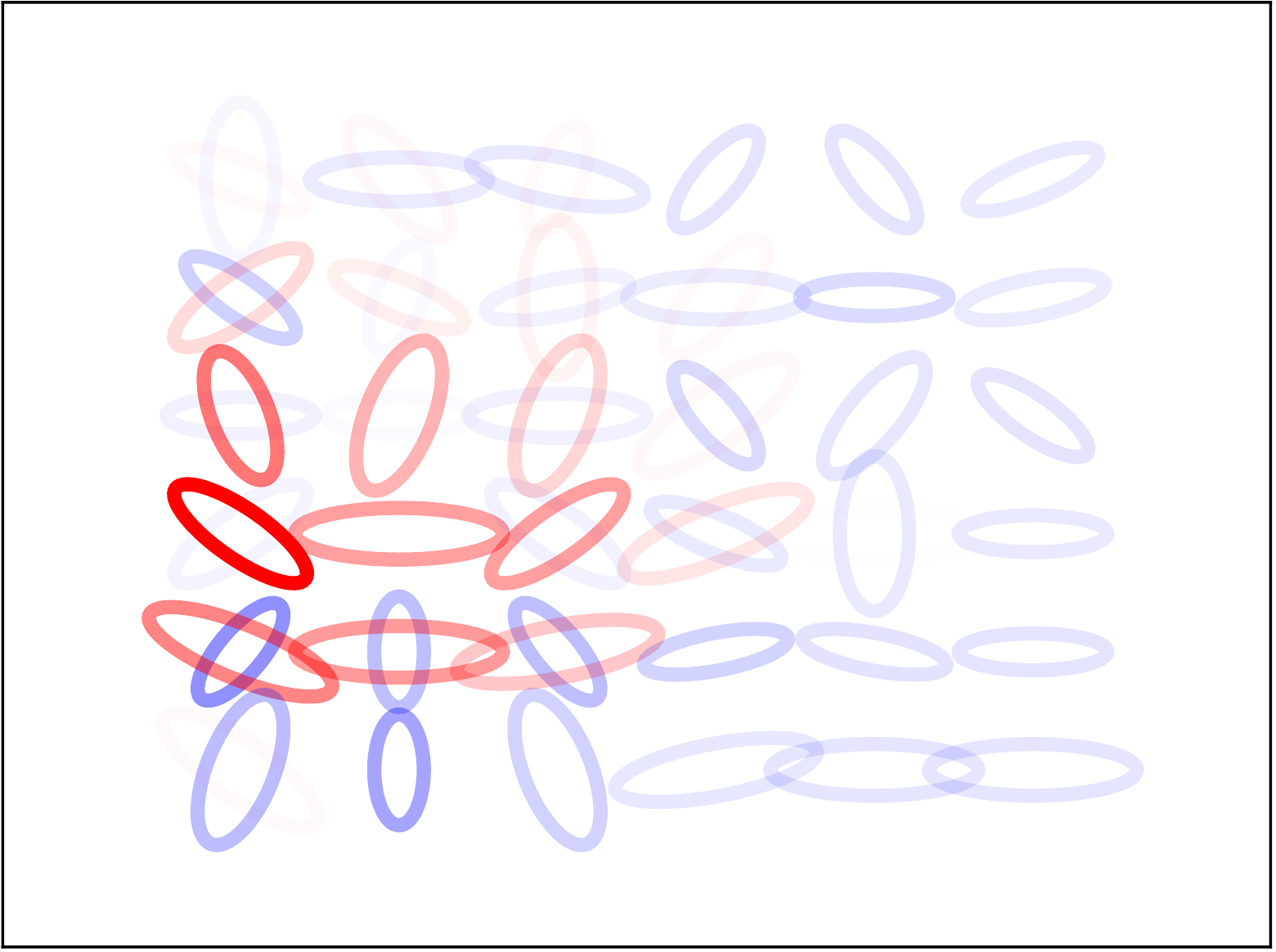}
	\includegraphics[width=0.19\linewidth]{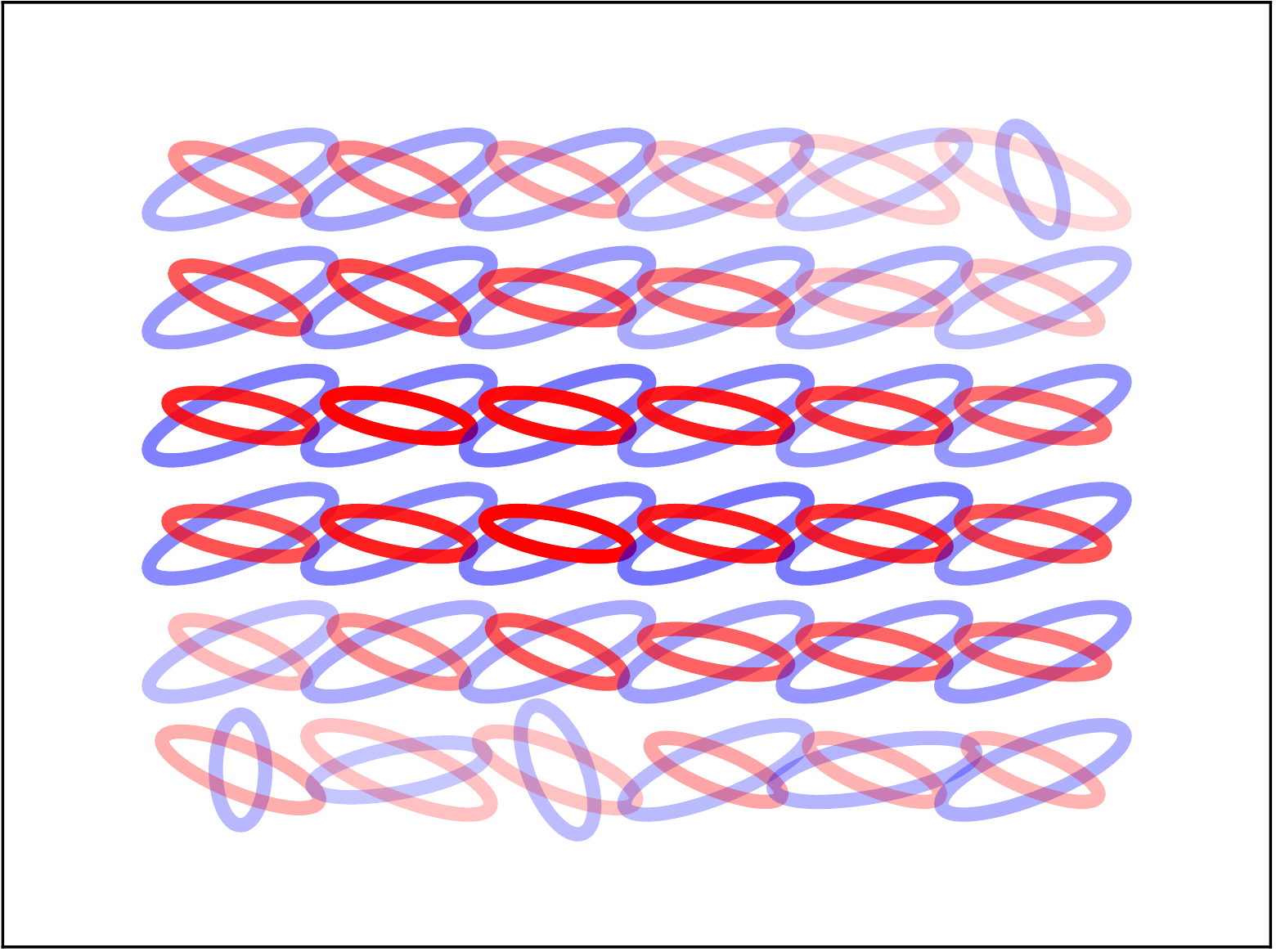}
	\includegraphics[width=0.19\linewidth]{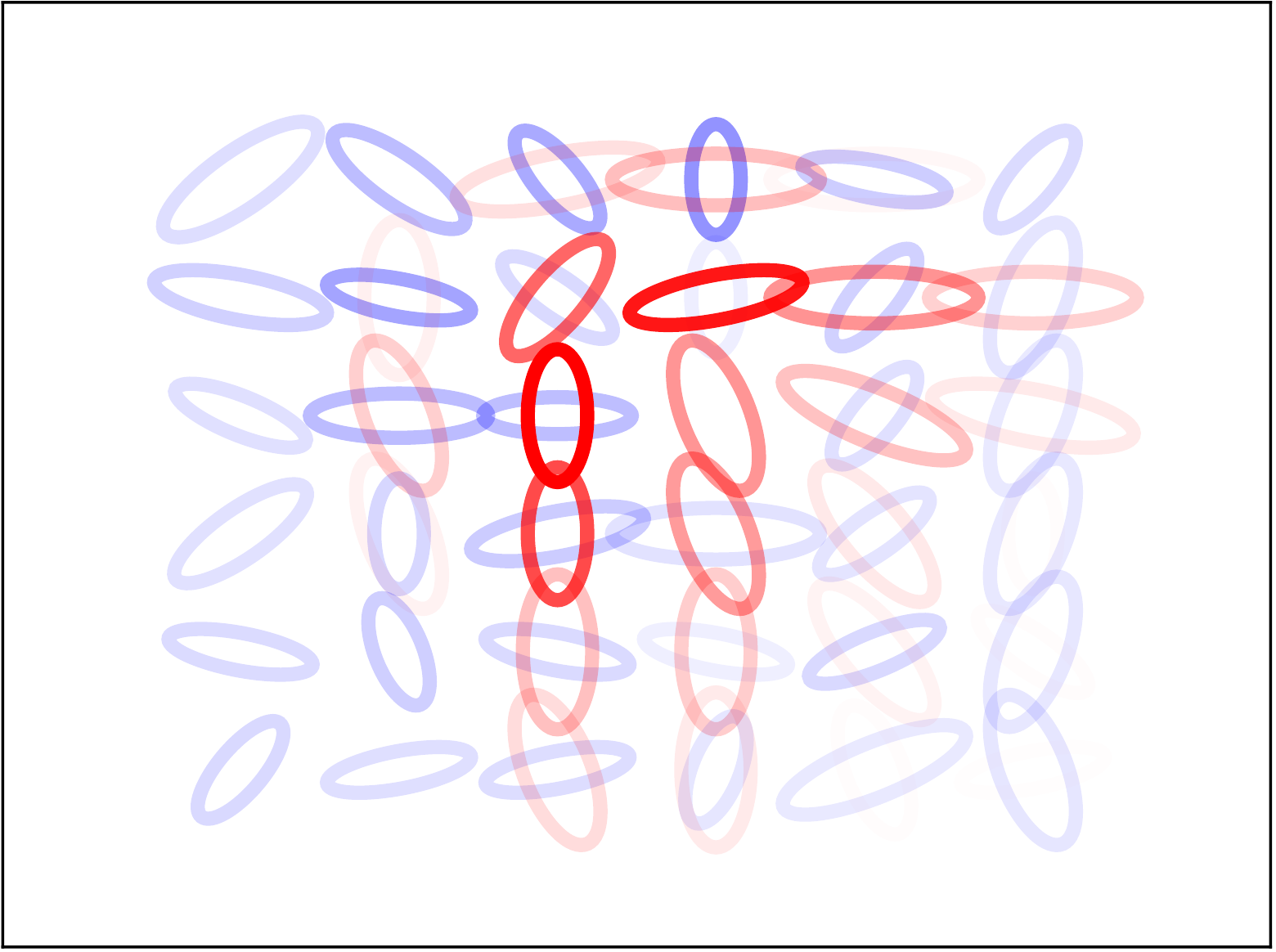} \\
	\phantomsubcaption
	\label{fig:6x6gpsa}
\end{subfigure}
\Large \textbf{(b)} \\
\begin{subfigure}[t]{\linewidth}
	\centering
	\includegraphics[width=0.19\linewidth]{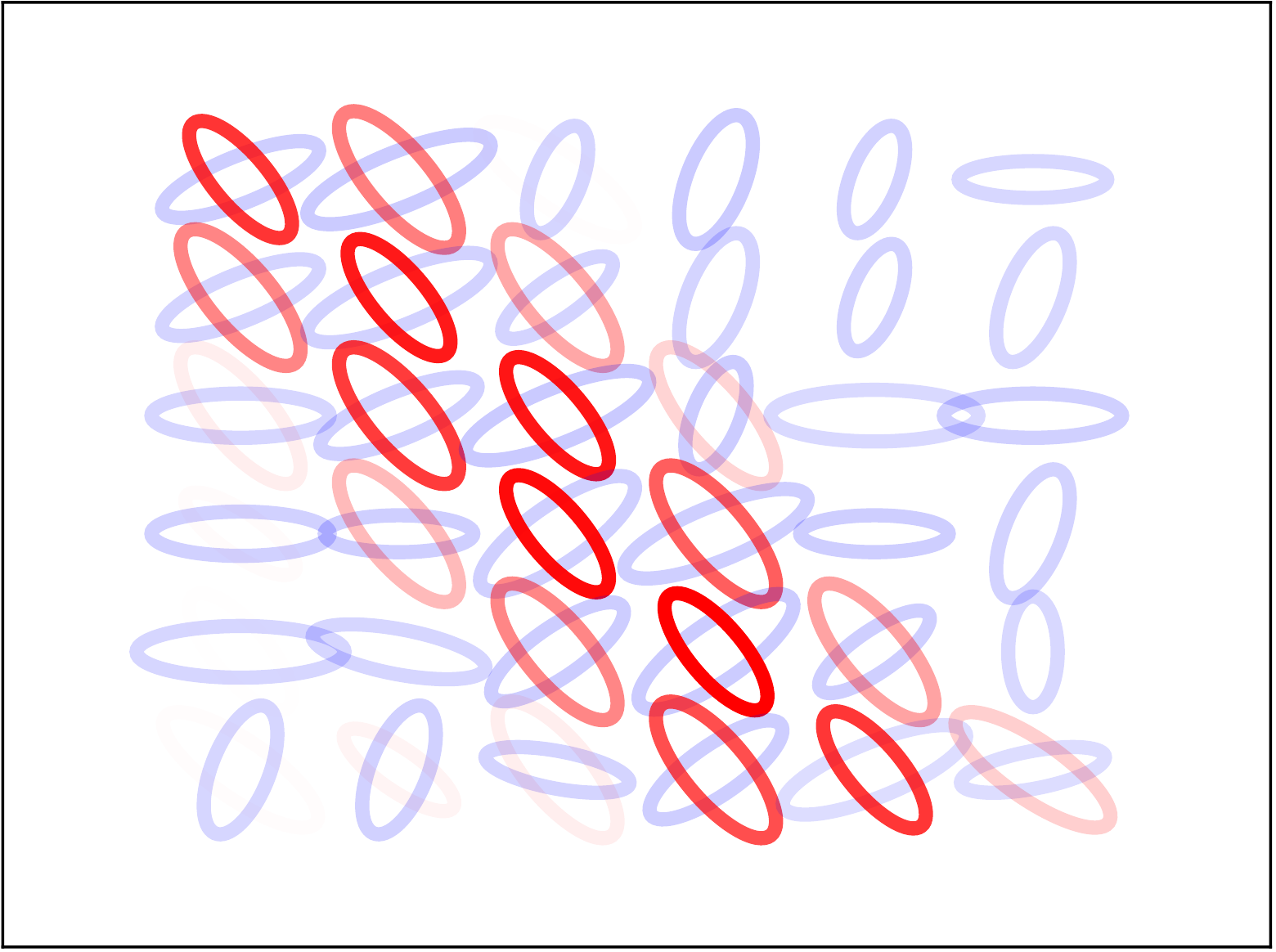}
	\includegraphics[width=0.19\linewidth]{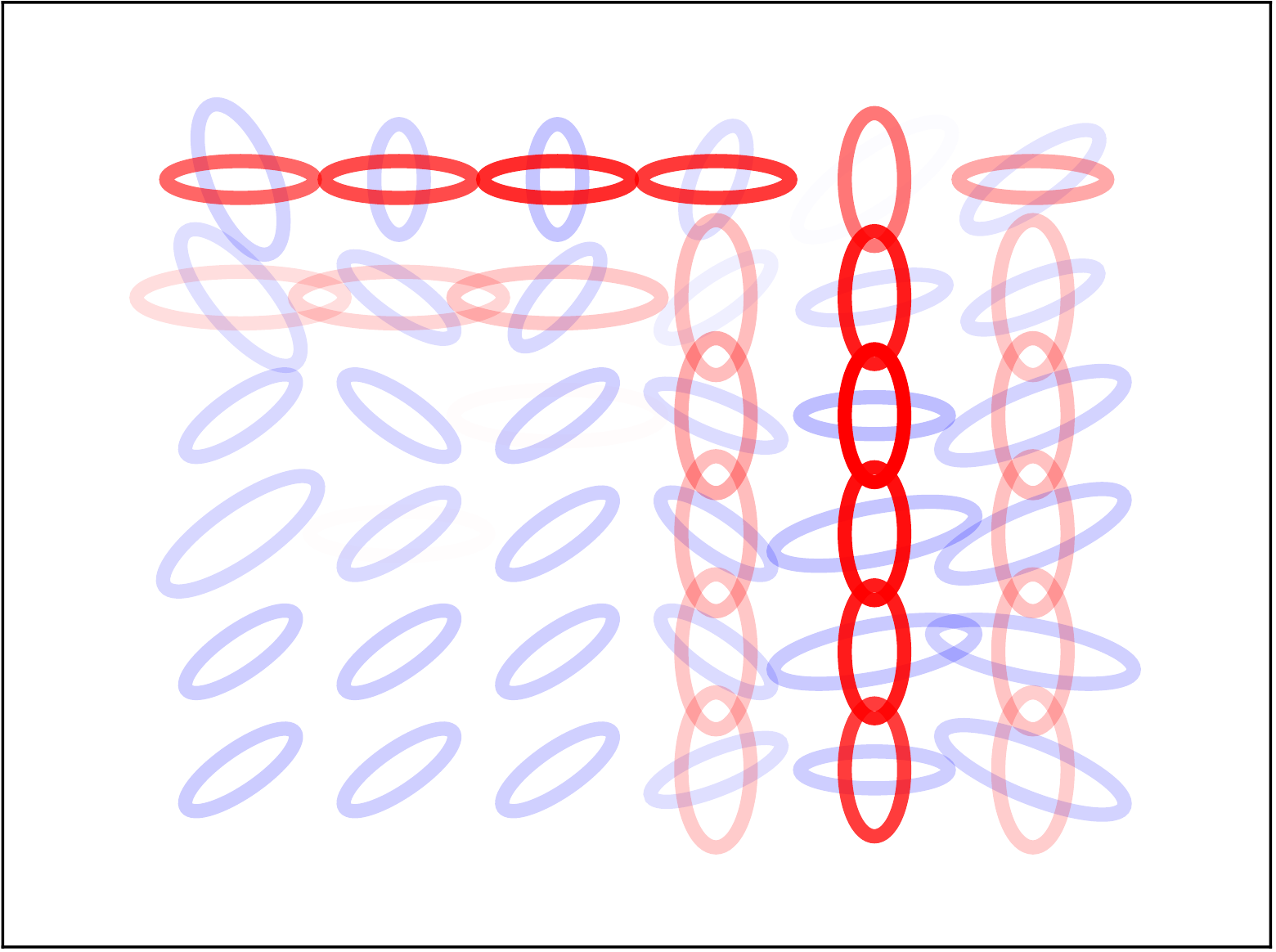}
	\includegraphics[width=0.19\linewidth]{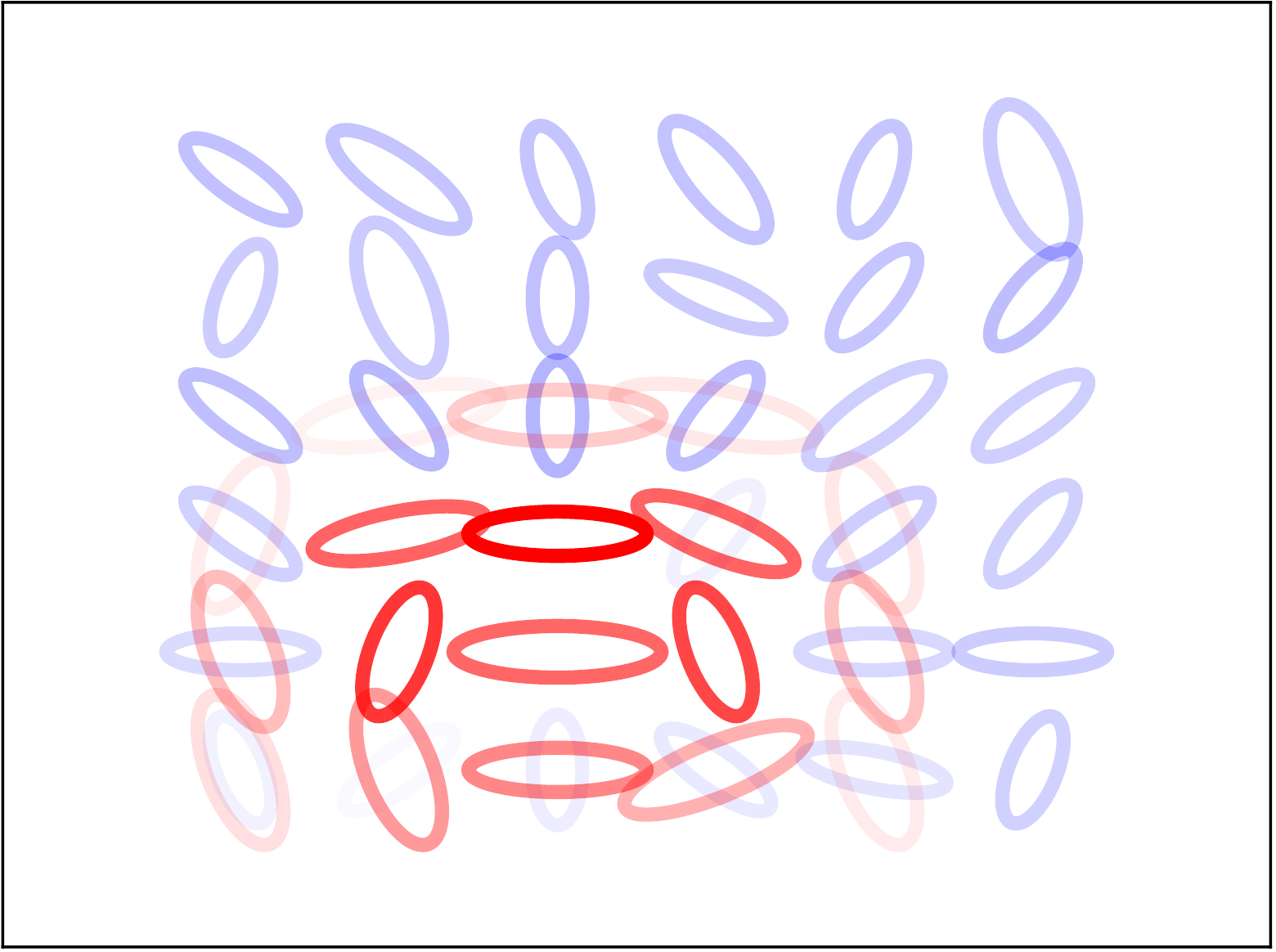}
	\includegraphics[width=0.19\linewidth]{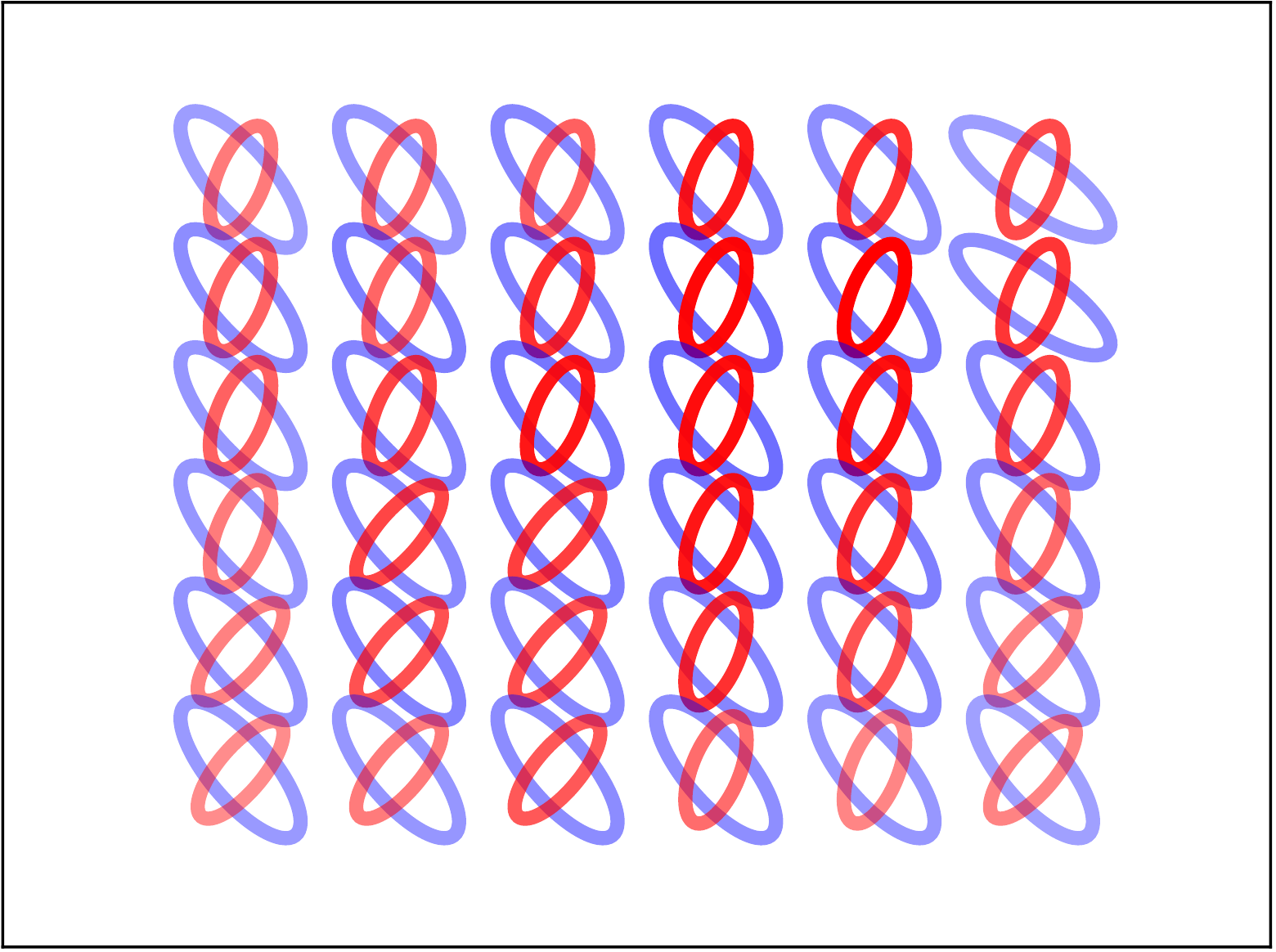}
	\includegraphics[width=0.19\linewidth]{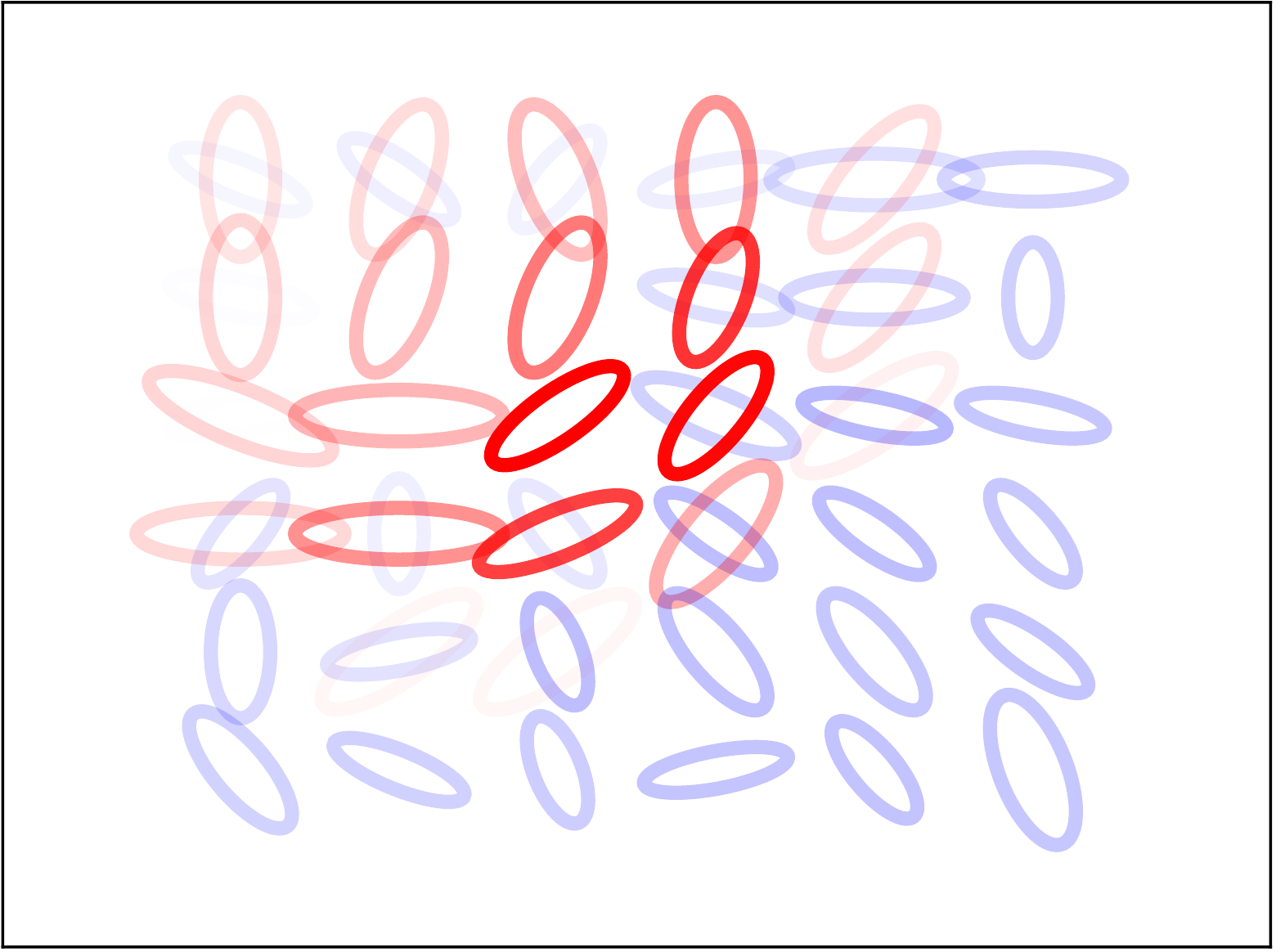}
	\includegraphics[width=0.19\linewidth]{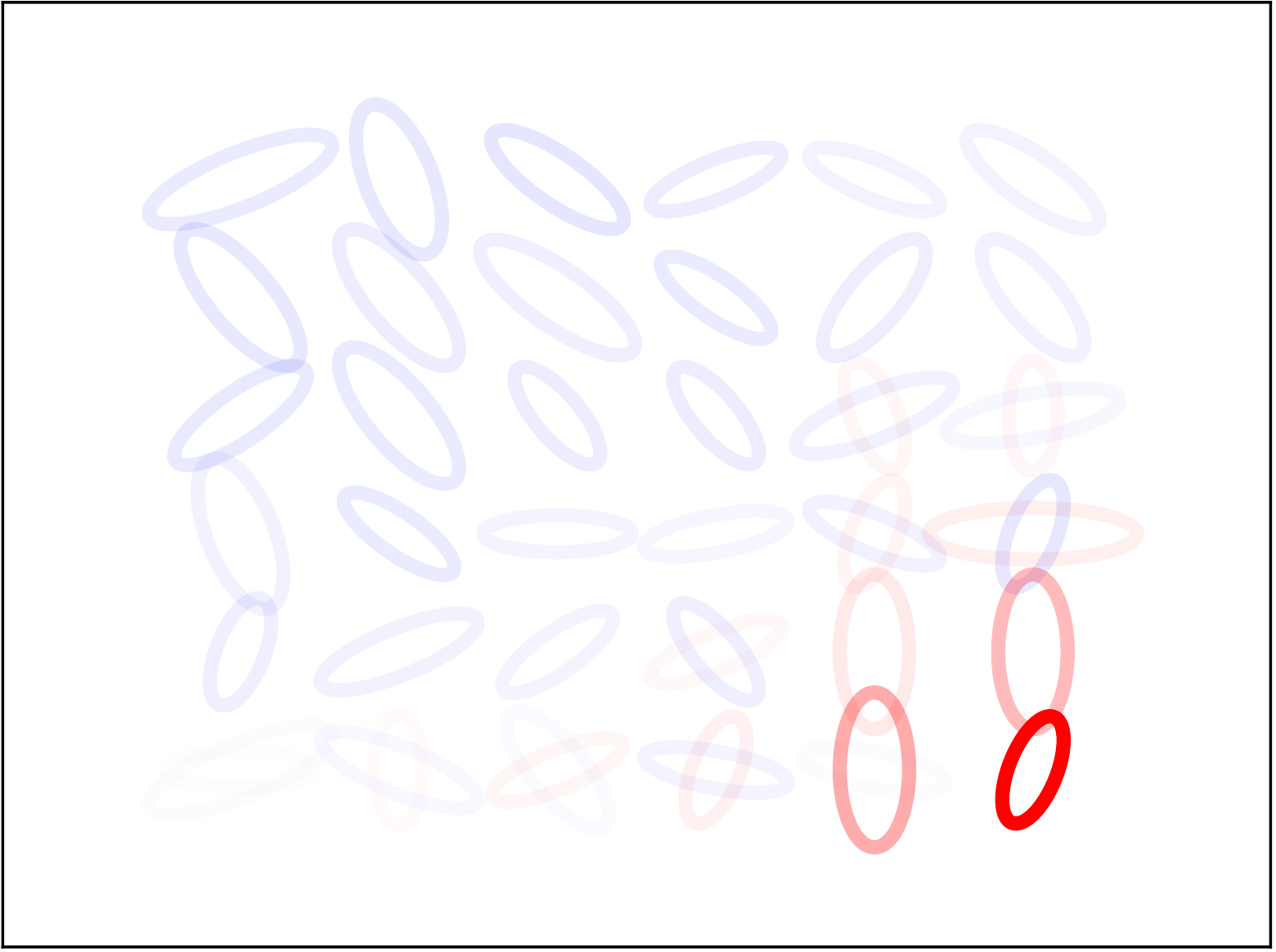}
	\includegraphics[width=0.19\linewidth]{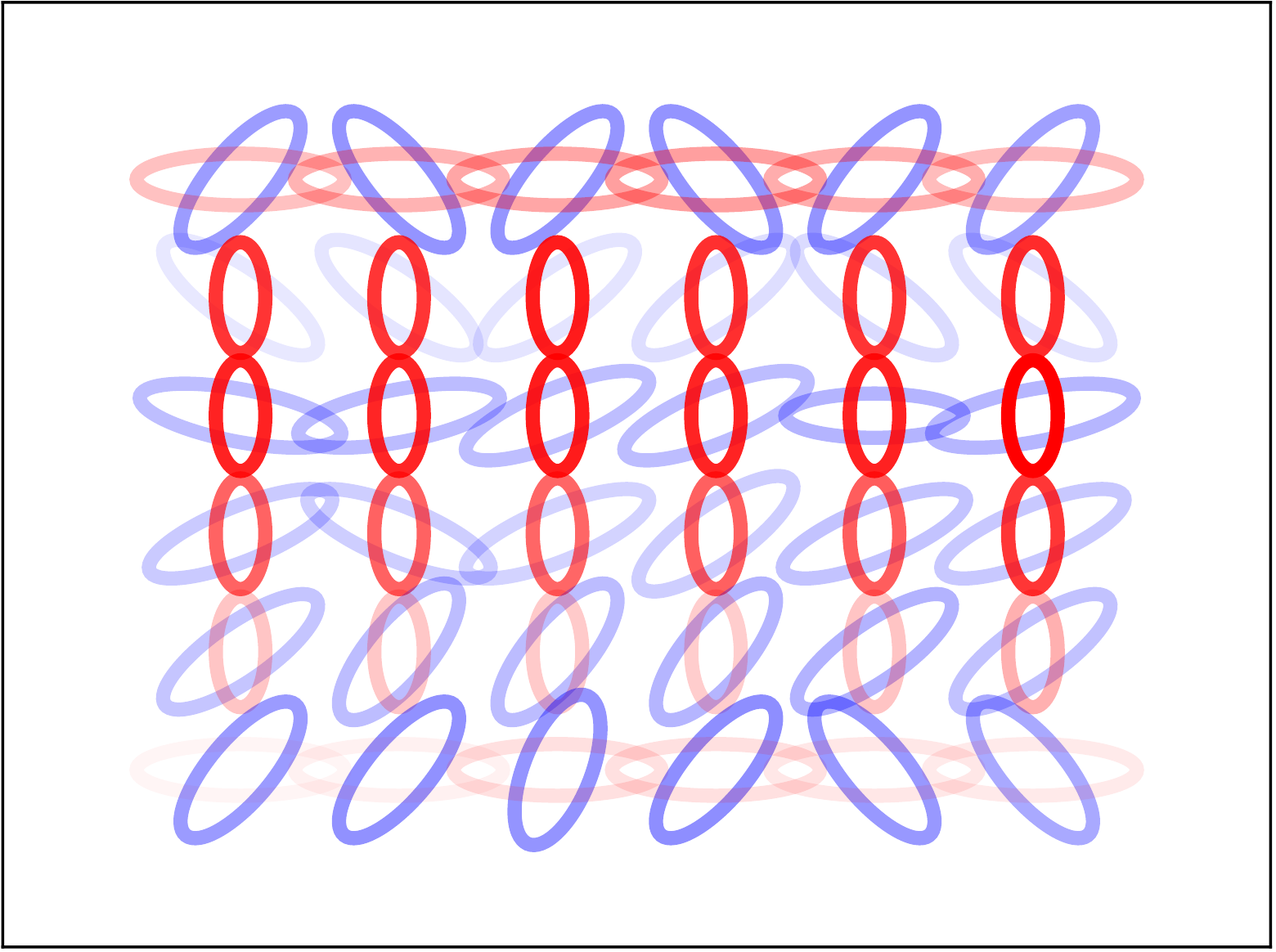}
	\includegraphics[width=0.19\linewidth]{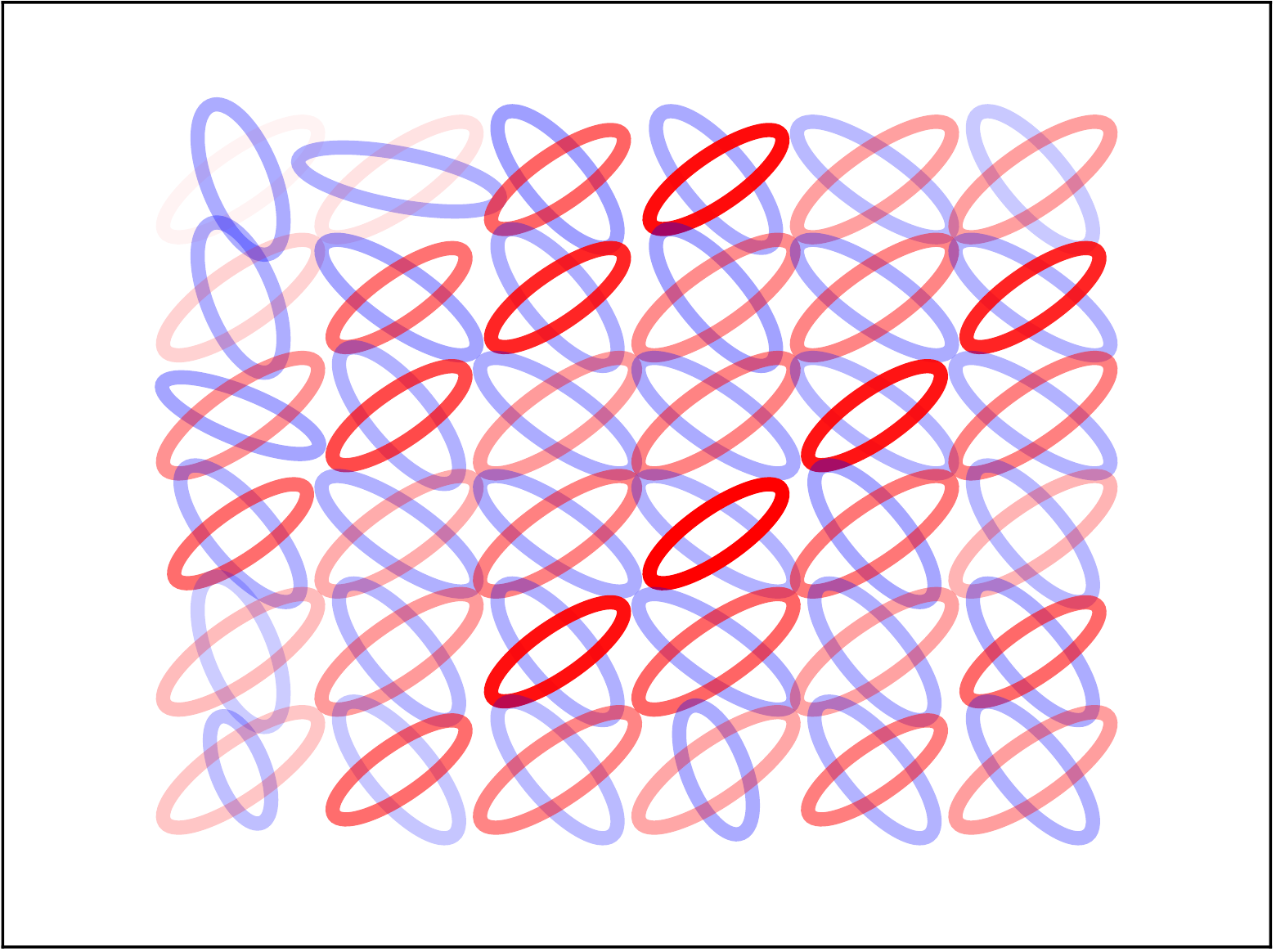}
	\includegraphics[width=0.19\linewidth]{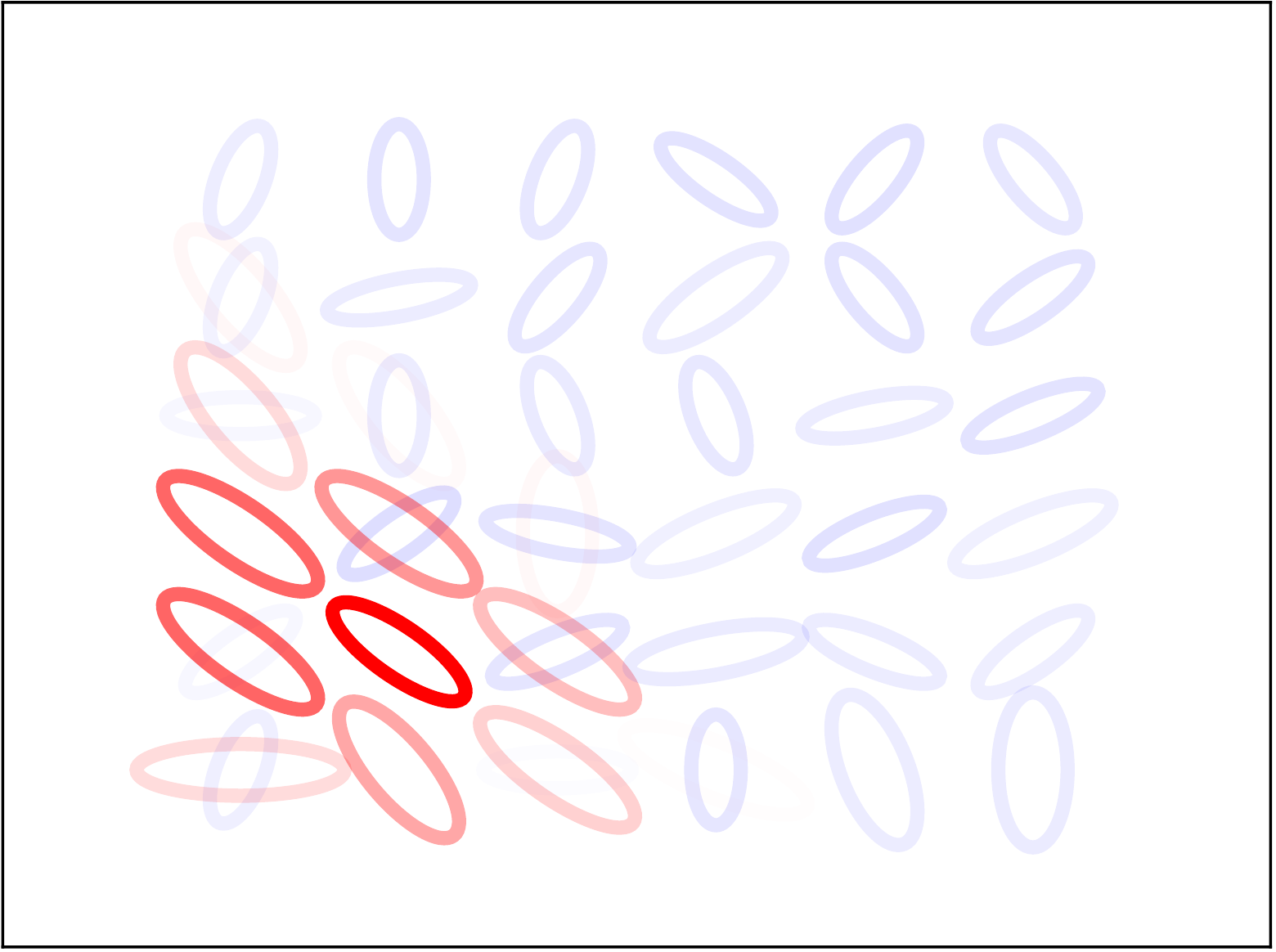}
	\includegraphics[width=0.19\linewidth]{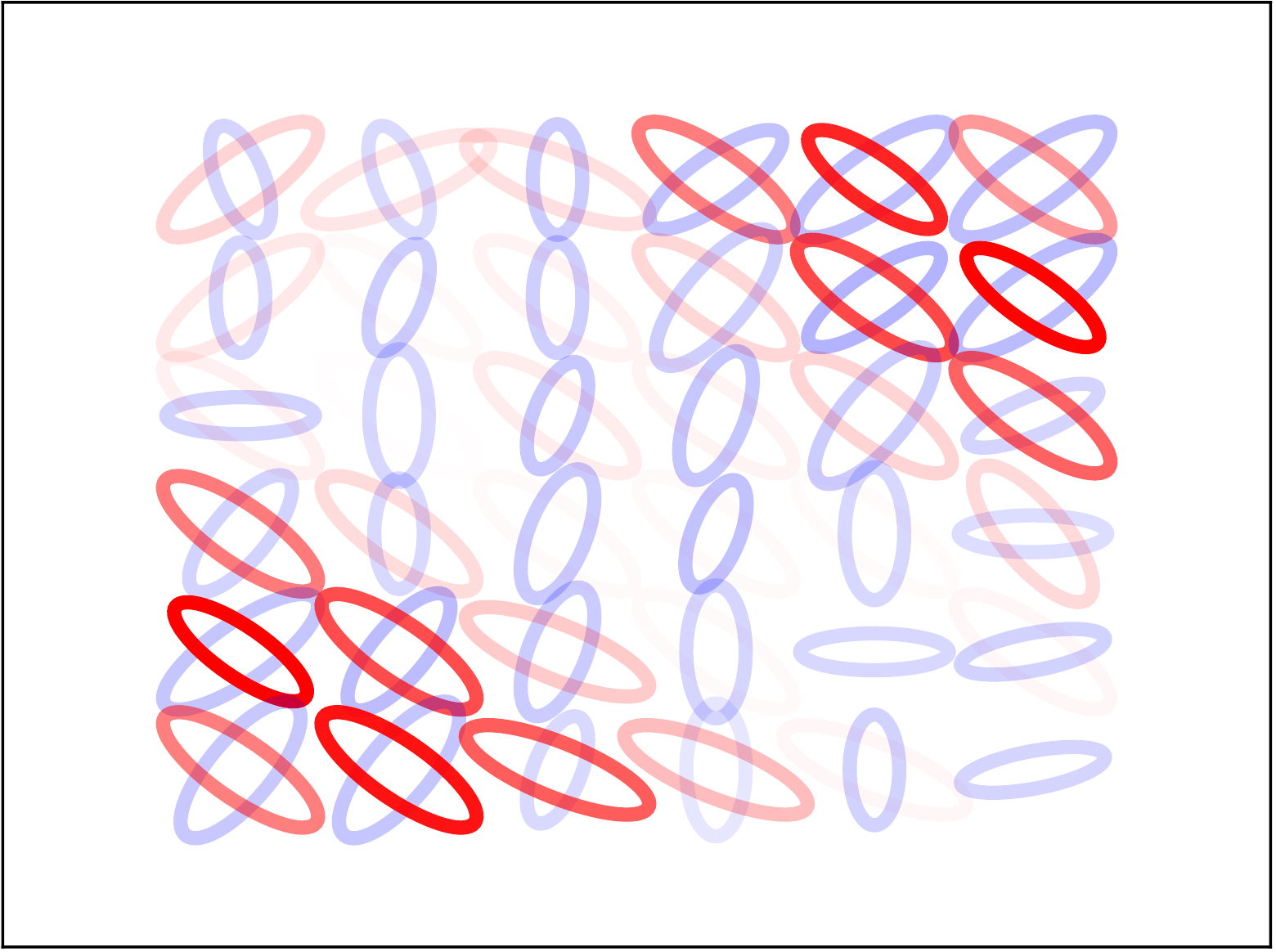}
	\includegraphics[width=0.19\linewidth]{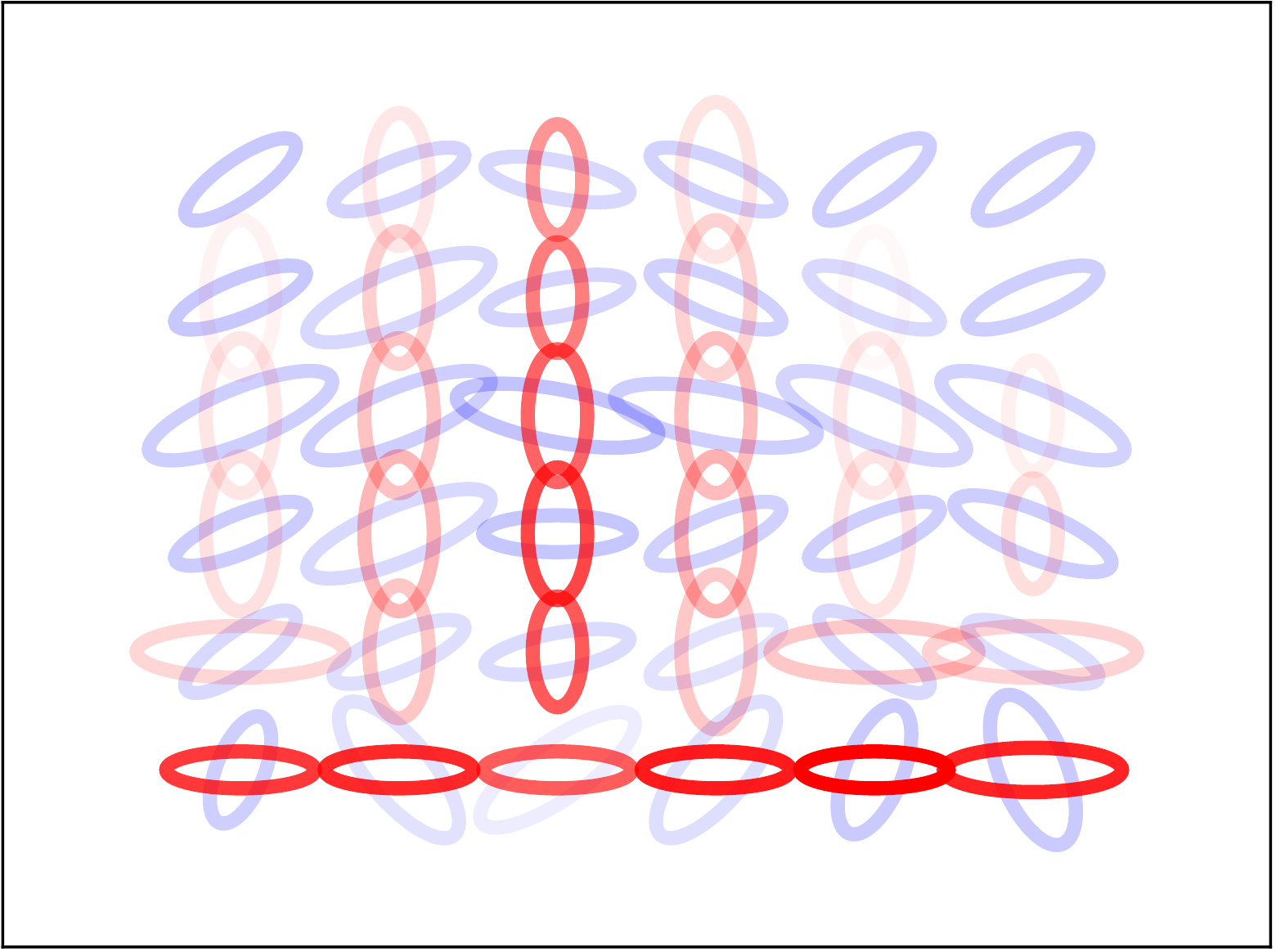}
	\includegraphics[width=0.19\linewidth]{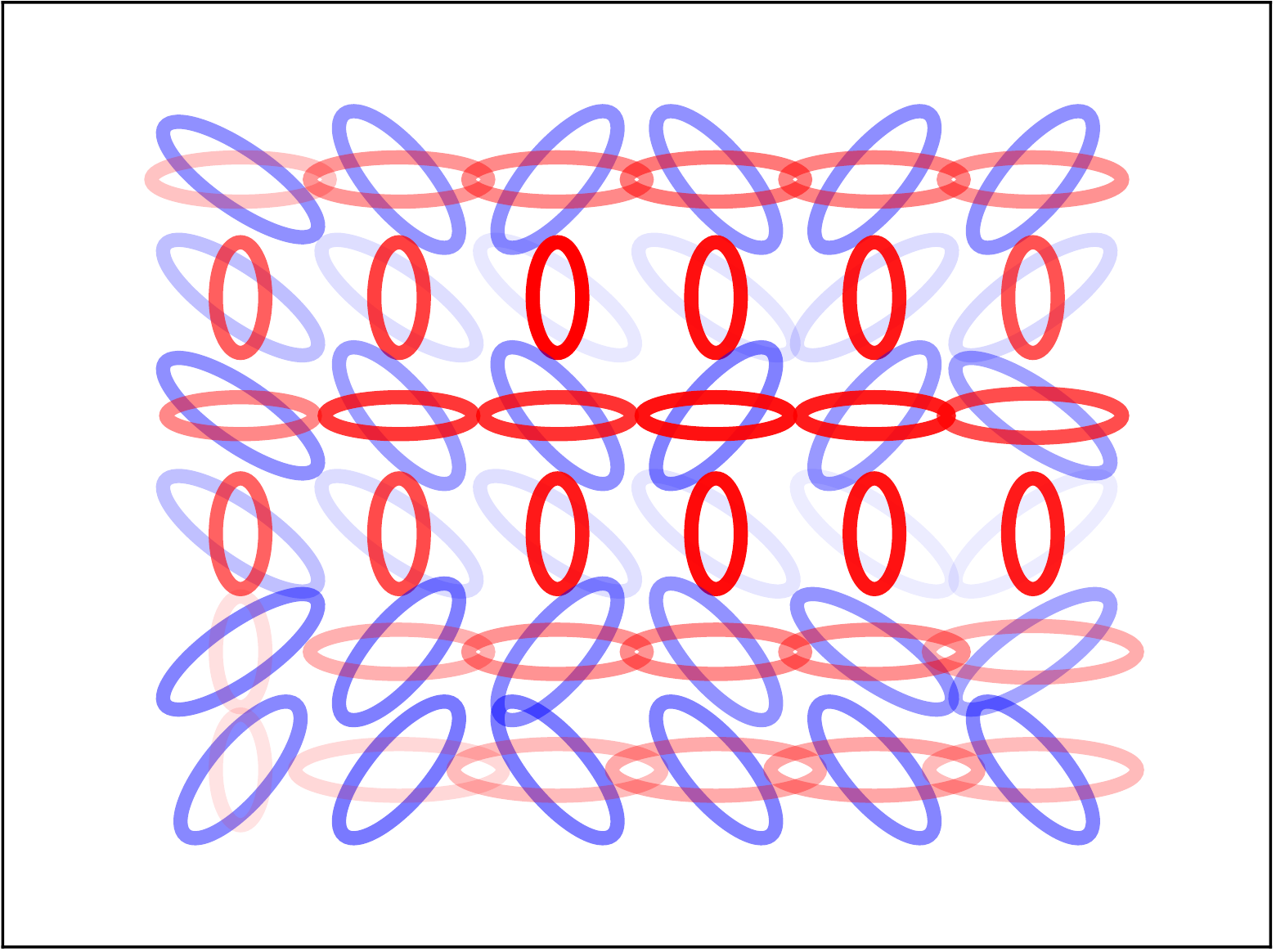}
	\includegraphics[width=0.19\linewidth]{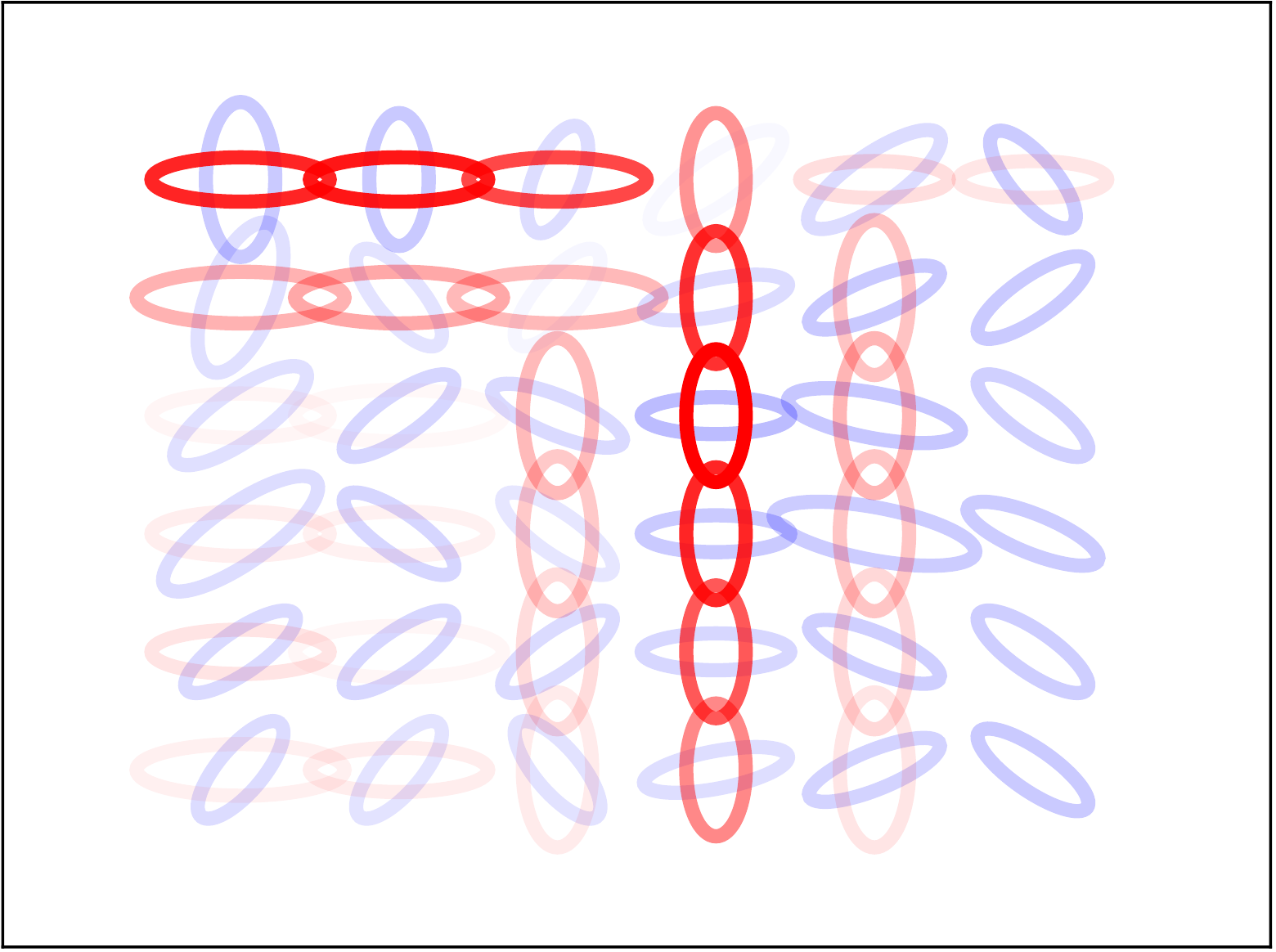}
	\includegraphics[width=0.19\linewidth]{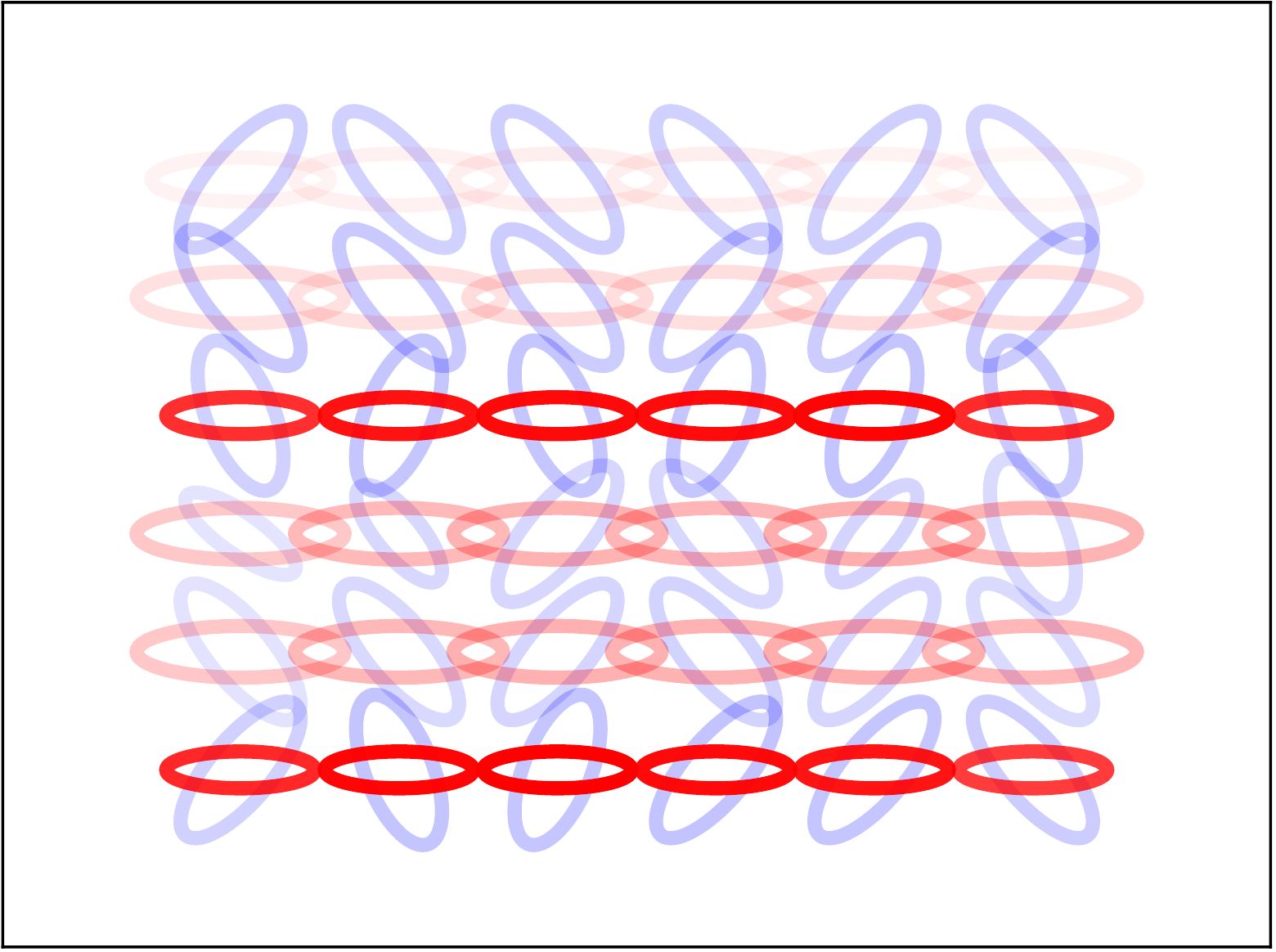}
	\includegraphics[width=0.19\linewidth]{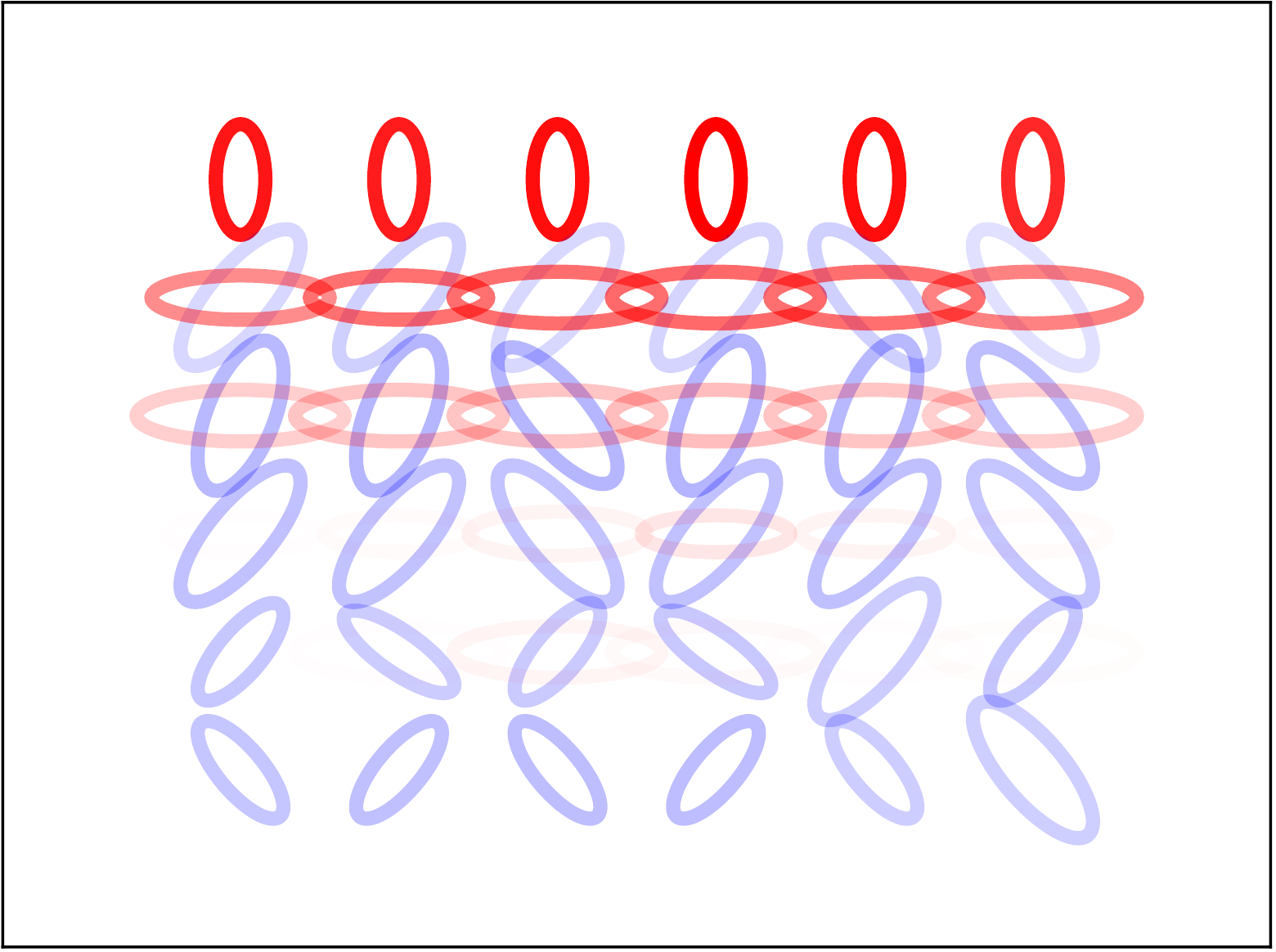}
	\includegraphics[width=0.19\linewidth]{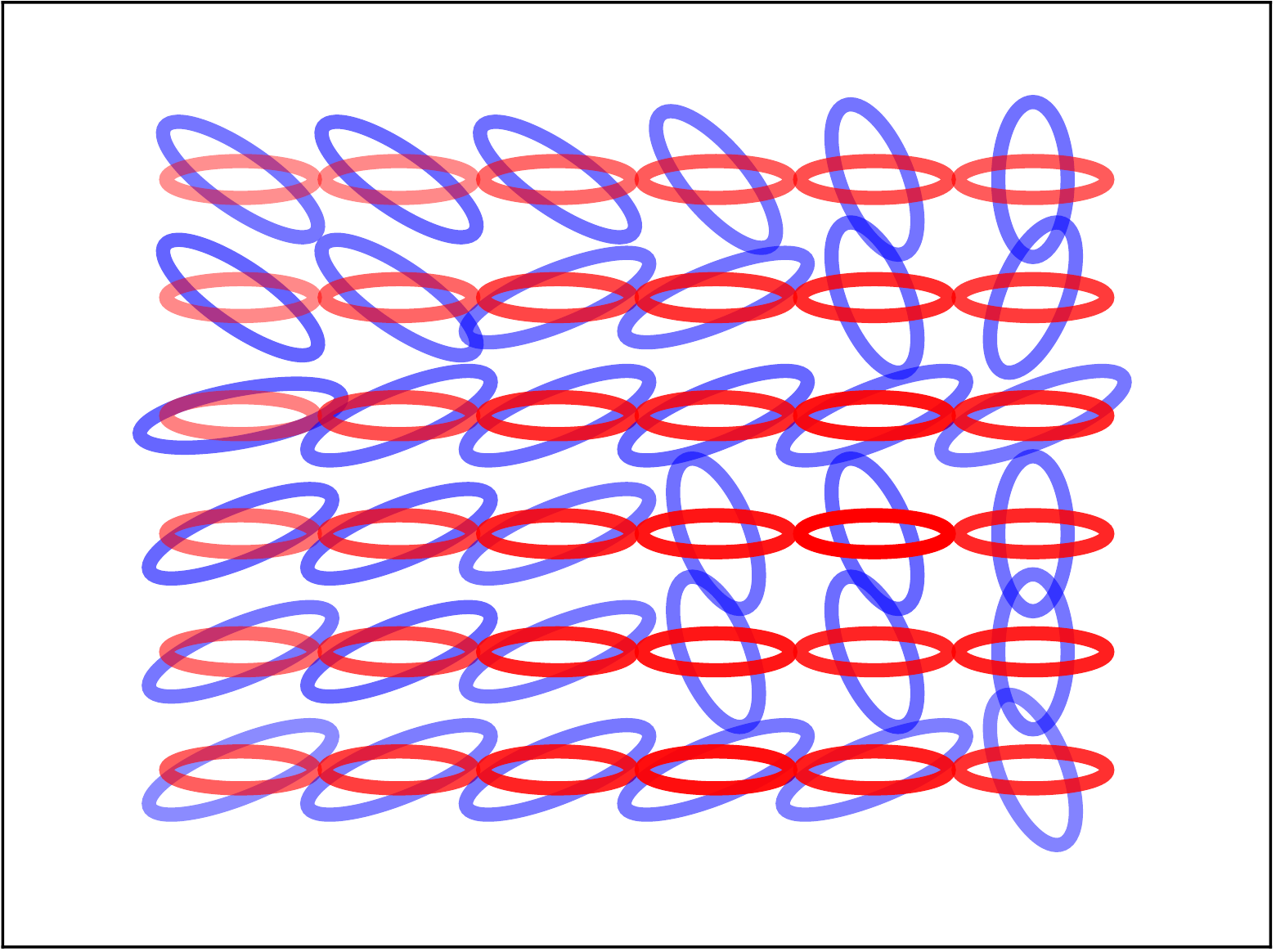}
	\includegraphics[width=0.19\linewidth]{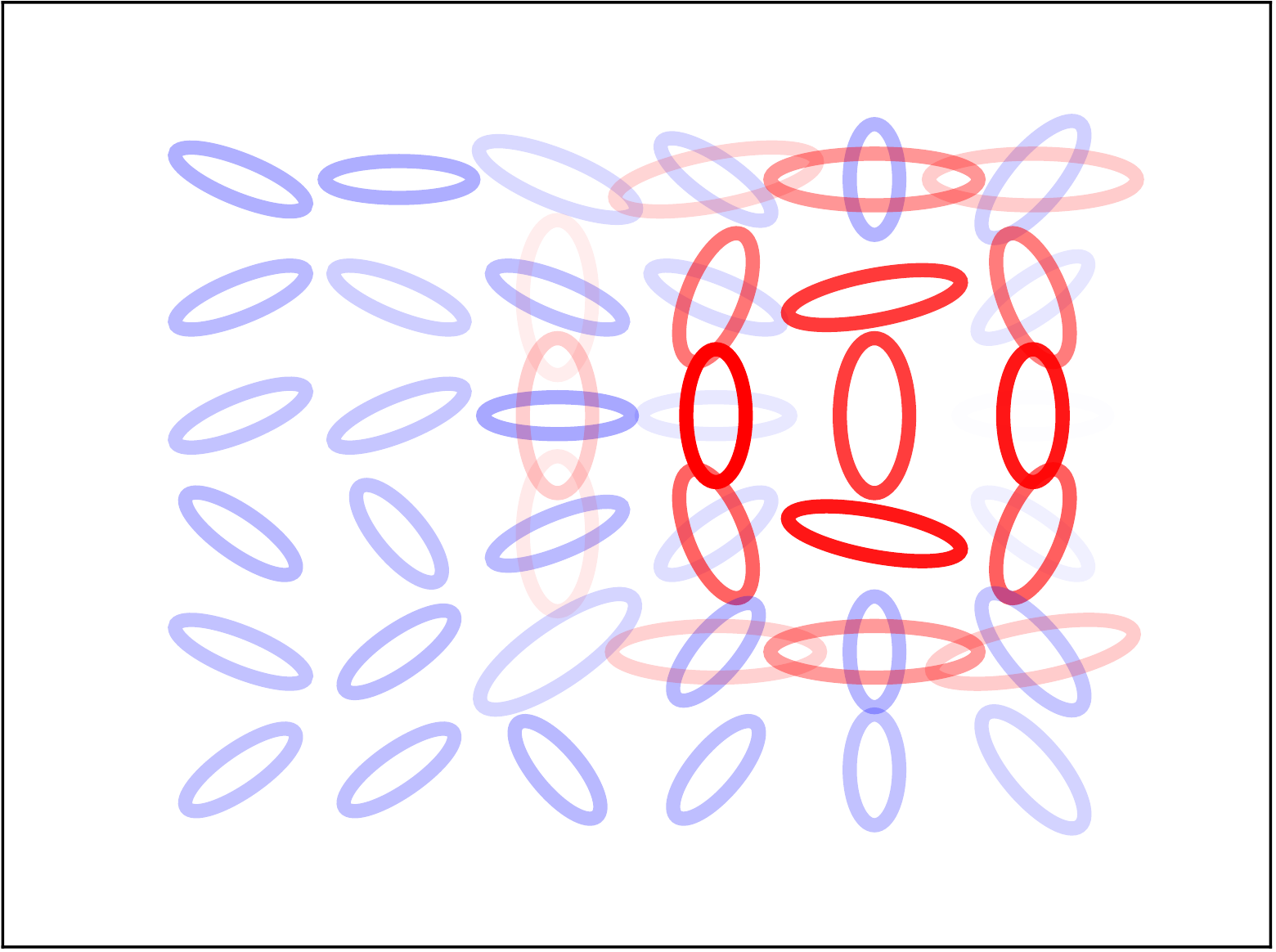}
	\includegraphics[width=0.19\linewidth]{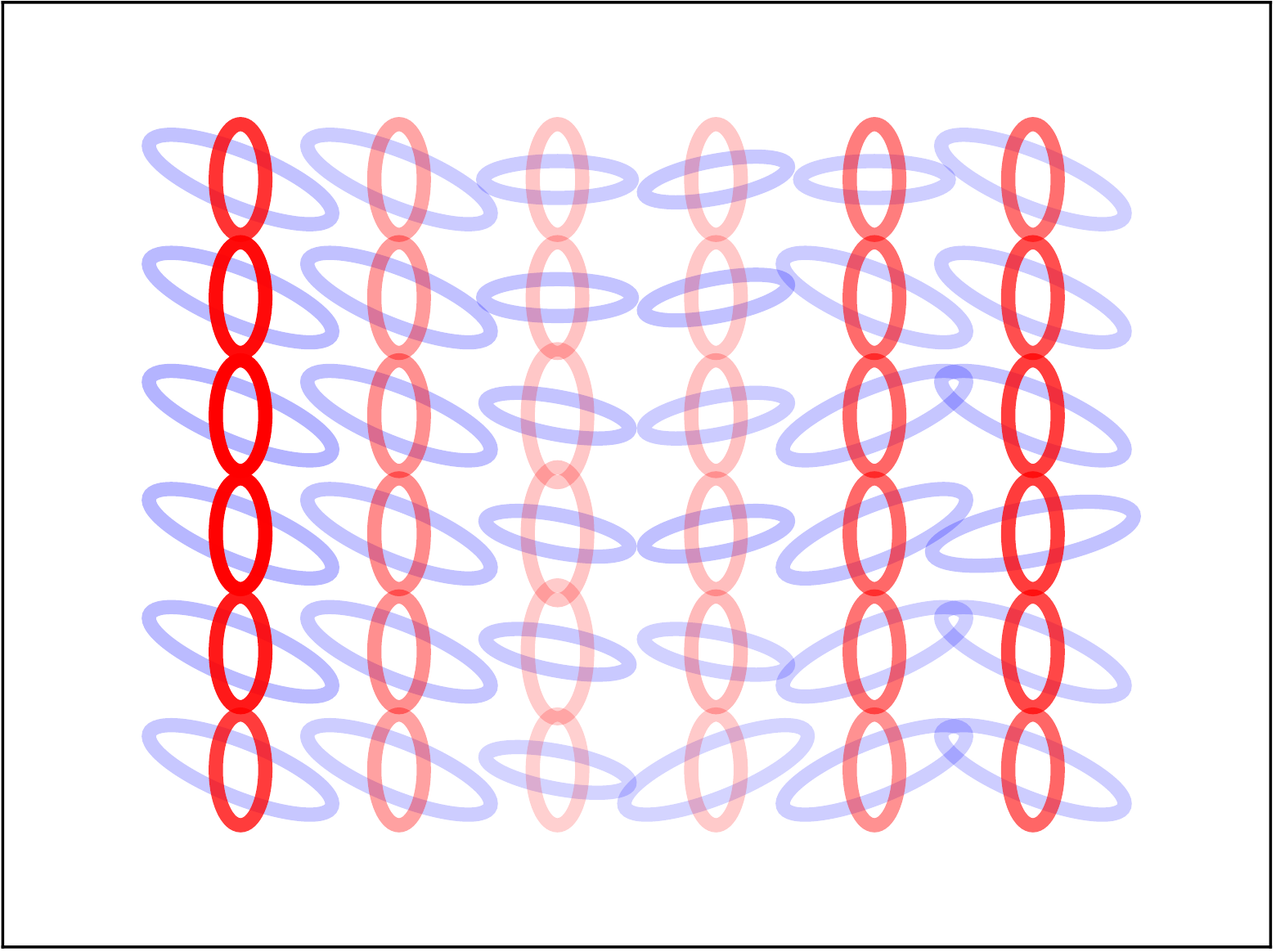}
	\includegraphics[width=0.19\linewidth]{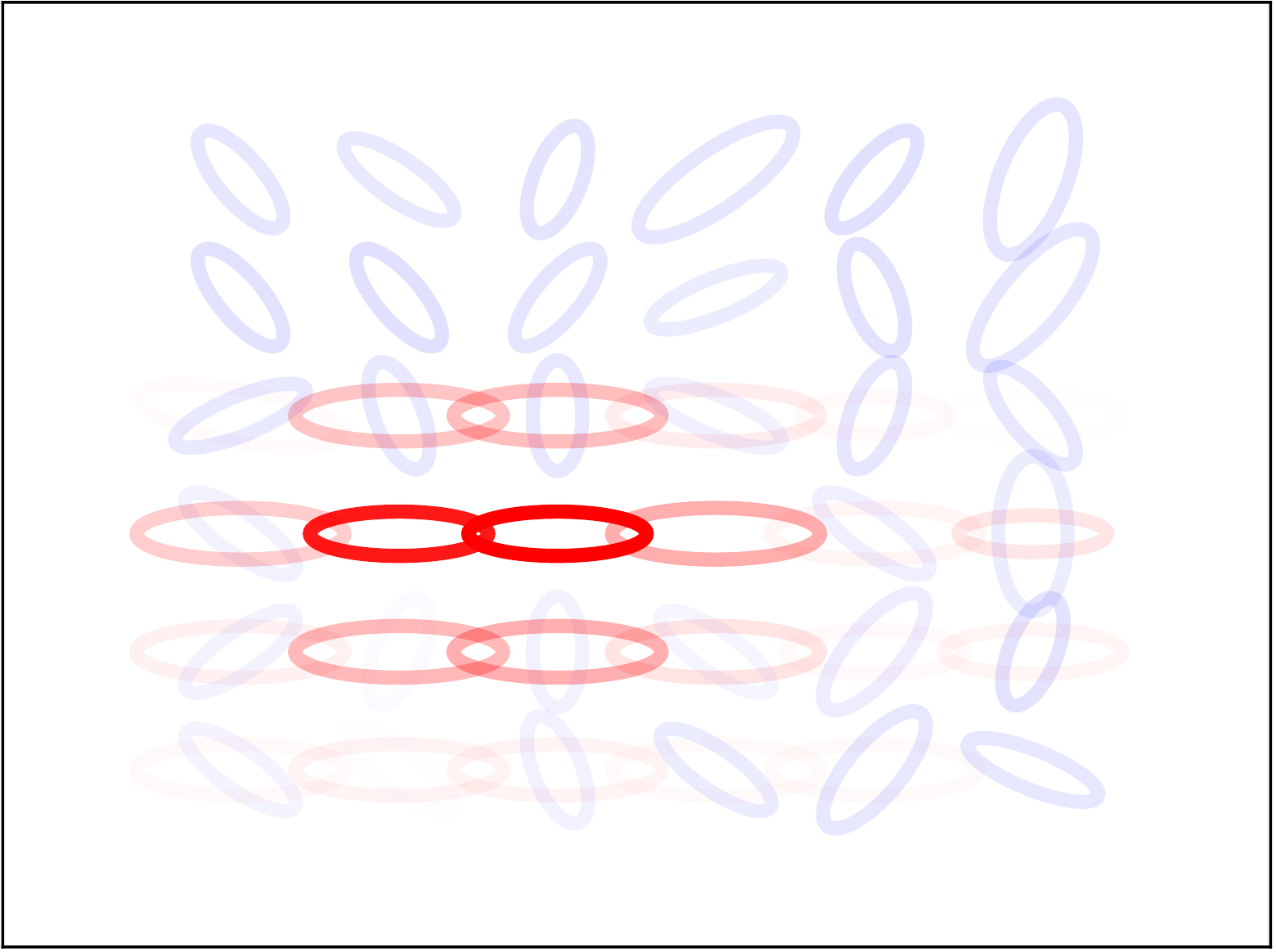}
	\includegraphics[width=0.19\linewidth]{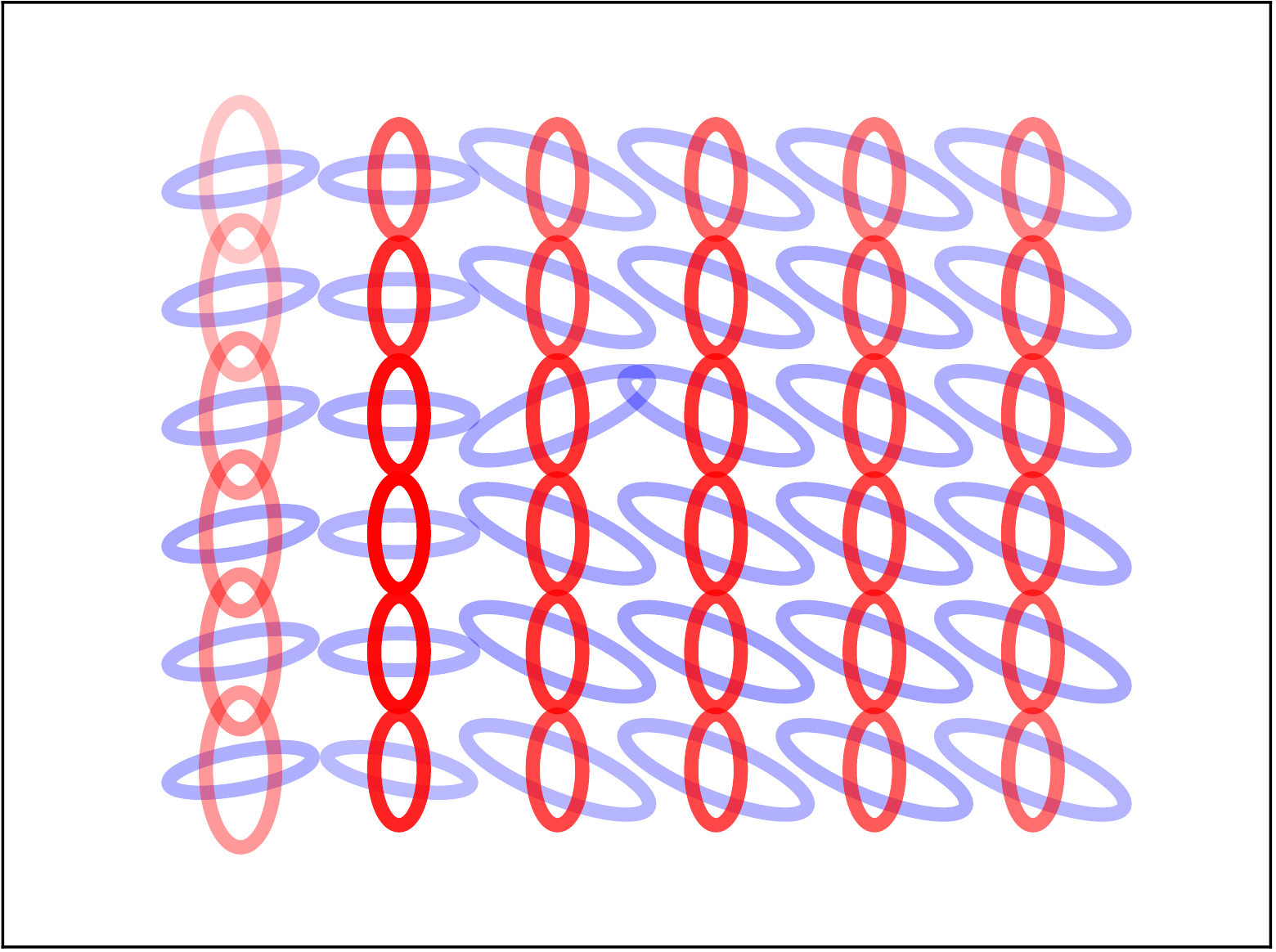} \\
	\phantomsubcaption
	\label{fig:6x6gpsb}
\end{subfigure}
\end{figure}

\begin{figure}
\ContinuedFloat
\Large \textbf{(c)} \\
\begin{subfigure}[t]{\linewidth}
	\centering
	\includegraphics[width=0.19\linewidth]{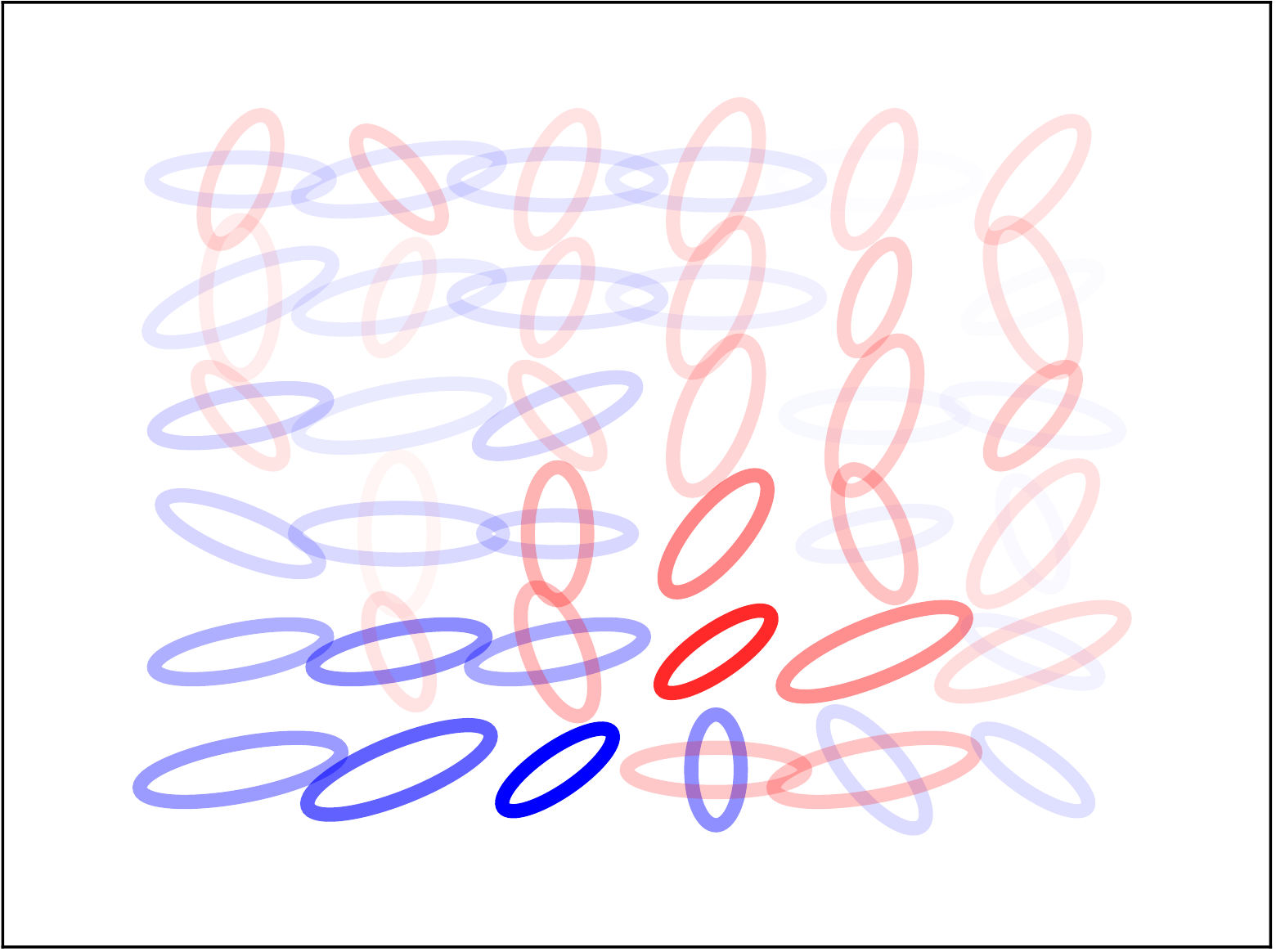}
	\includegraphics[width=0.19\linewidth]{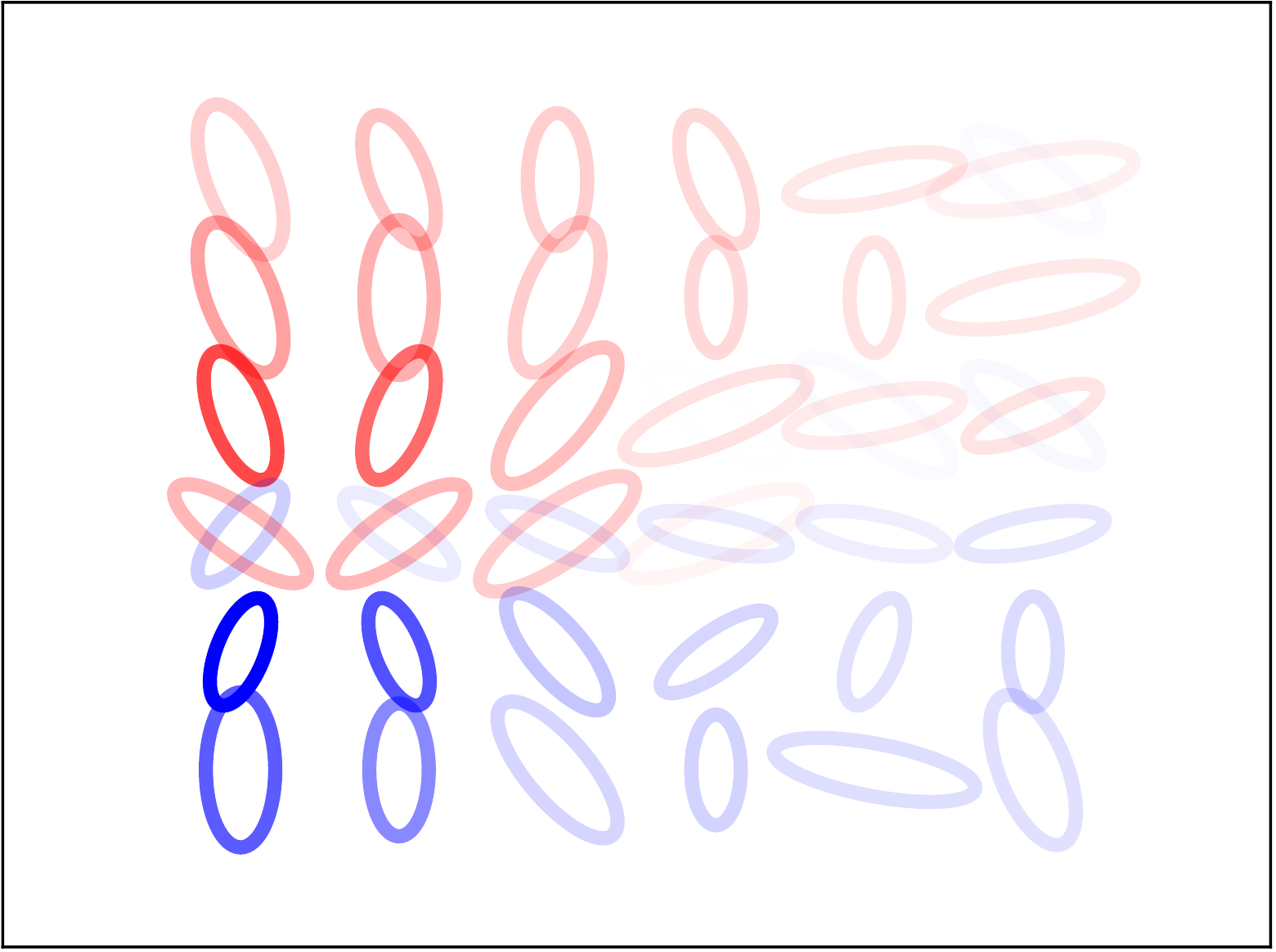}
	\includegraphics[width=0.19\linewidth]{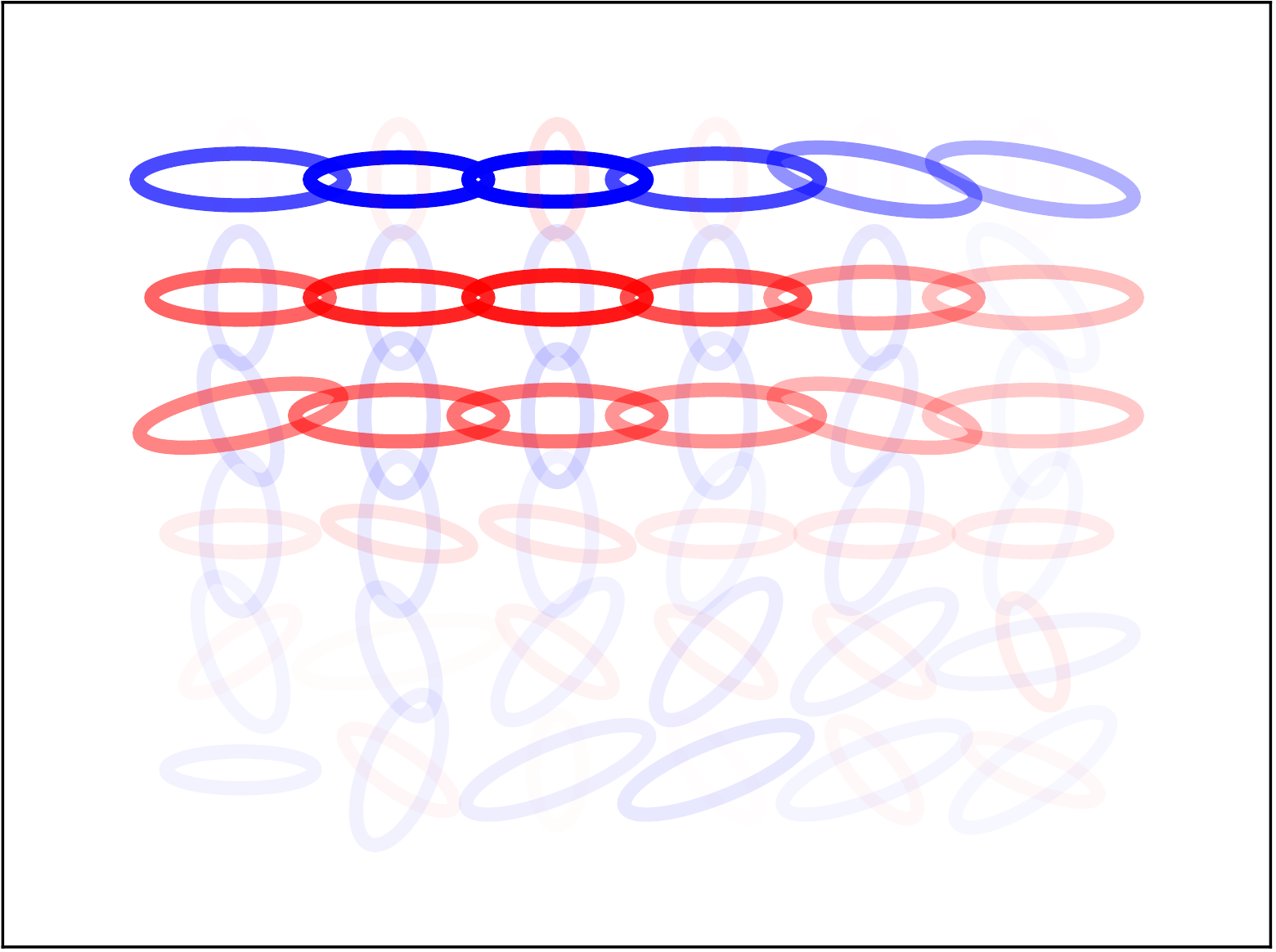}
	\includegraphics[width=0.19\linewidth]{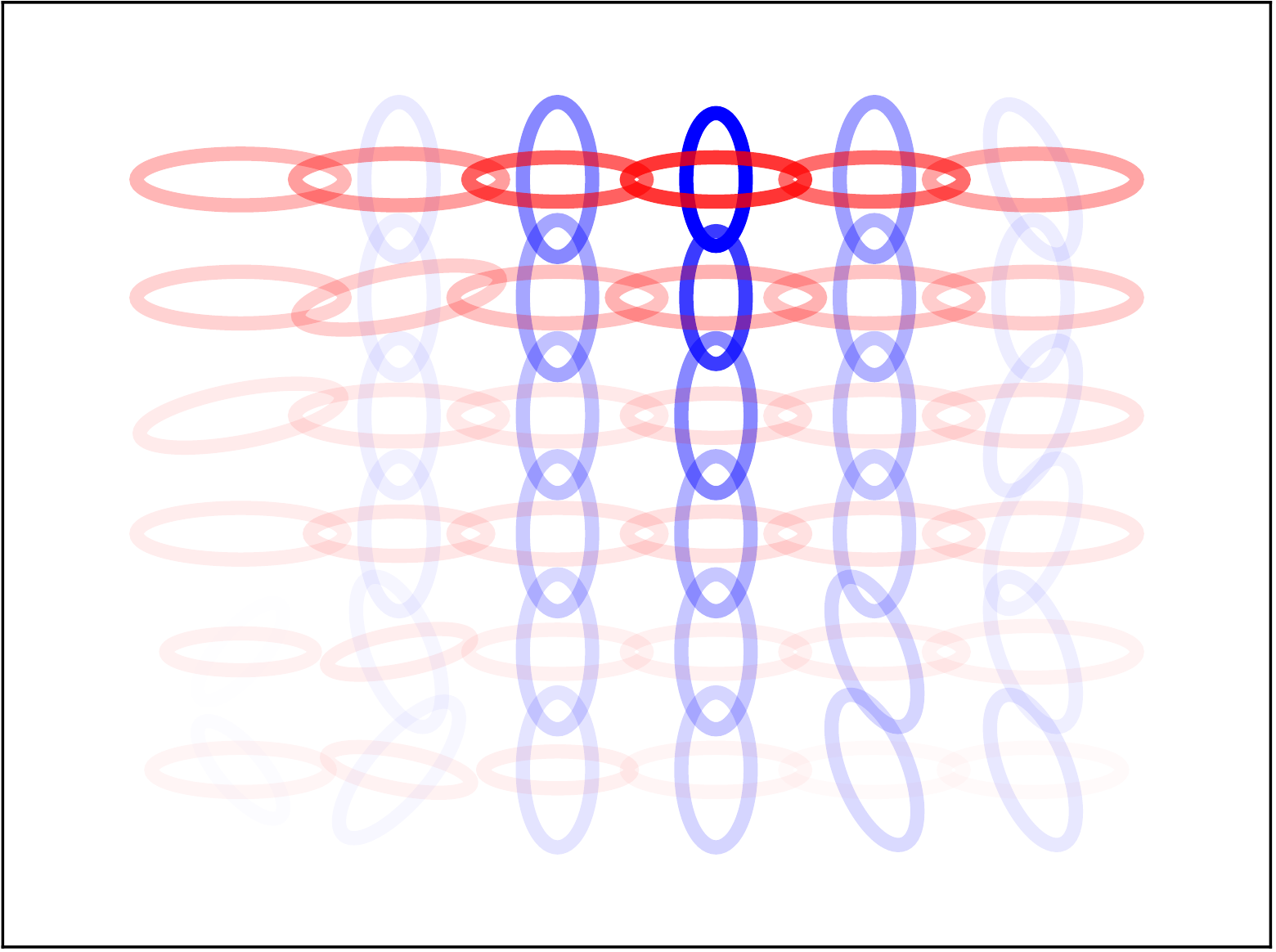}
	\includegraphics[width=0.19\linewidth]{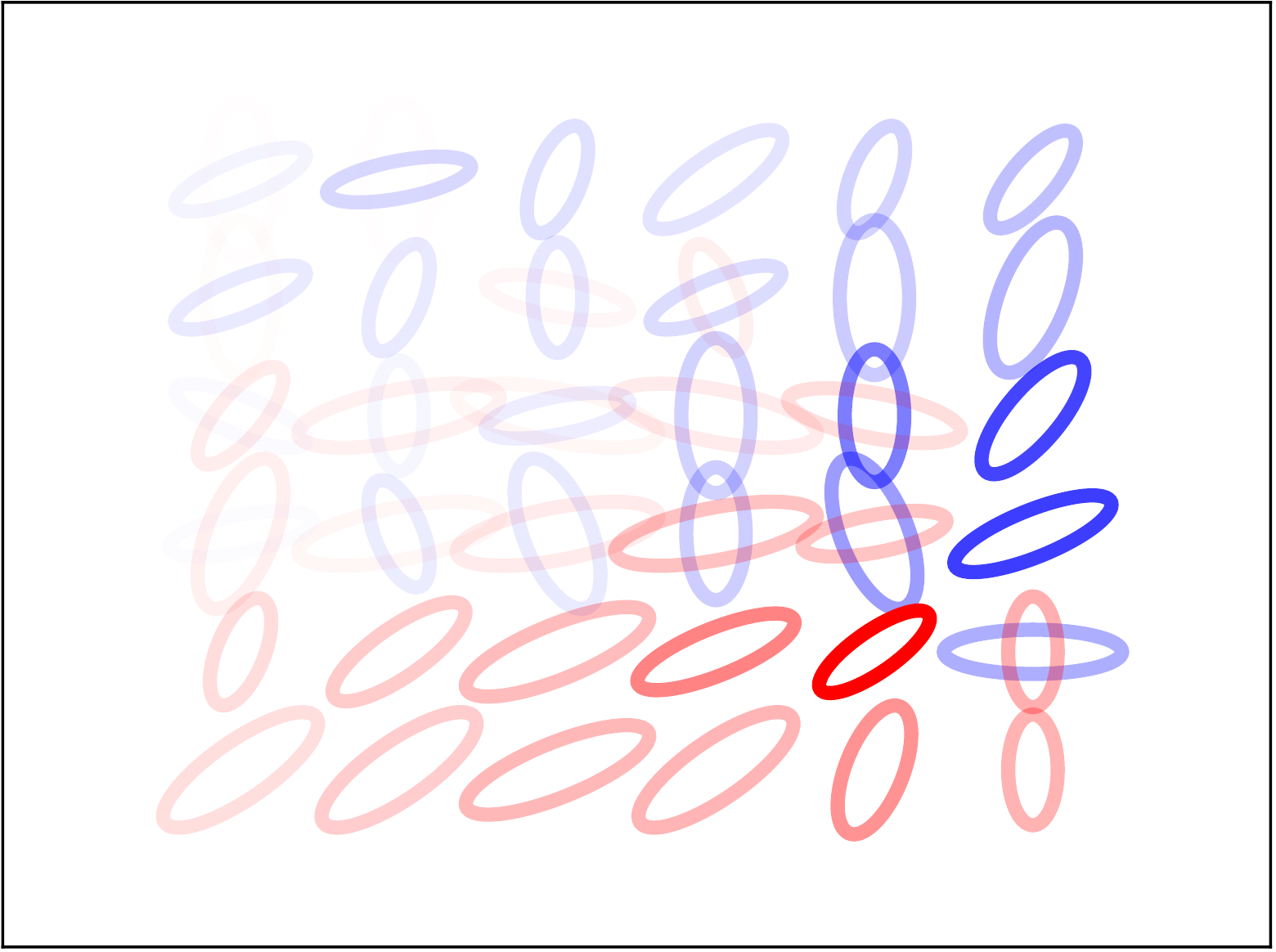}
	\includegraphics[width=0.19\linewidth]{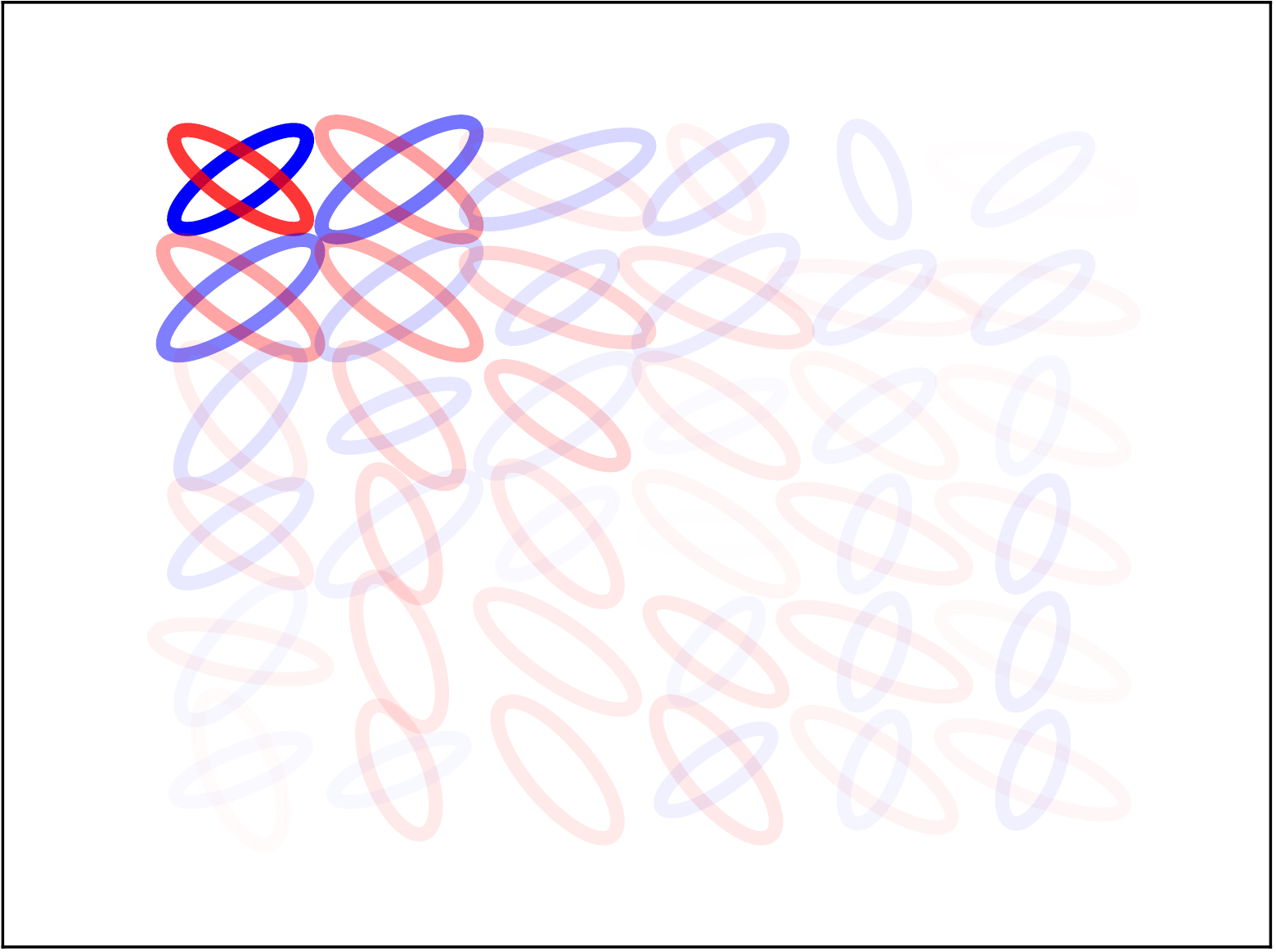}
	\includegraphics[width=0.19\linewidth]{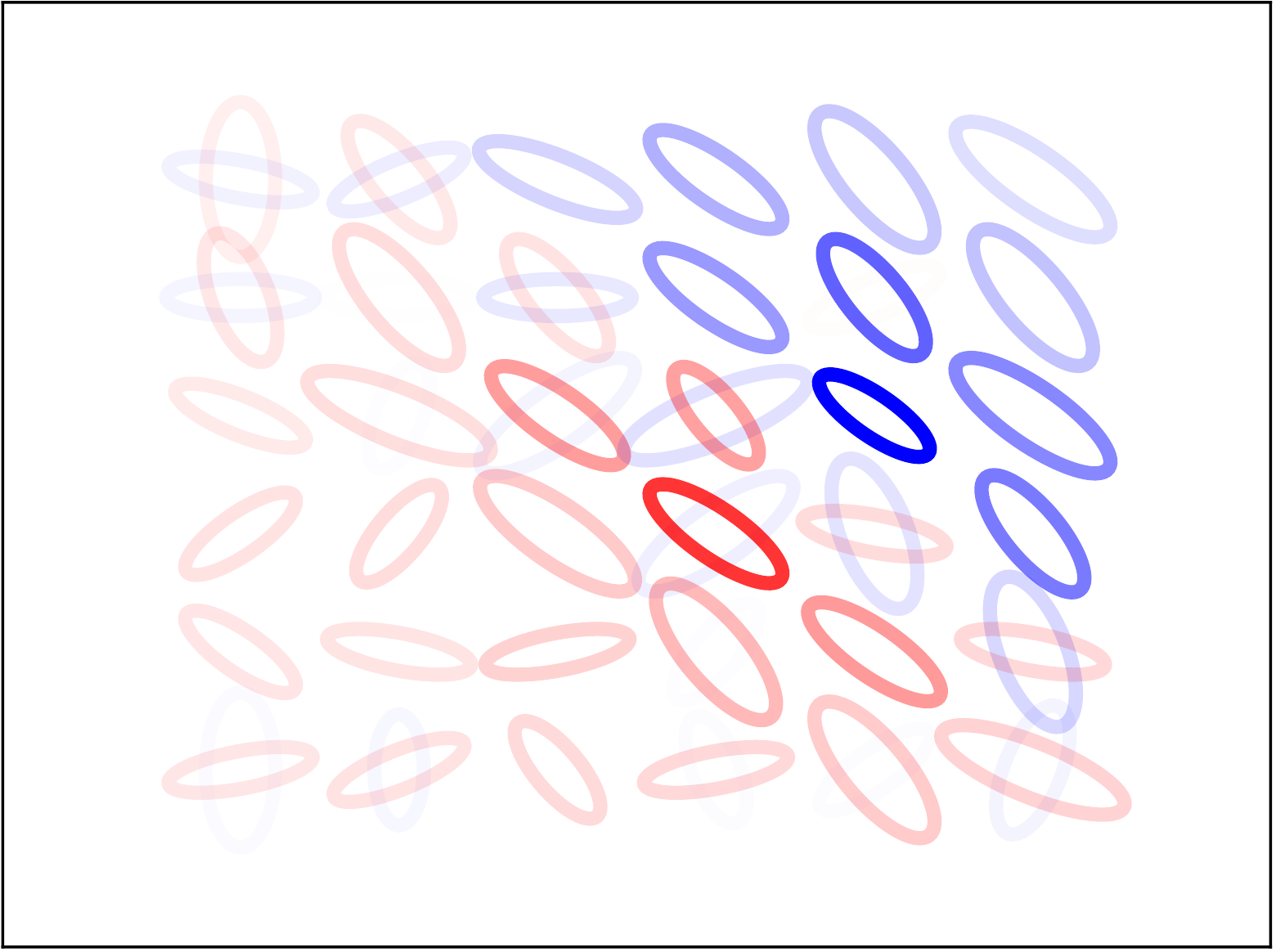}
	\includegraphics[width=0.19\linewidth]{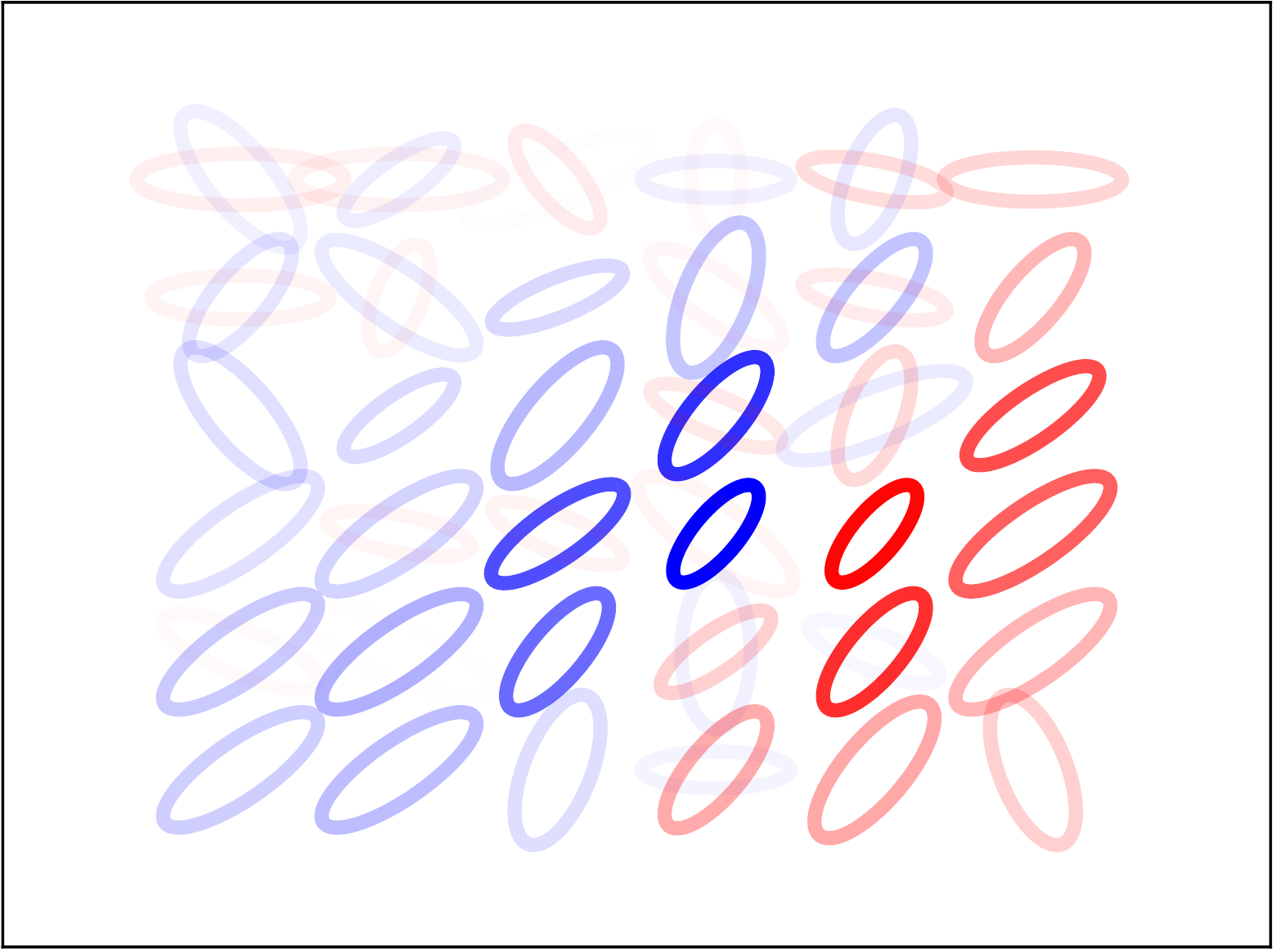}
	\includegraphics[width=0.19\linewidth]{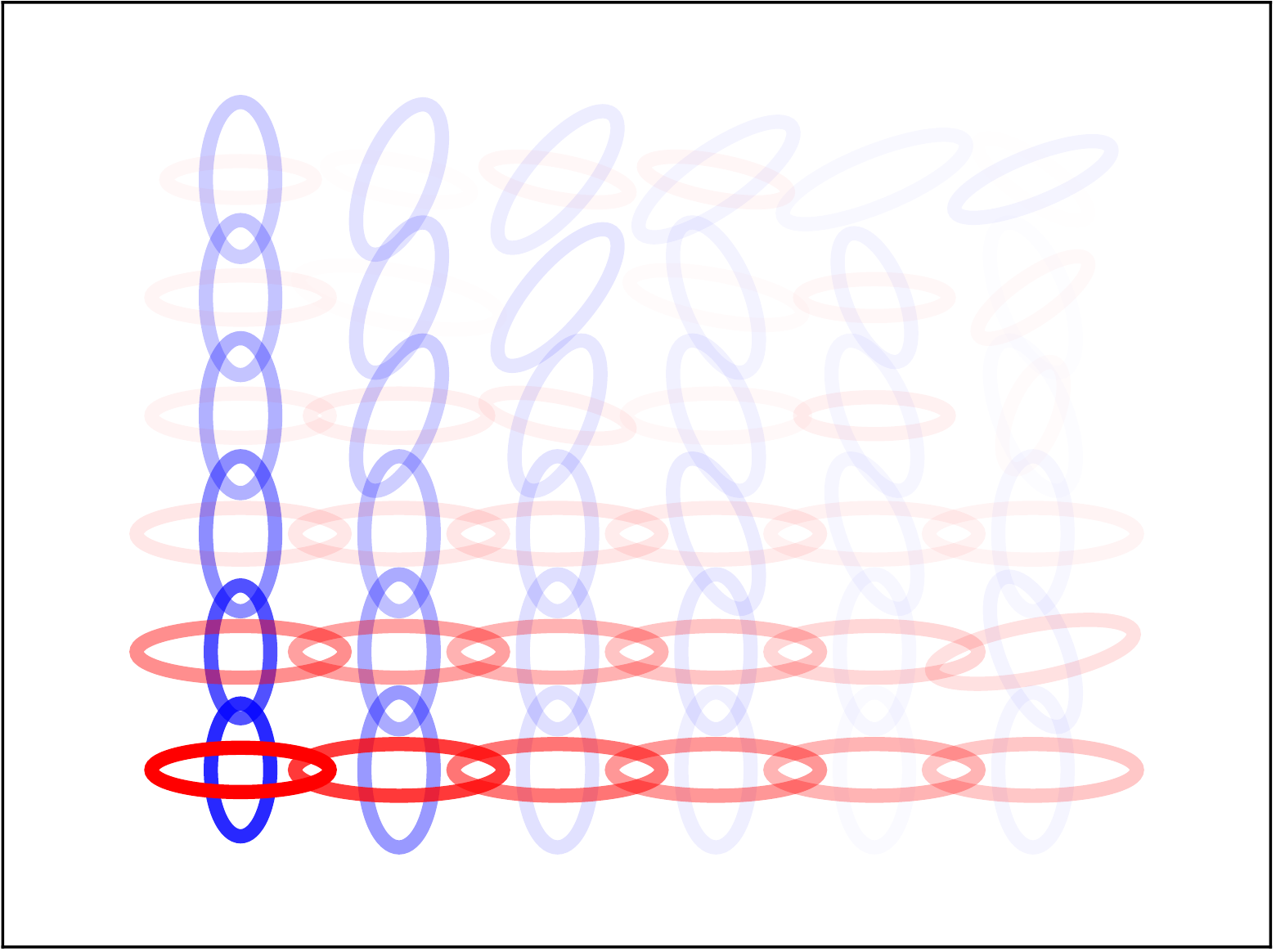}
	\includegraphics[width=0.19\linewidth]{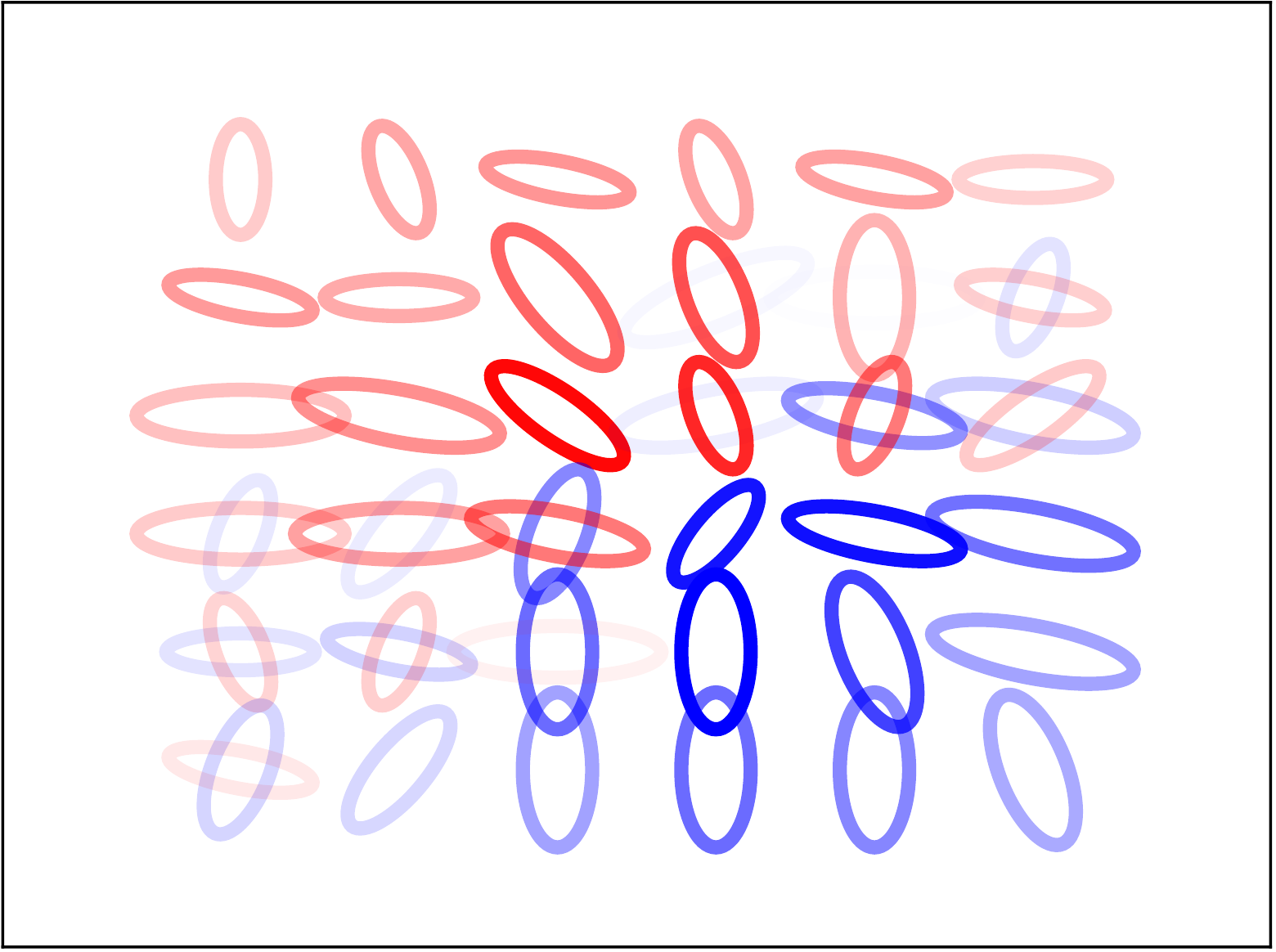}
	\includegraphics[width=0.19\linewidth]{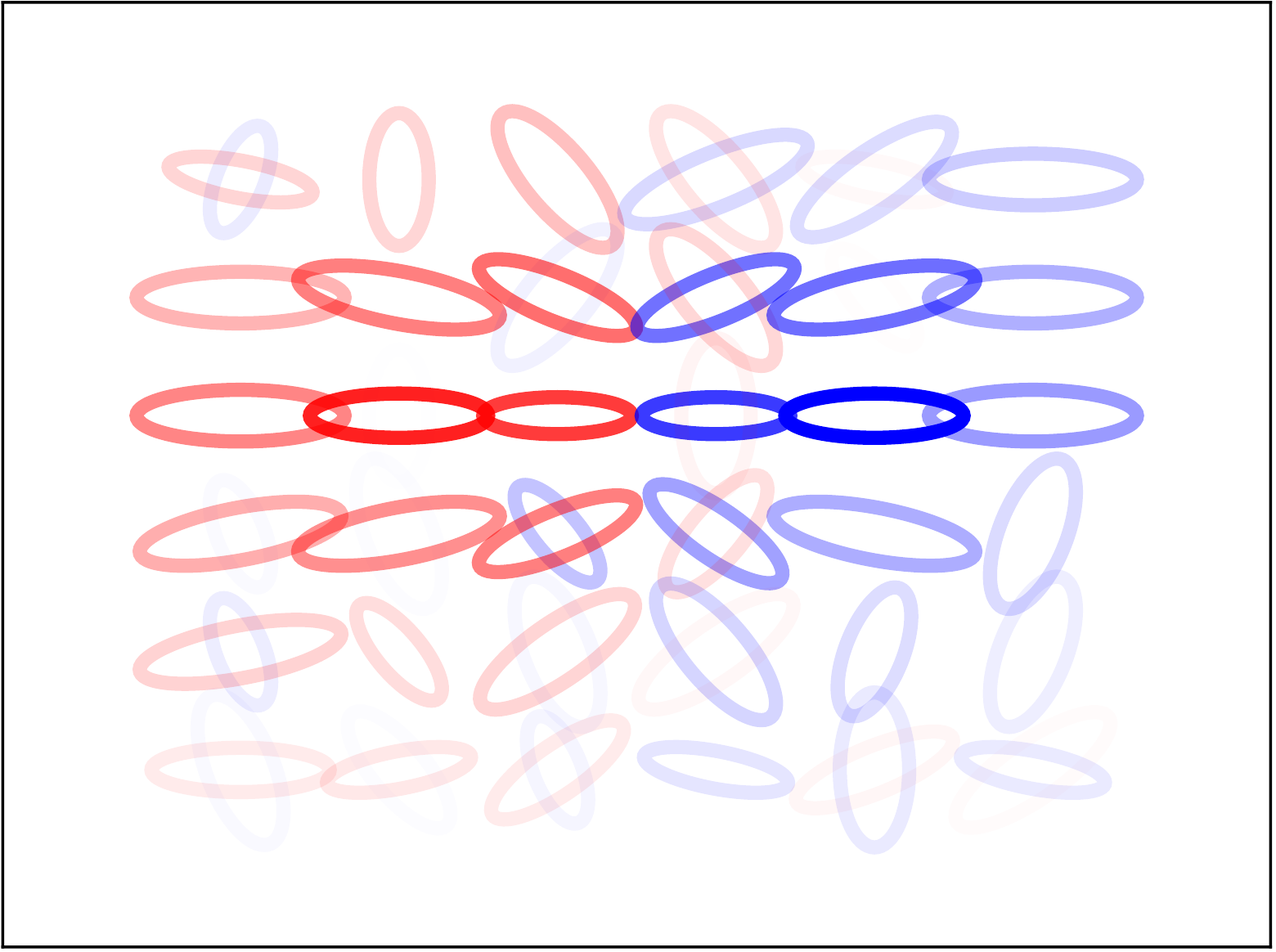}
	\includegraphics[width=0.19\linewidth]{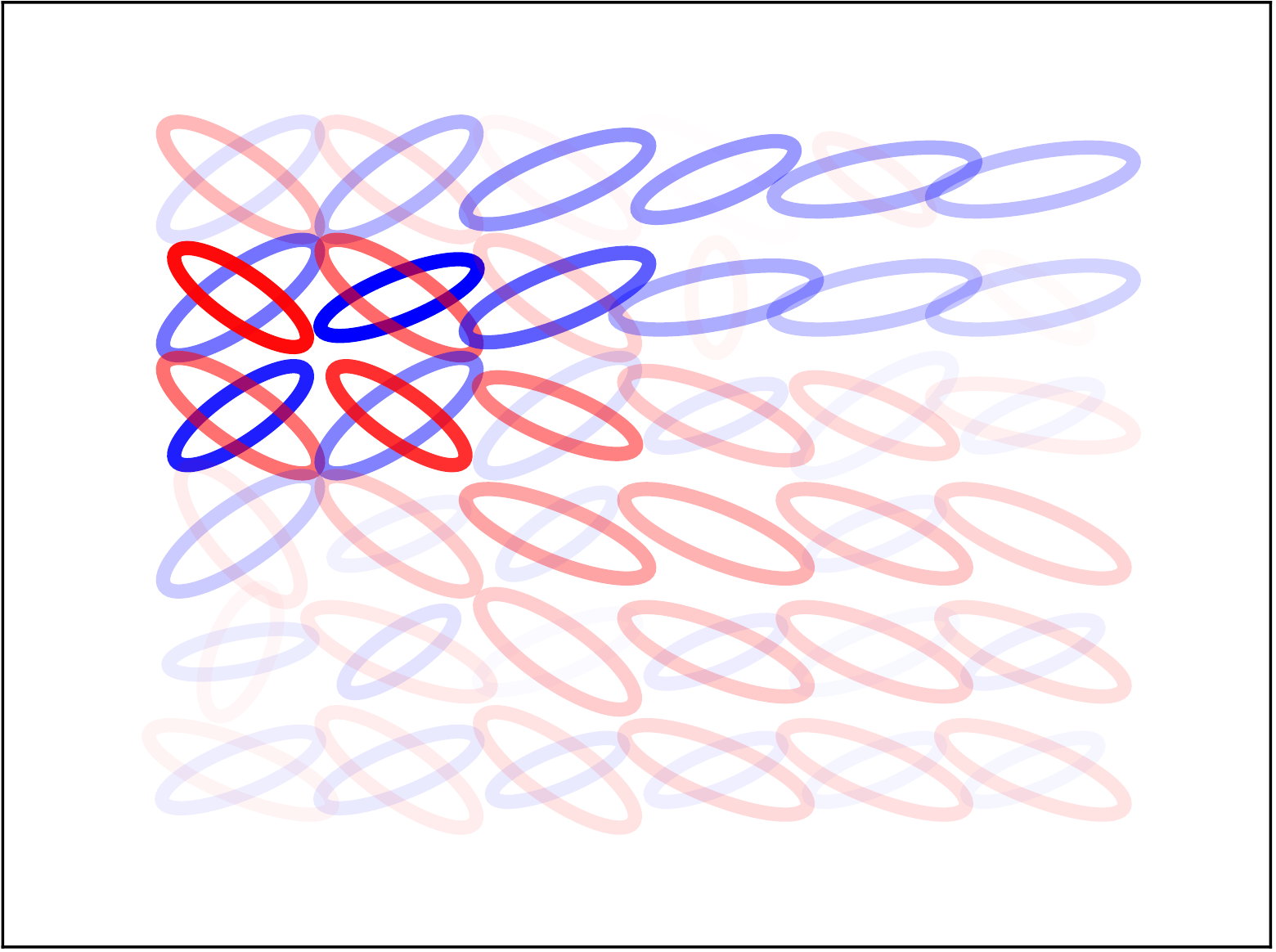}
	\includegraphics[width=0.19\linewidth]{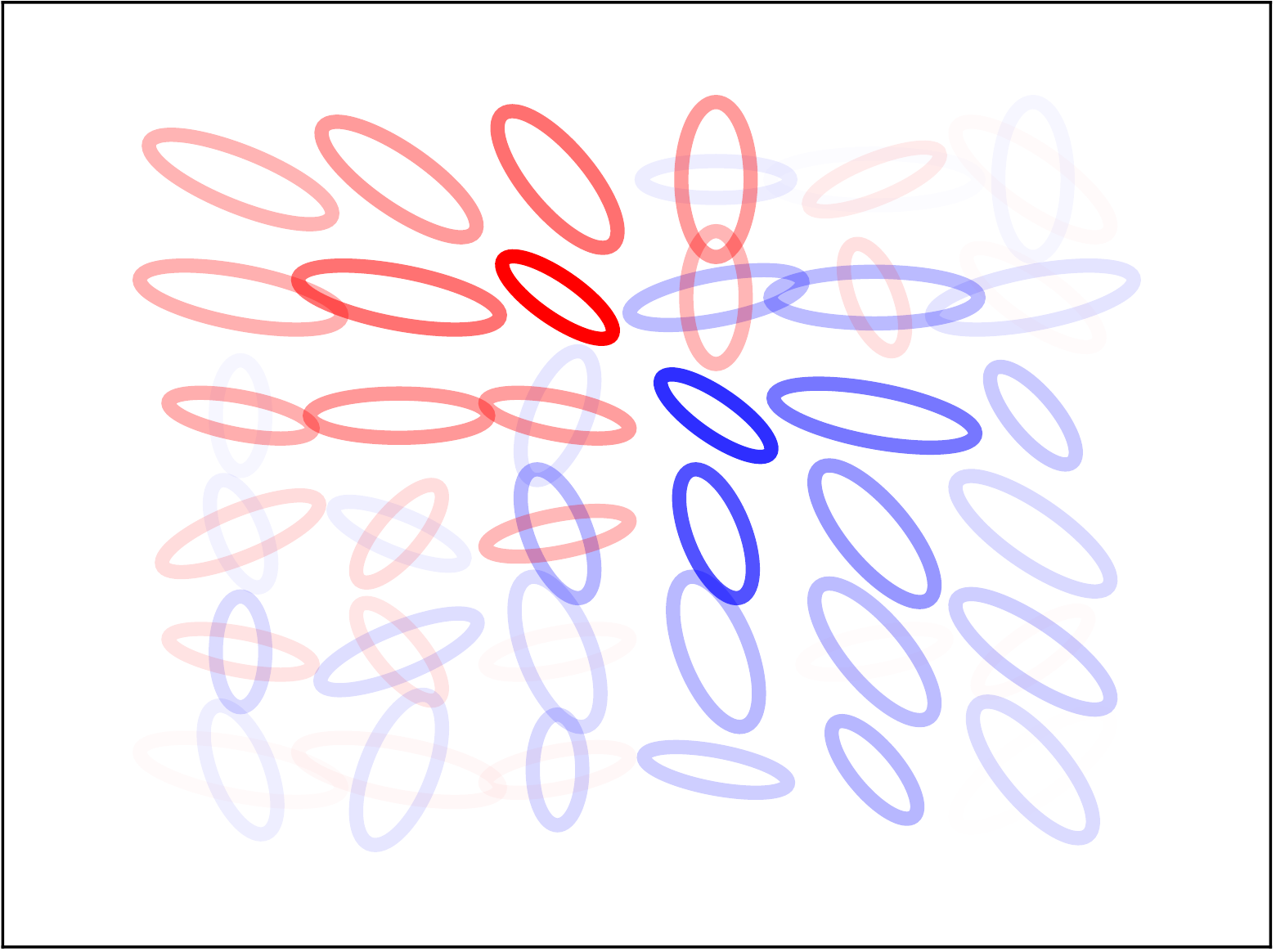}
	\includegraphics[width=0.19\linewidth]{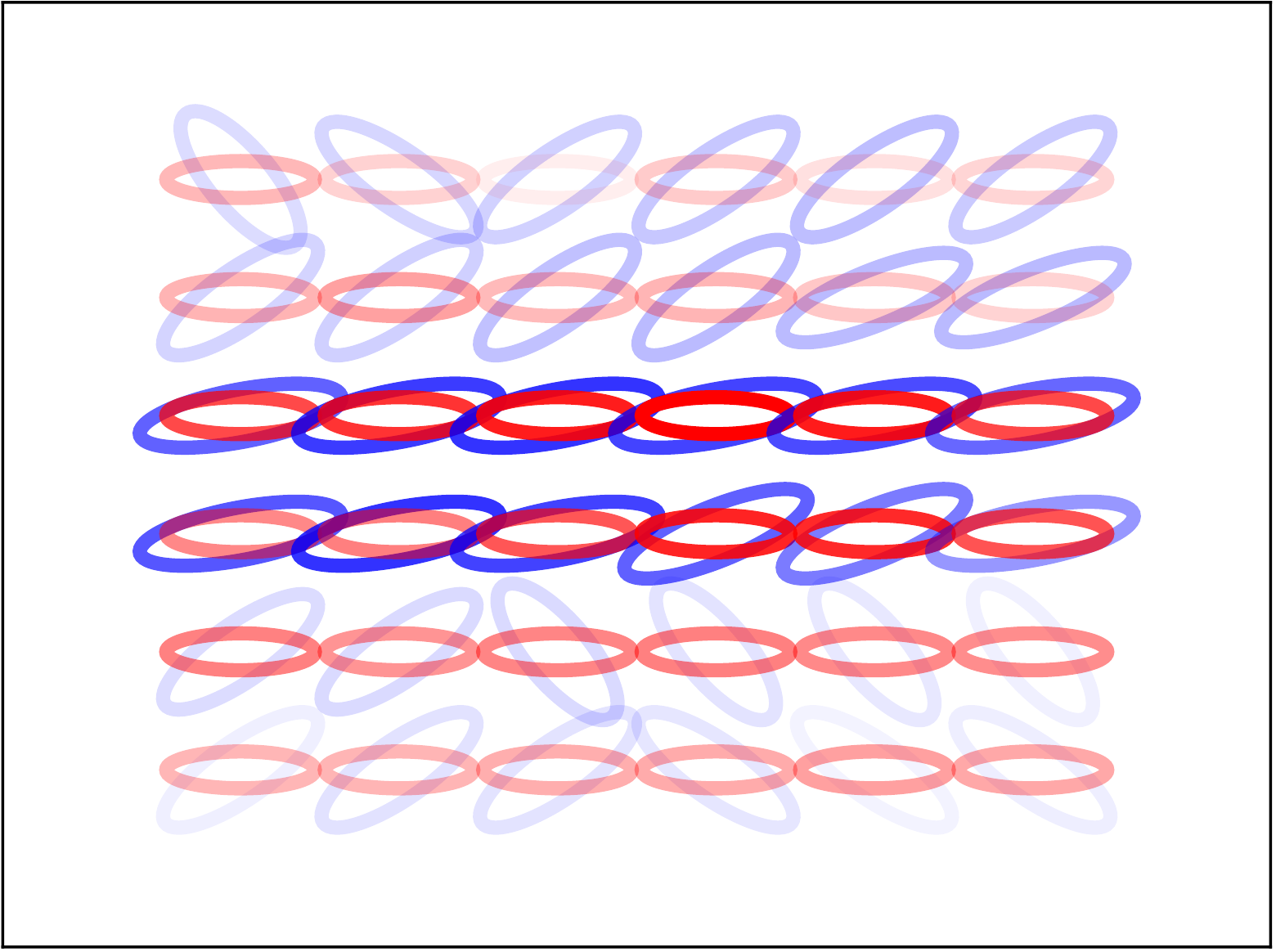}
	\includegraphics[width=0.19\linewidth]{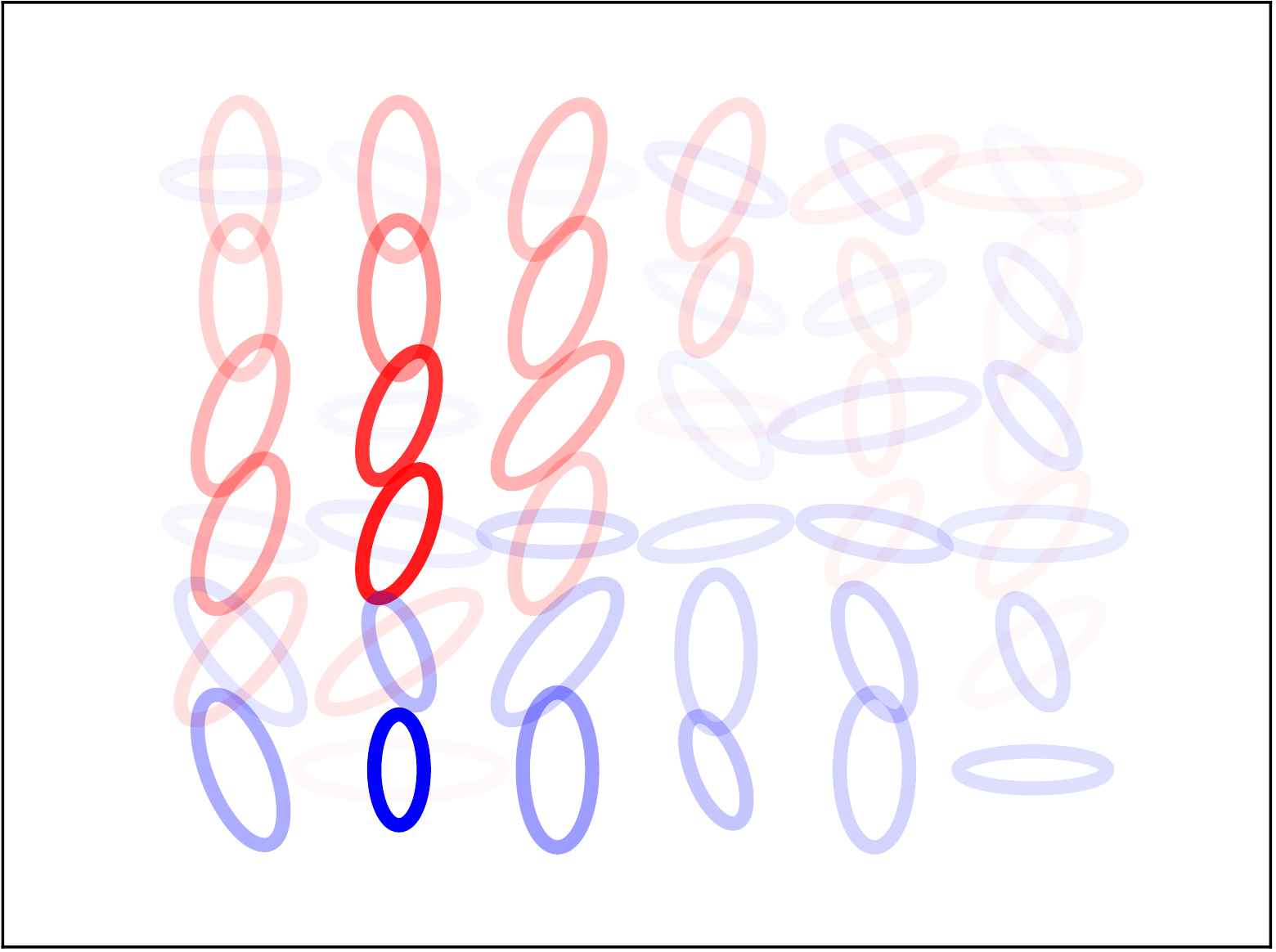}
	\includegraphics[width=0.19\linewidth]{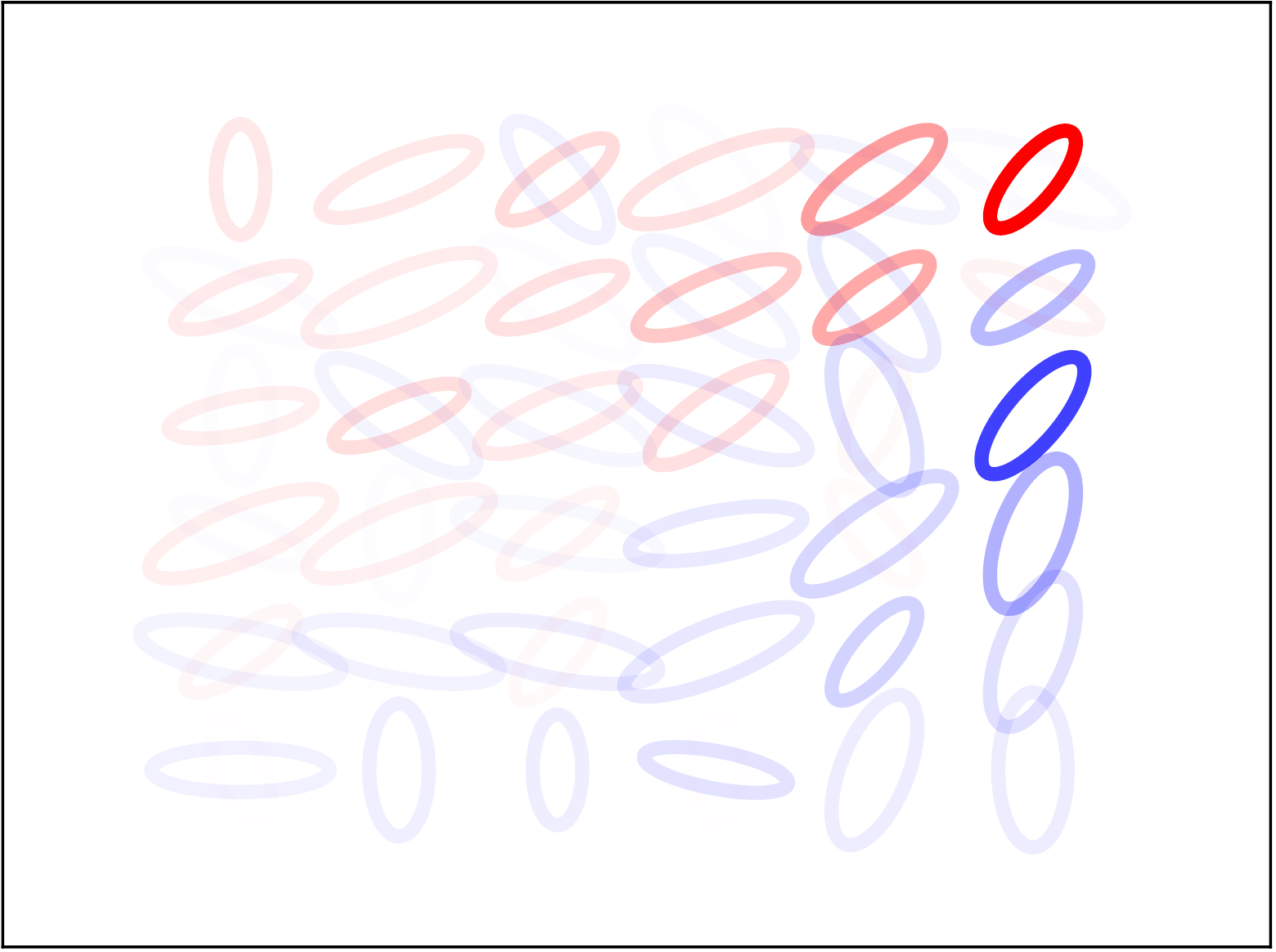}
	\includegraphics[width=0.19\linewidth]{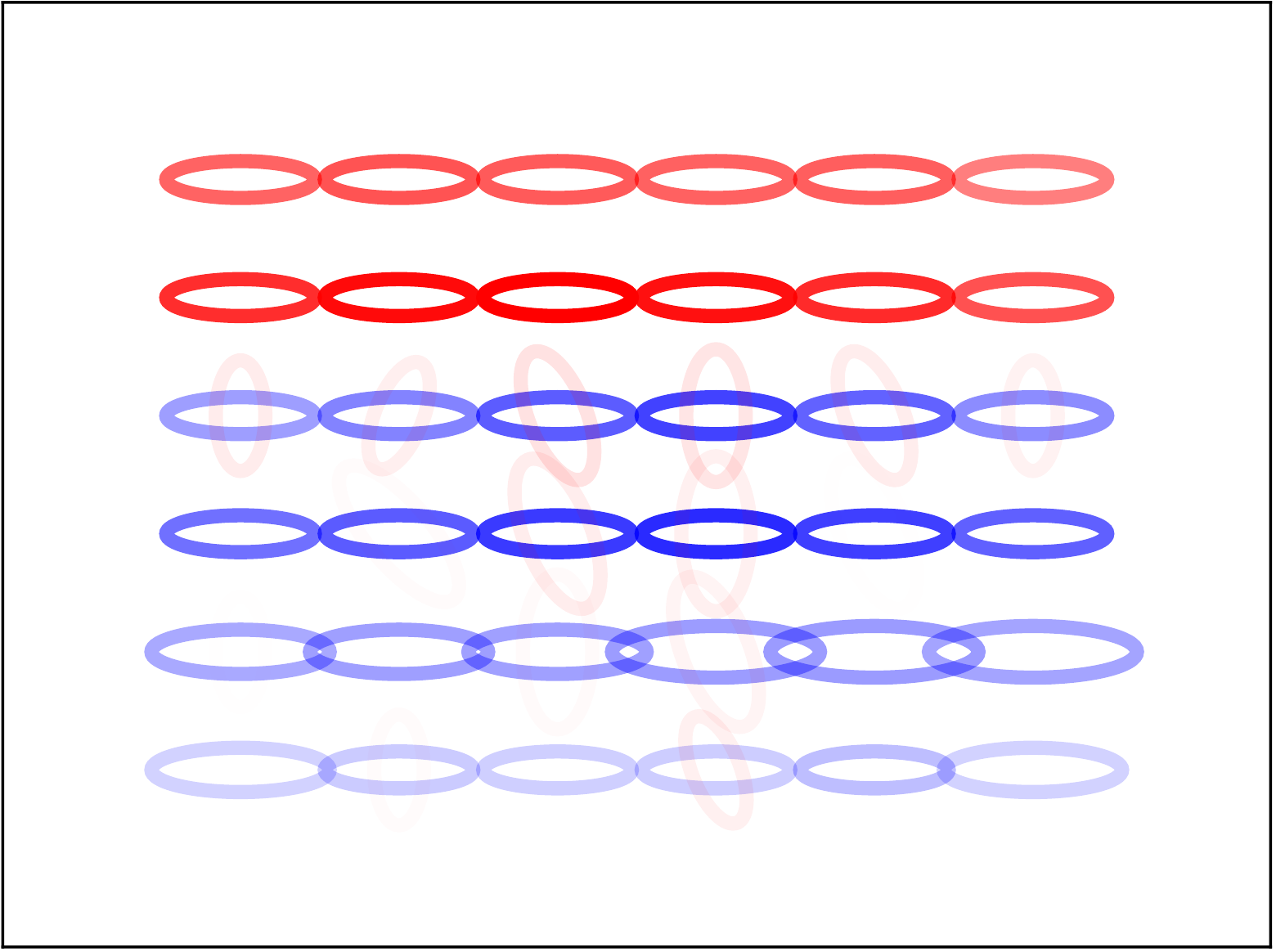}
	\includegraphics[width=0.19\linewidth]{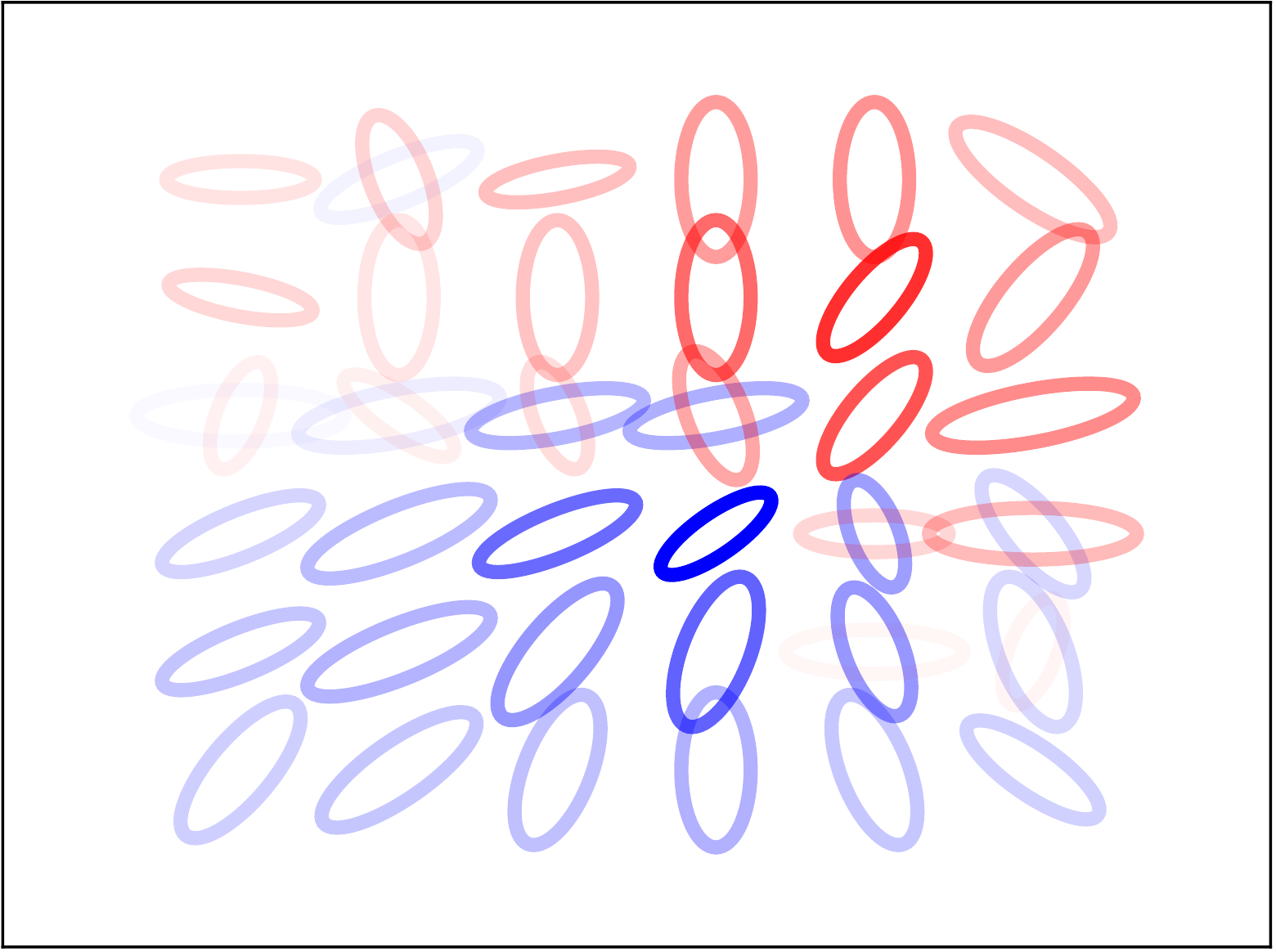}
	\includegraphics[width=0.19\linewidth]{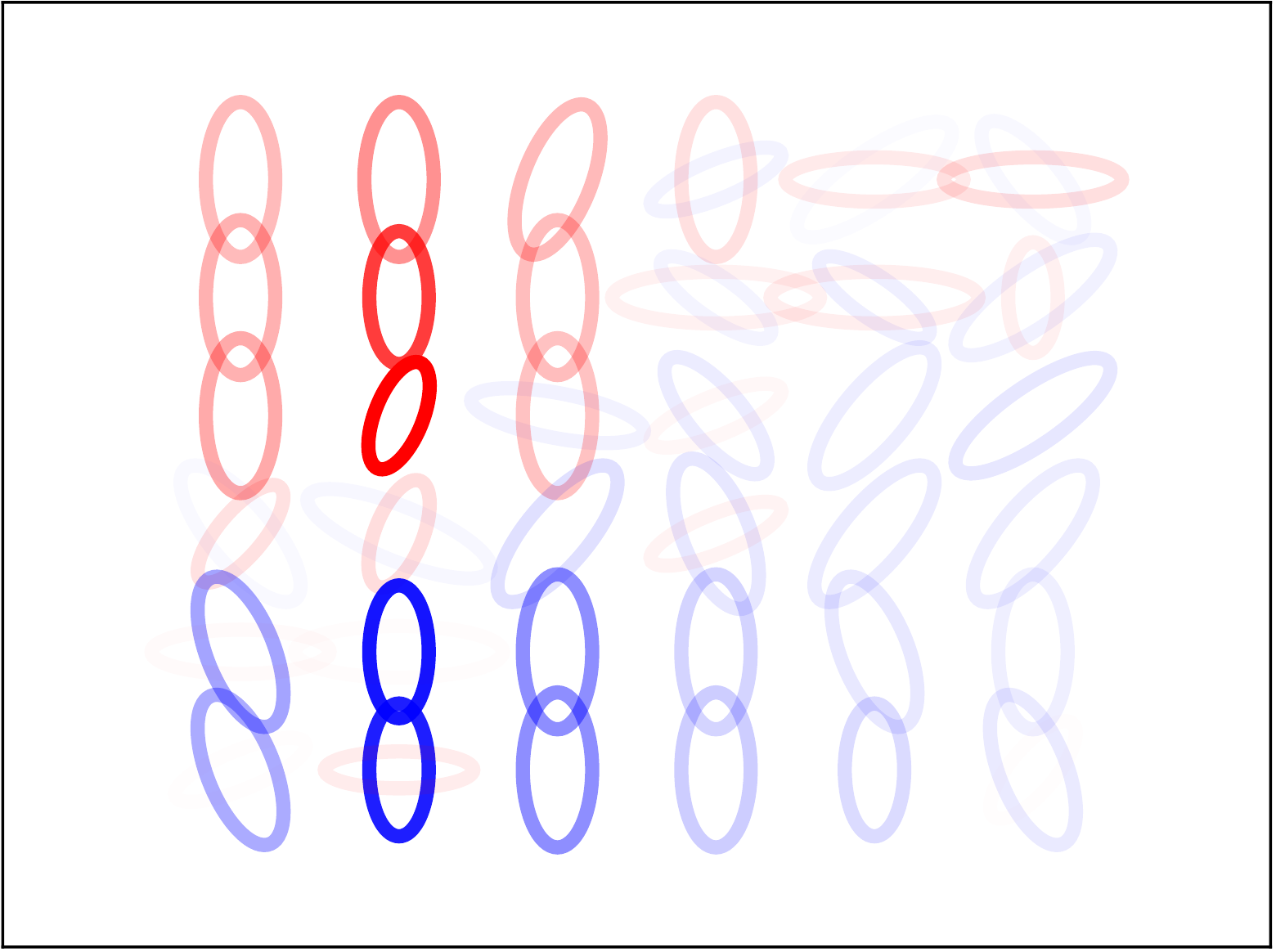}
	\includegraphics[width=0.19\linewidth]{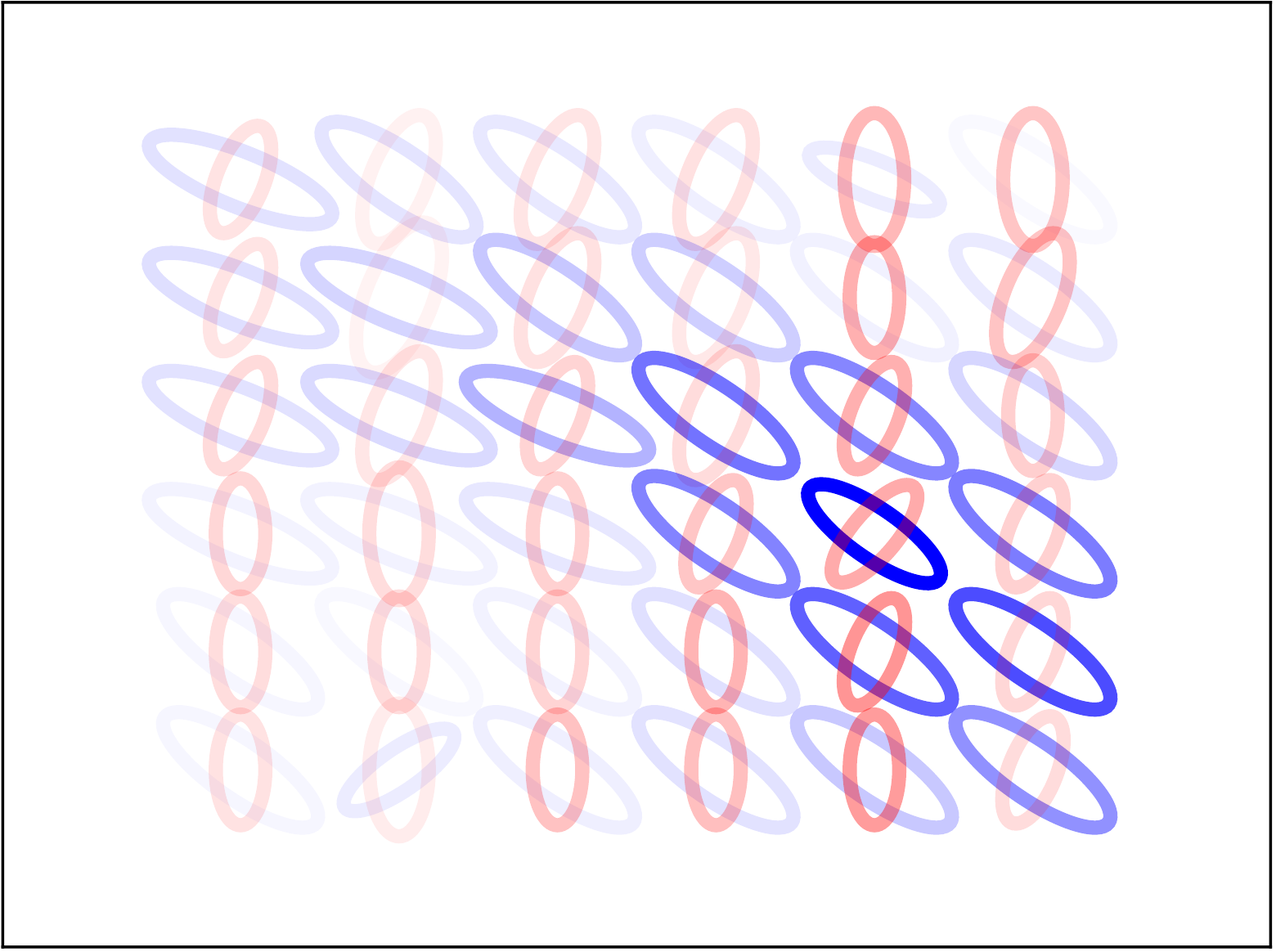} \\
	\phantomsubcaption
	\label{fig:6x6gpsc}
\end{subfigure}
\caption{\textbf{Visualization of V2 Model Units With 6x6 Spatial Locations and 100 principal components.} (a) Sparse coding with a regularization coefficient of 0.5. (b) Sparse coding with a regularization coefficient of 4.0. (c) ICA. Values outside the central 6x6 region do not have a response. Opacity reflects the response intensity, color reflects the sign of the response (red for positive and blue for negative), and size reflects frequency.}
\label{fig:6x6gps}
\end{figure}

\begin{figure}
\Large \textbf{(a)} \\
\begin{subfigure}[t]{\linewidth}
	\centering
	\includegraphics[width=0.19\linewidth]{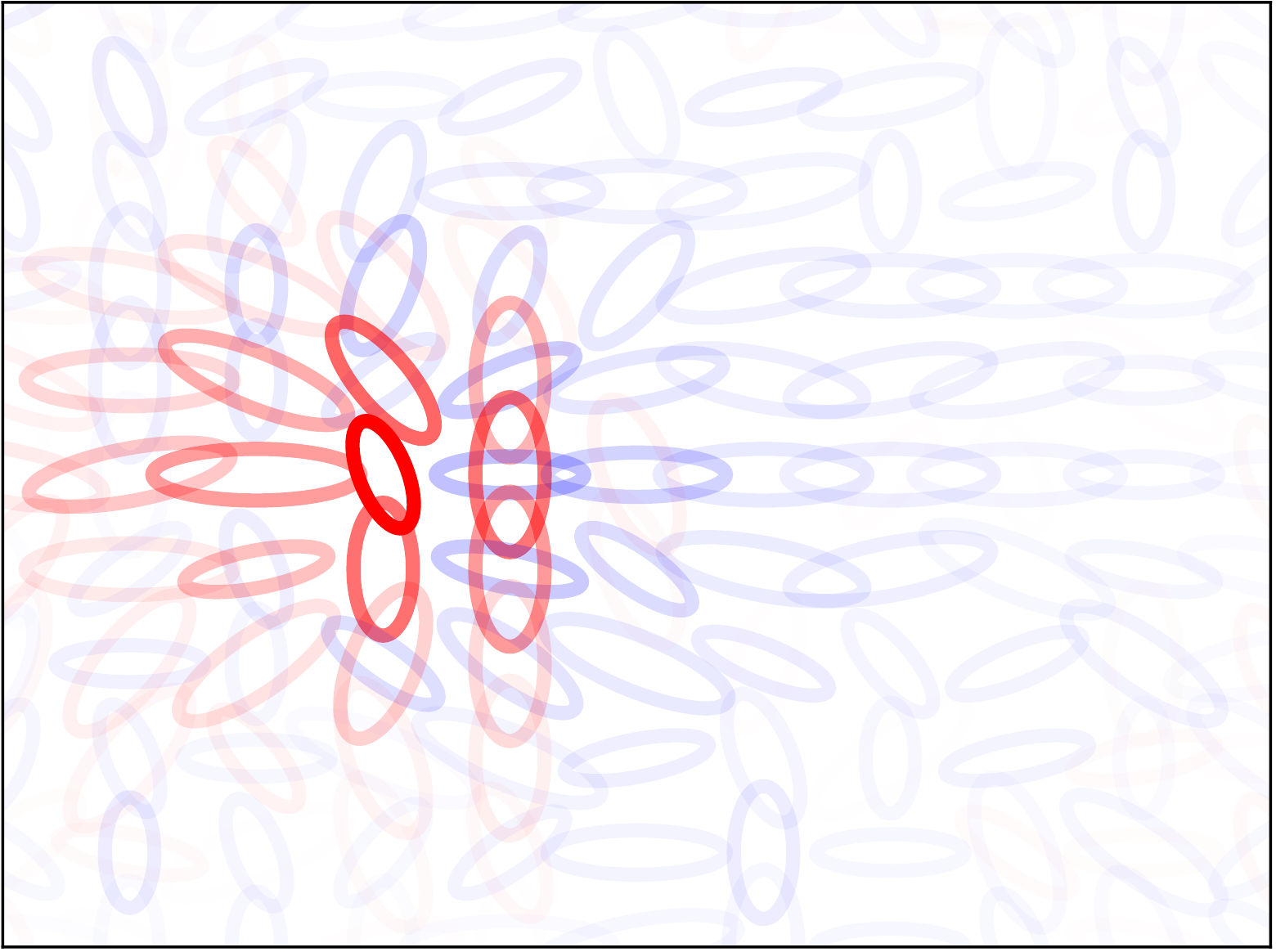}
	\includegraphics[width=0.19\linewidth]{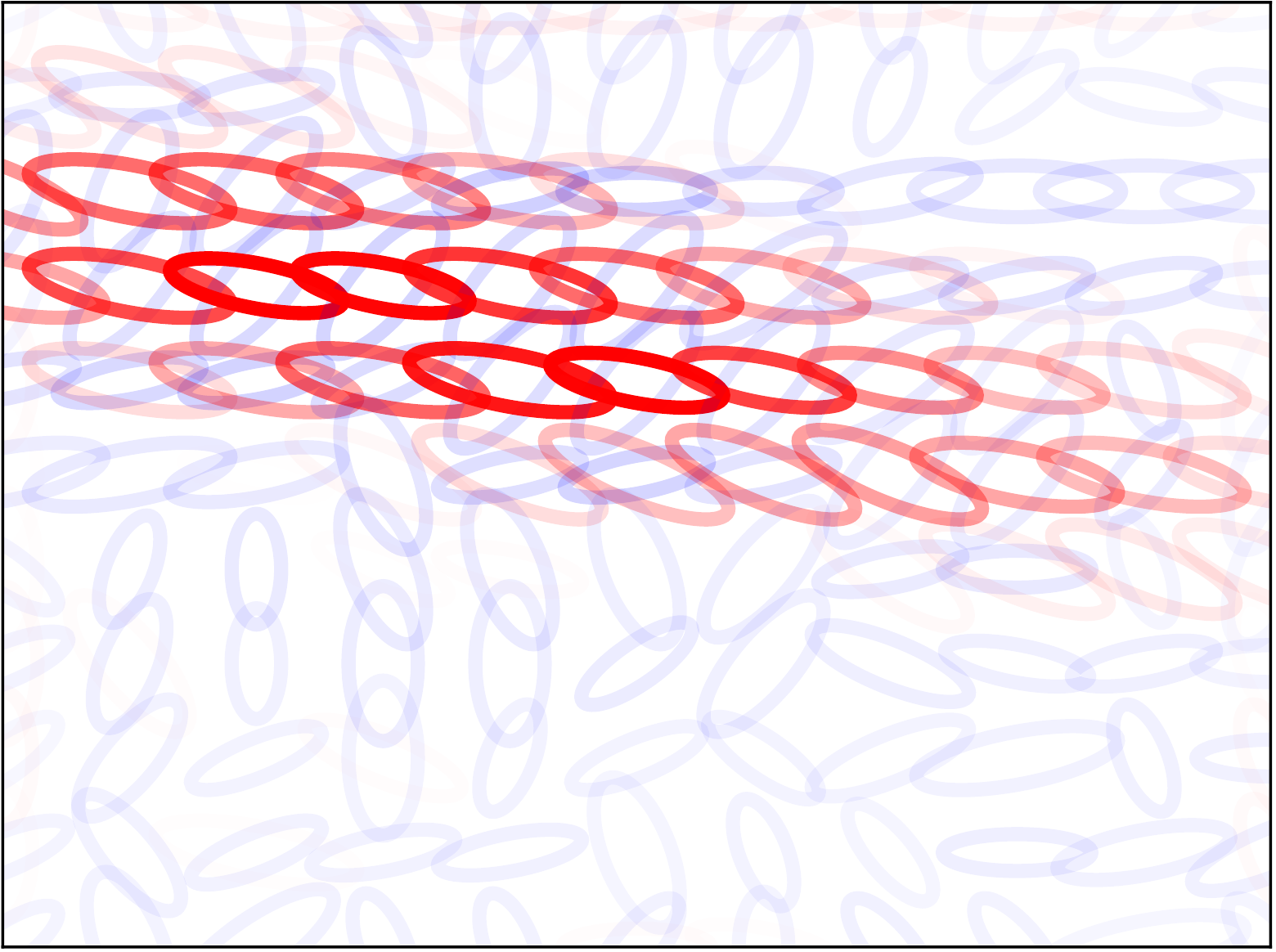}
	\includegraphics[width=0.19\linewidth]{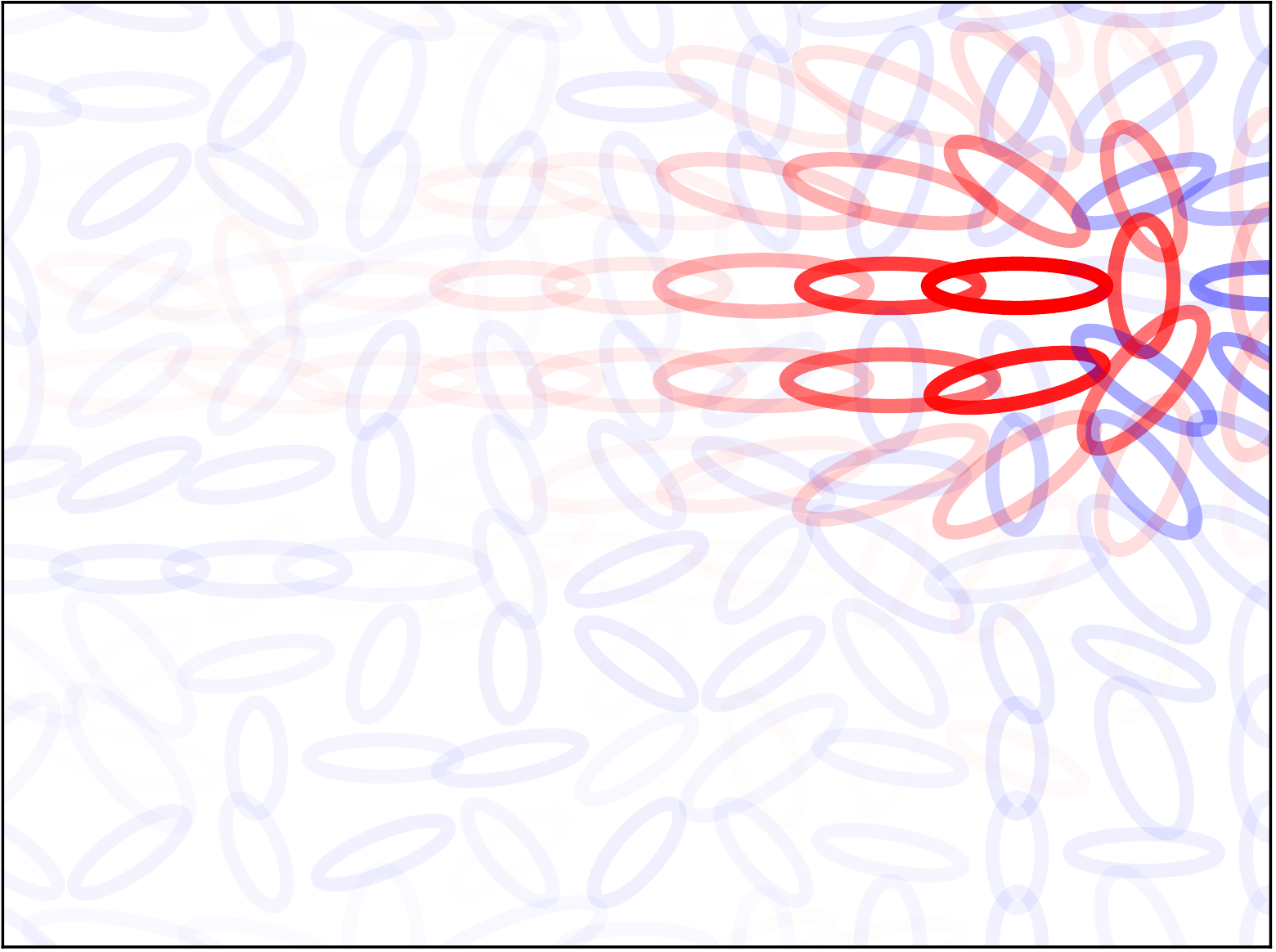}
	\includegraphics[width=0.19\linewidth]{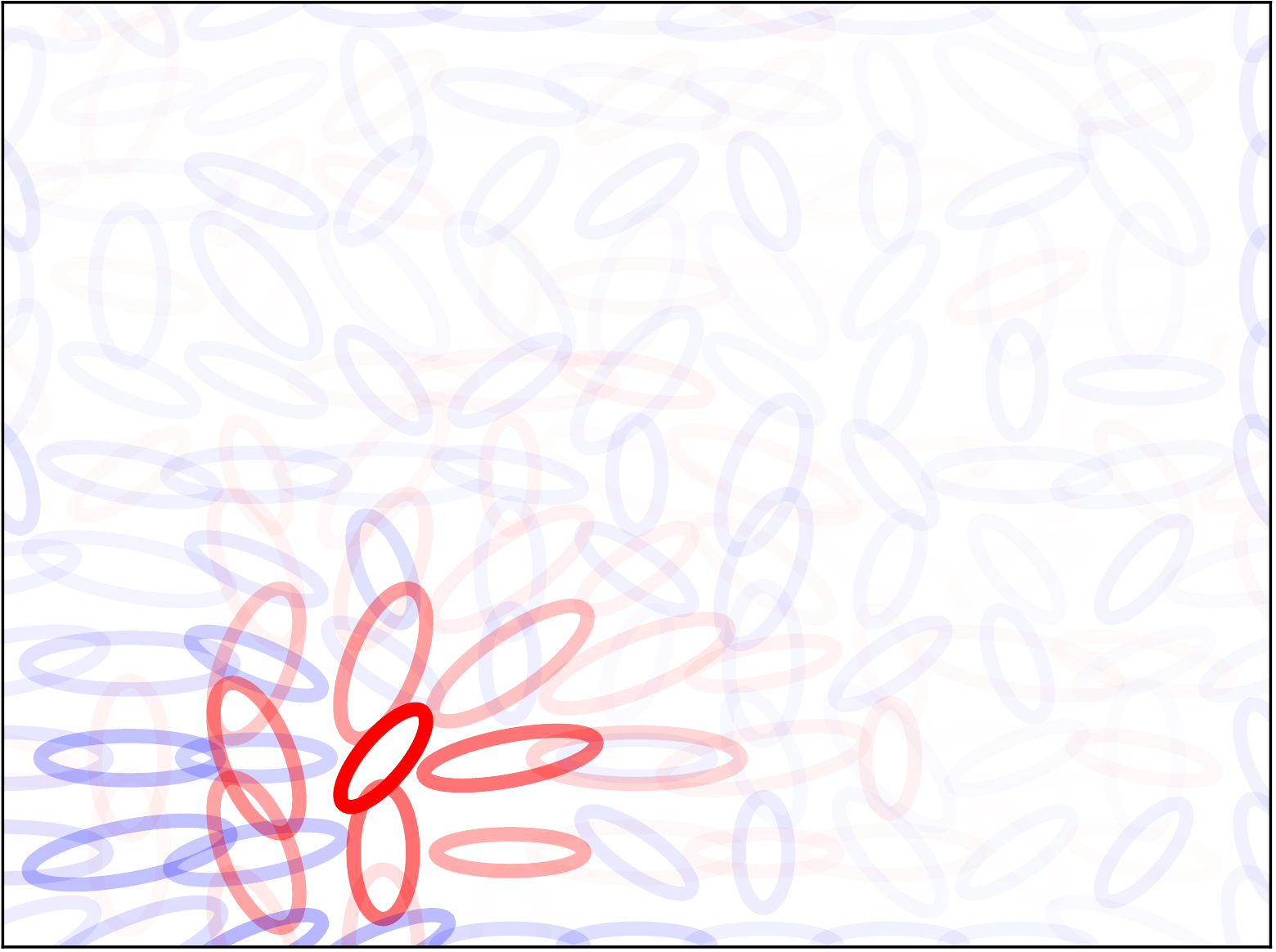}
	\includegraphics[width=0.19\linewidth]{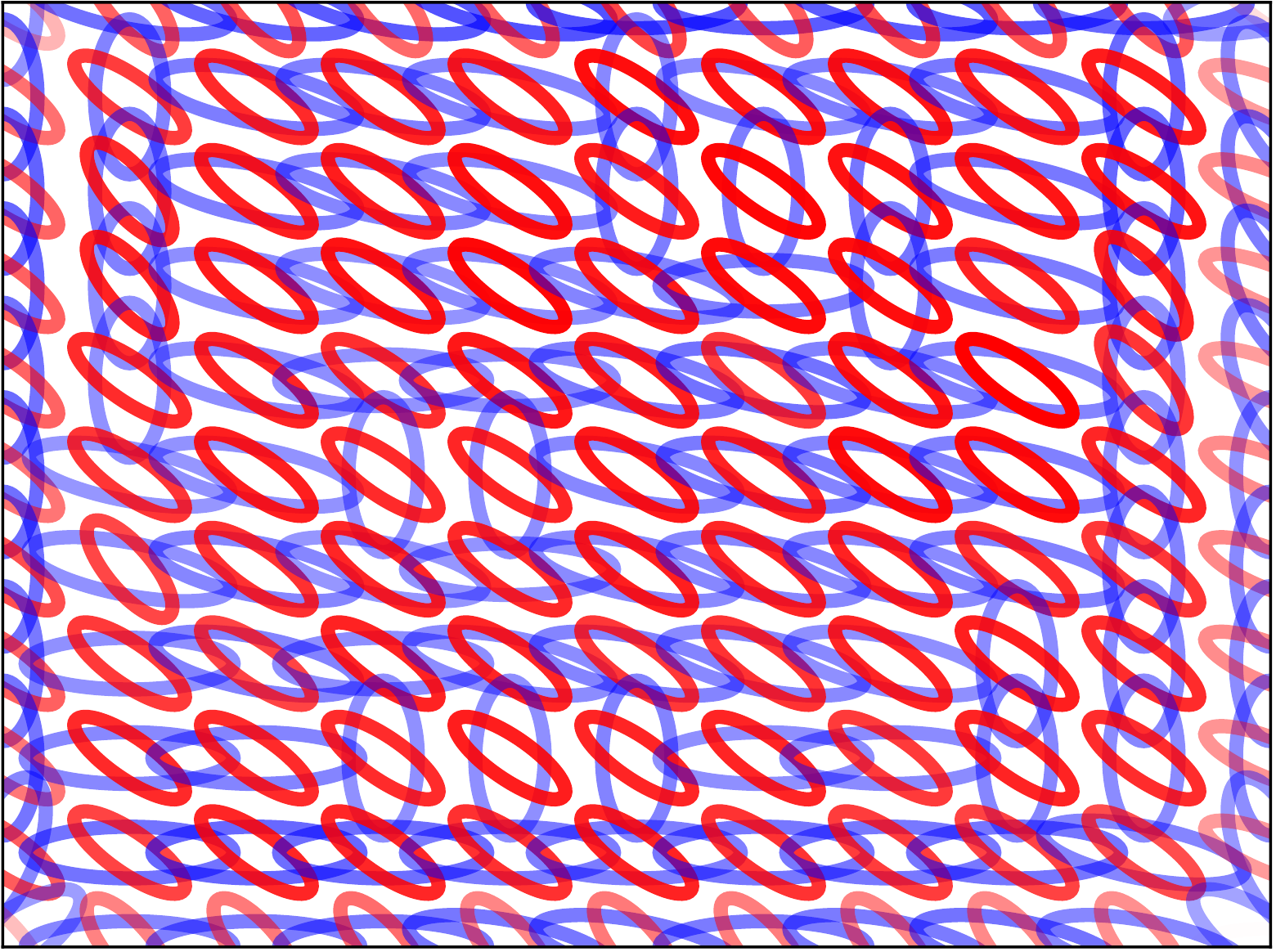}
	\includegraphics[width=0.19\linewidth]{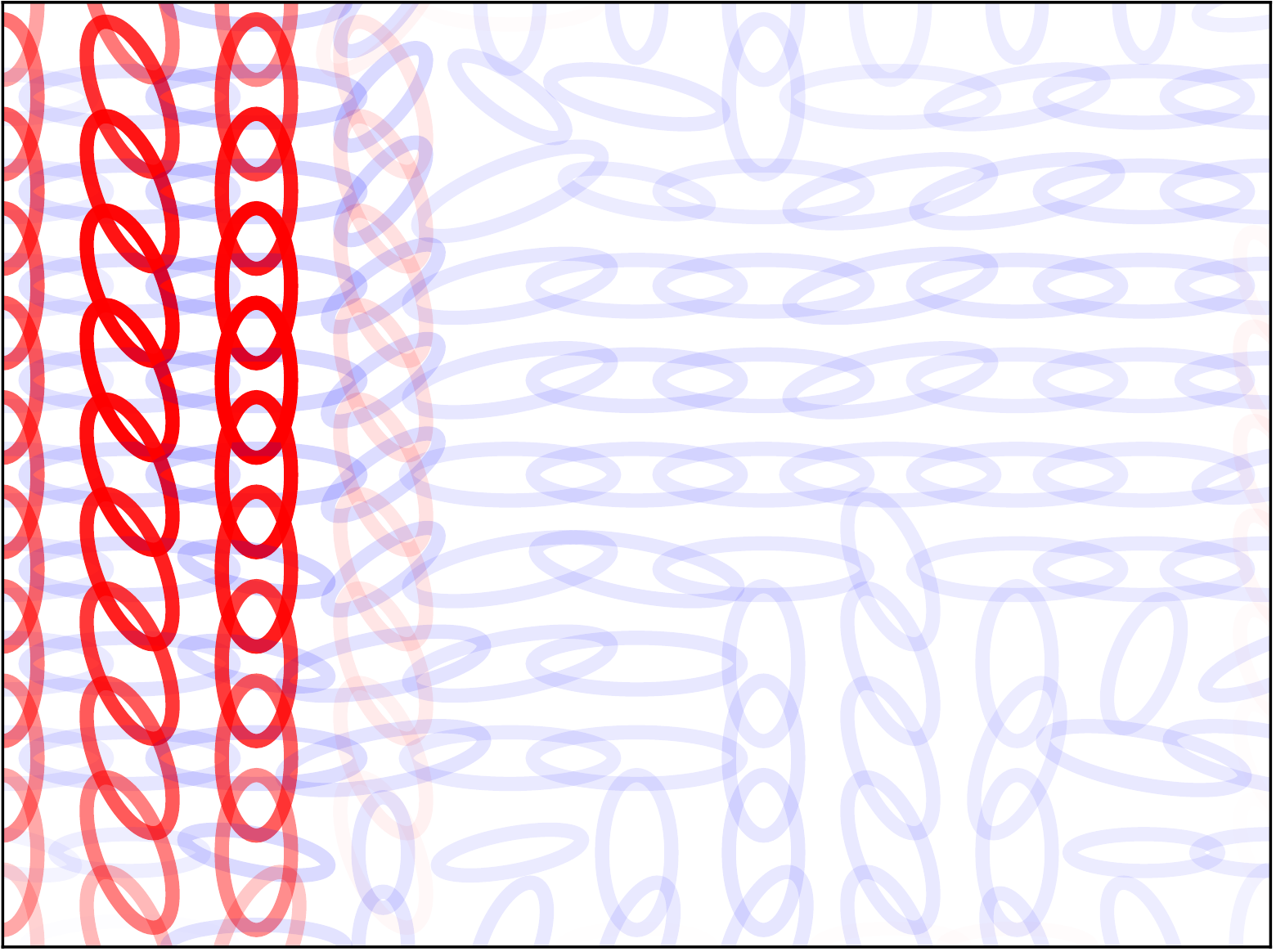}
	\includegraphics[width=0.19\linewidth]{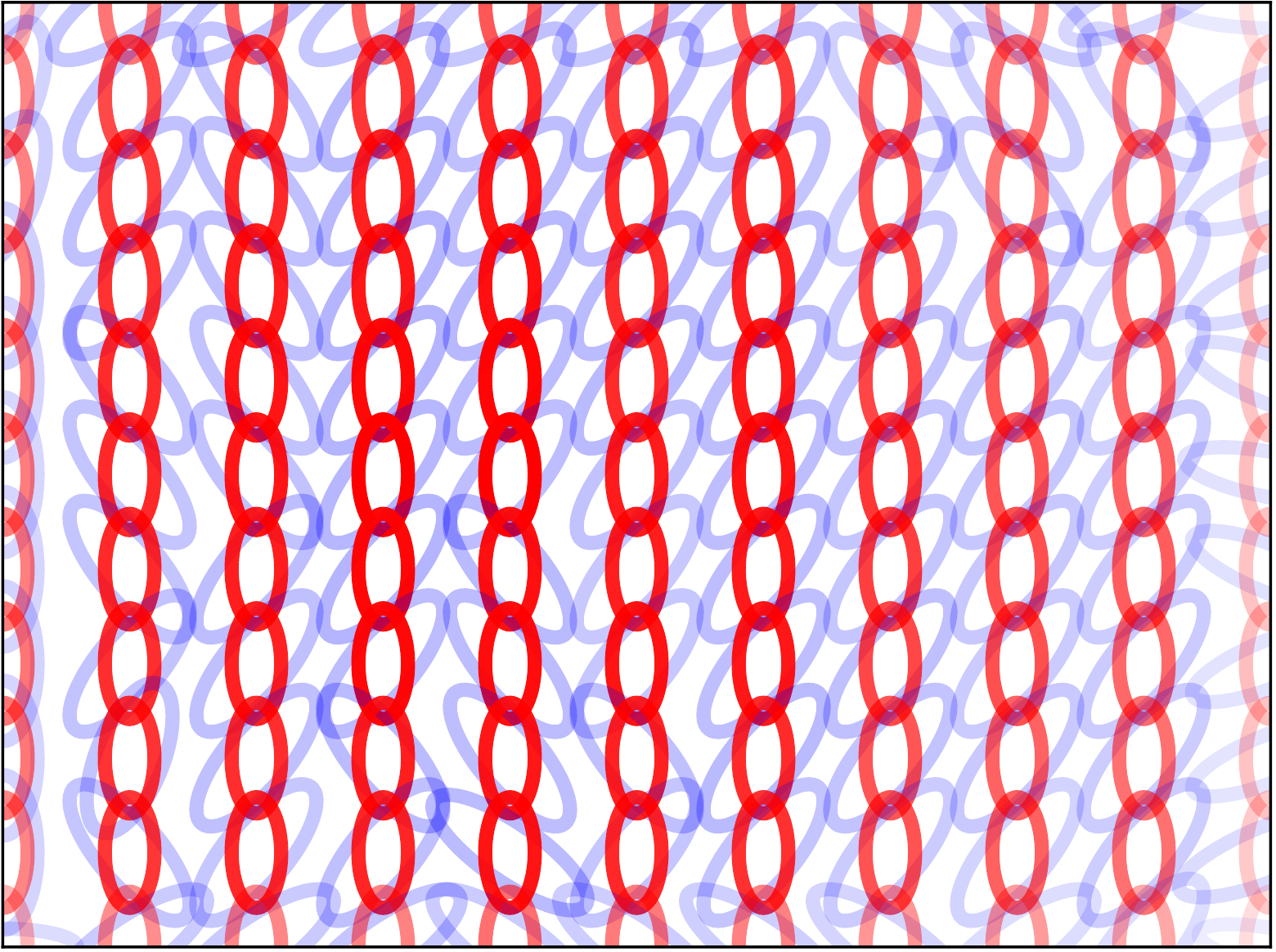}
	\includegraphics[width=0.19\linewidth]{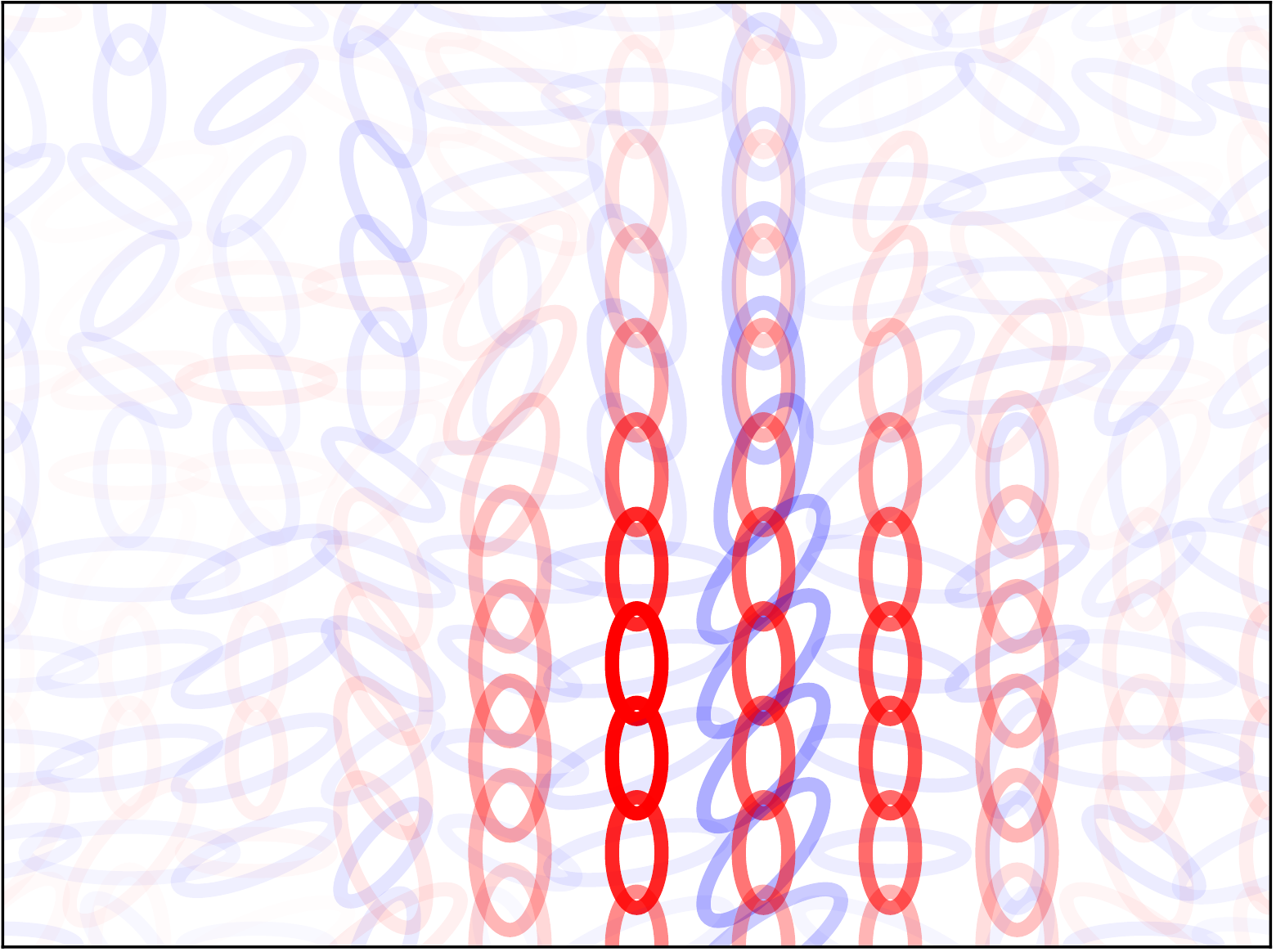}
	\includegraphics[width=0.19\linewidth]{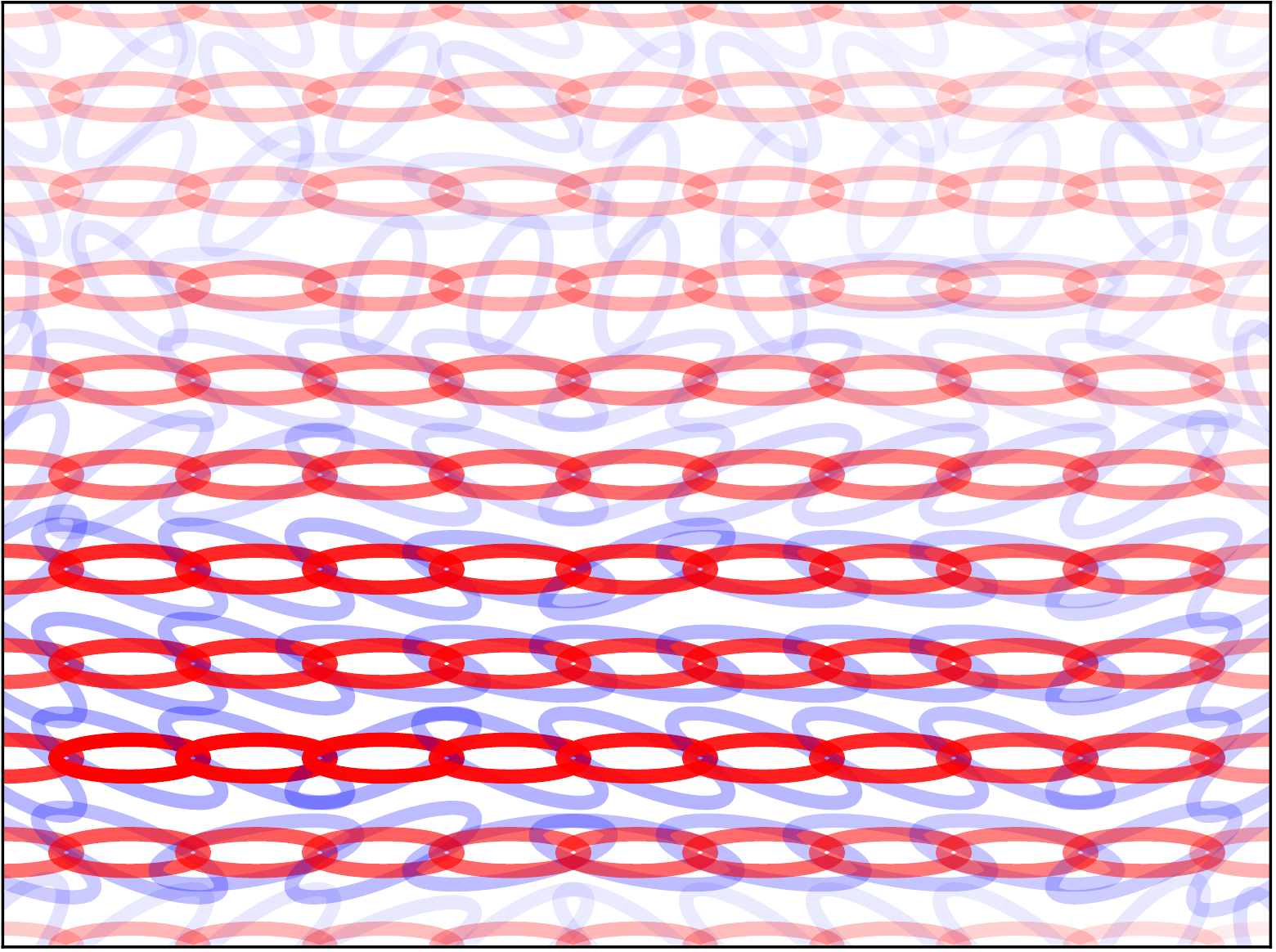}
	\includegraphics[width=0.19\linewidth]{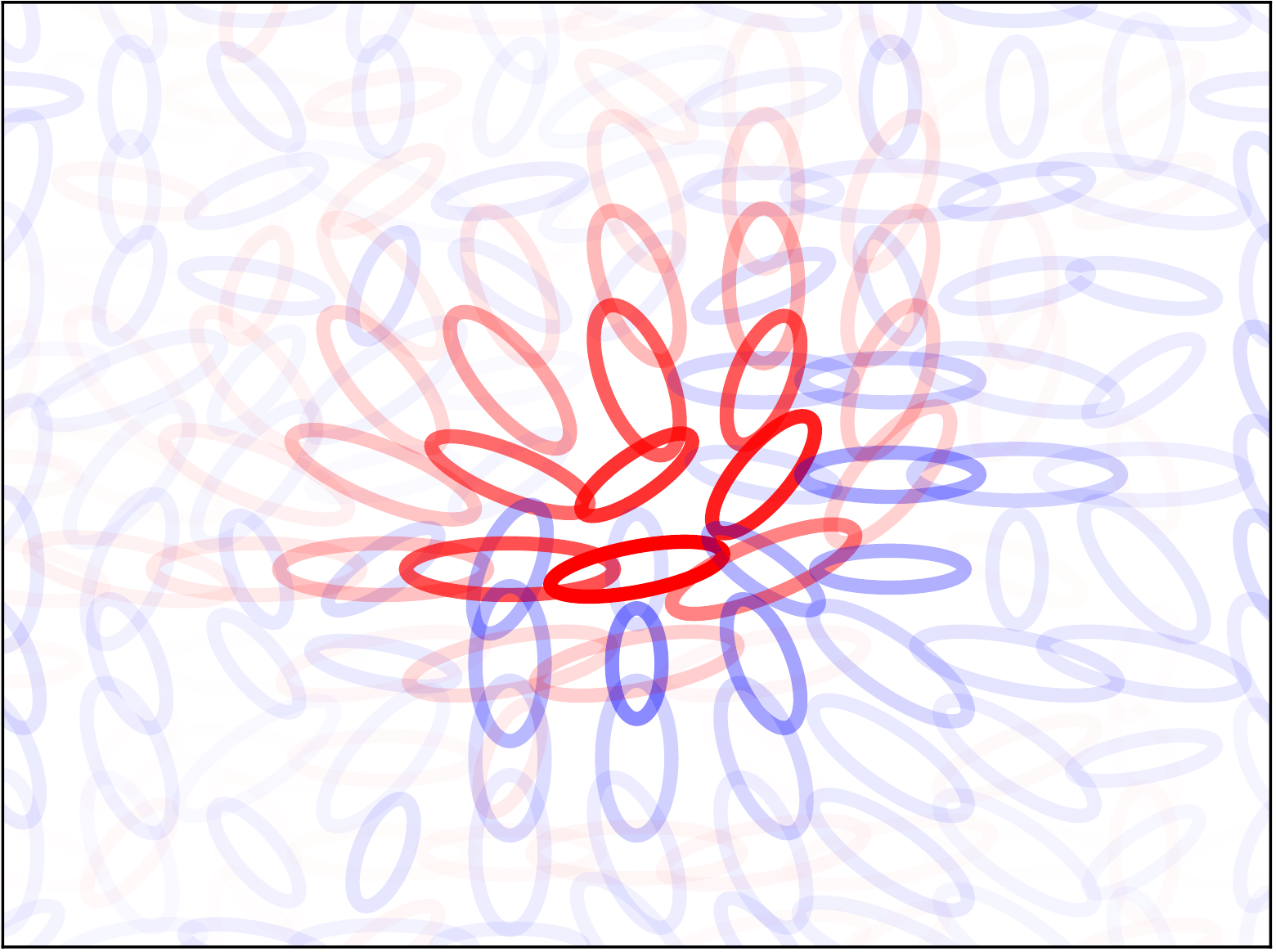}
	\includegraphics[width=0.19\linewidth]{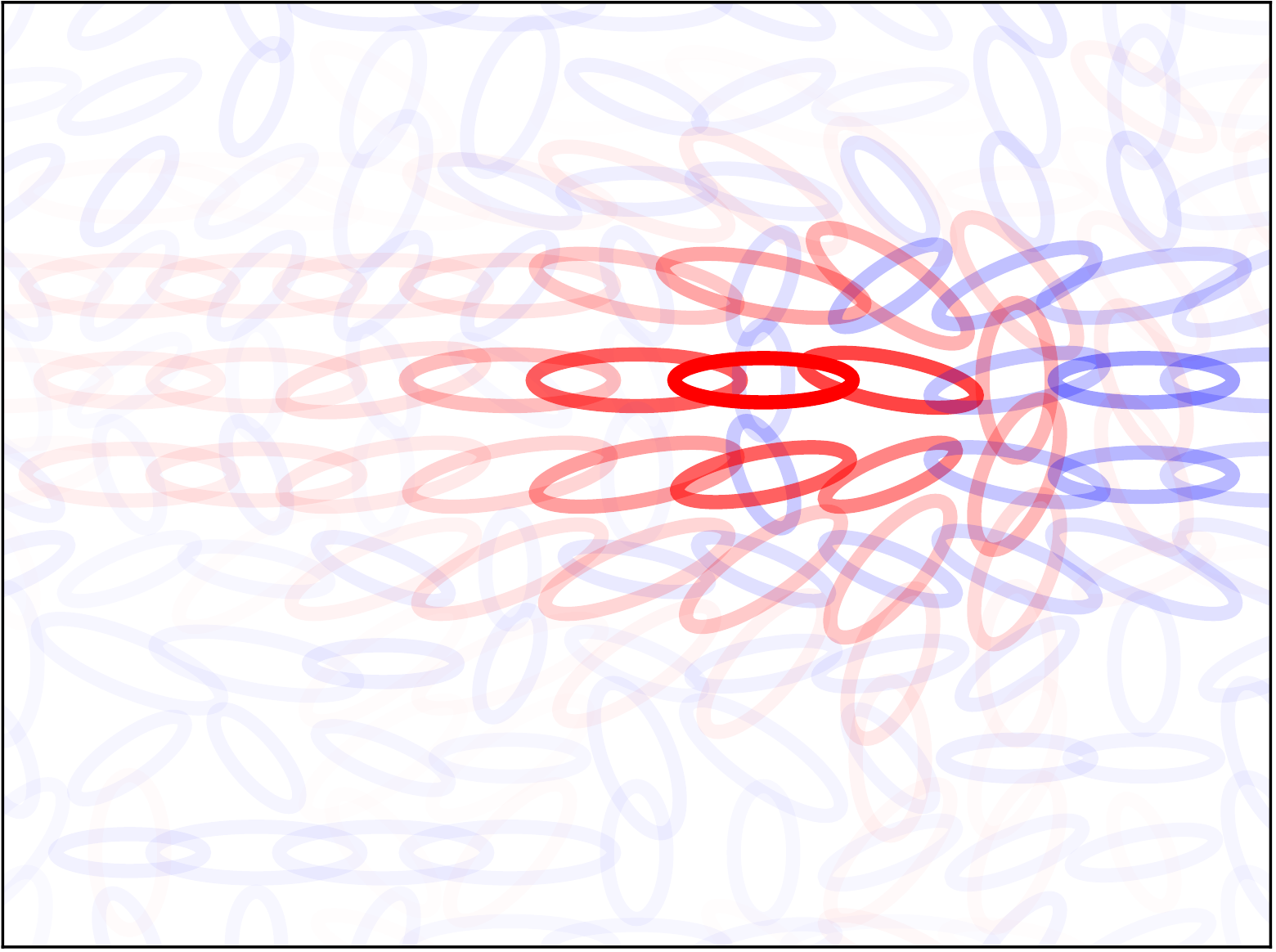}
	\includegraphics[width=0.19\linewidth]{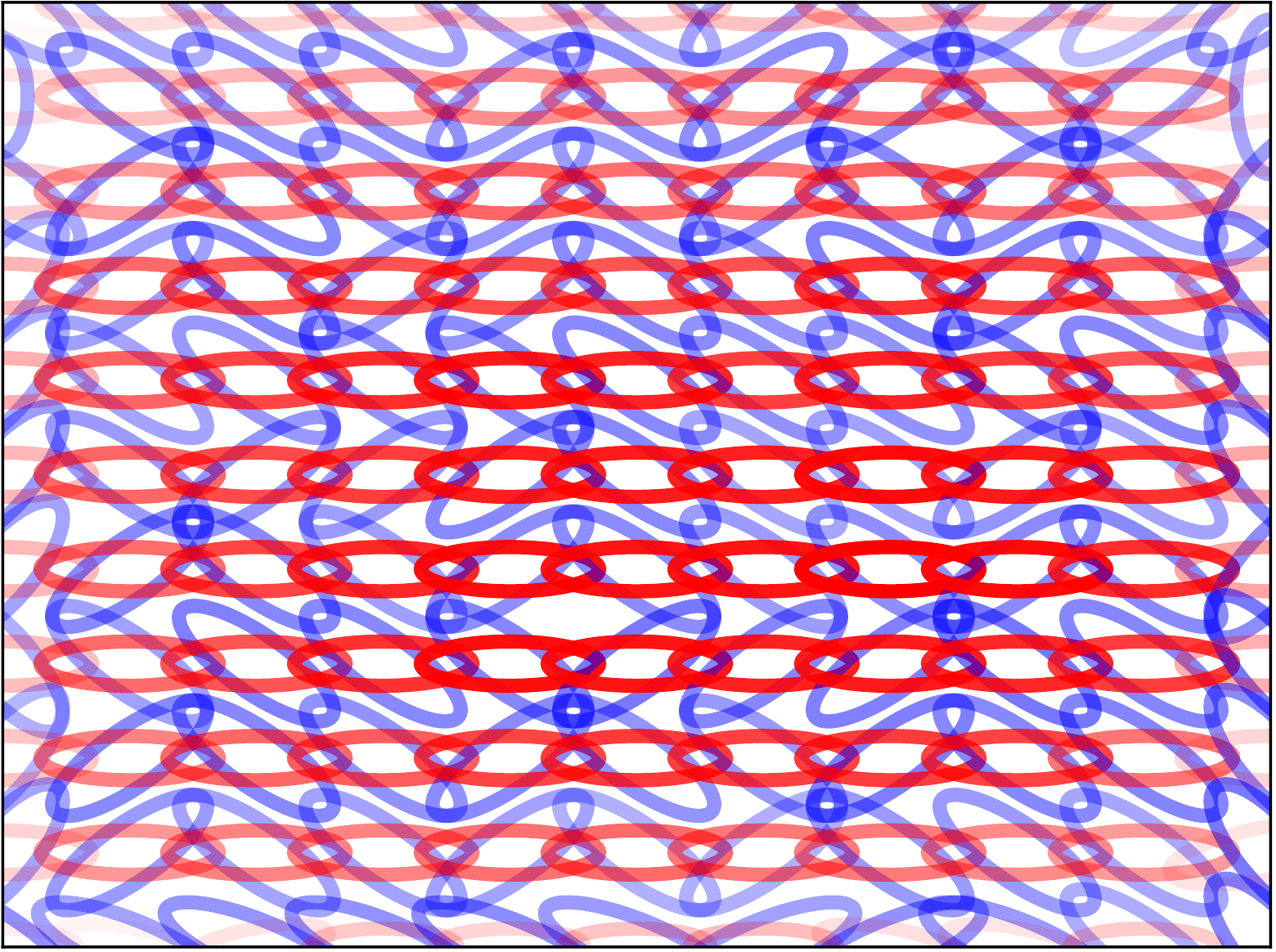}
	\includegraphics[width=0.19\linewidth]{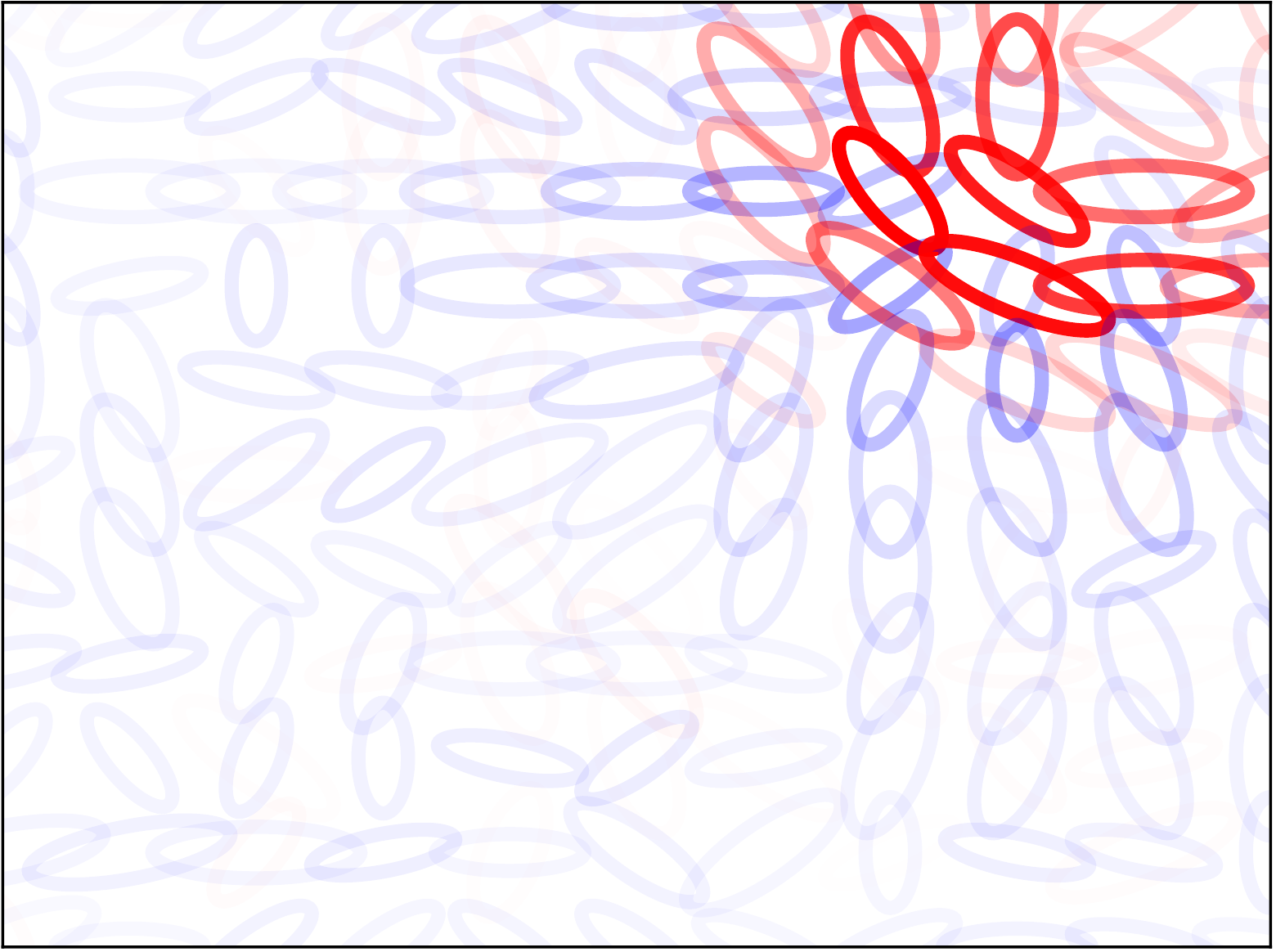}
	\includegraphics[width=0.19\linewidth]{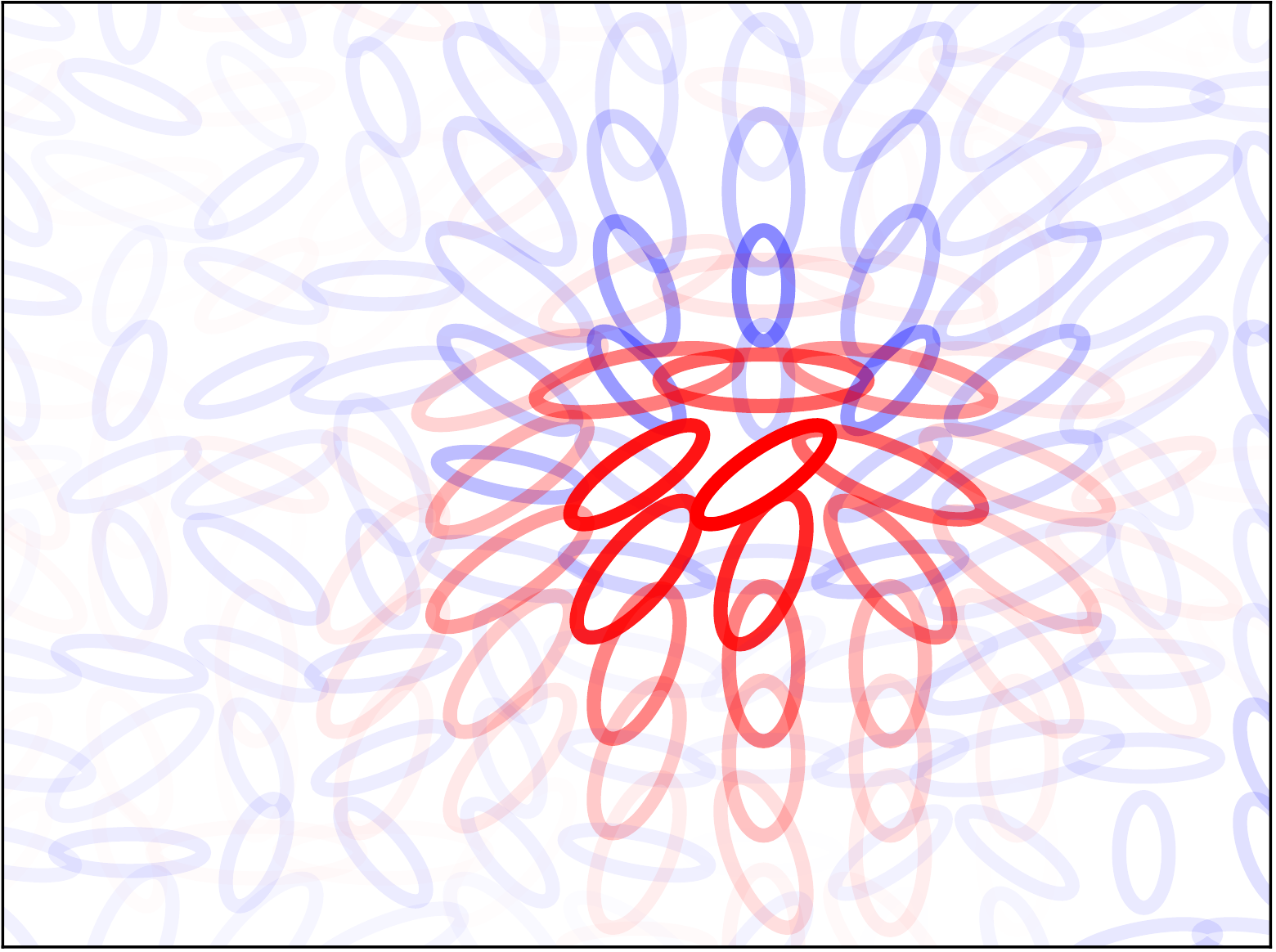}
	\includegraphics[width=0.19\linewidth]{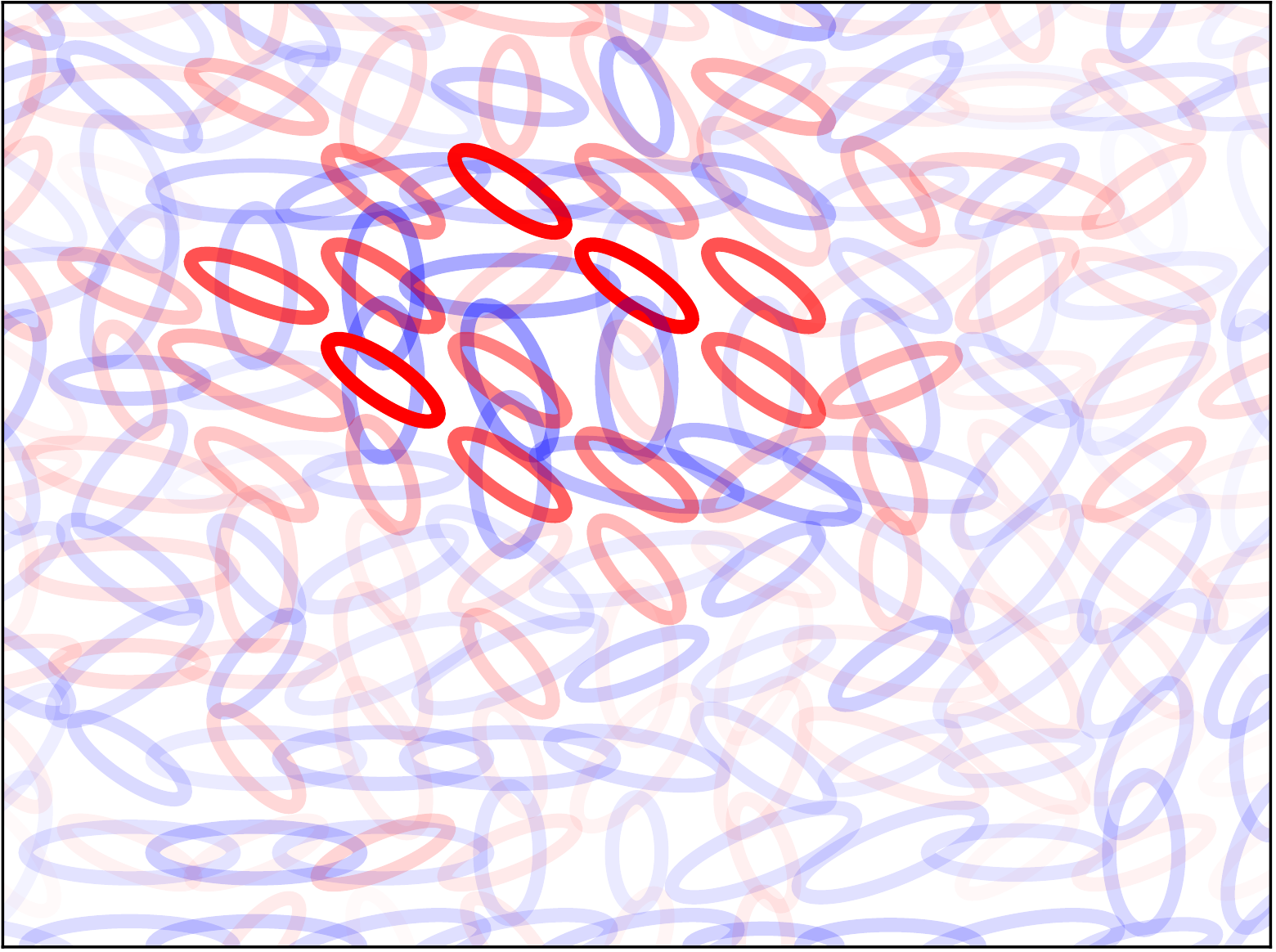}
	\includegraphics[width=0.19\linewidth]{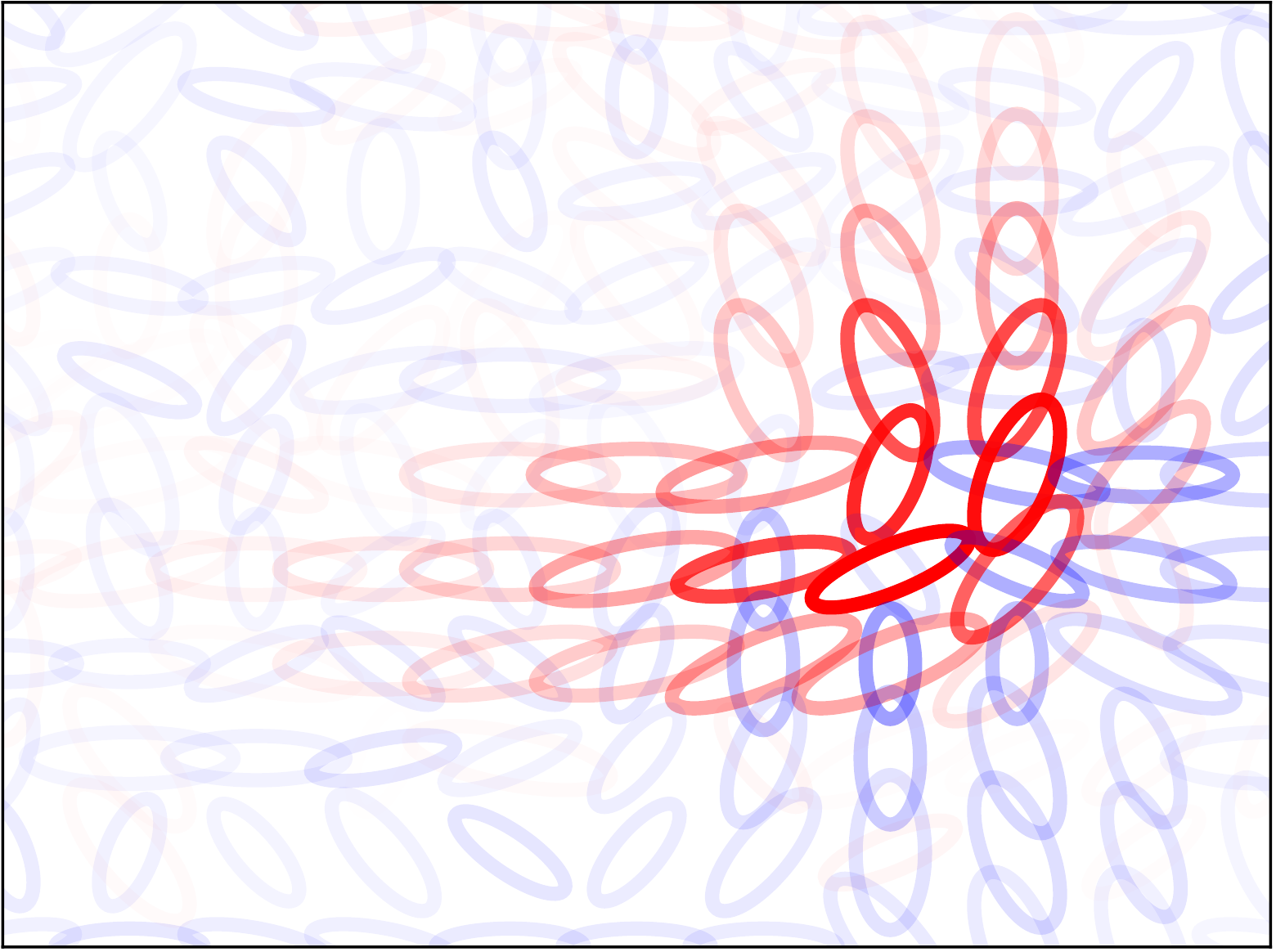}
	\includegraphics[width=0.19\linewidth]{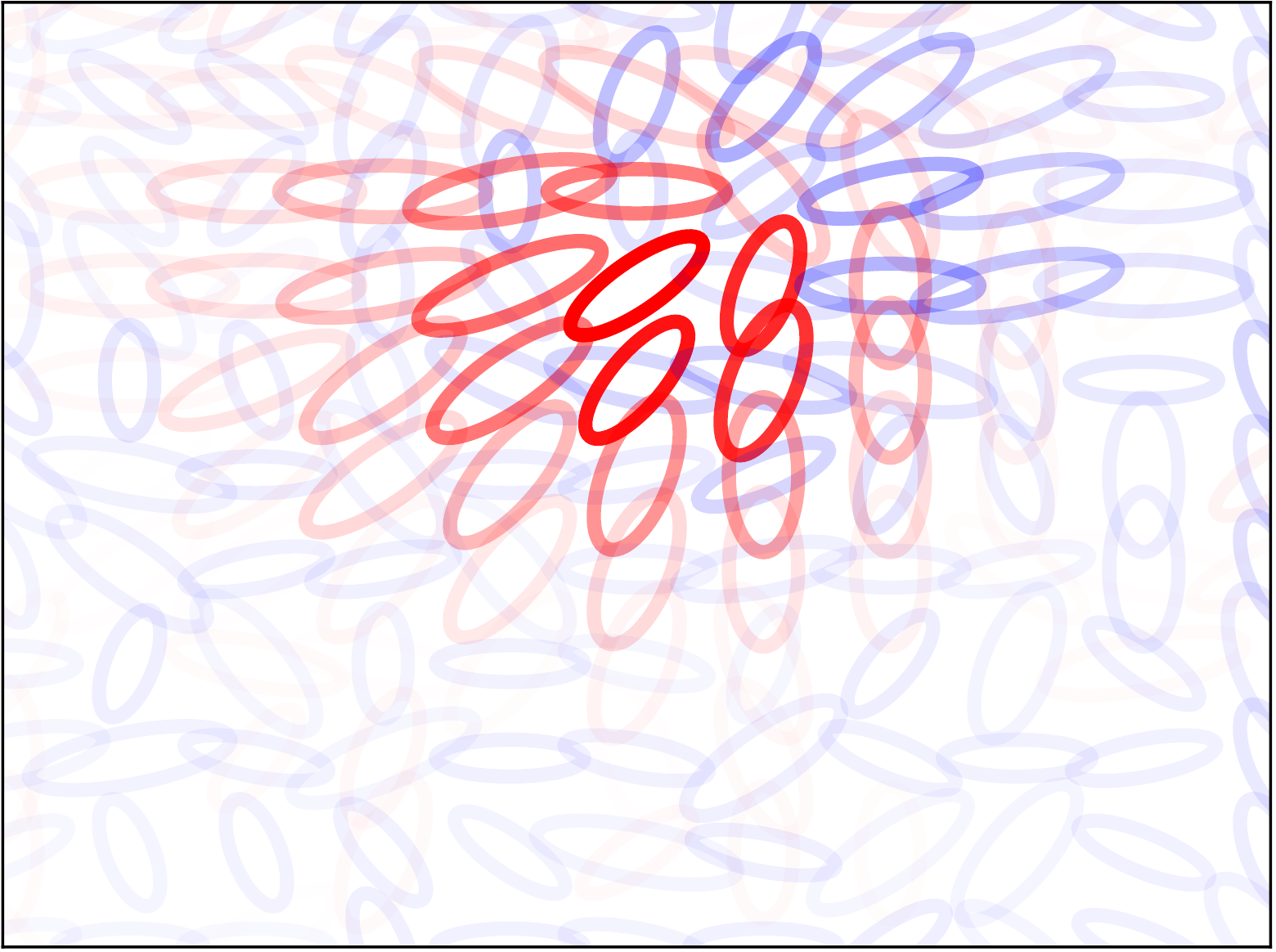}
	\includegraphics[width=0.19\linewidth]{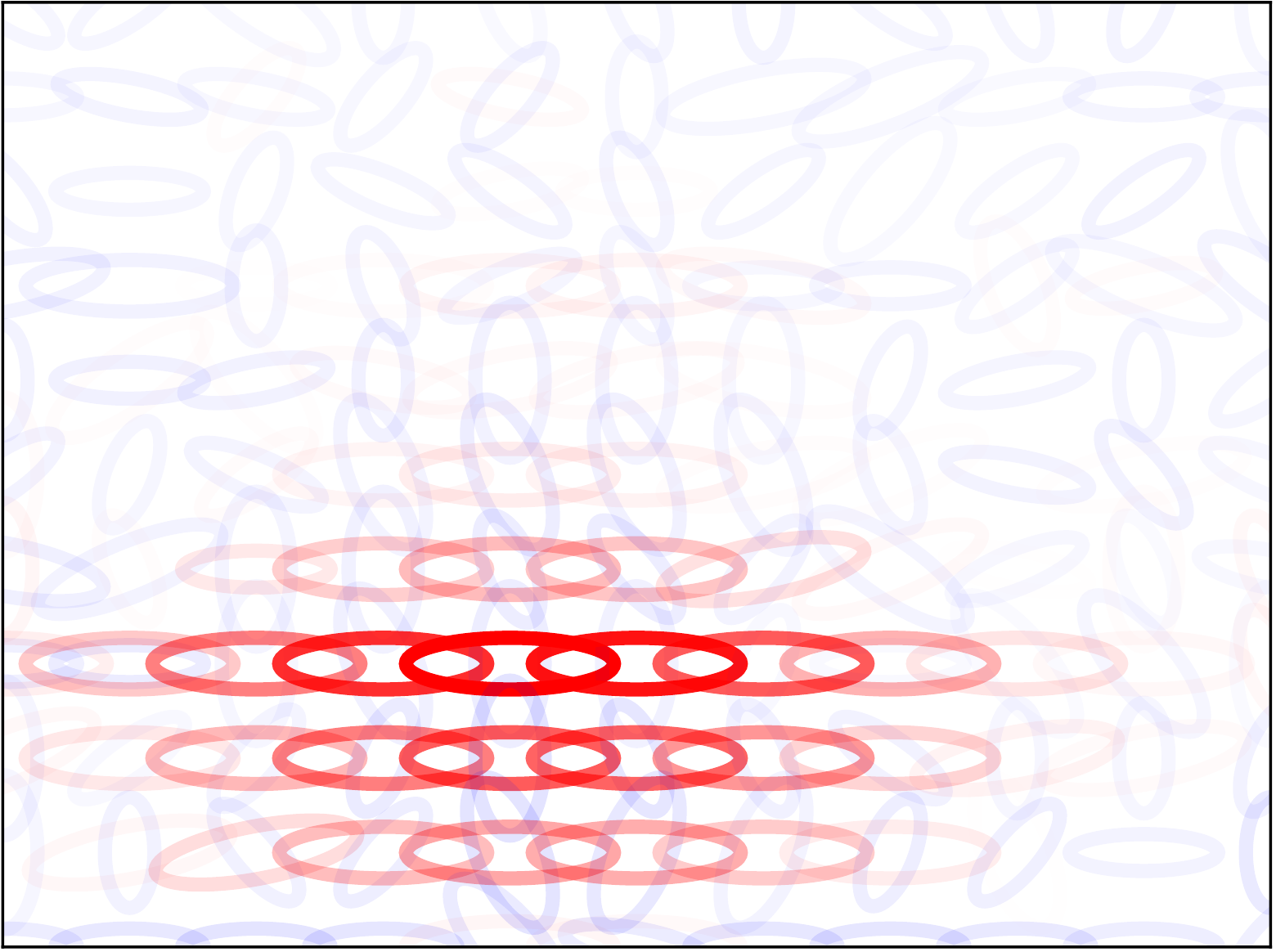}
	\includegraphics[width=0.19\linewidth]{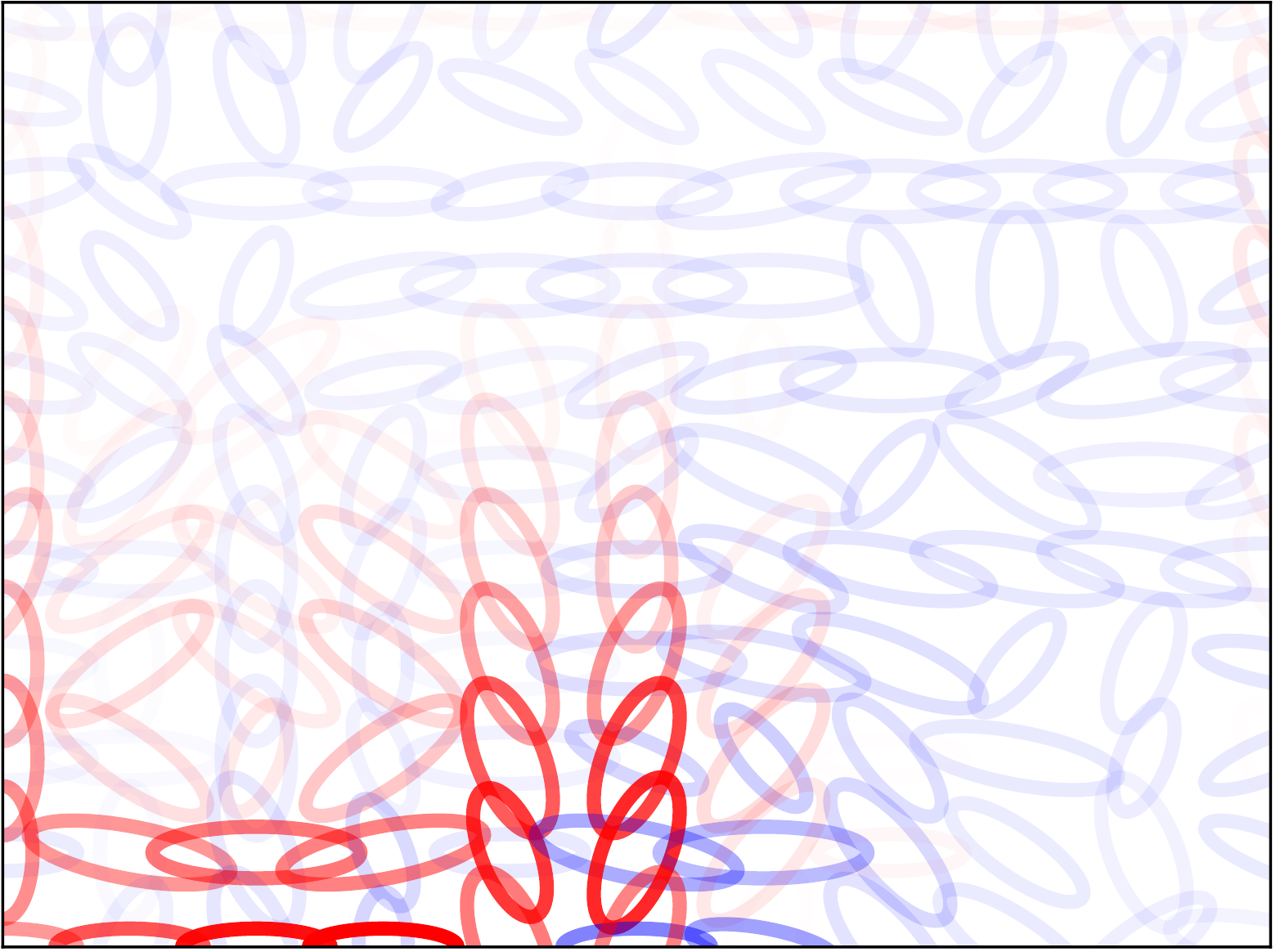}
	\includegraphics[width=0.19\linewidth]{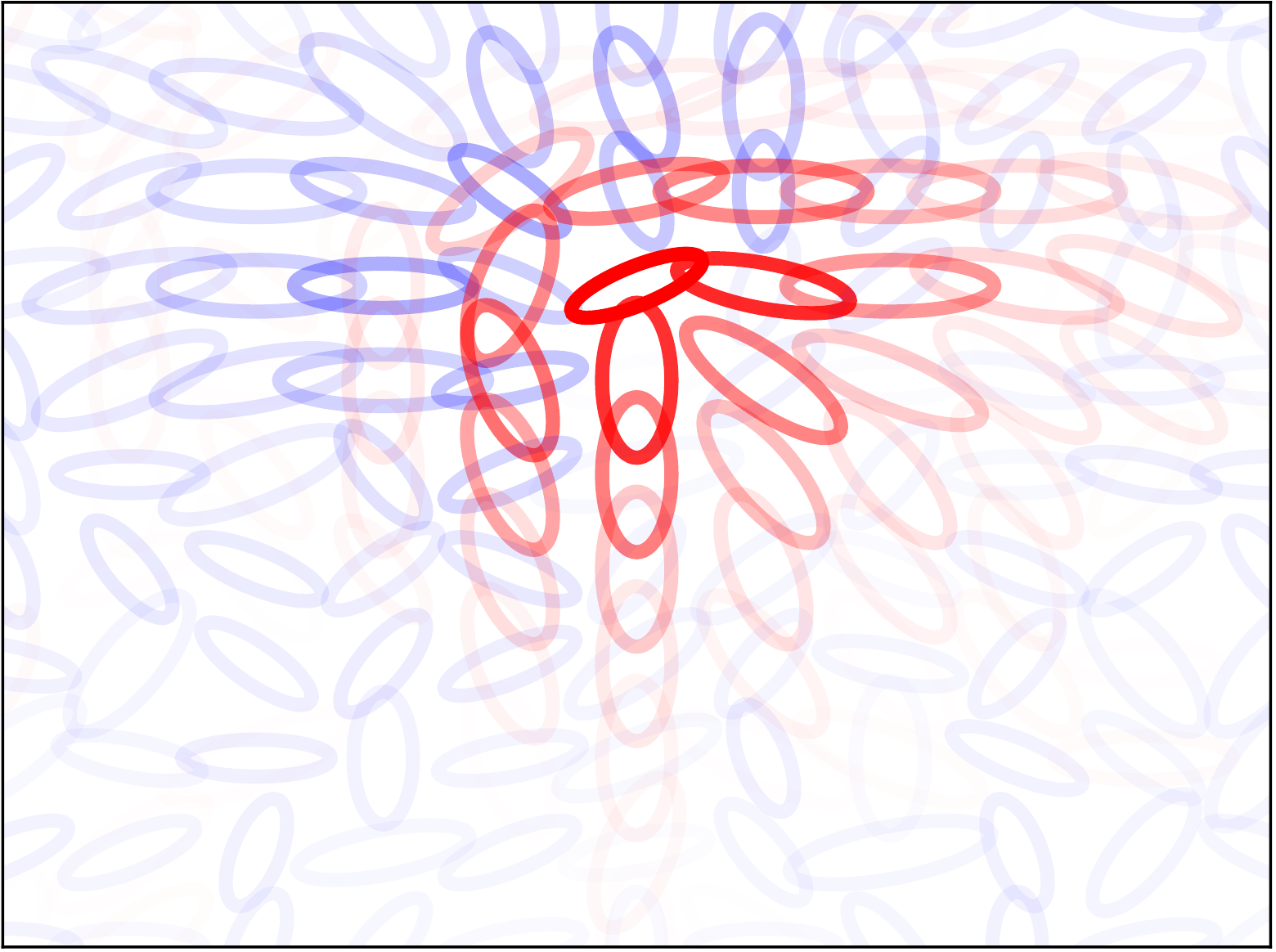} \\
	\phantomsubcaption
	\label{fig:11x11gpsa}
\end{subfigure}
\Large \textbf{(b)} \\
\begin{subfigure}[t]{\linewidth}
	\centering
	\includegraphics[width=0.19\linewidth]{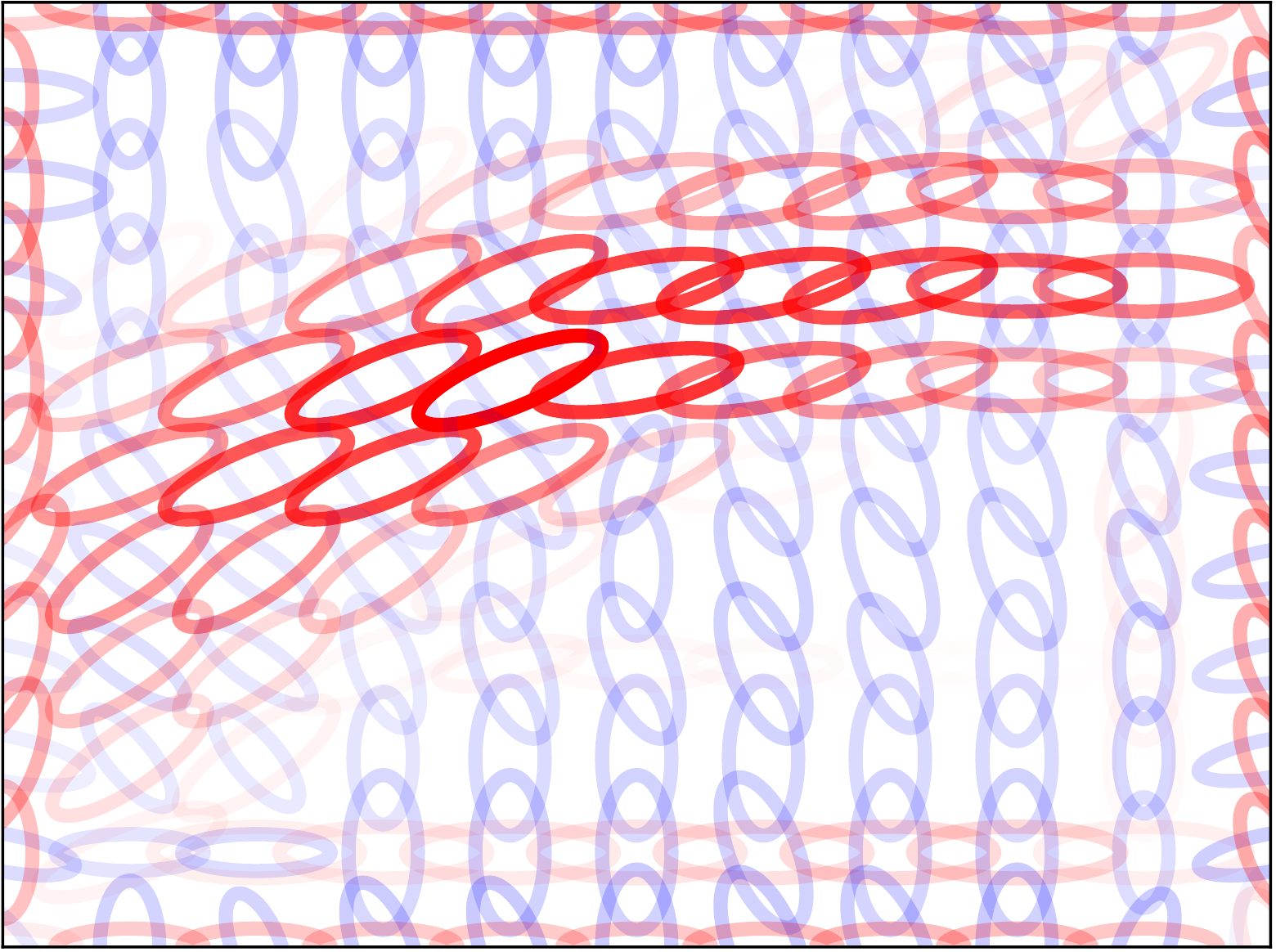}
	\includegraphics[width=0.19\linewidth]{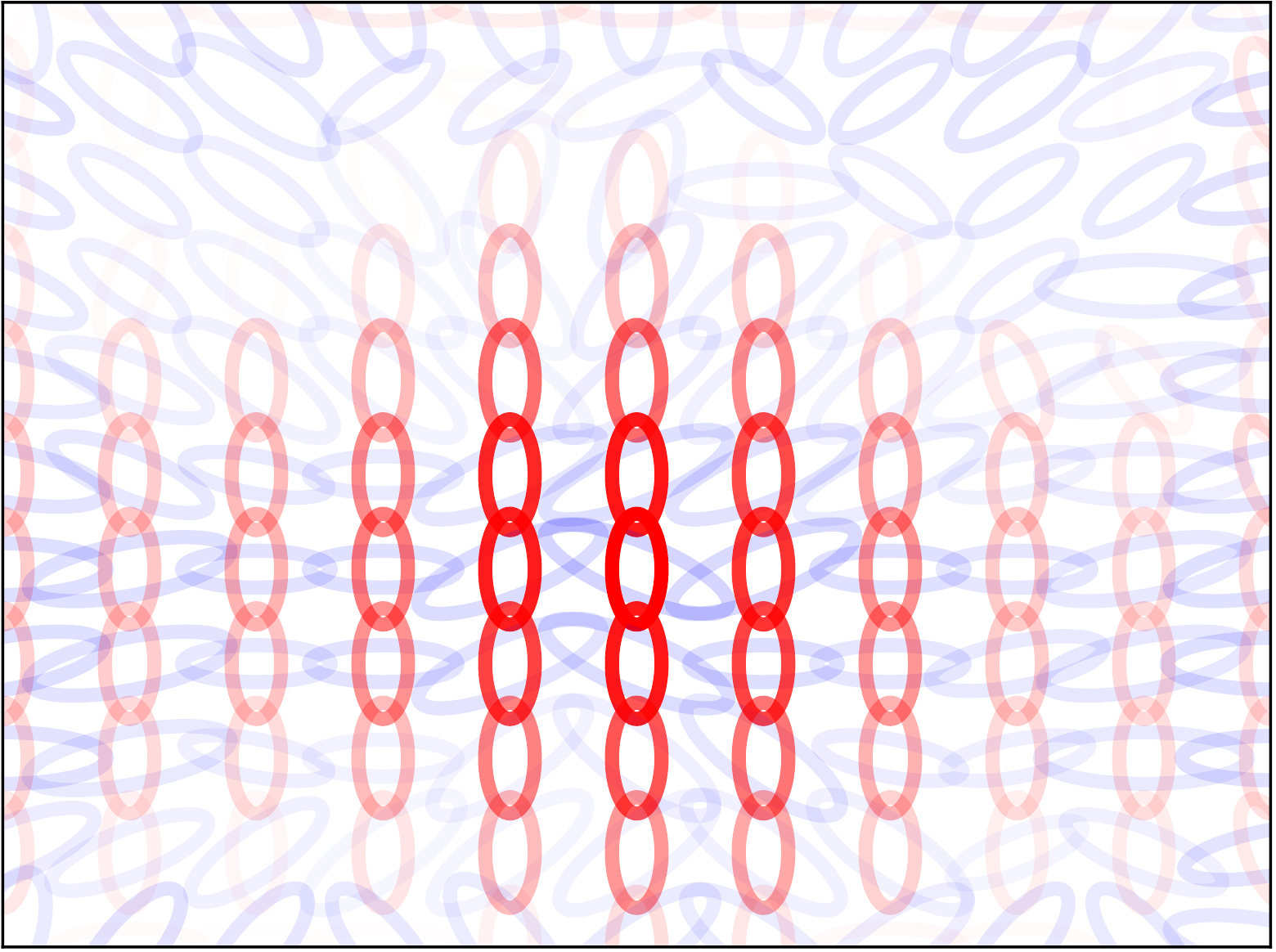}
	\includegraphics[width=0.19\linewidth]{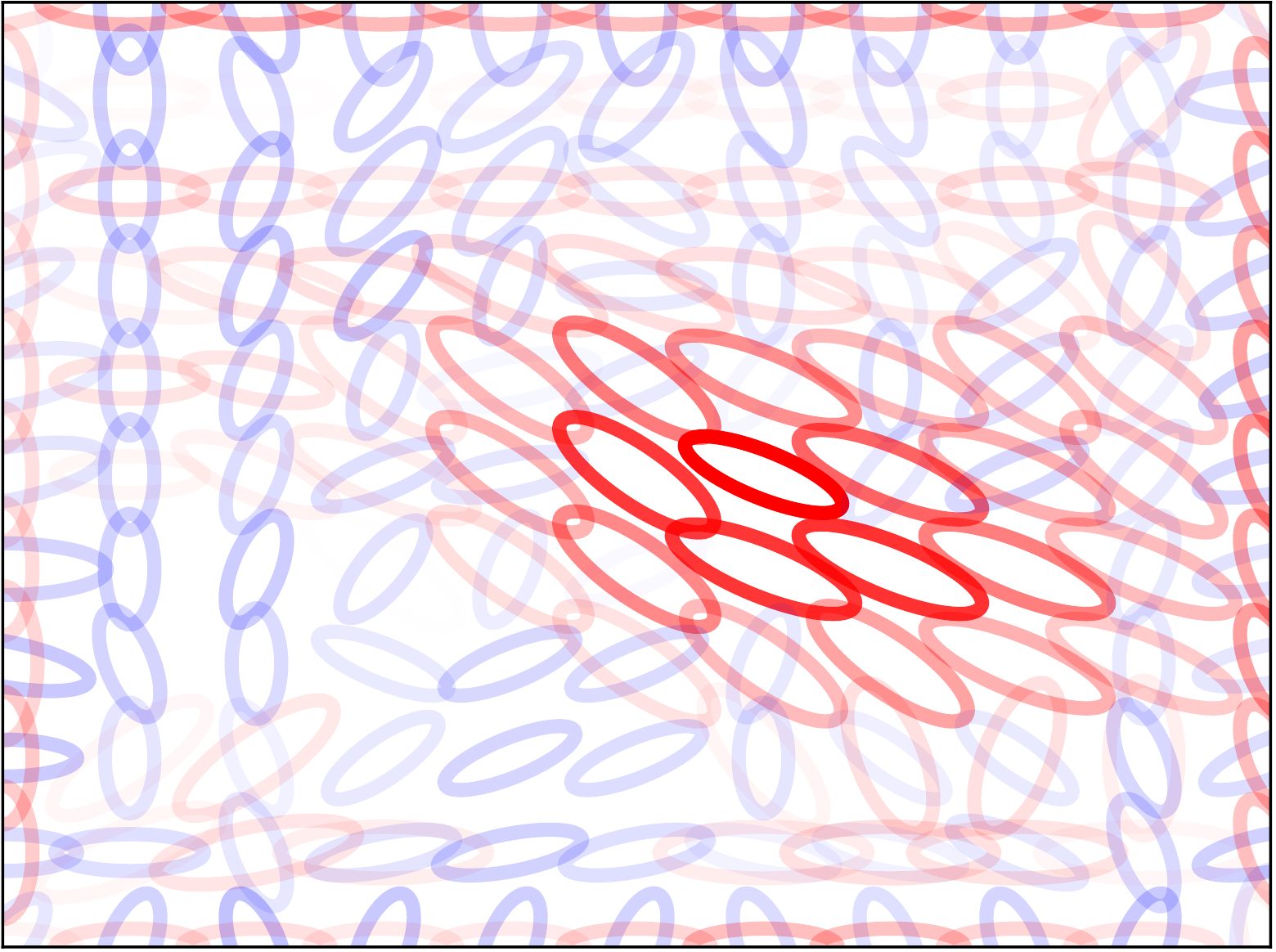}
	\includegraphics[width=0.19\linewidth]{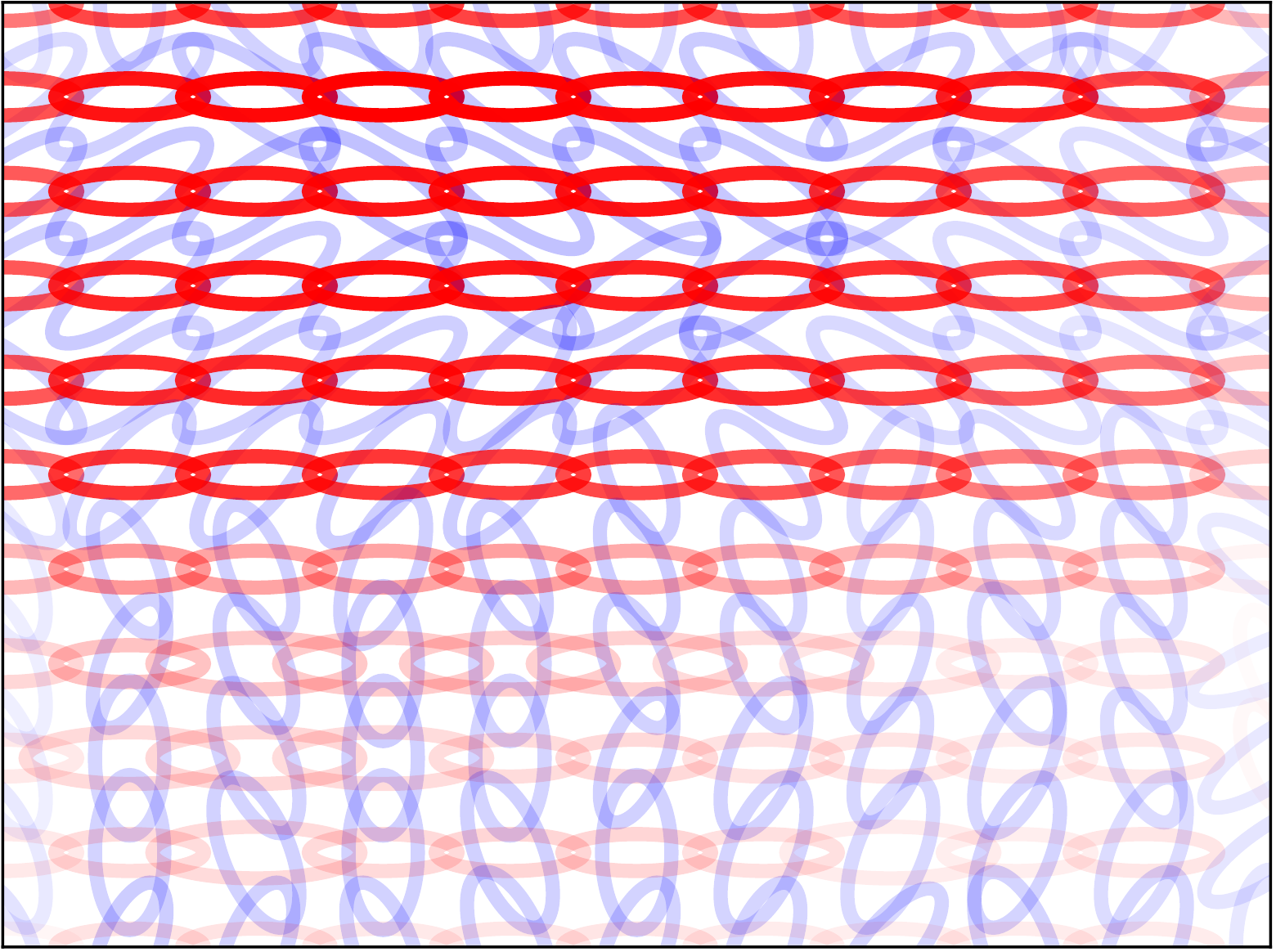}
	\includegraphics[width=0.19\linewidth]{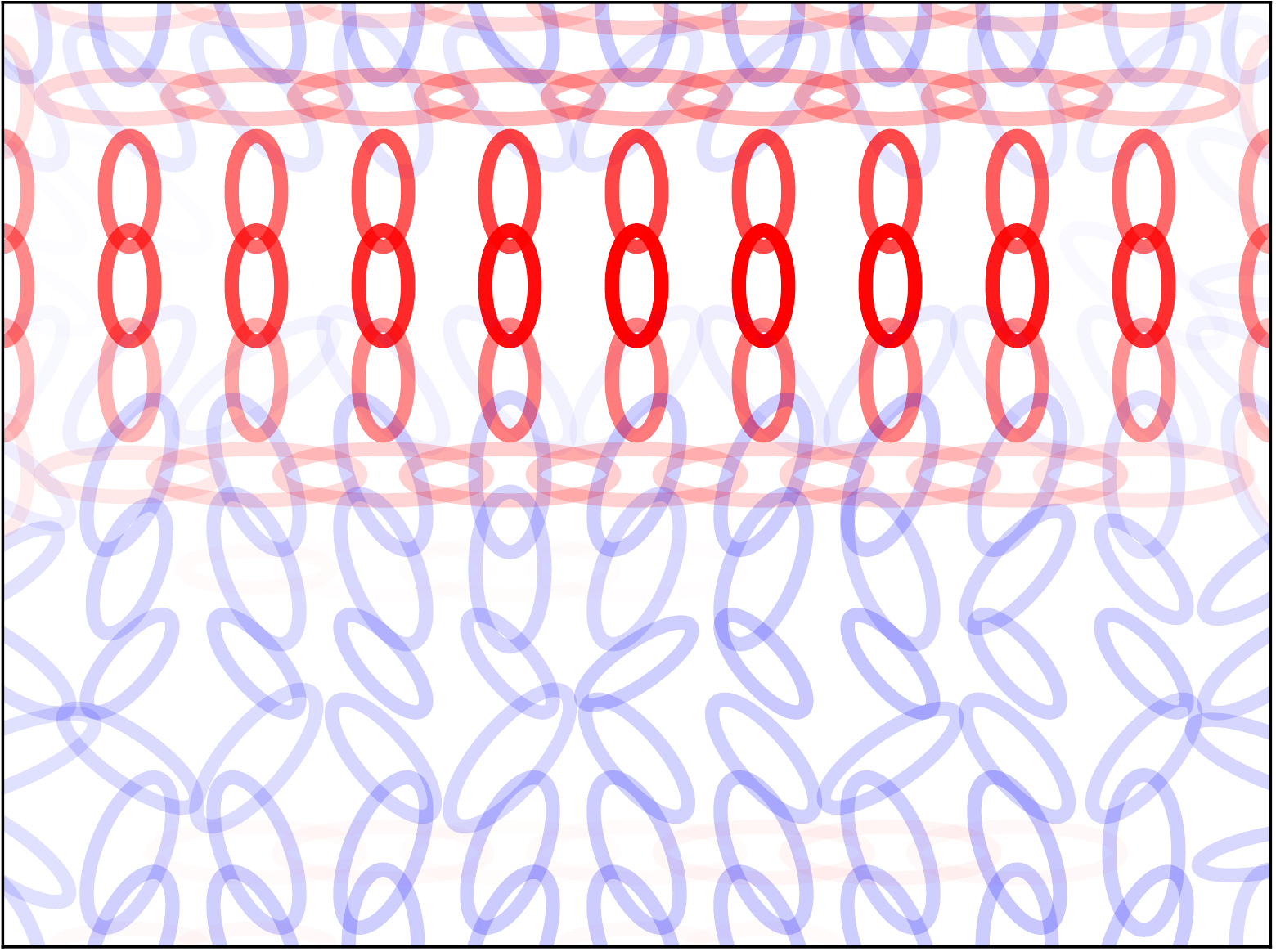}
	\includegraphics[width=0.19\linewidth]{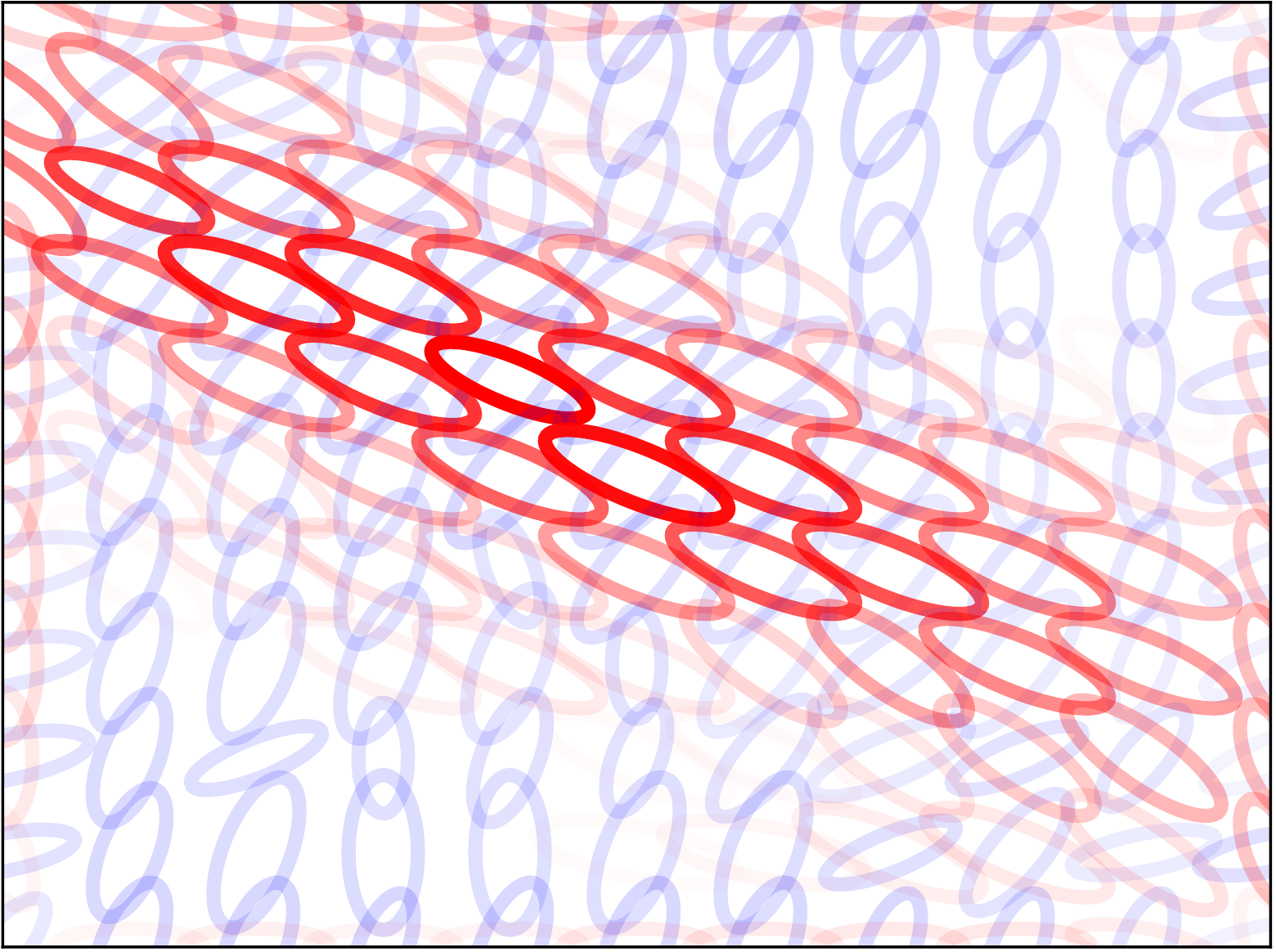}
	\includegraphics[width=0.19\linewidth]{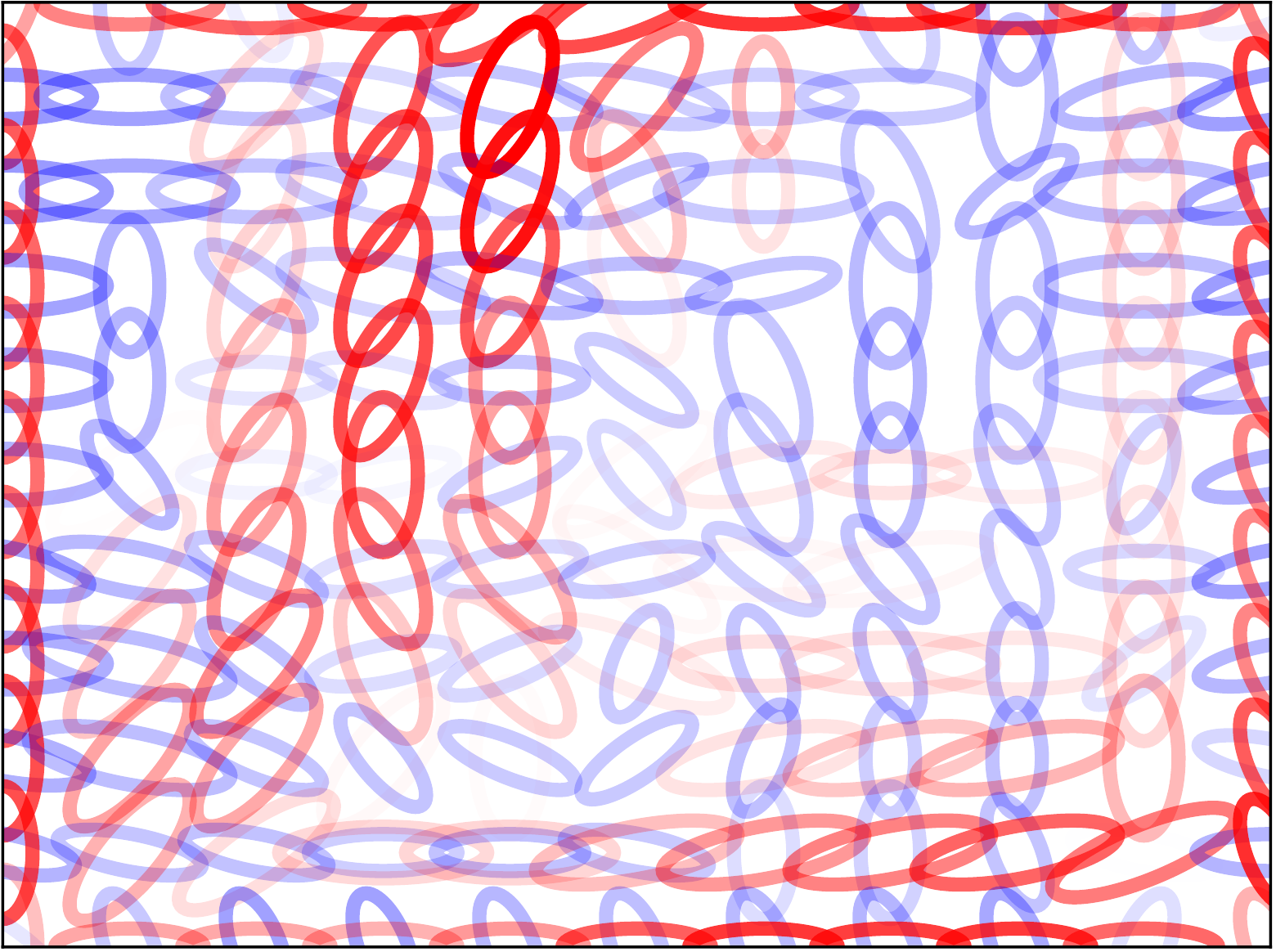}
	\includegraphics[width=0.19\linewidth]{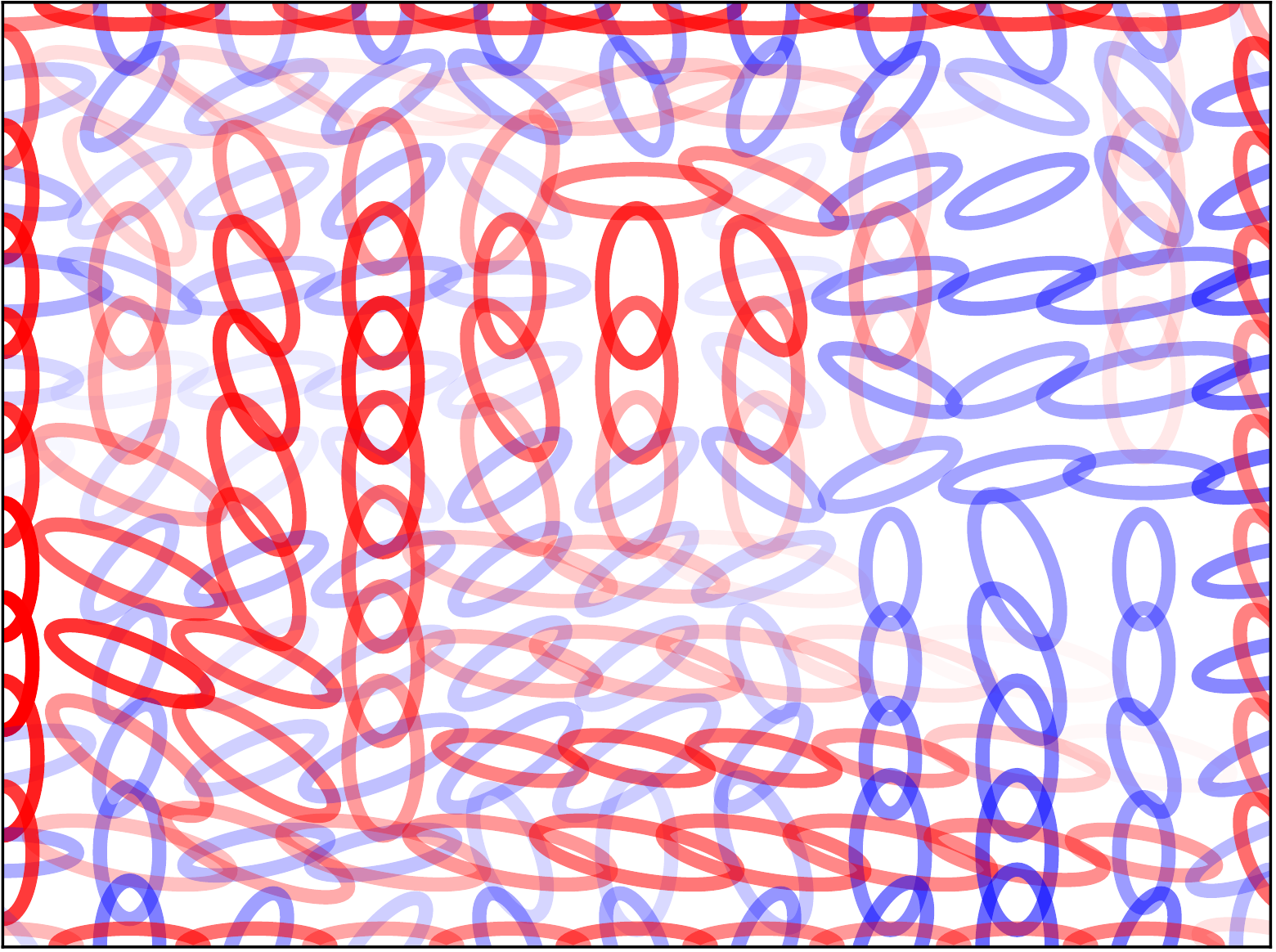}
	\includegraphics[width=0.19\linewidth]{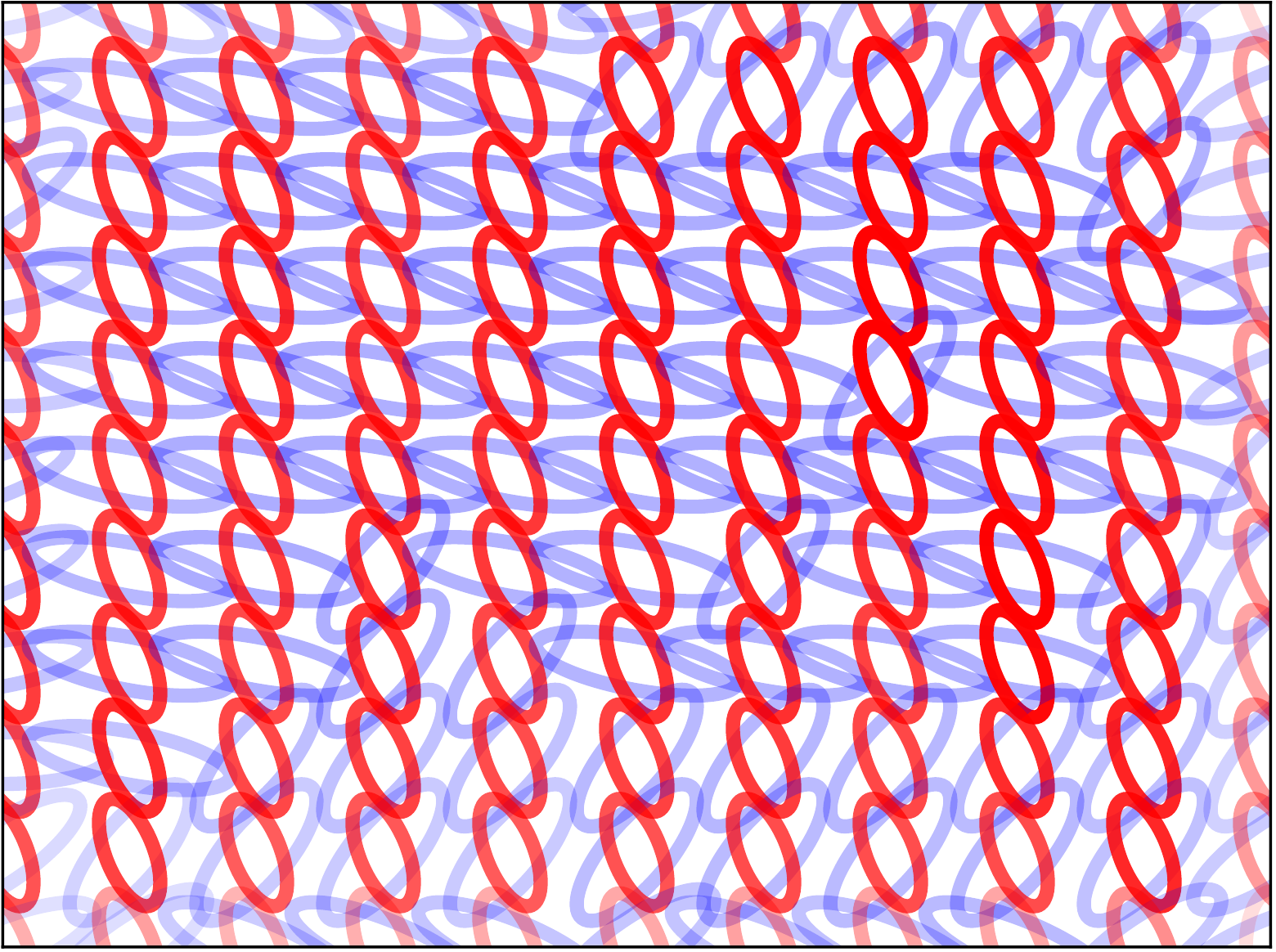}
	\includegraphics[width=0.19\linewidth]{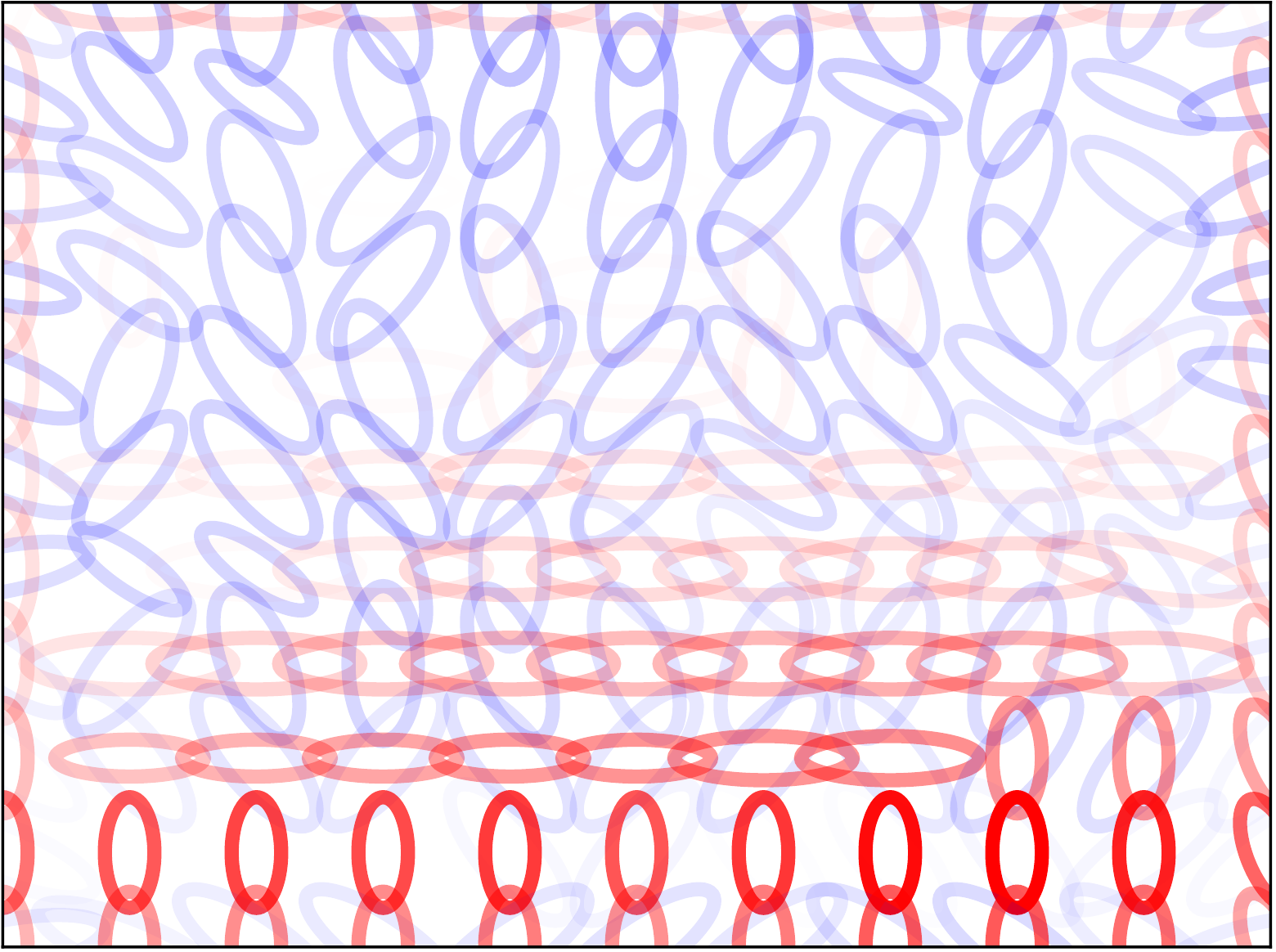}
	\includegraphics[width=0.19\linewidth]{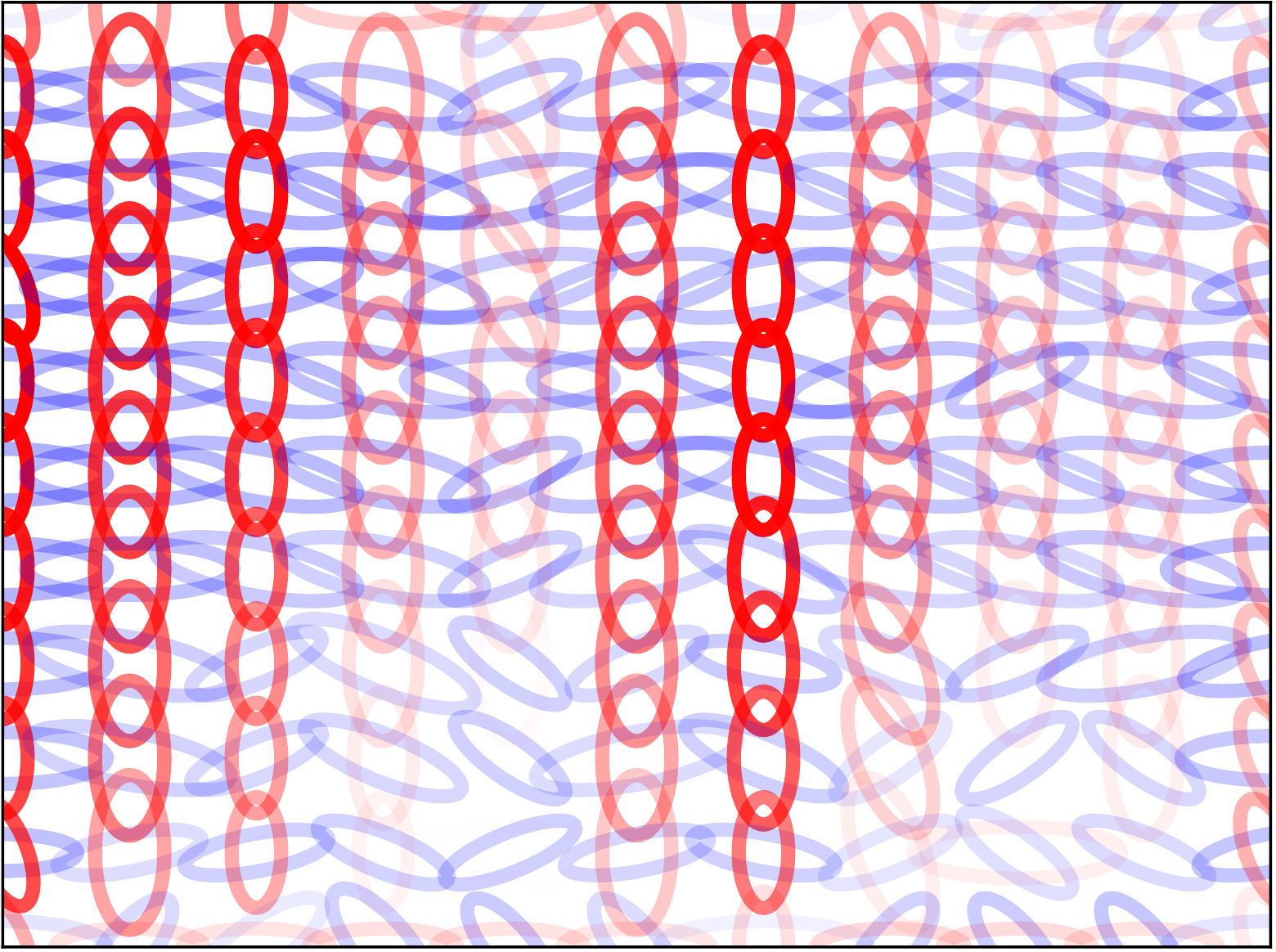}
	\includegraphics[width=0.19\linewidth]{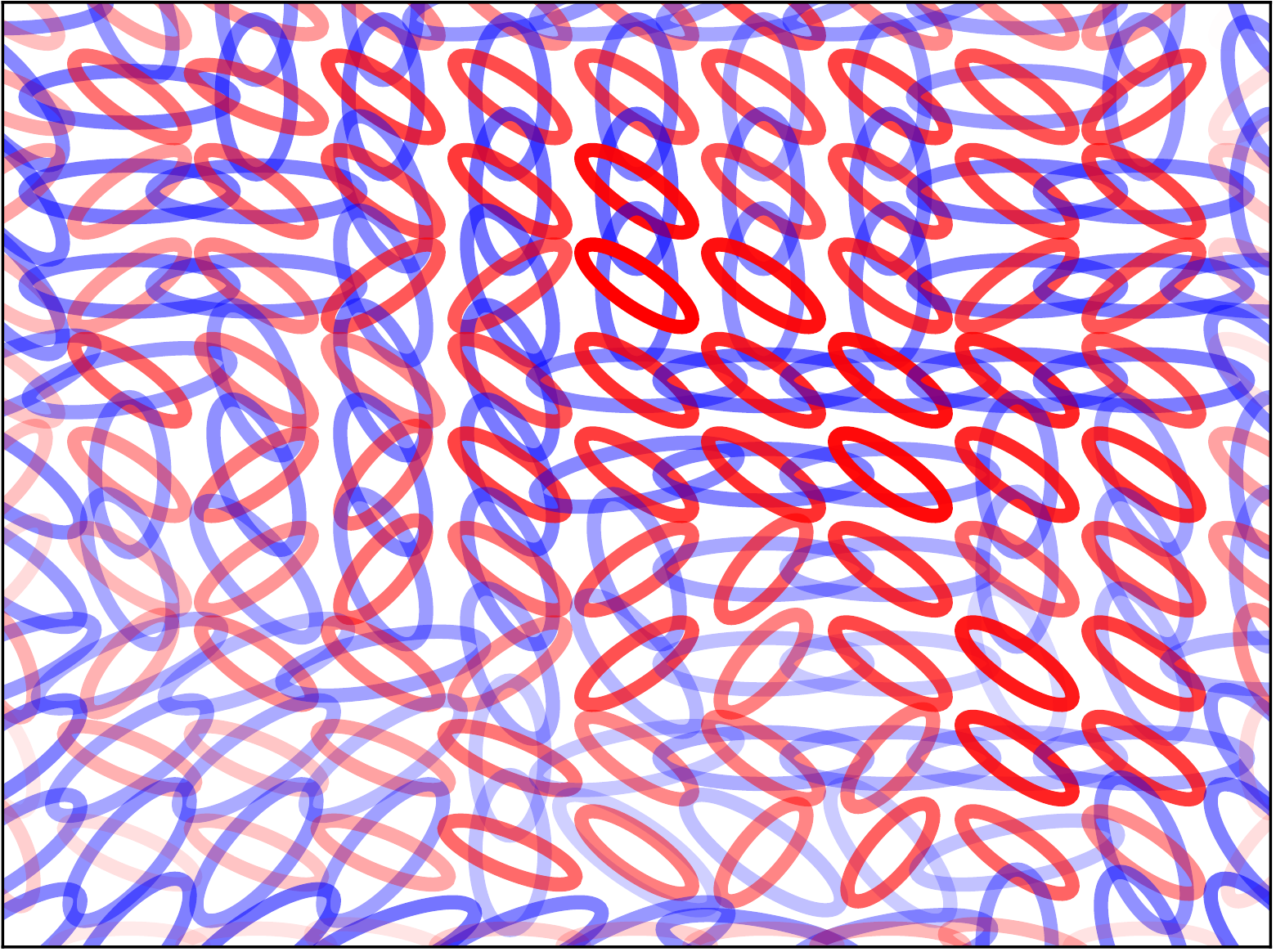}
	\includegraphics[width=0.19\linewidth]{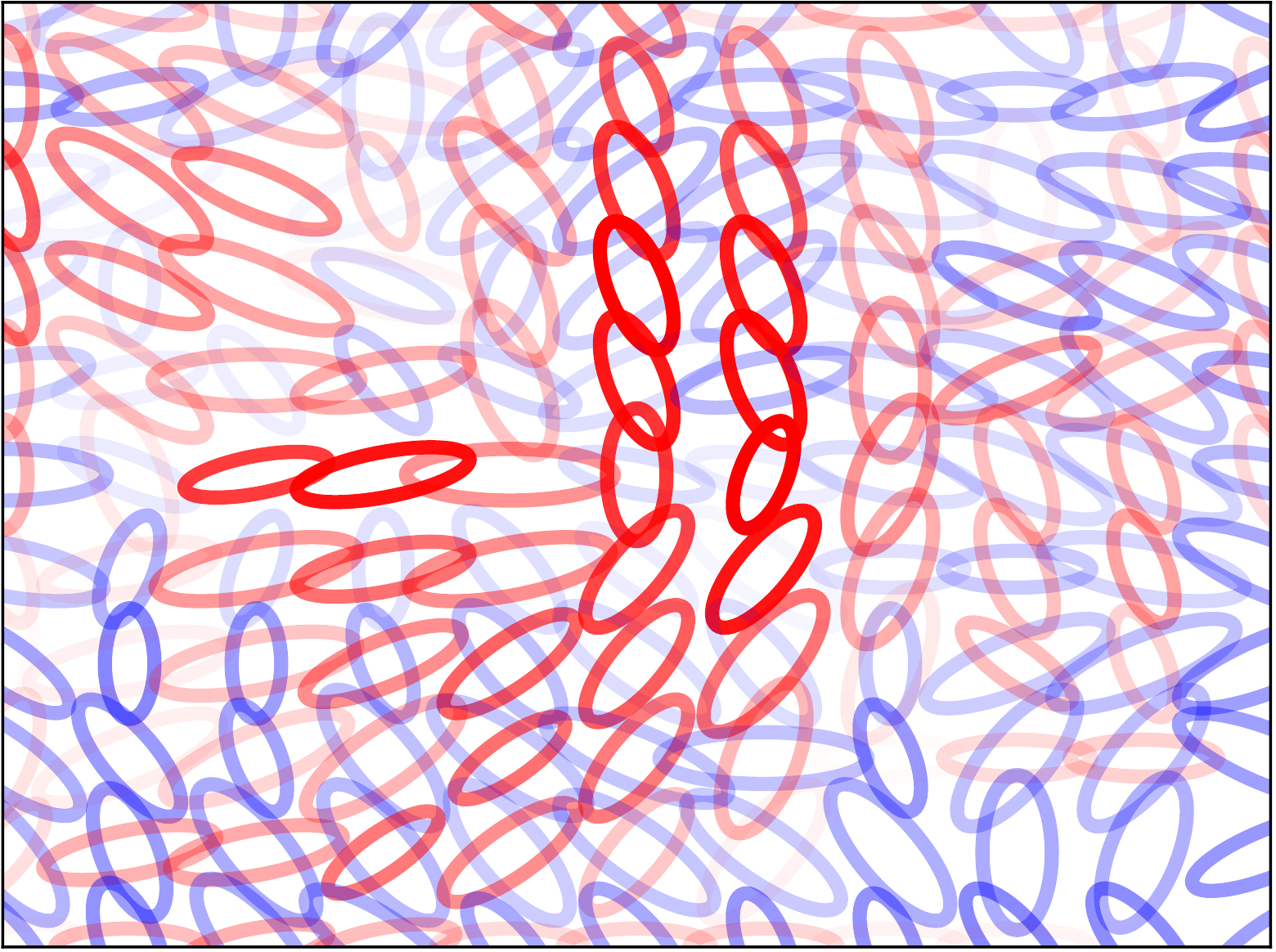}
	\includegraphics[width=0.19\linewidth]{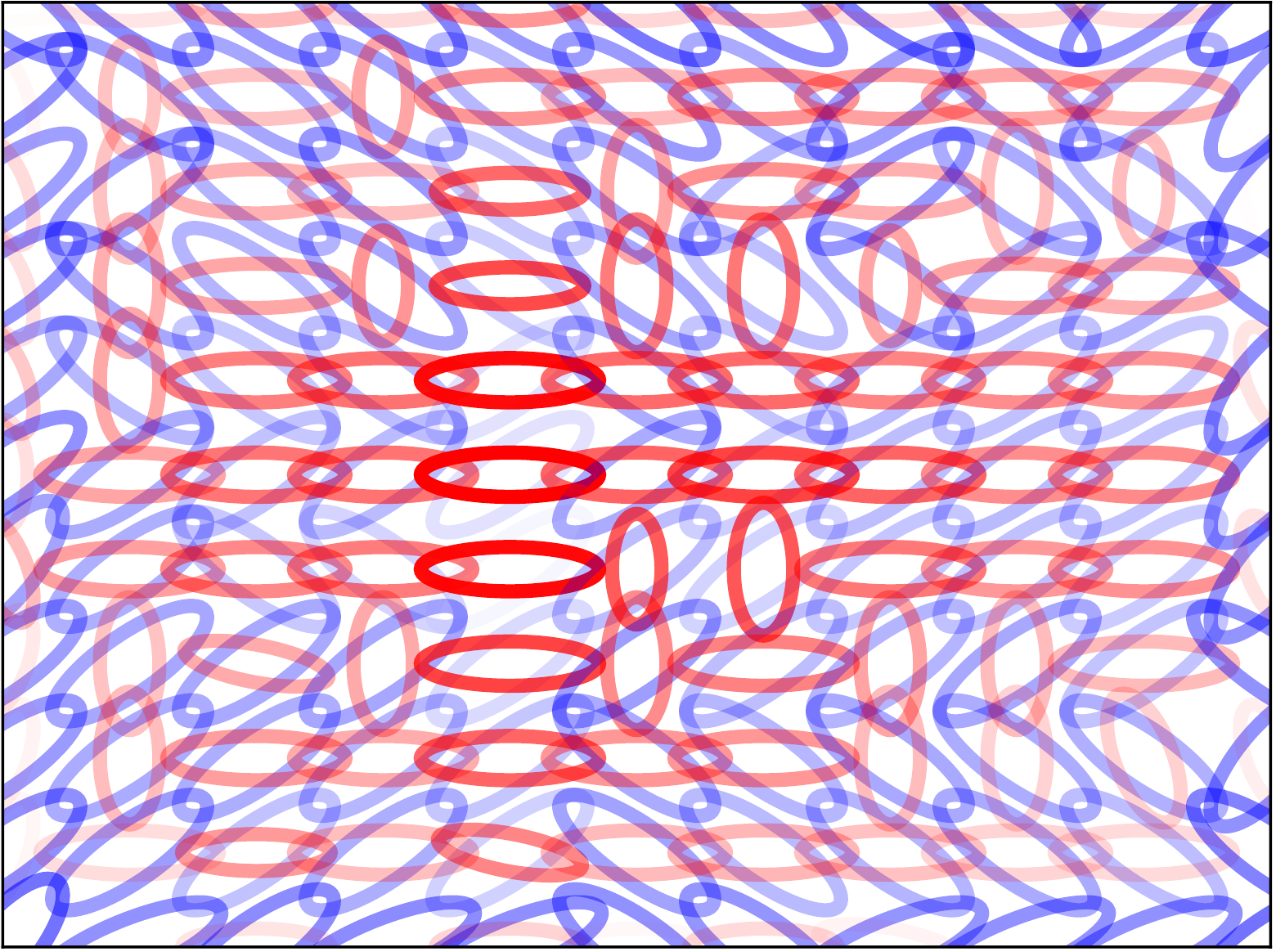}
	\includegraphics[width=0.19\linewidth]{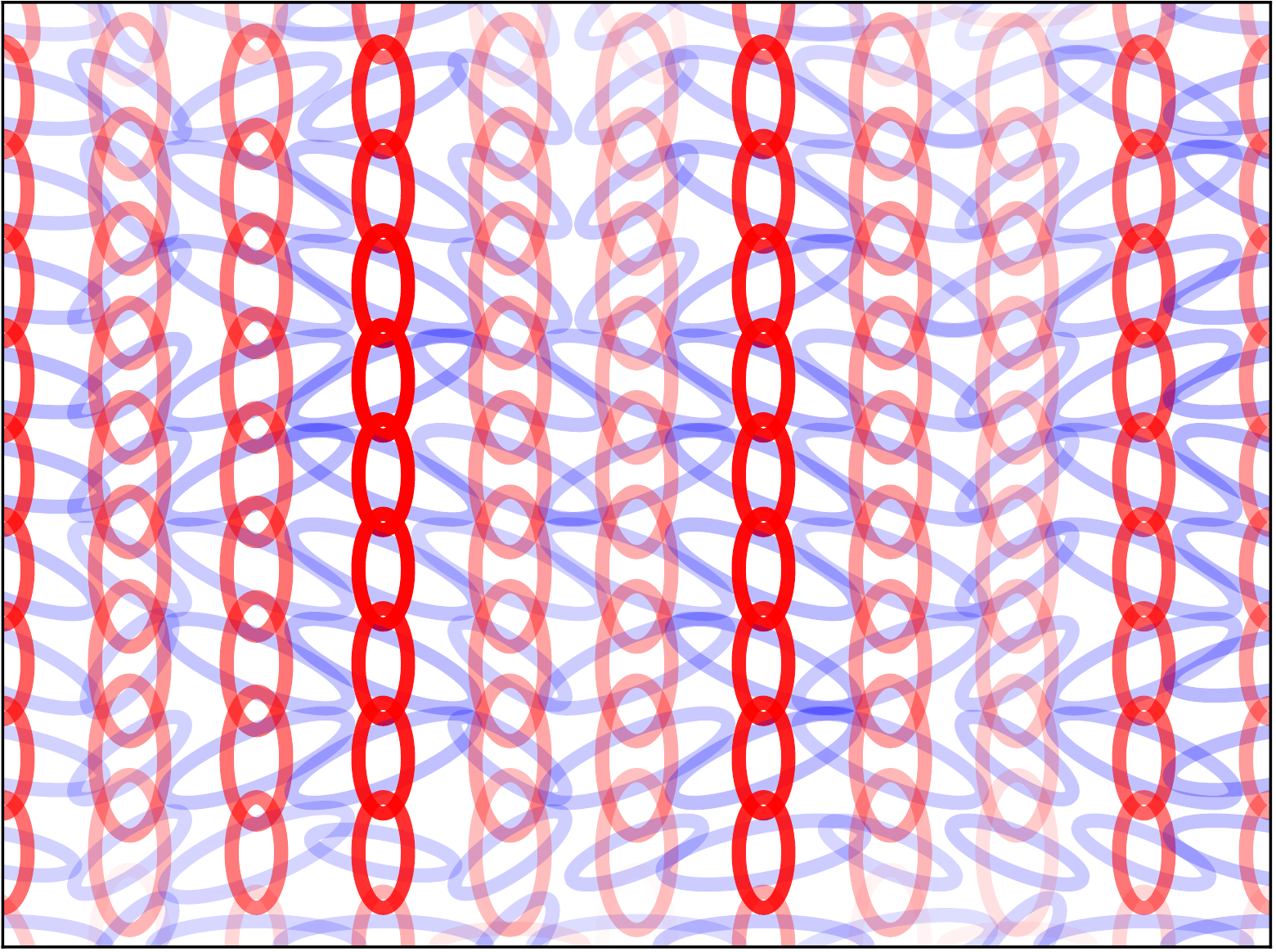}
	\includegraphics[width=0.19\linewidth]{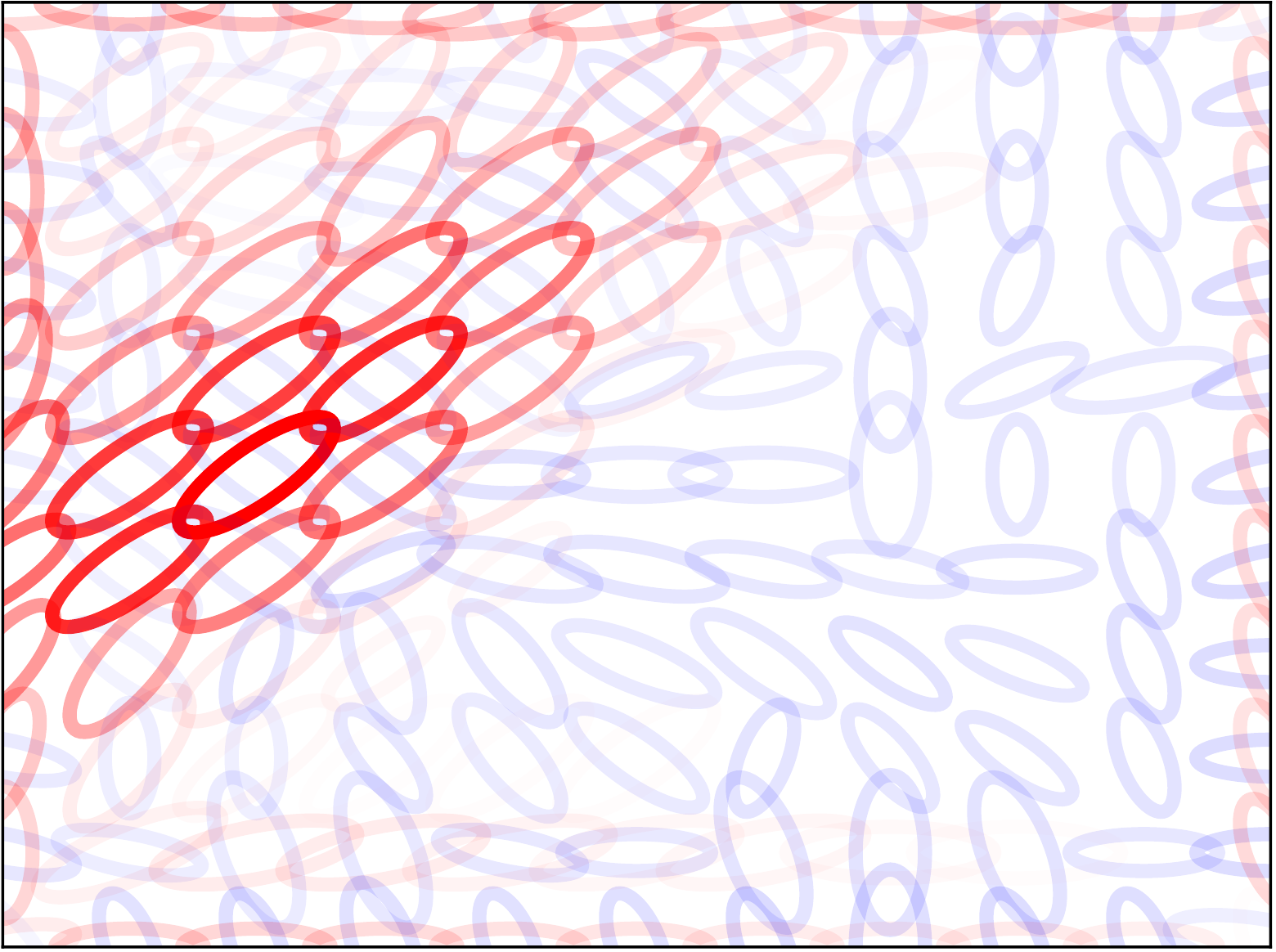}
	\includegraphics[width=0.19\linewidth]{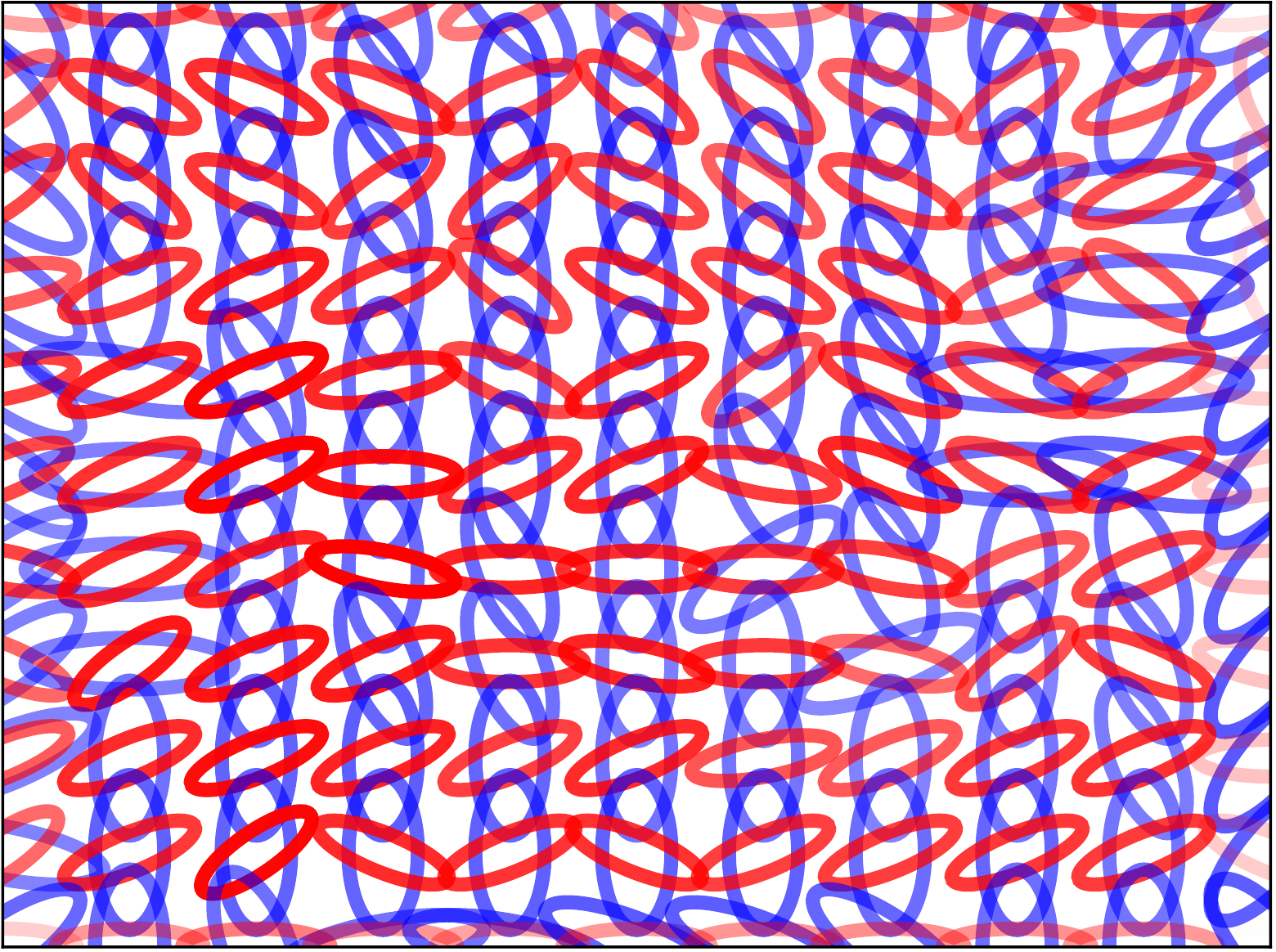}
	\includegraphics[width=0.19\linewidth]{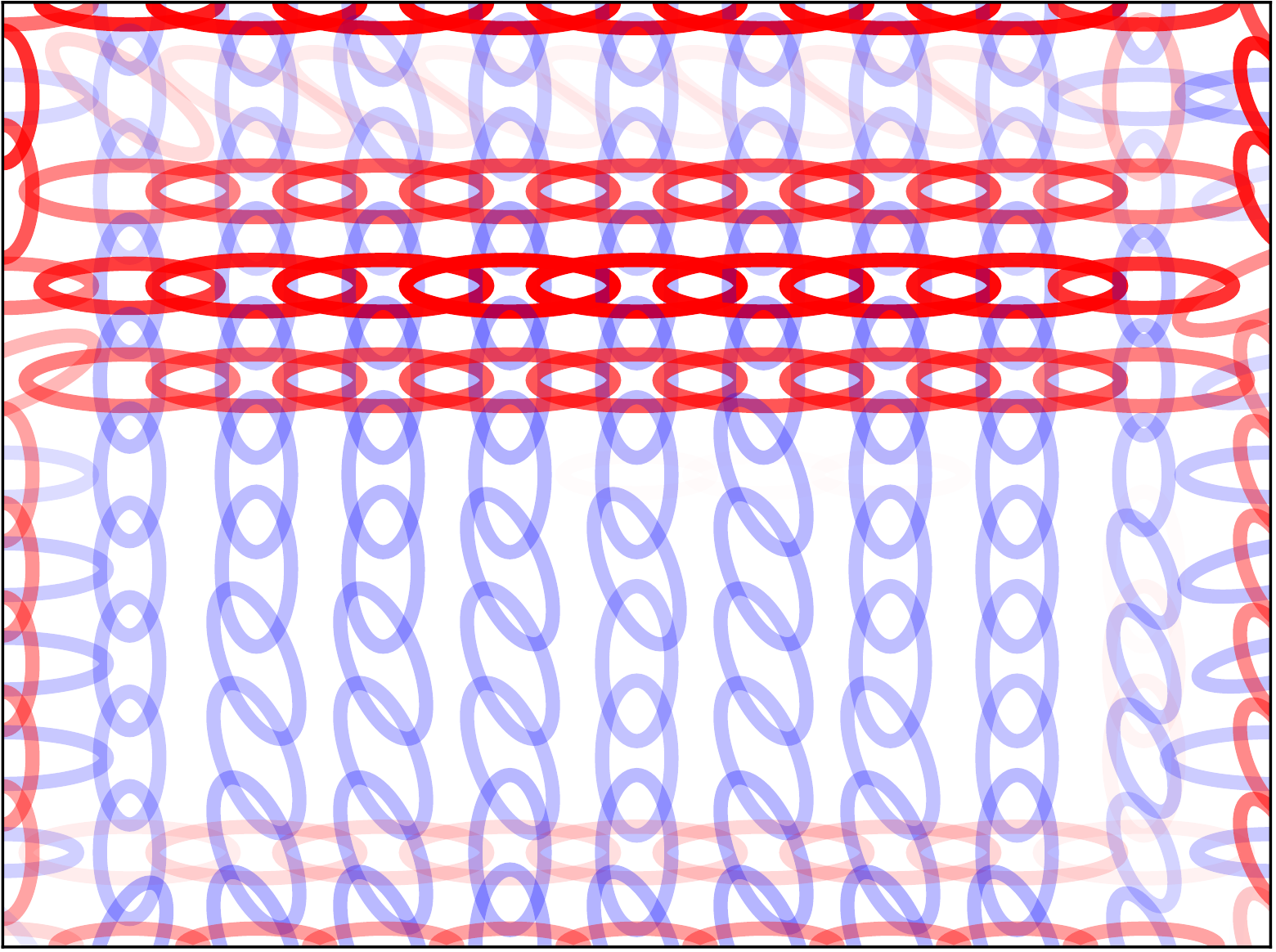}
	\includegraphics[width=0.19\linewidth]{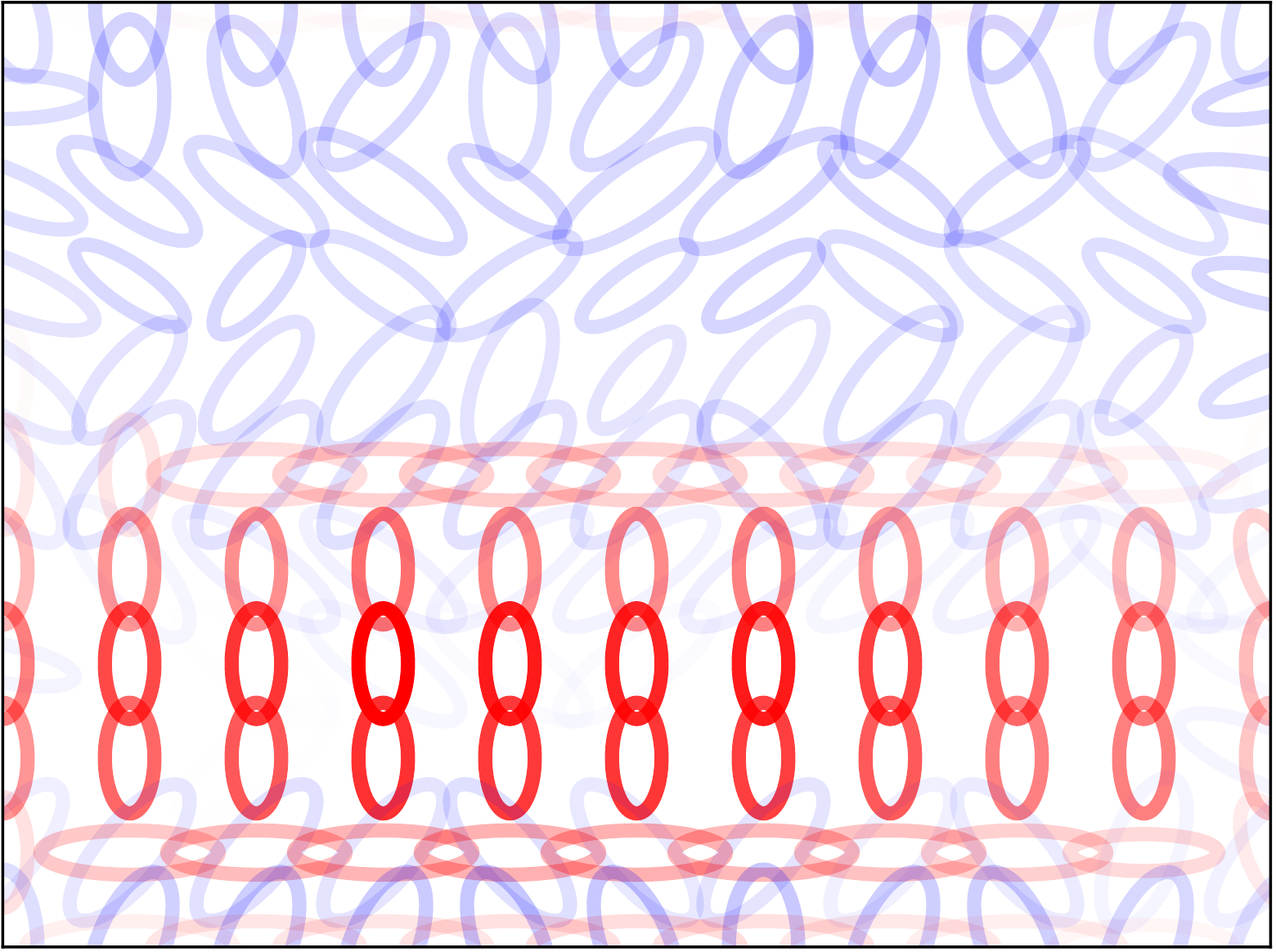}
	\includegraphics[width=0.19\linewidth]{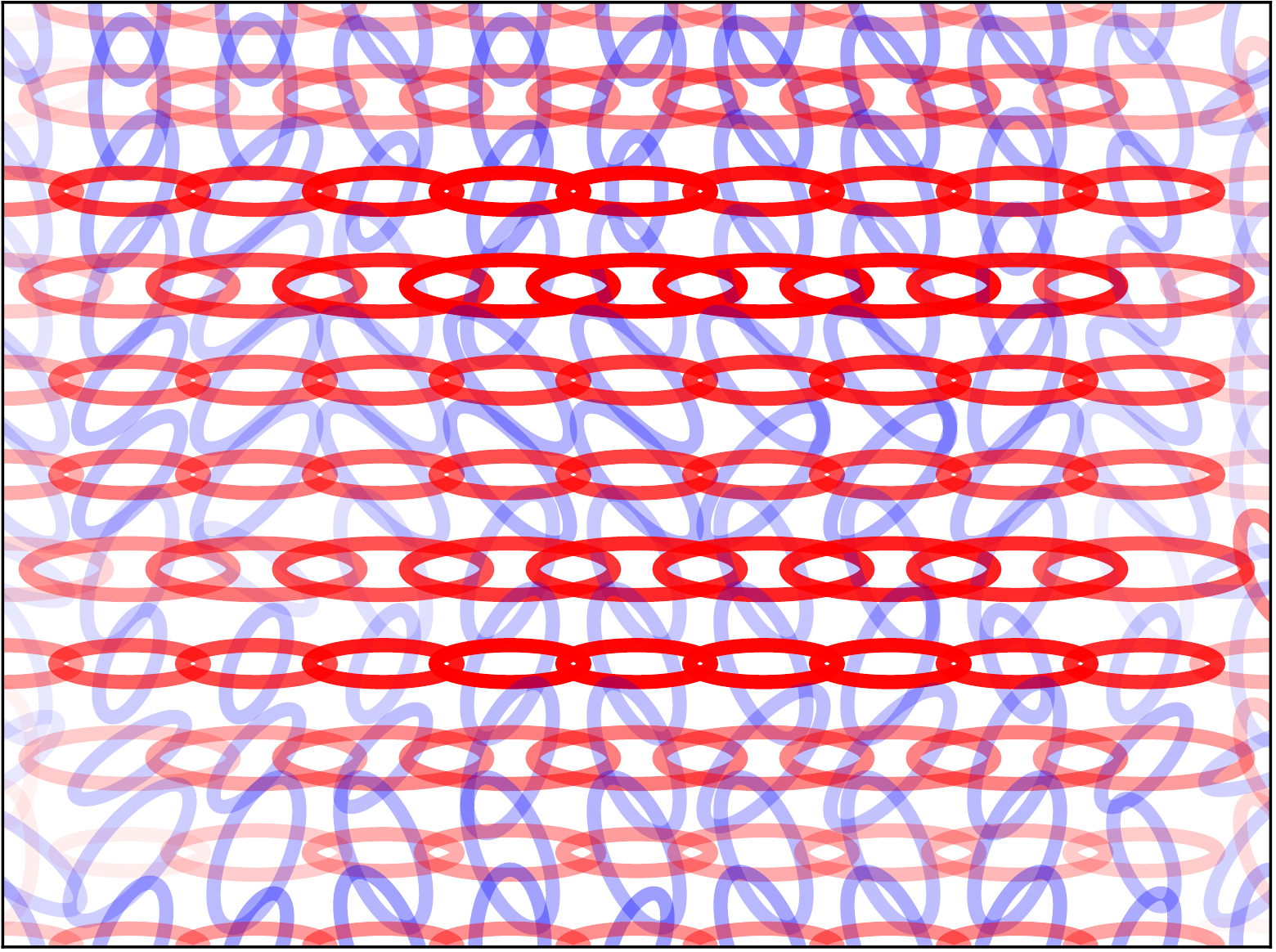} \\
	\phantomsubcaption
	\label{fig:11x11gpsb}
\end{subfigure}
\end{figure}

\begin{figure}
\ContinuedFloat
\Large \textbf{(c)} \\
\begin{subfigure}[t]{\linewidth}
	\centering
	\includegraphics[width=0.19\linewidth]{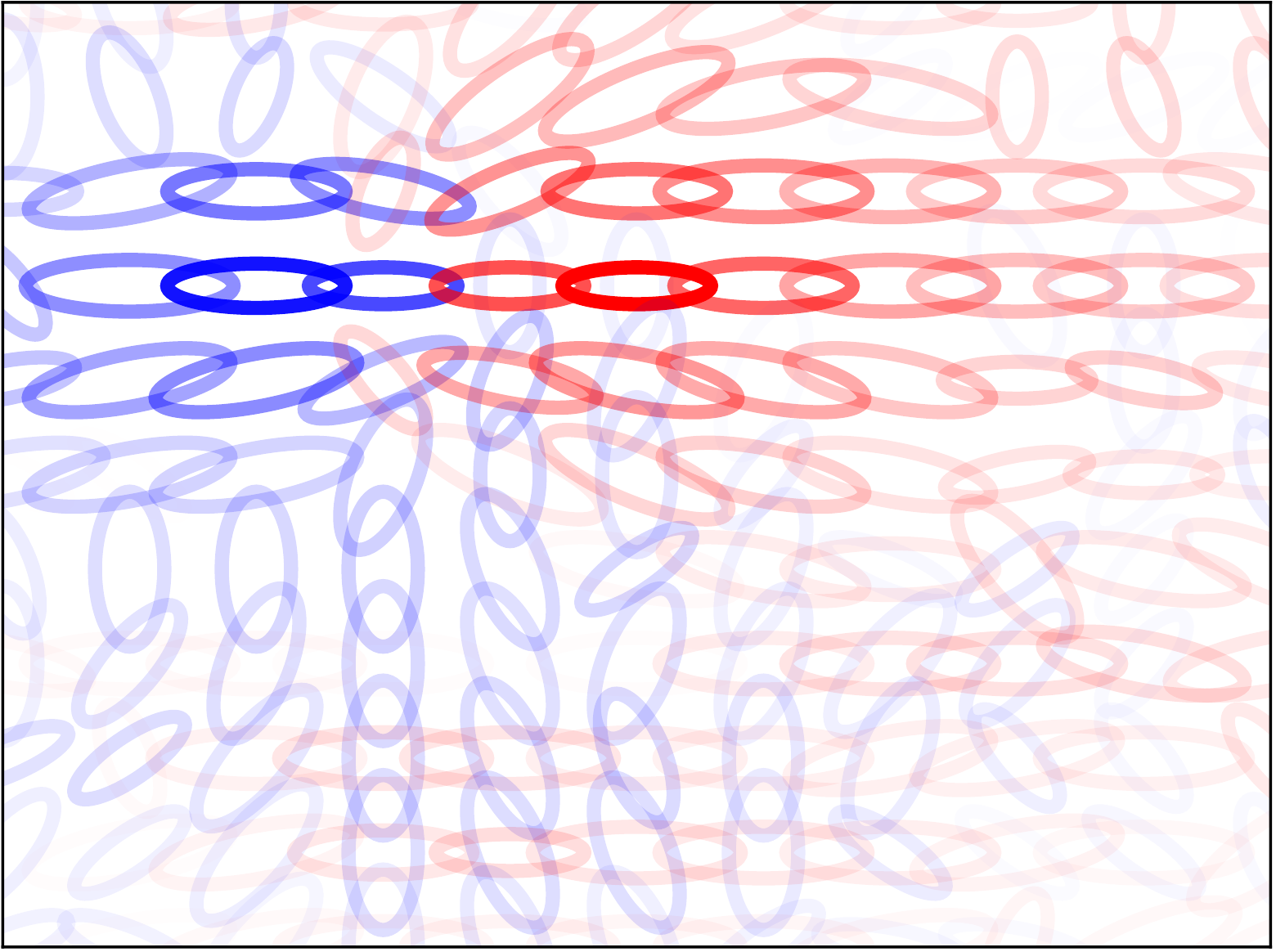}
	\includegraphics[width=0.19\linewidth]{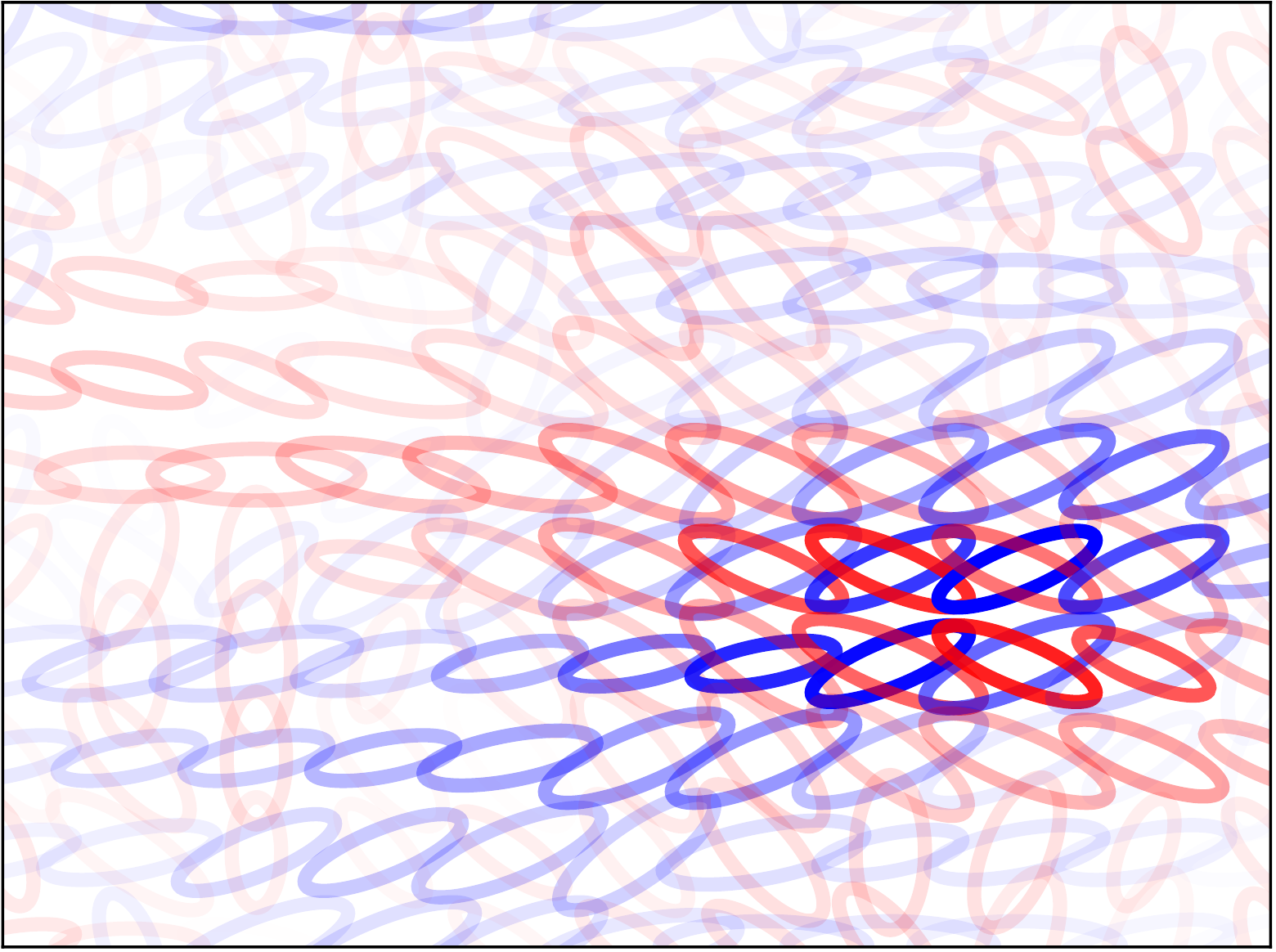}
	\includegraphics[width=0.19\linewidth]{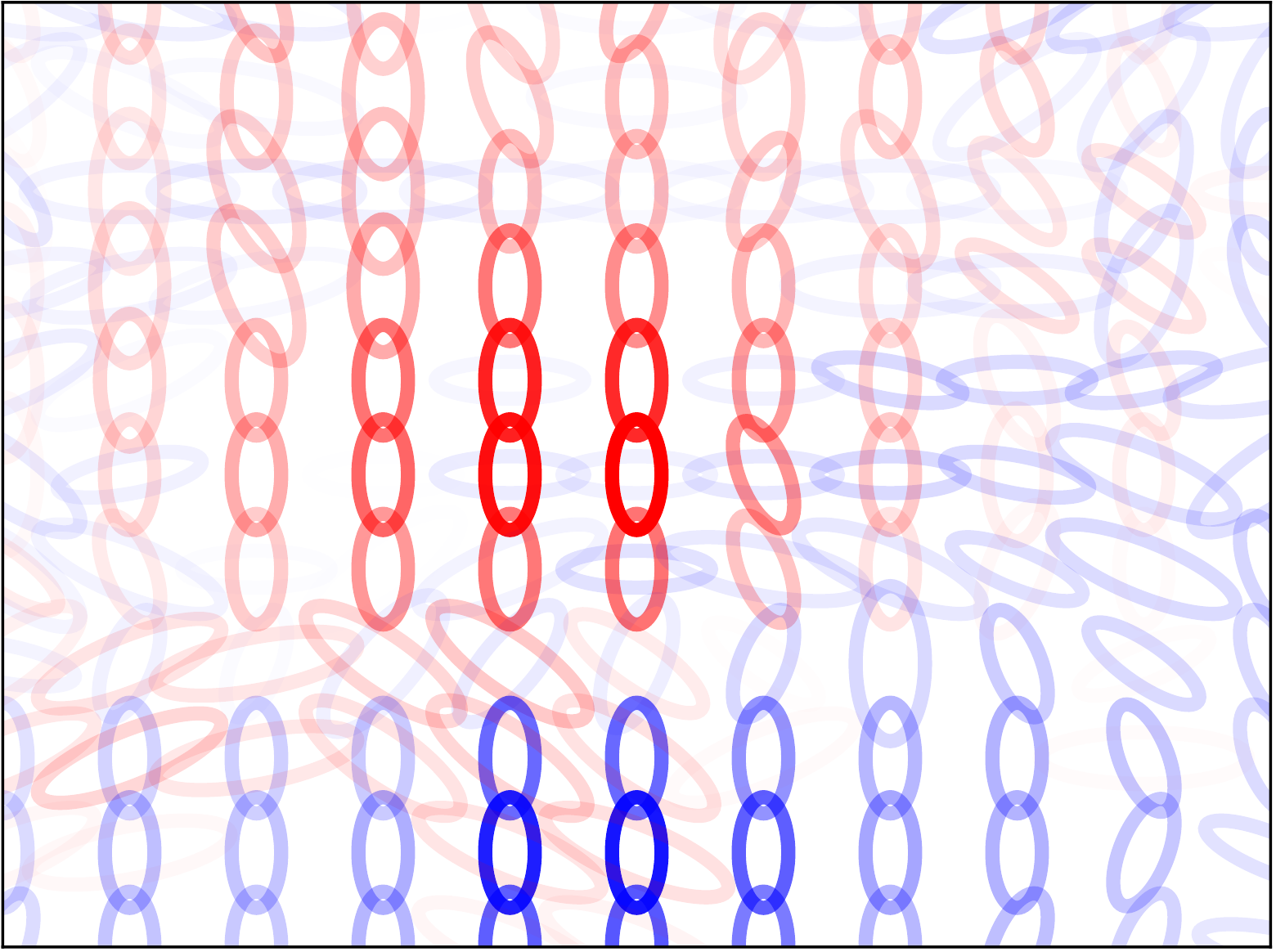}
	\includegraphics[width=0.19\linewidth]{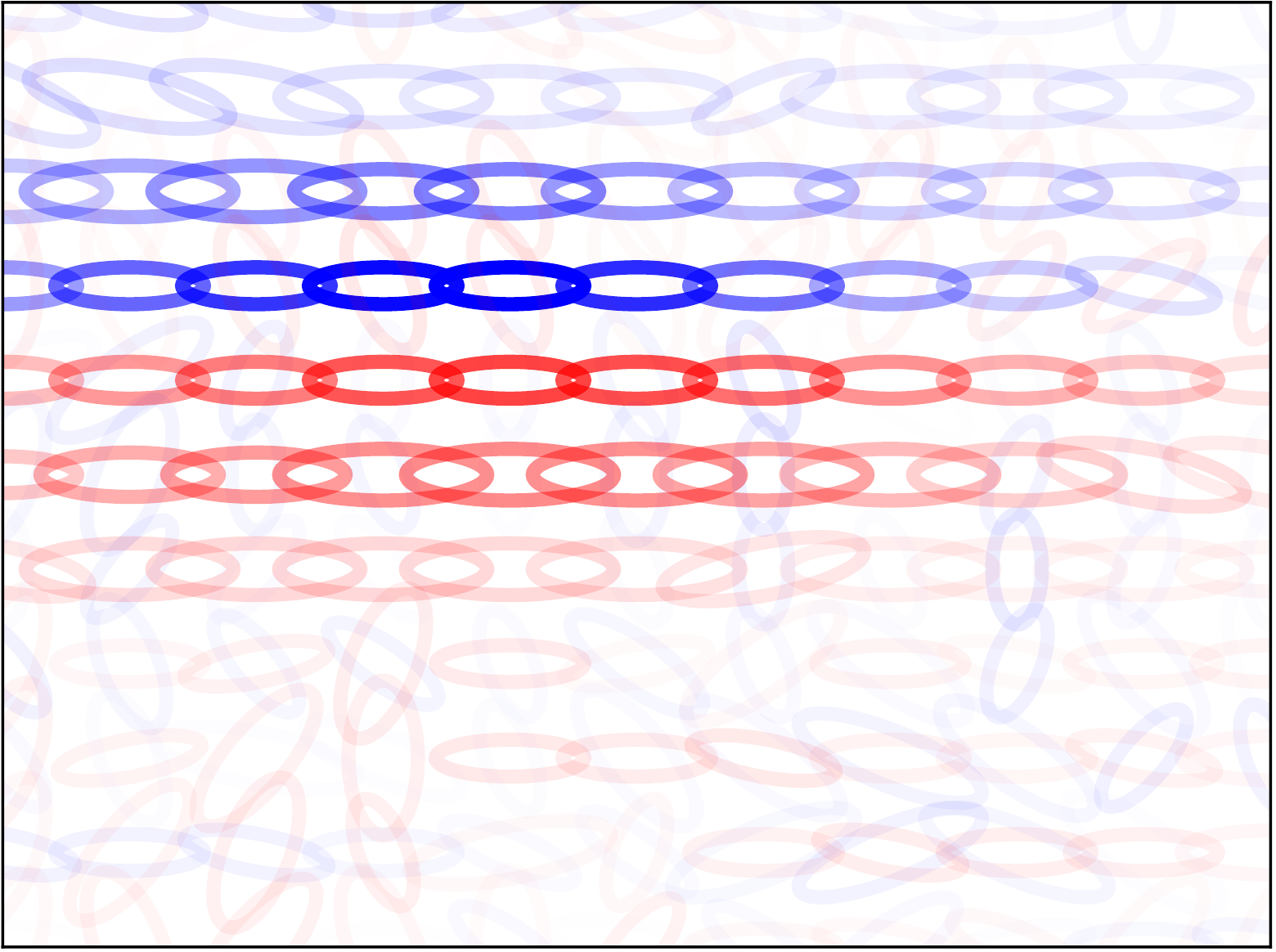}
	\includegraphics[width=0.19\linewidth]{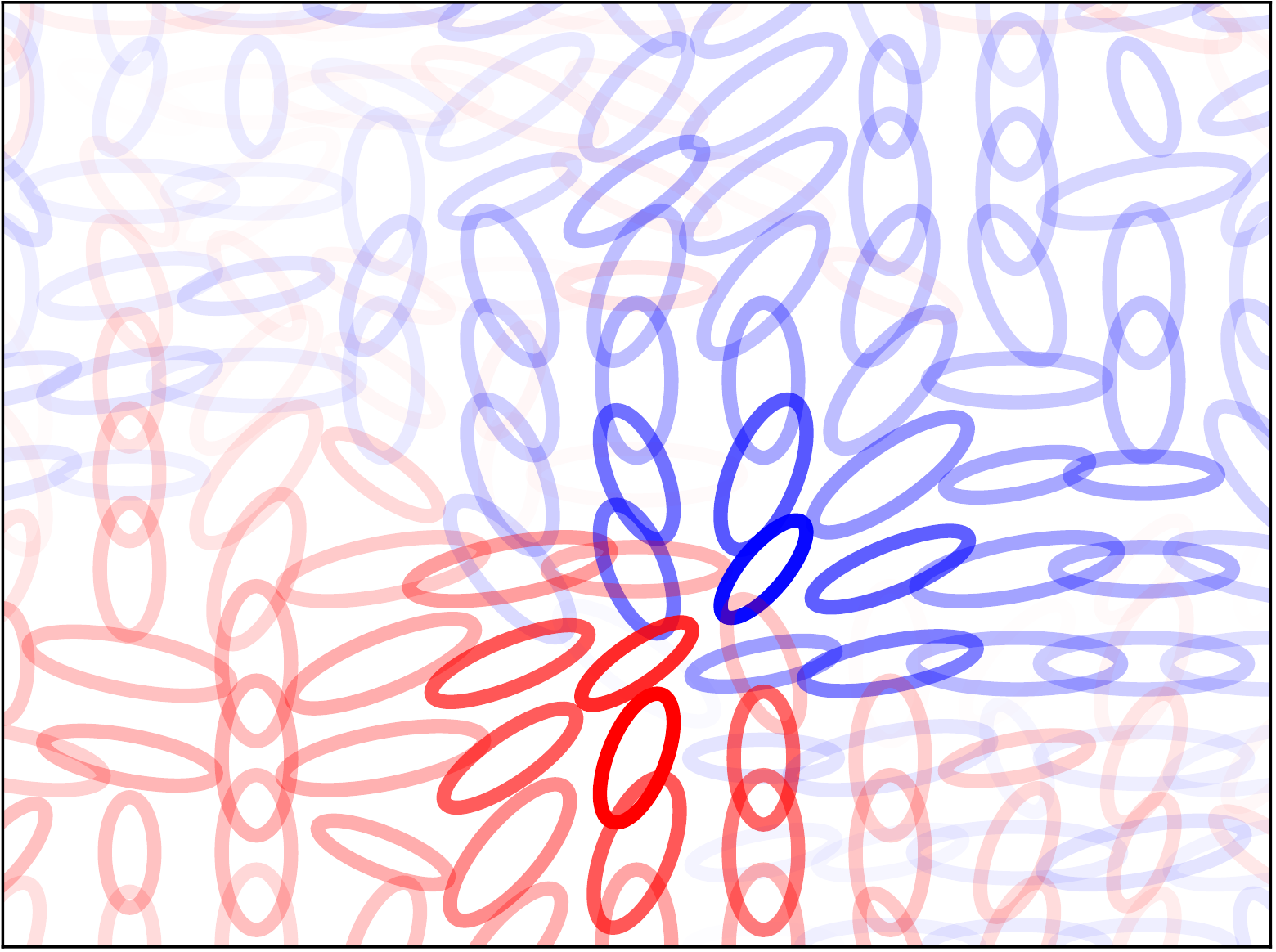}
	\includegraphics[width=0.19\linewidth]{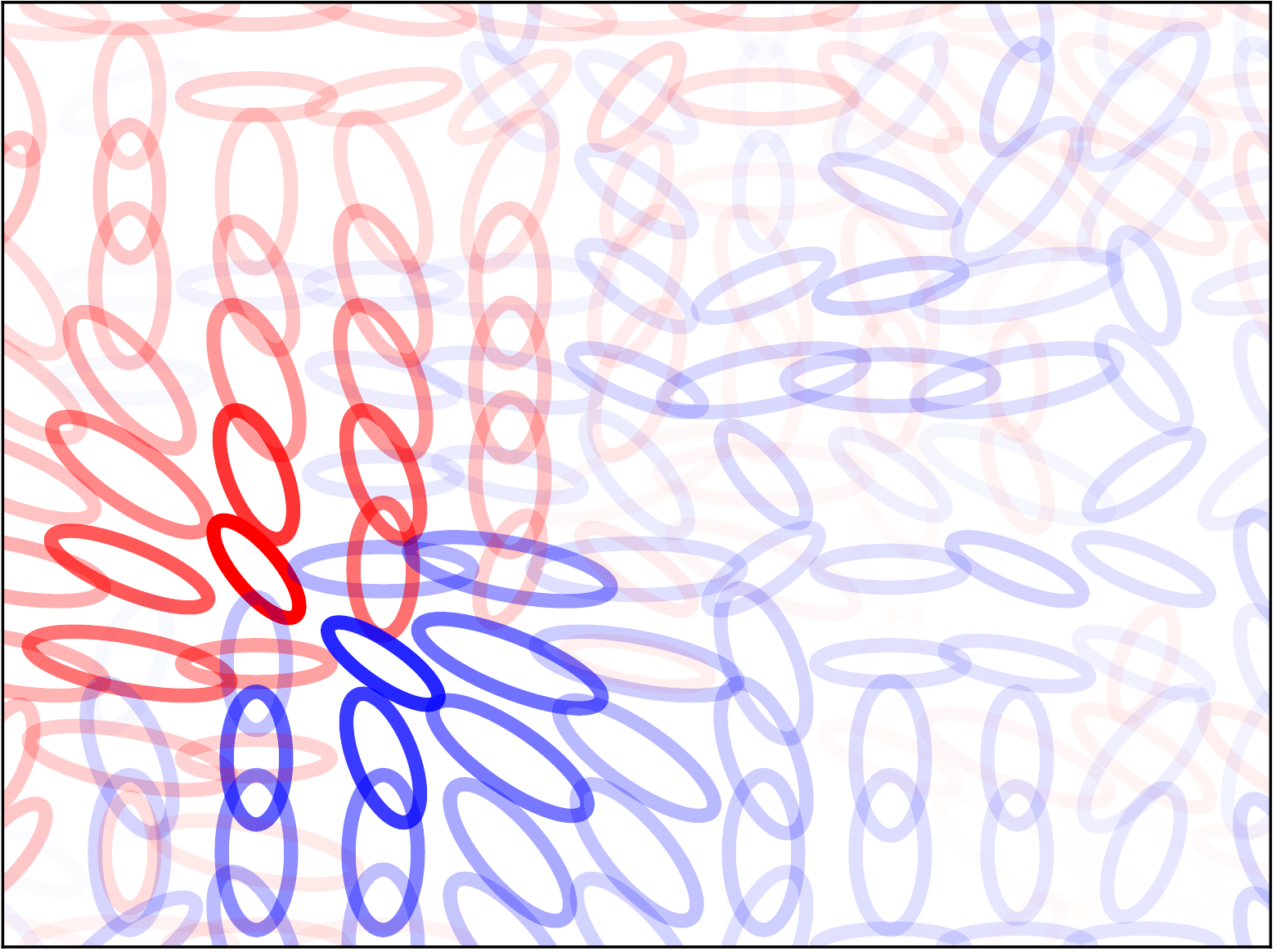}
	\includegraphics[width=0.19\linewidth]{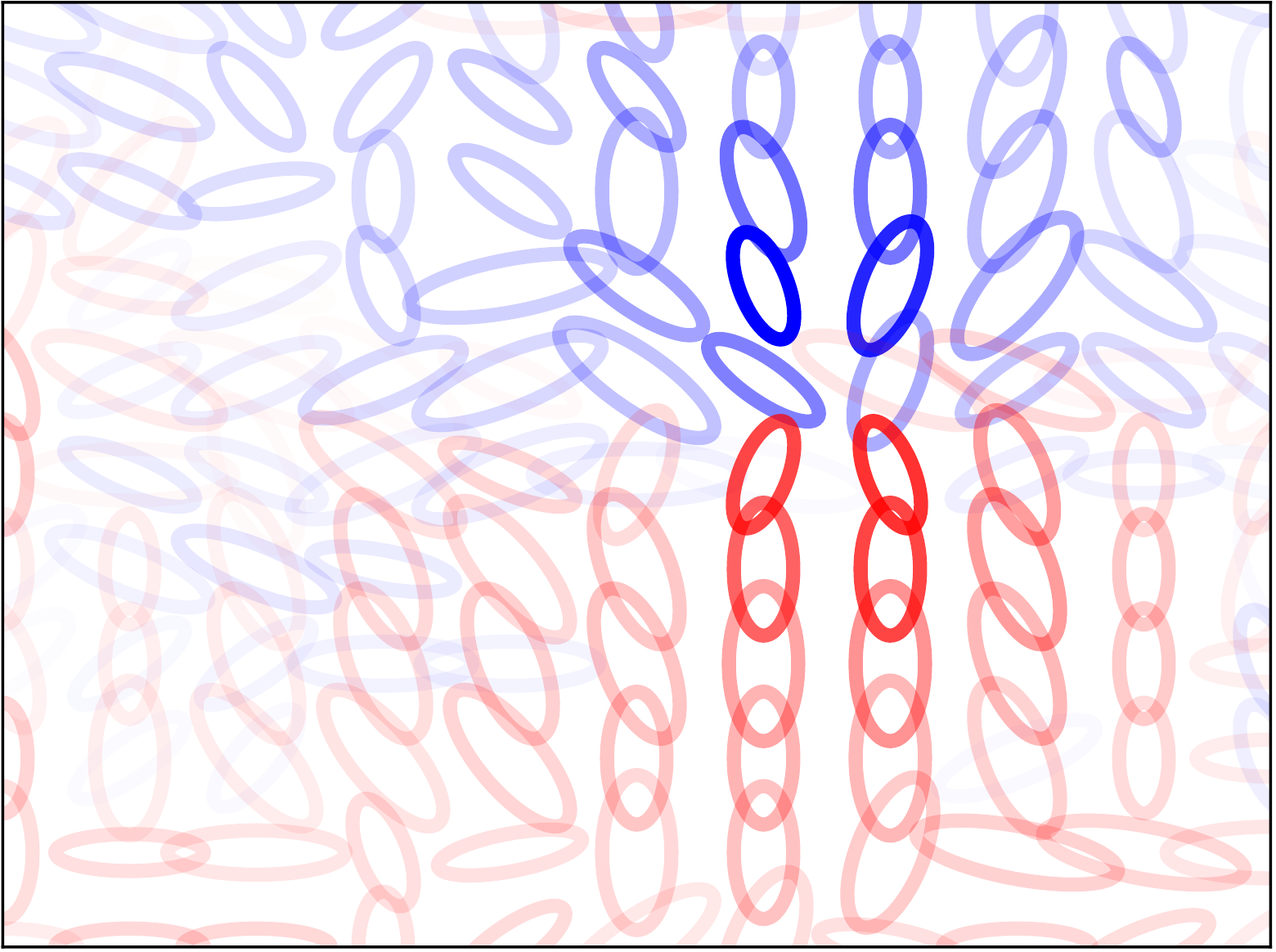}
	\includegraphics[width=0.19\linewidth]{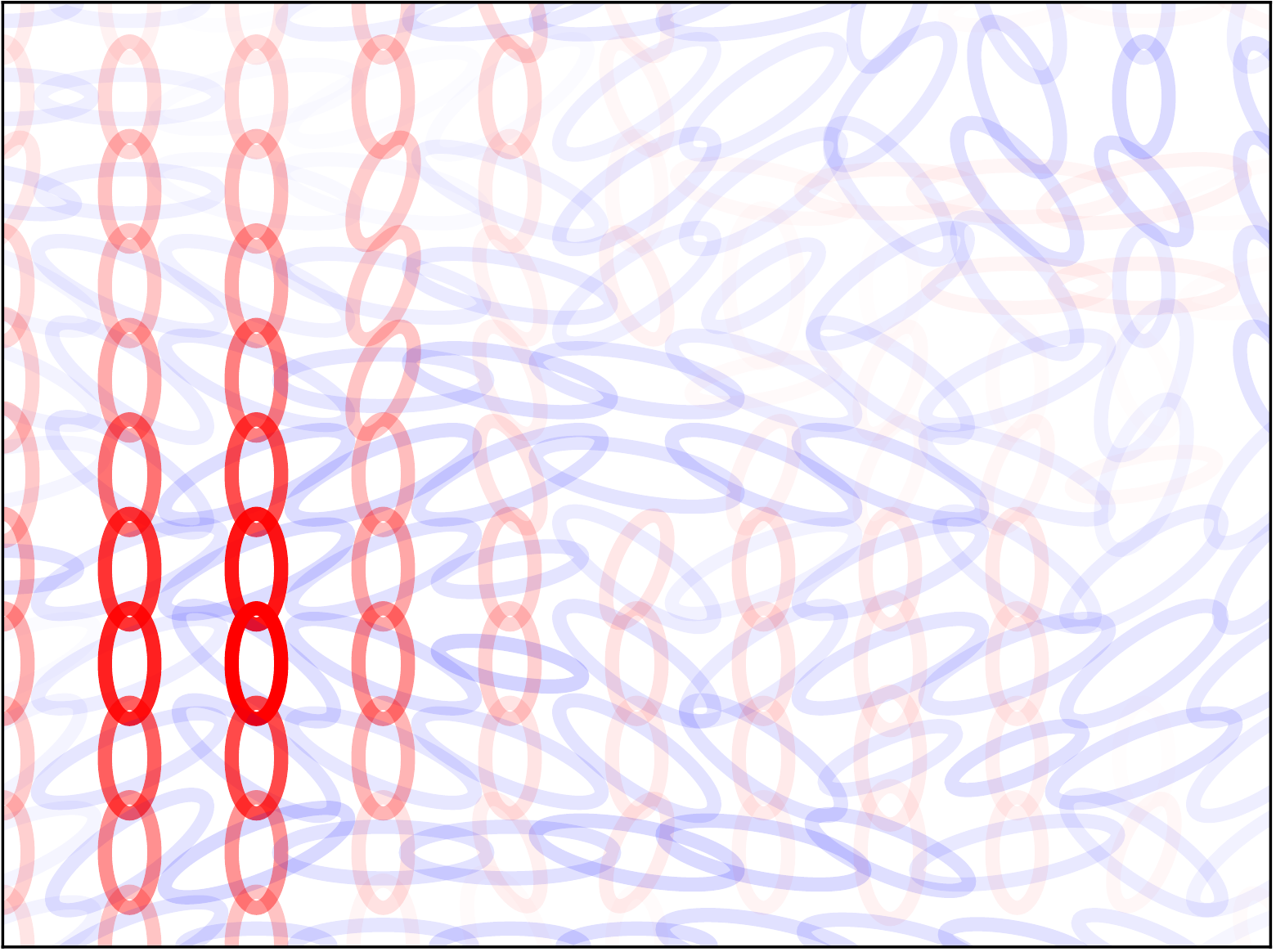}
	\includegraphics[width=0.19\linewidth]{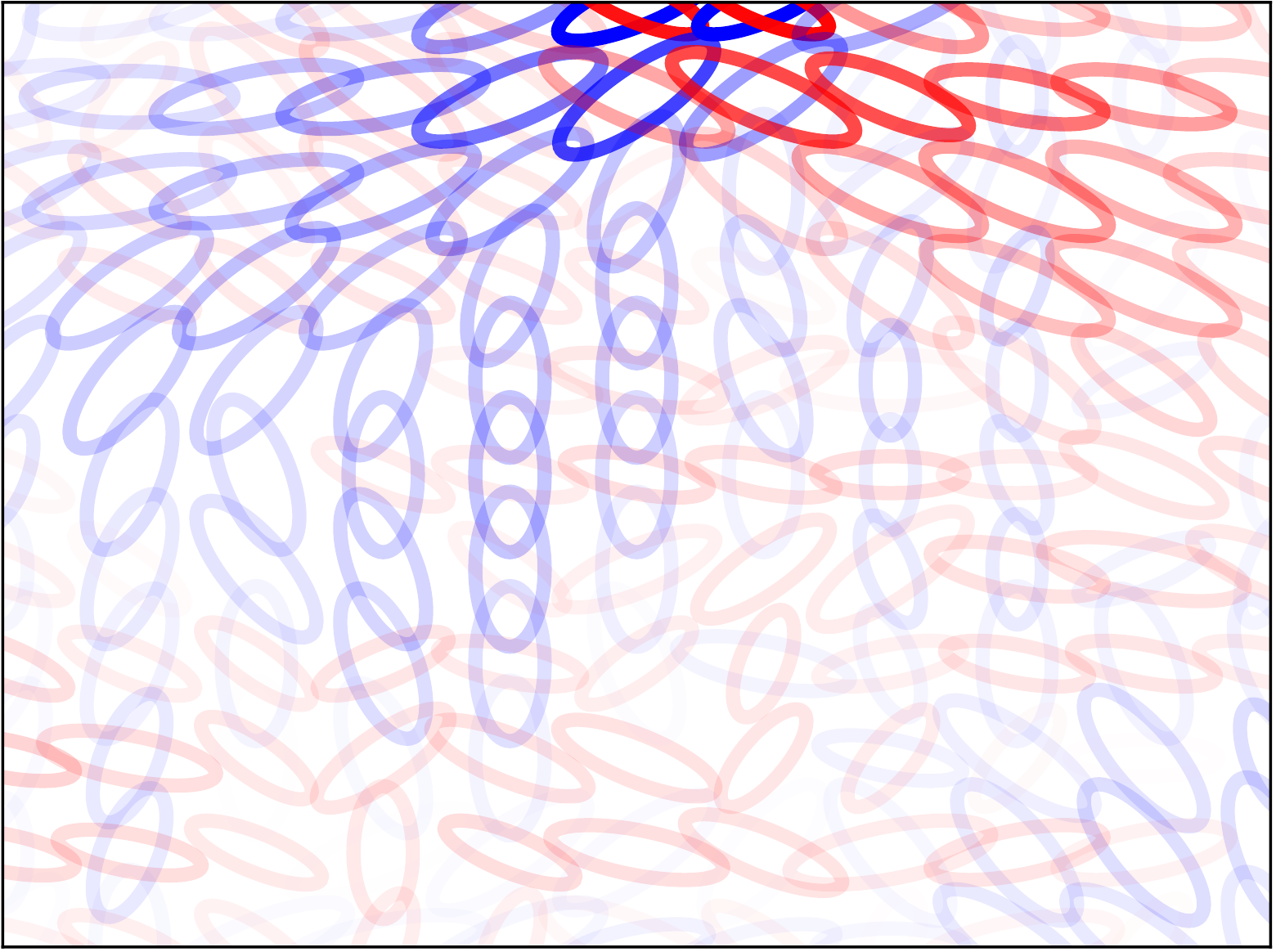}
	\includegraphics[width=0.19\linewidth]{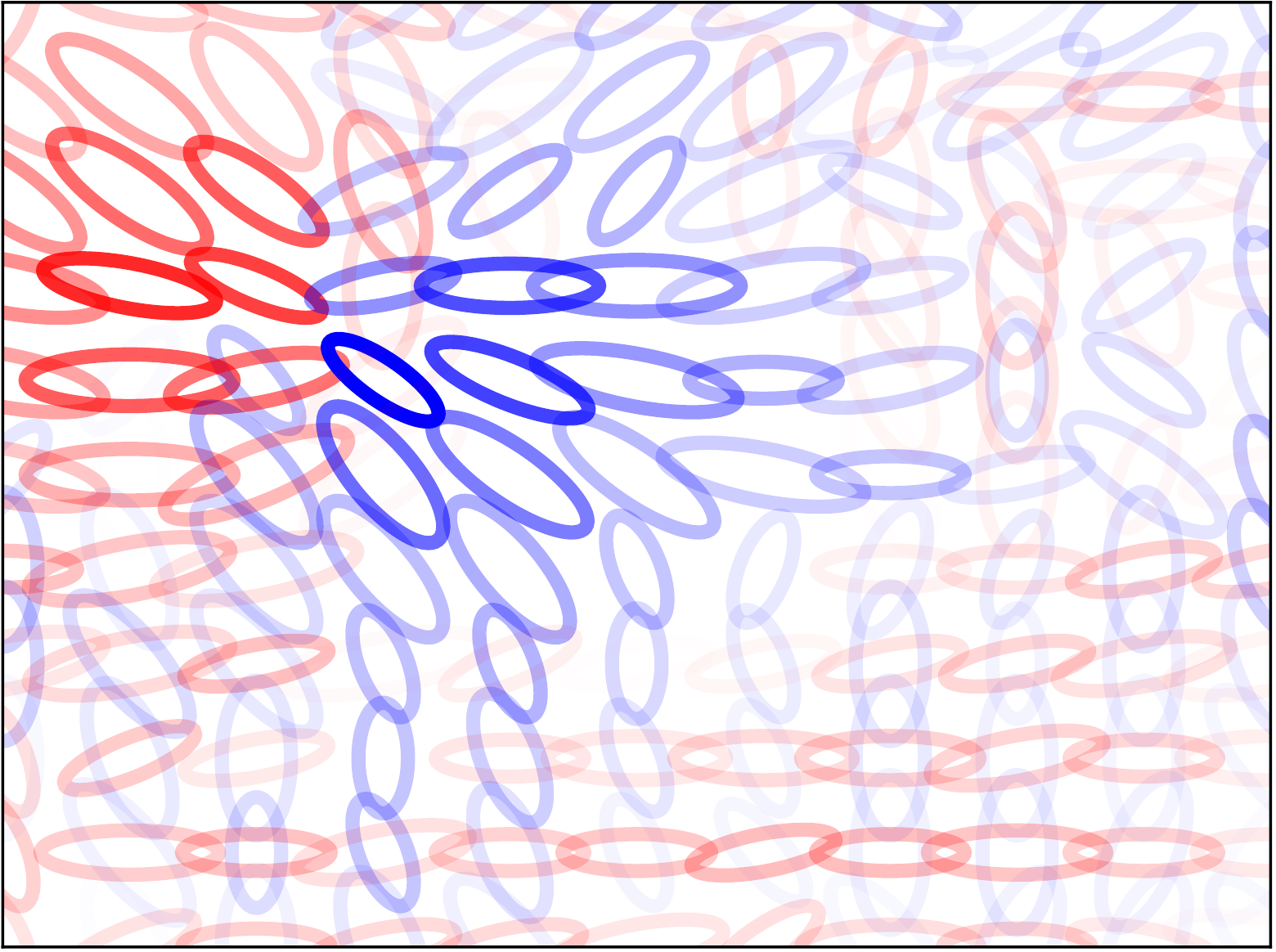}
	\includegraphics[width=0.19\linewidth]{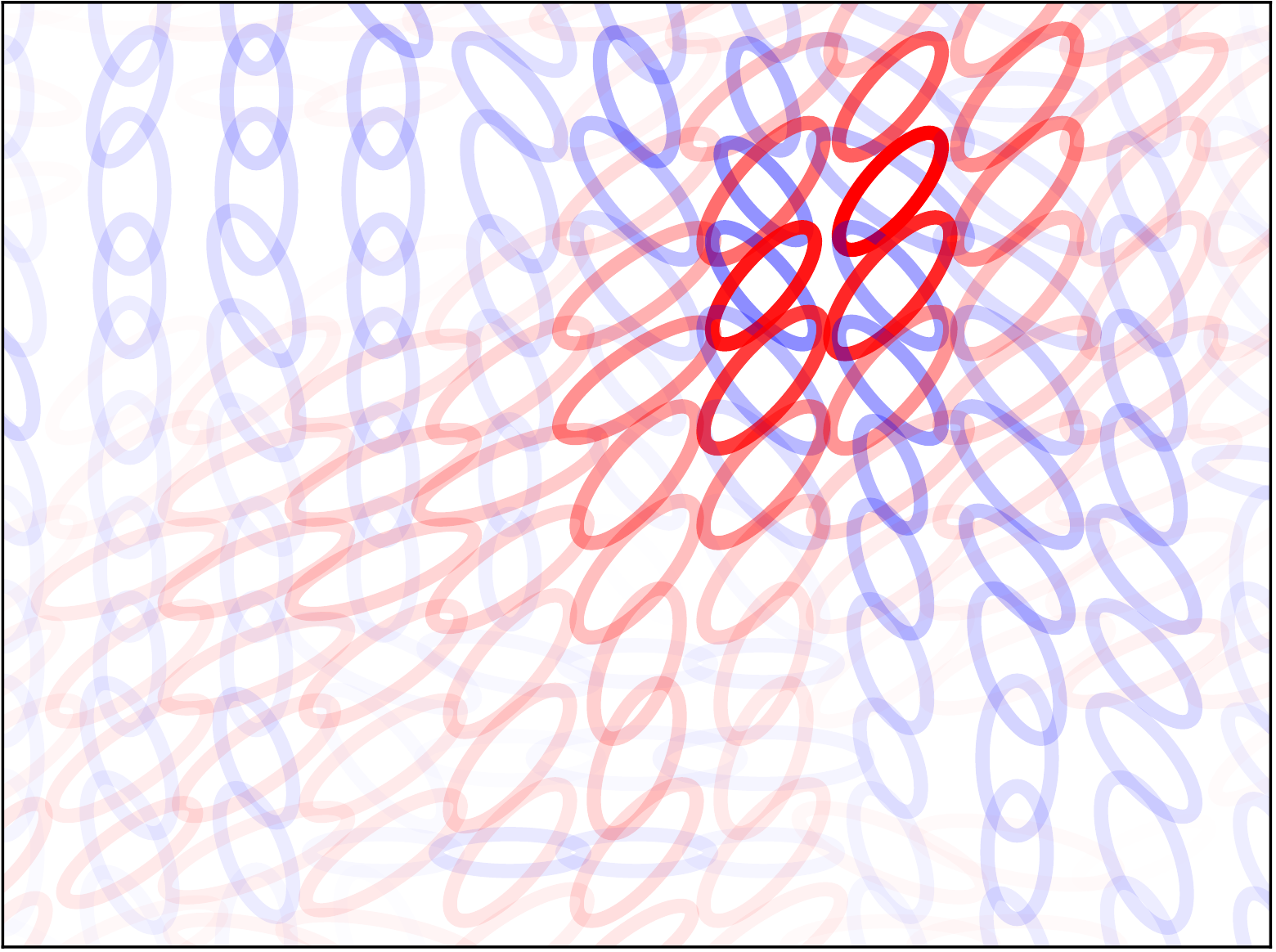}
	\includegraphics[width=0.19\linewidth]{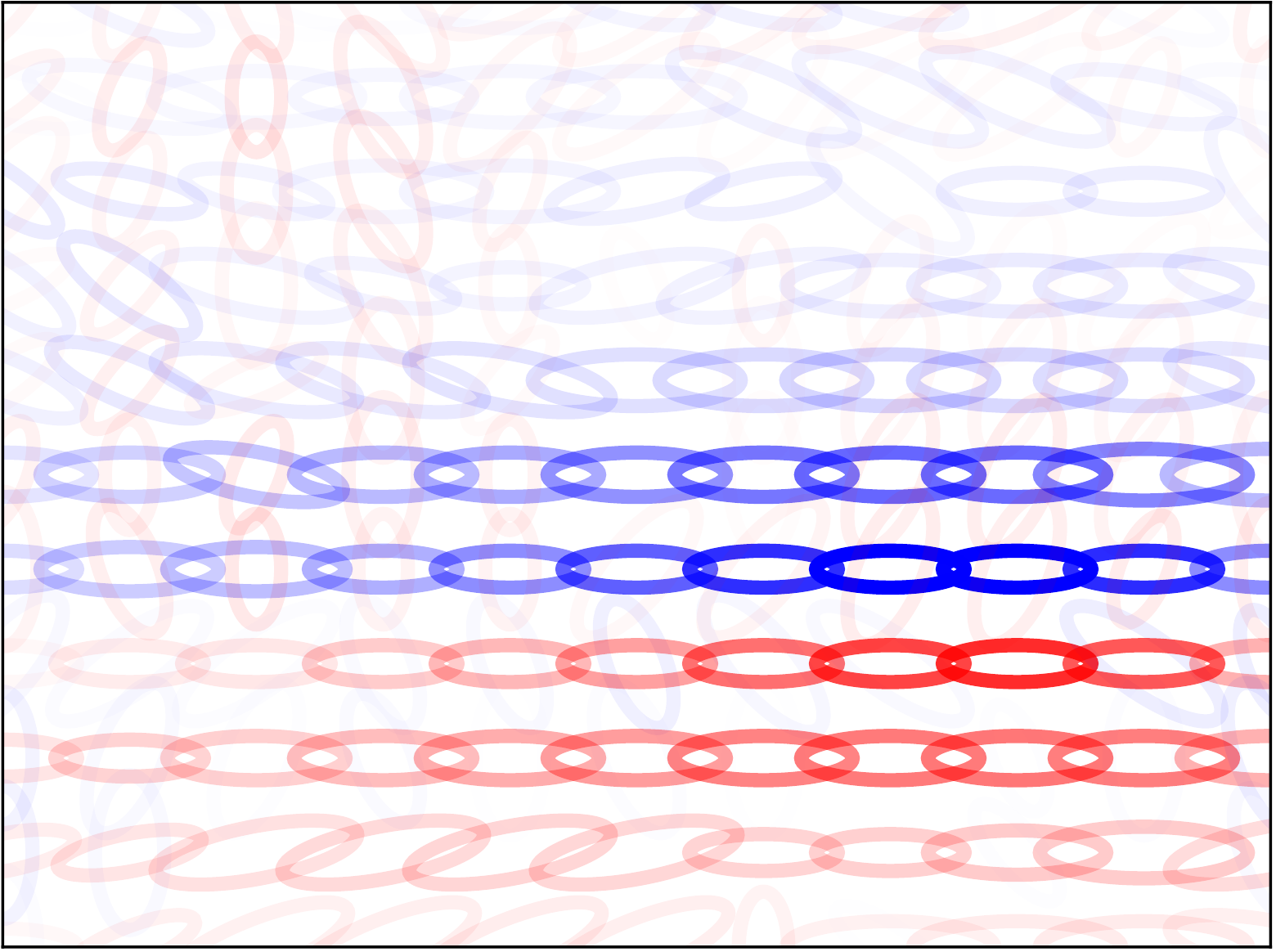}
	\includegraphics[width=0.19\linewidth]{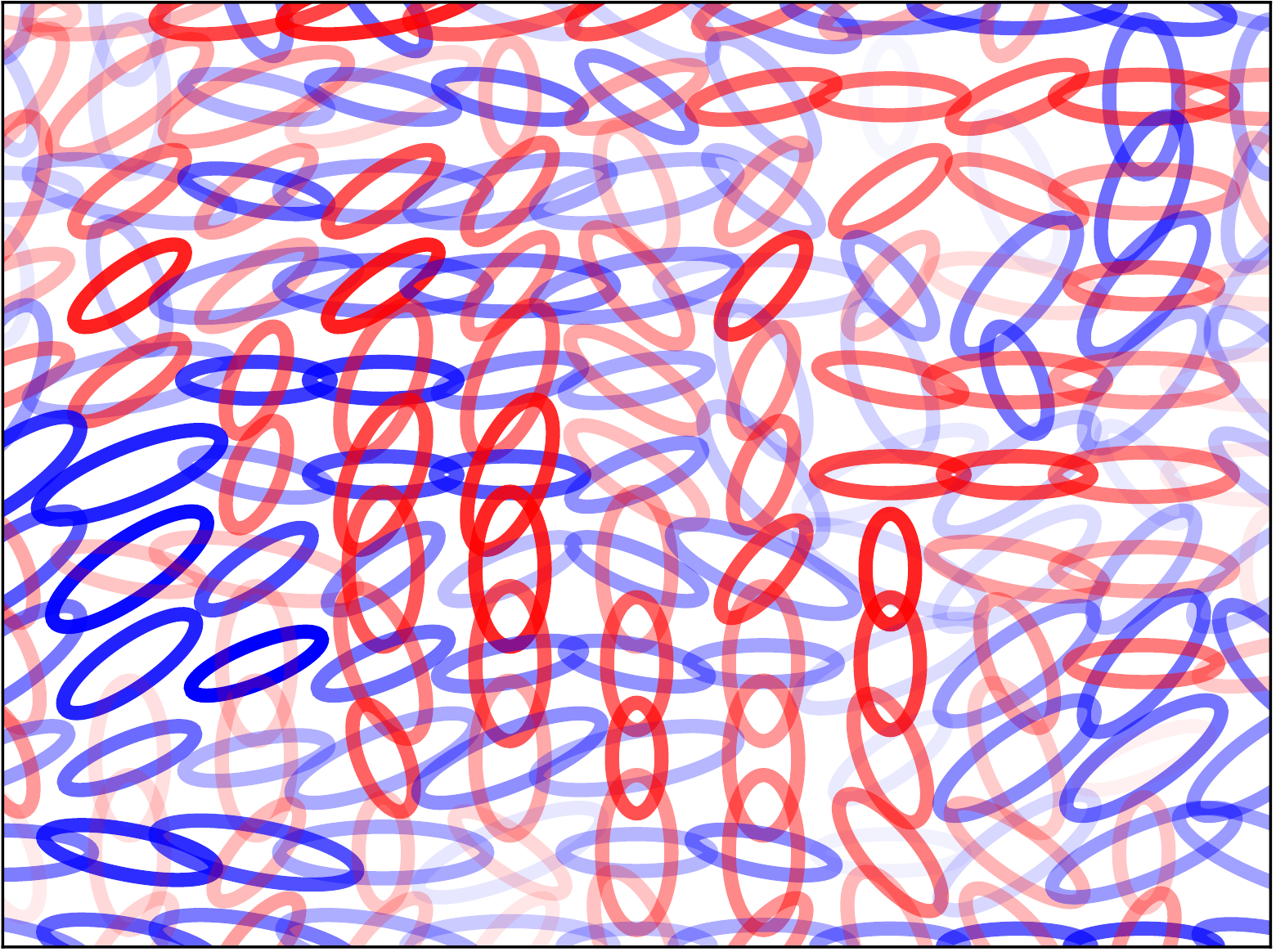}
	\includegraphics[width=0.19\linewidth]{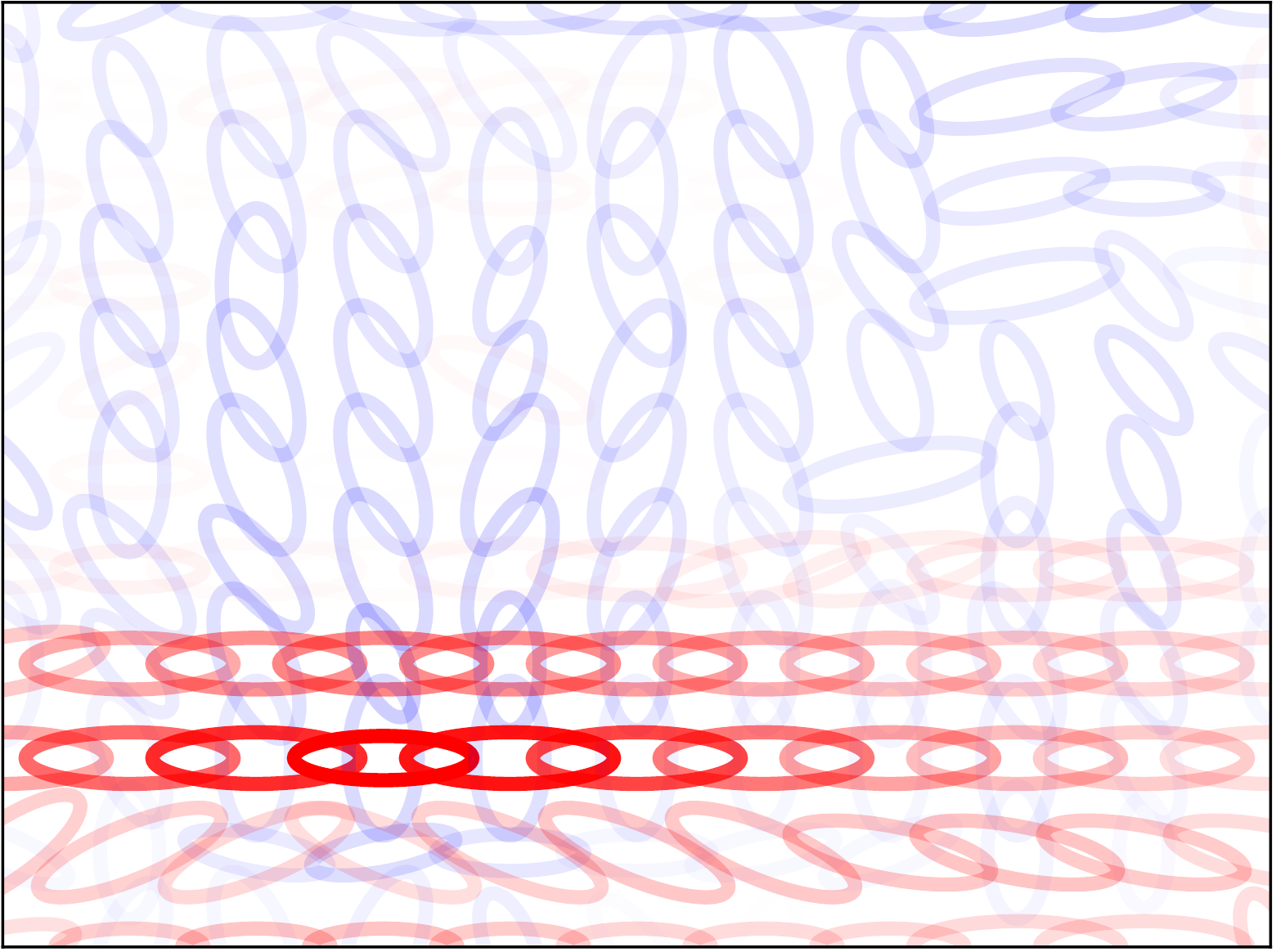}
	\includegraphics[width=0.19\linewidth]{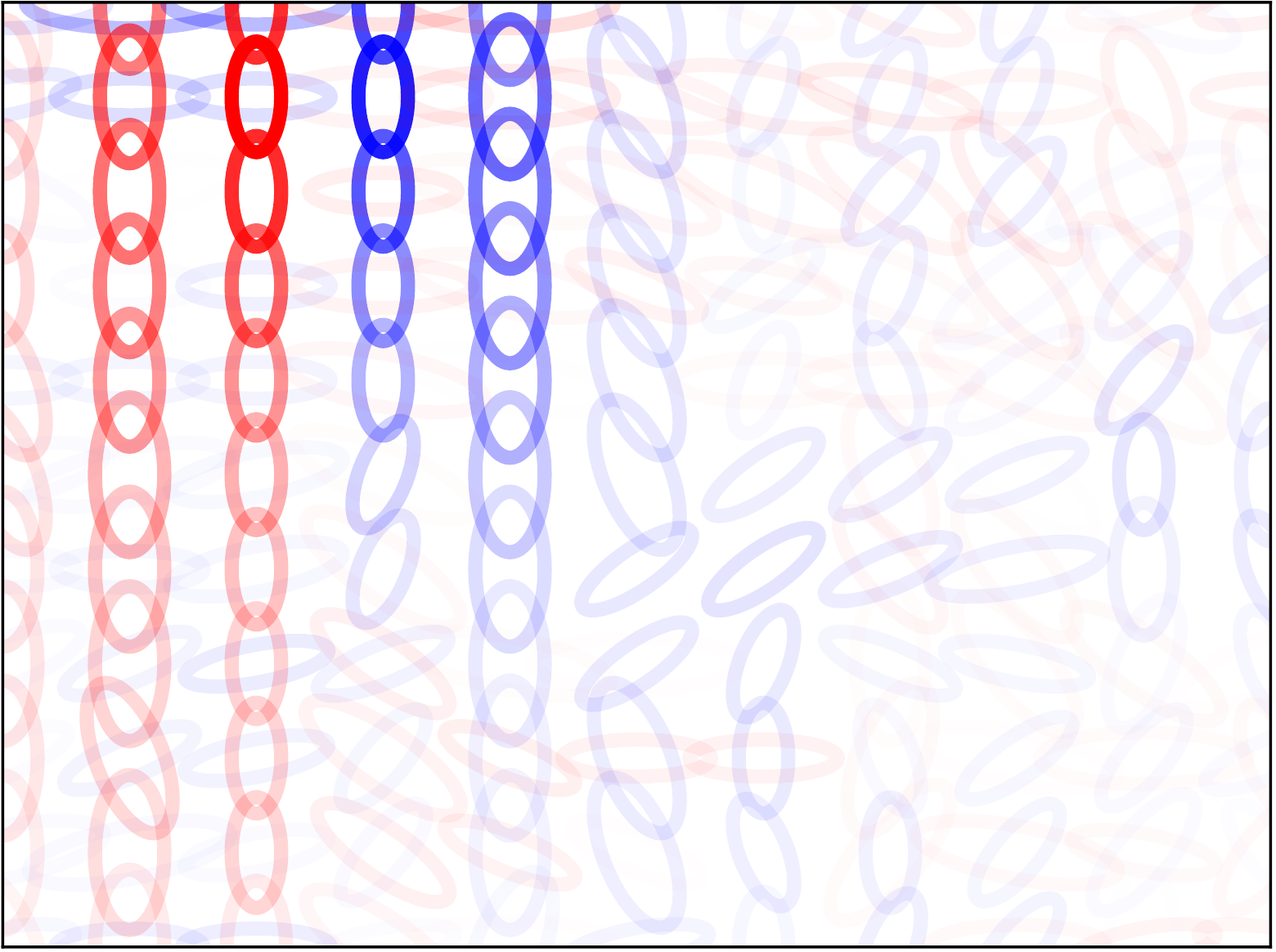}
	\includegraphics[width=0.19\linewidth]{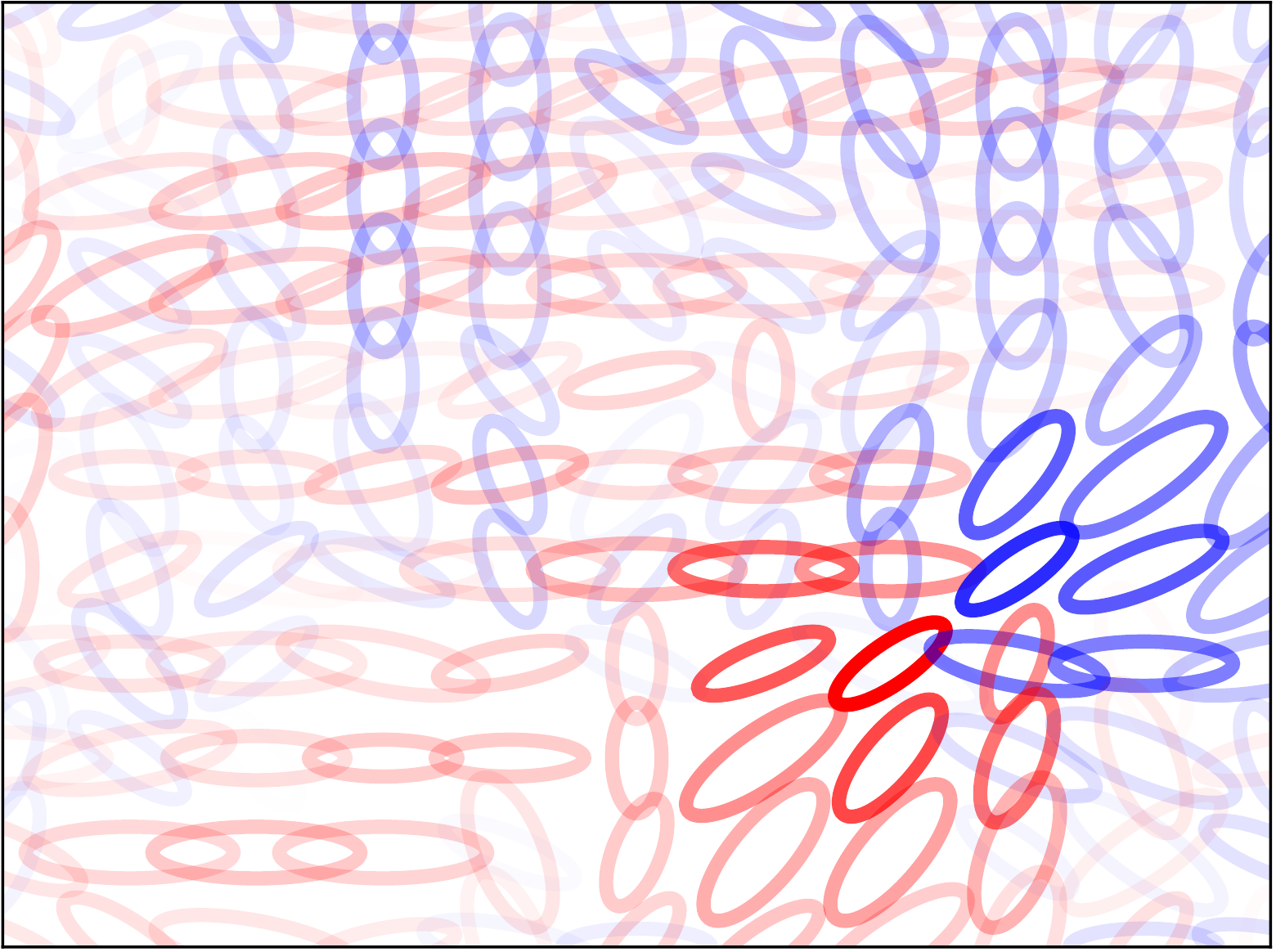}
	\includegraphics[width=0.19\linewidth]{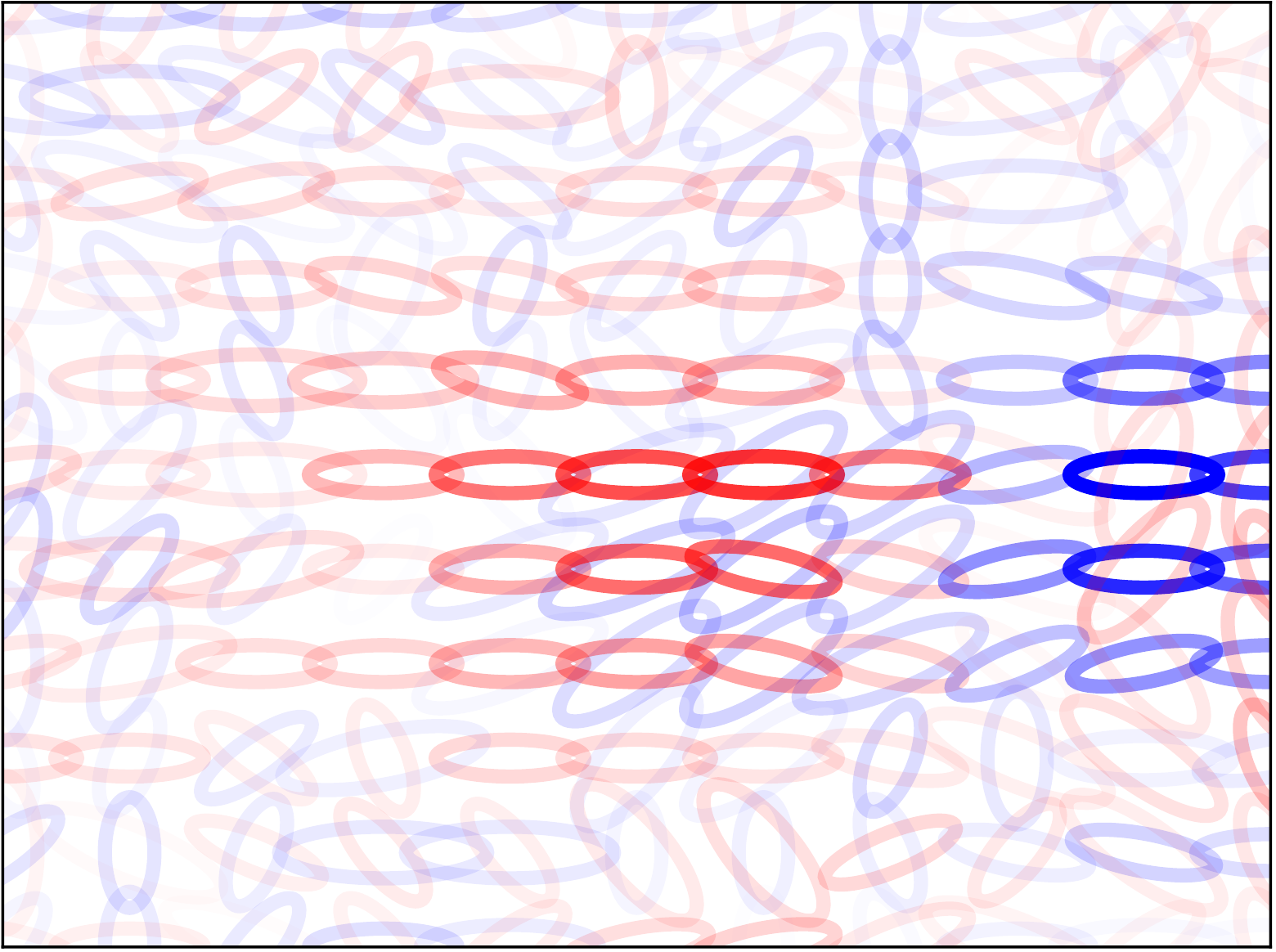}
	\includegraphics[width=0.19\linewidth]{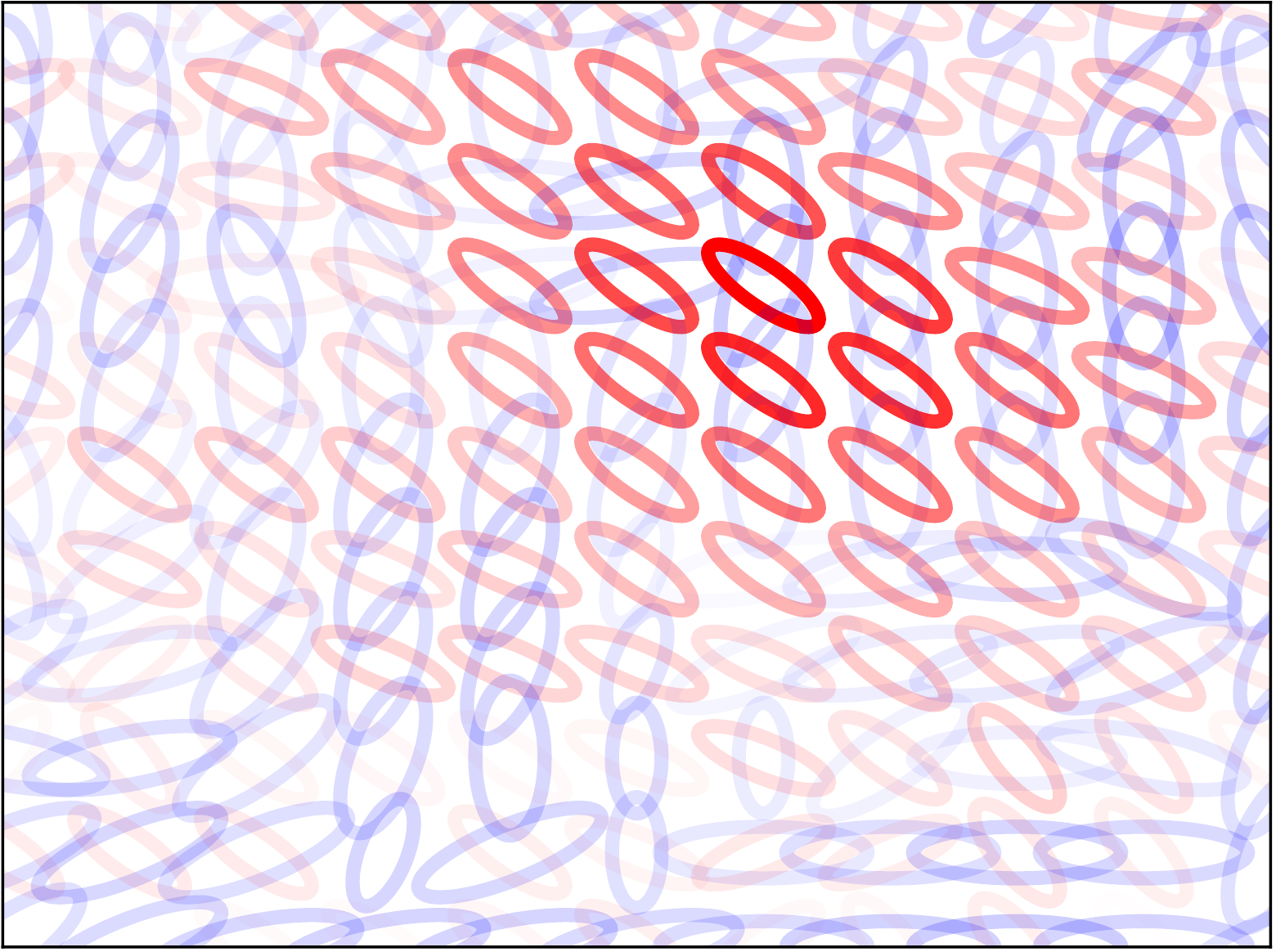}
	\includegraphics[width=0.19\linewidth]{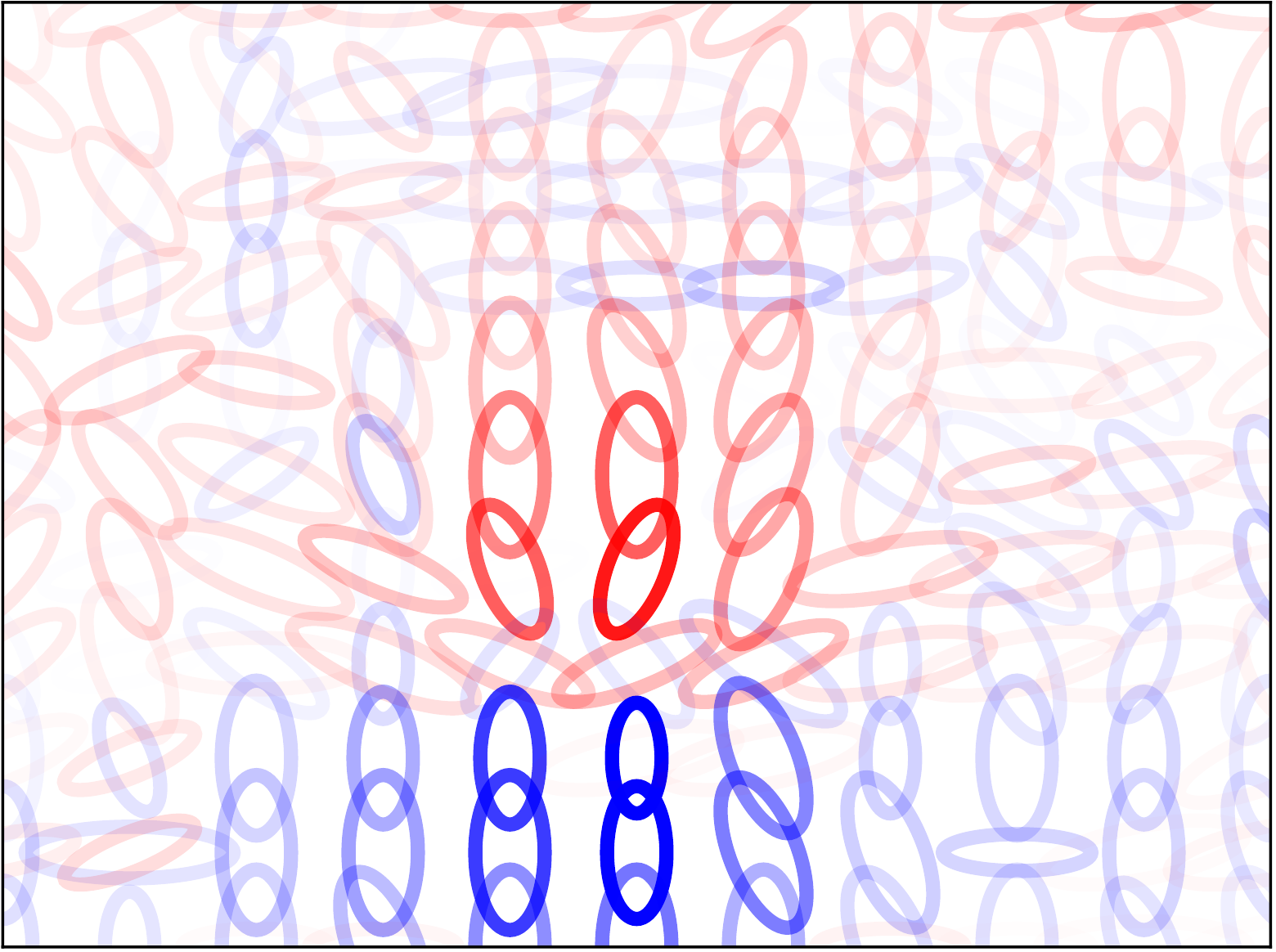}
	\includegraphics[width=0.19\linewidth]{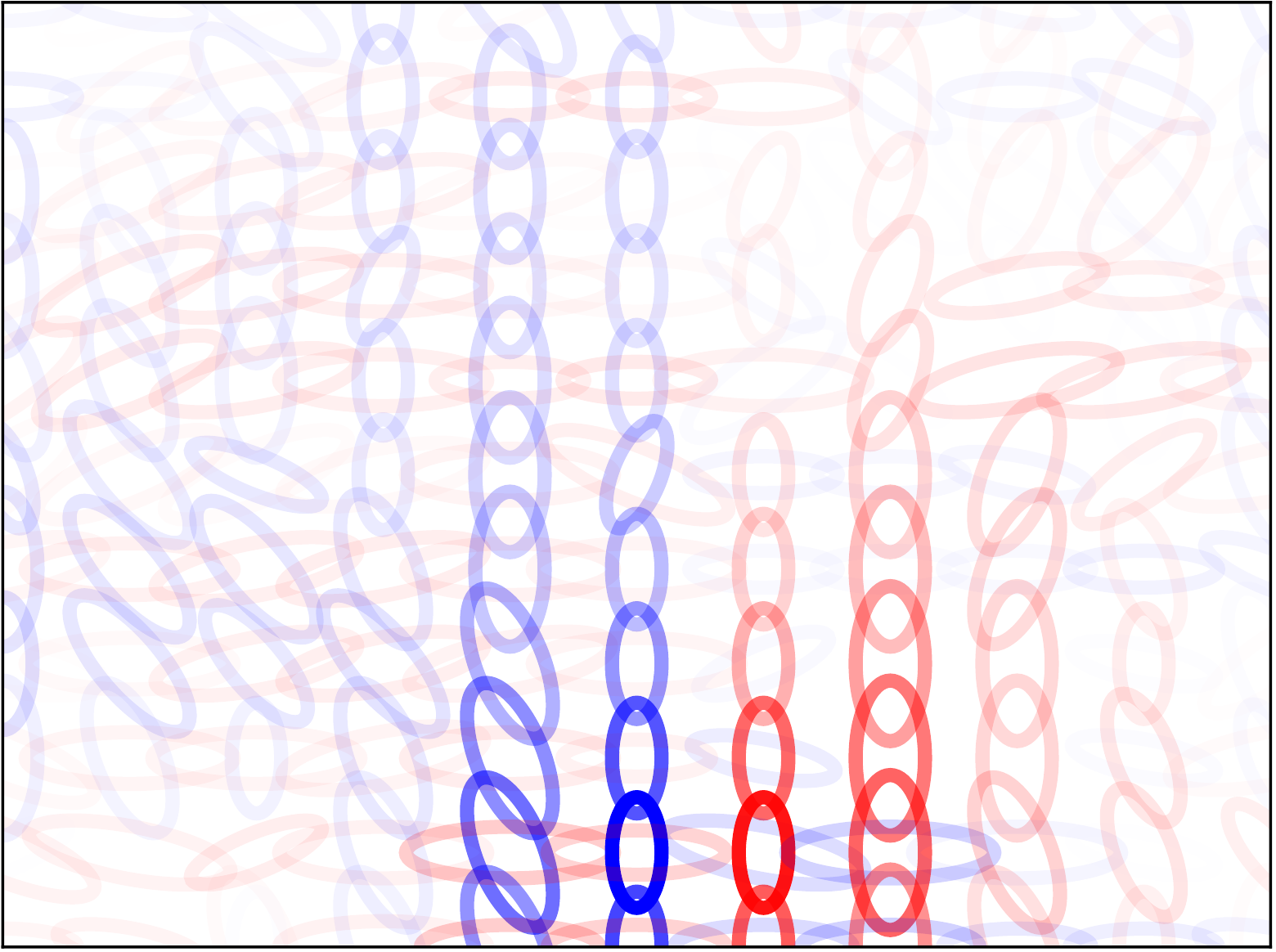} \\
	\phantomsubcaption
	\label{fig:11x11gpsc}
\end{subfigure}
\caption{\textbf{Visualization of Sparse Coding Model V2 Units With 11x11 Spatial Locations and 350 Principal Components.} (a) Sparse coding with a regularization coefficient of 2.0. (b) Sparse coding with a regularization coefficient of 4.0. (c) ICA. Opacity reflects the response intensity, color reflects the sign of the response (red for positive and blue for negative), and size reflects frequency.}
\label{fig:11x11gps}
\end{figure}

The patches that maximally excited selected units for non-negative sparse coding and overcomplete ICA with 6x6 spatial locations are shown in Figure \ref{fig:maxpatches}. The patches for non-negative sparse coding (for both values of the regularization coefficient) reveal texture-like selectively in certain units (the second and third in Figure \ref{fig:maxpatchesa} and the third in Figure \ref{fig:maxpatchesb}) that are not easily described by common geometric primitives. The second unit in Figure \ref{fig:maxpatchesa} could be described as horizontal lines with gaps in between, though it was also activated by images of text. The third unit in Figure \ref{fig:maxpatchesa} appears as repeating small circles, and the third unit in Figure \ref{fig:maxpatchesb} appears as repeating curved lines. Corners appeared as lines connected at 90 degree angles sometimes with other geometries nearby. One corner unit continued in both directions and was activated by crosses. Curves appeared mostly in circles. For the overcomplete ICA units, iso-oriented excitation units with side and cross inhibition appeared as lines, and iso-oriented excitation units with end inhibition appeared as lines stopping at a point. Orientation-convergent units with end inhibition appeared as blobs stopping at a point. Iso-oriented excitation with broad inhibition units varied, but often appeared as lines.

\begin{figure}
	\Large \textbf{(a)} \\
	\begin{subfigure}[t]{\linewidth}
		\centering
		\includegraphics[width=0.15\columnwidth]{vis/6x6/sc0.5/540.pdf}
		\includegraphics[width=0.75\columnwidth]{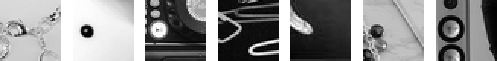} \\
		\includegraphics[width=0.15\columnwidth]{vis/6x6/sc0.5/12.pdf}
		\includegraphics[width=0.75\columnwidth]{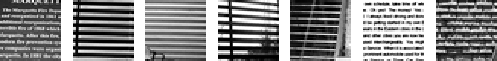} \\
		\includegraphics[width=0.15\columnwidth]{vis/6x6/sc0.5/473.pdf}
		\includegraphics[width=0.75\columnwidth]{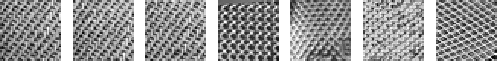} \\
		\includegraphics[width=0.15\columnwidth]{vis/6x6/sc0.5/111.pdf}
		\includegraphics[width=0.75\columnwidth]{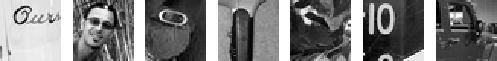} \\
		\phantomsubcaption
		\label{fig:maxpatchesa}
	\end{subfigure}
	\Large \textbf{(b)} \\
	\begin{subfigure}[t]{\linewidth}
		\centering
		\includegraphics[width=0.15\columnwidth]{vis/6x6/sc4.0/8.pdf}
		\includegraphics[width=0.75\columnwidth]{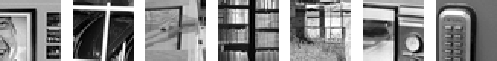} \\
		\includegraphics[width=0.15\columnwidth]{vis/6x6/sc4.0/10.pdf}
		\includegraphics[width=0.75\columnwidth]{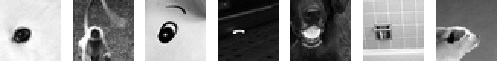} \\
		\includegraphics[width=0.15\columnwidth]{vis/6x6/sc4.0/14.pdf}
		\includegraphics[width=0.75\columnwidth]{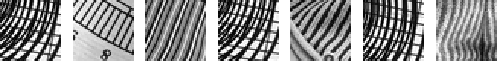} \\
		\includegraphics[width=0.15\columnwidth]{vis/6x6/sc4.0/138.pdf}
		\includegraphics[width=0.75\columnwidth]{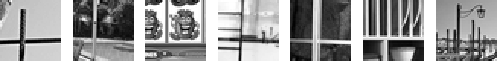} \\
		\phantomsubcaption
		\label{fig:maxpatchesb}
	\end{subfigure}
\end{figure}

\begin{figure}
	\ContinuedFloat
	\Large \textbf{(c)} \\
	\begin{subfigure}[t]{\linewidth}
		\centering
		\includegraphics[width=0.15\columnwidth]{vis/6x6/ica/0.pdf}
		\includegraphics[width=0.75\columnwidth]{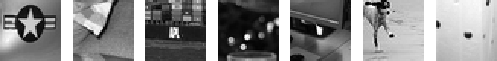} \\
		\includegraphics[width=0.15\columnwidth]{vis/6x6/ica/8.pdf}
		\includegraphics[width=0.75\columnwidth]{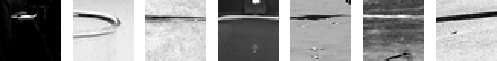} \\
		\includegraphics[width=0.15\columnwidth]{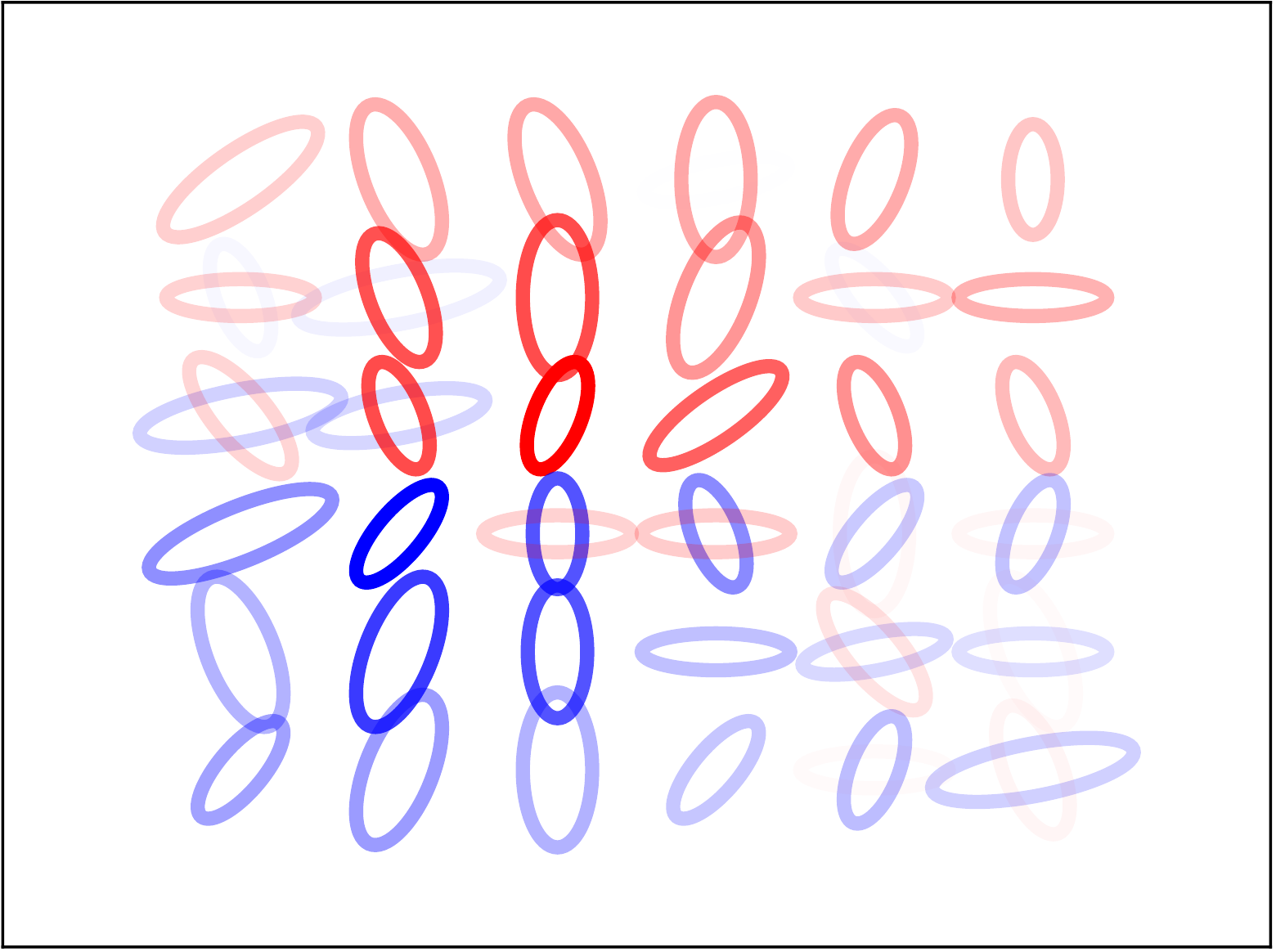}
		\includegraphics[width=0.75\columnwidth]{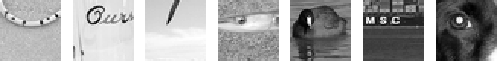} \\
		\includegraphics[width=0.15\columnwidth]{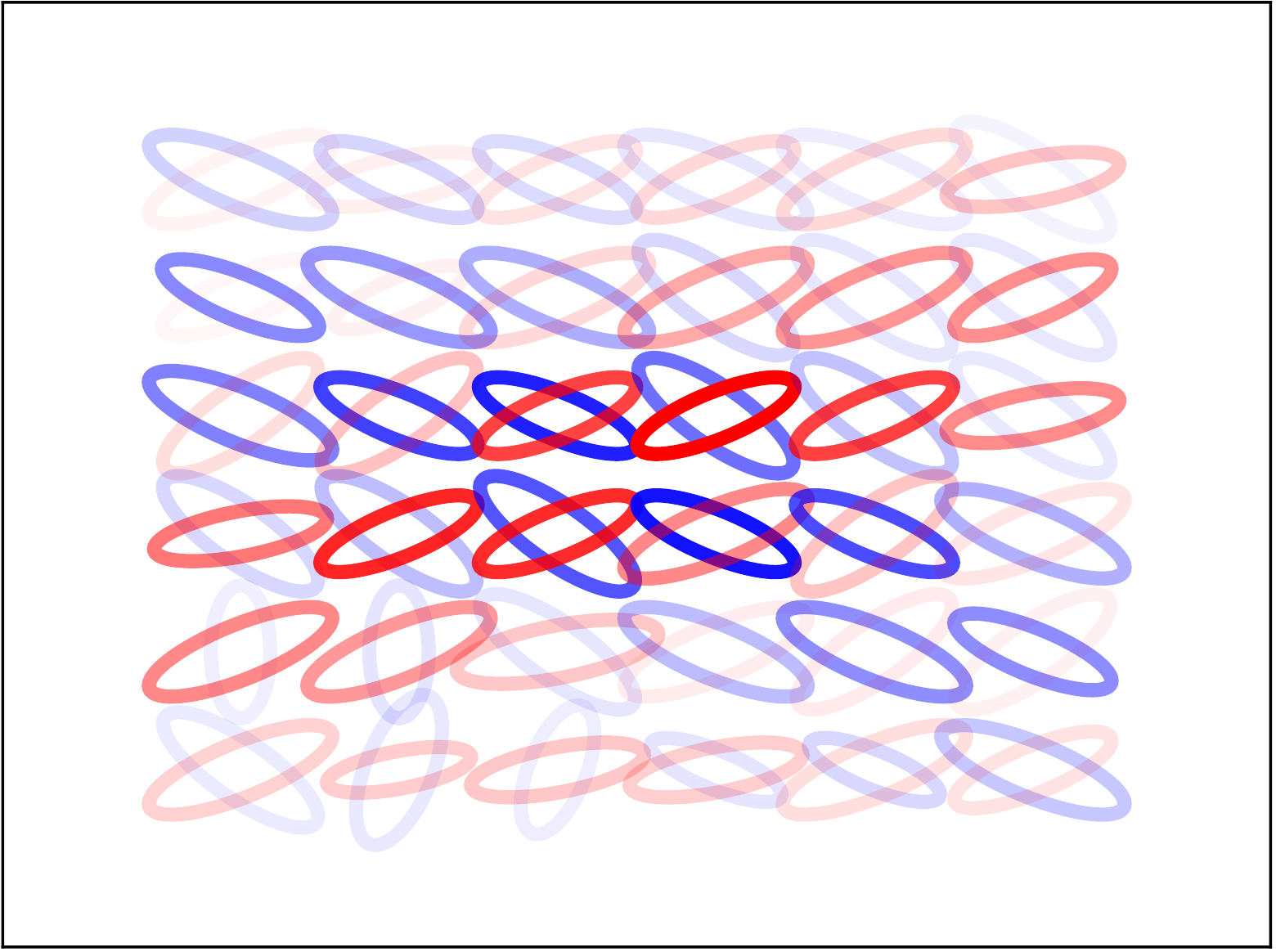}
		\includegraphics[width=0.75\columnwidth]{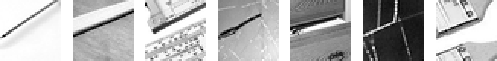} \\
		\phantomsubcaption
		\label{fig:maxpatchesc}
	\end{subfigure}
	\caption{\textbf{Maximum Response Patches.}  (a) Sparse coding with a regularization coefficient of 0.5. (b) Sparse coding with a regularization coefficient of 4.0. (c) ICA. The patches that maximally activated each V2 unit are shown to the right of its visualization. The response strength decreases from left to right. }
	\label{fig:maxpatches}
\end{figure}

\begin{figure}
	\centering
	\includegraphics[width=0.65\columnwidth]{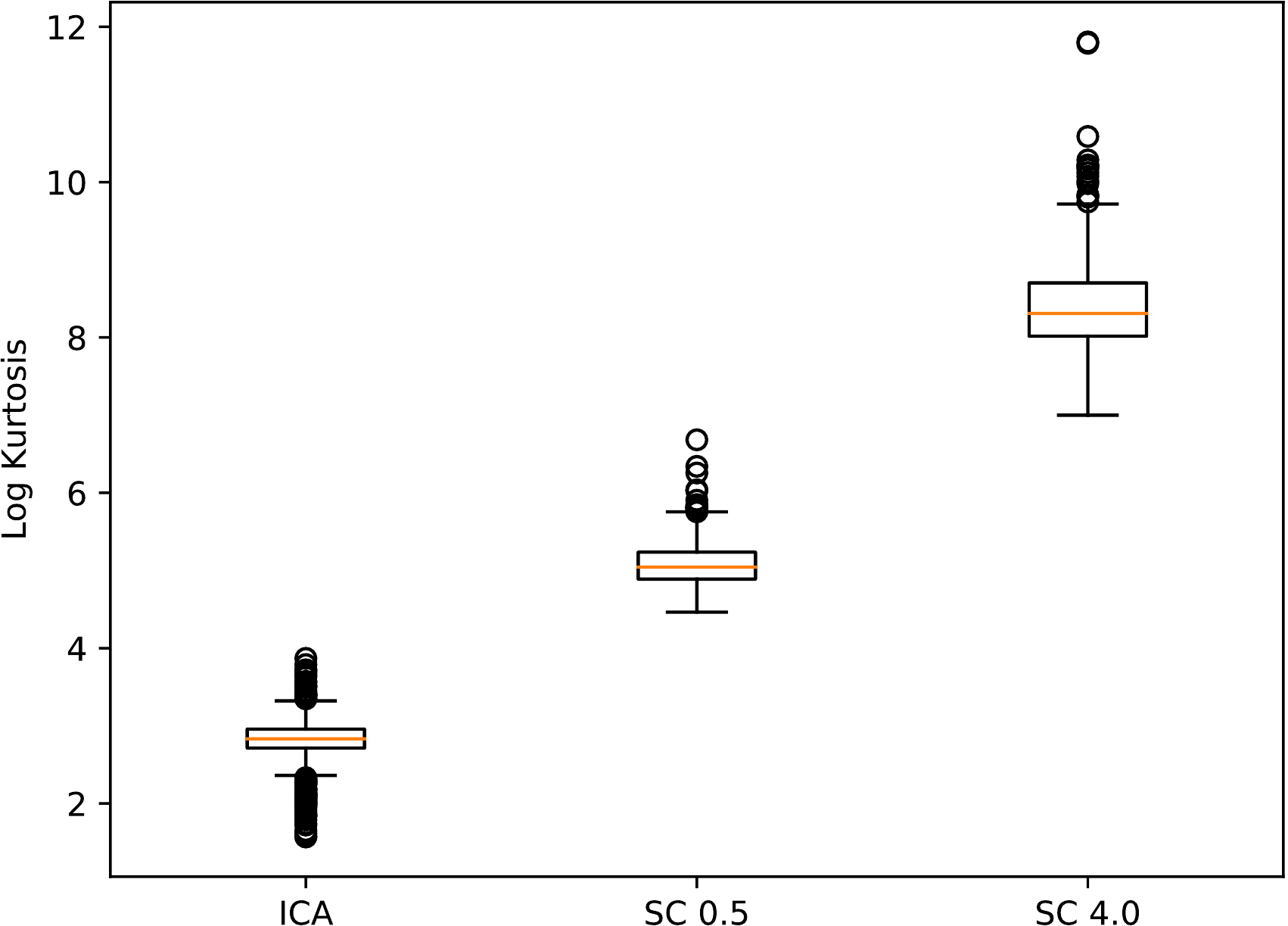}
	\caption{\textbf{Box and Whisker Plot of Kurtosis for All Models.} Plots are generated from the kurtosis over all 400,000 ImageNet patches for each unit. Circles represent outliers. Non-negative sparse coding with a regularization coefficient of 4.0 (SC 4.0) had the highest overall kurtosis, then non-negative sparse coding with an regularization coefficient of 0.5 (SC 0.5), then ICA. Each model finds a different sparse representation. }
	\label{fig:boxplot}
\end{figure}

\begin{figure}
	\Large \textbf{(a)} \\
	\begin{subfigure}[t]{\linewidth}
		\centering
		\begin{subfigure}[t]{0.16\linewidth}
			\centering
			\includegraphics[width=\linewidth]{vis/6x6/sc0.5/0.pdf} \\
			\large \textbf{190.1}
		\end{subfigure}
		\begin{subfigure}[t]{0.16\linewidth}
			\centering
			\includegraphics[width=\linewidth]{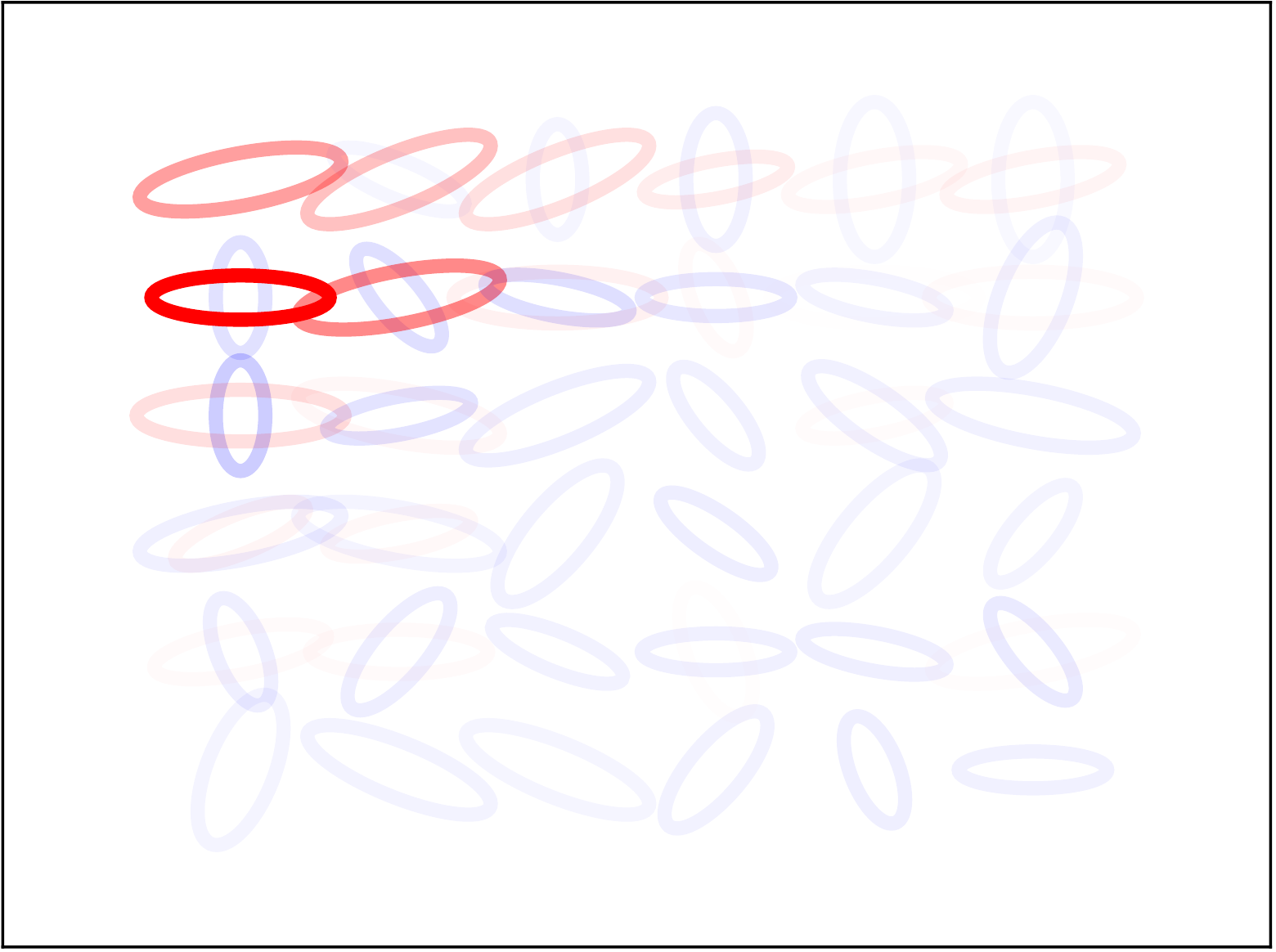}
			\large \textbf{114.5}
		\end{subfigure}
		\begin{subfigure}[t]{0.16\linewidth}
			\centering
			\includegraphics[width=\linewidth]{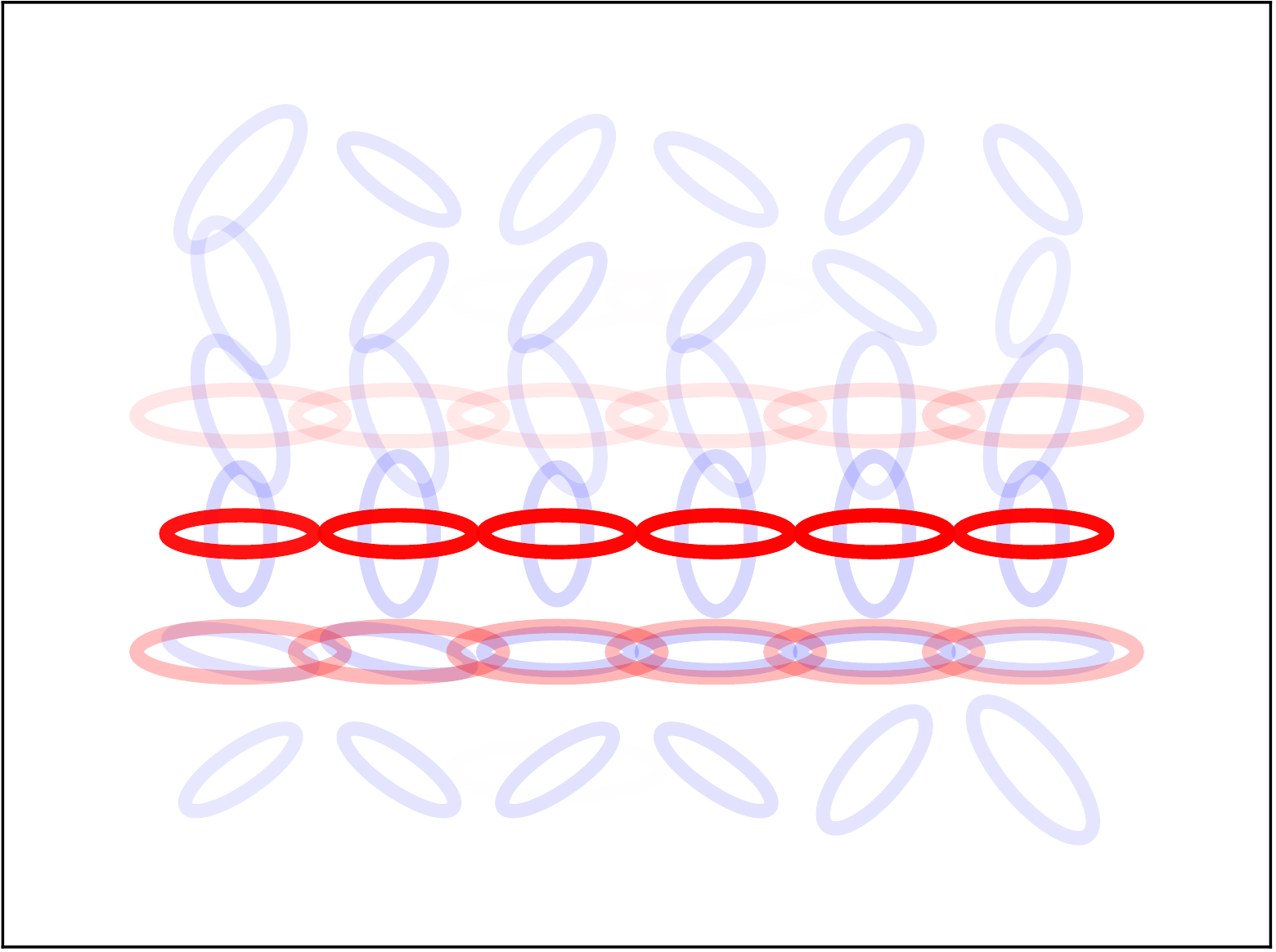}
			\large \textbf{224.0}
		\end{subfigure}
		\begin{subfigure}[t]{0.16\linewidth}
			\centering
			\includegraphics[width=\linewidth]{vis/6x6/sc0.5/12.pdf}
			\large \textbf{567.2}
		\end{subfigure}
		\begin{subfigure}[t]{0.16\linewidth}
			\centering
			\includegraphics[width=\linewidth]{vis/6x6/sc0.5/15.pdf}
			\large \textbf{140.5}
		\end{subfigure}
		\phantomsubcaption
		\label{fig:resppropa}
	\end{subfigure}
	\Large \textbf{(b)} \\
	\begin{subfigure}[t]{\linewidth}
		\centering
		\begin{subfigure}[t]{0.16\linewidth}
			\centering
			\includegraphics[width=\linewidth]{vis/6x6/sc4.0/0.pdf}
			\large \textbf{7860.7}
		\end{subfigure}
		\begin{subfigure}[t]{0.16\linewidth}
			\centering
			\includegraphics[width=\linewidth]{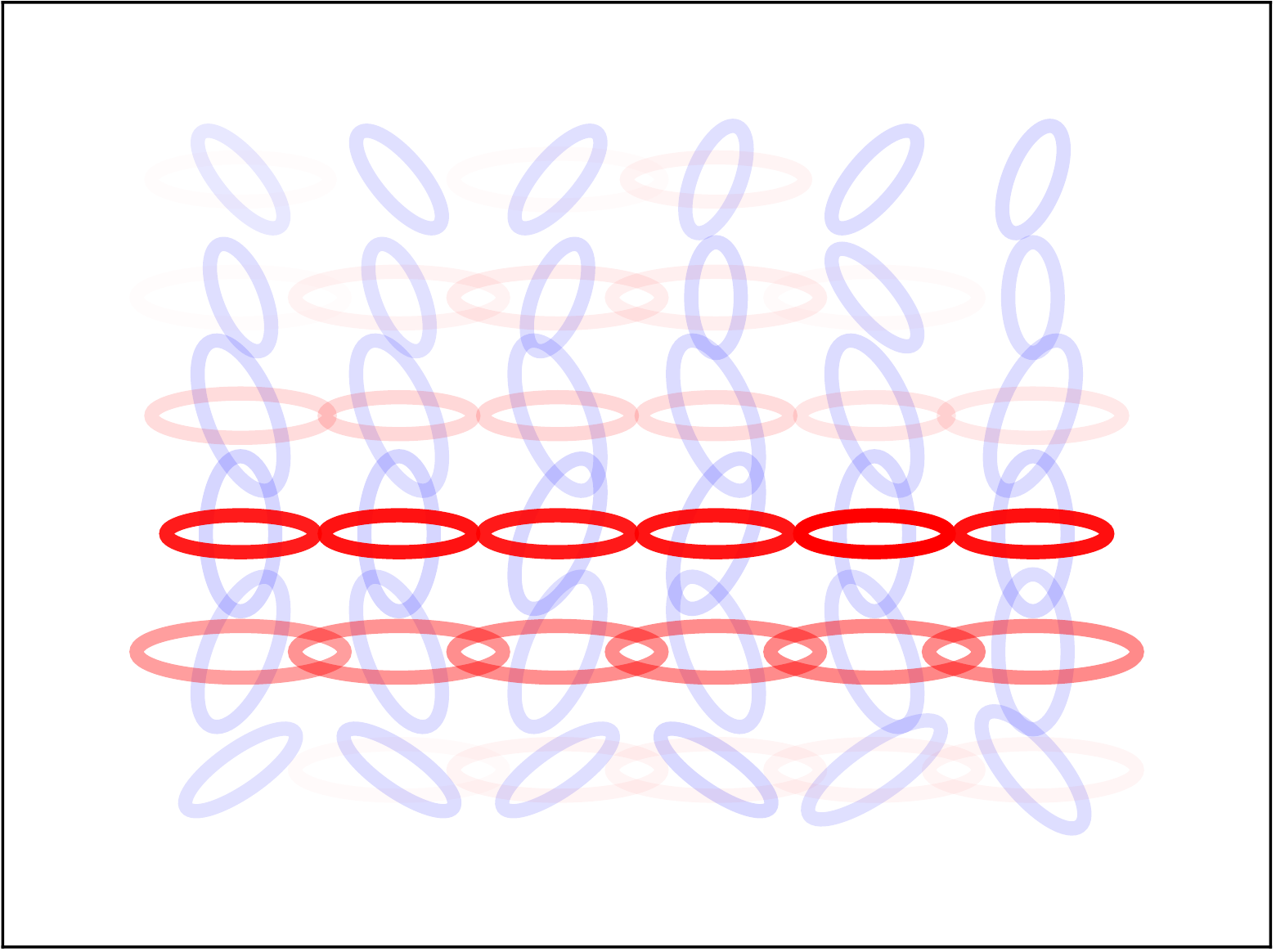}
			\large \textbf{7127.7}
		\end{subfigure}
		\begin{subfigure}[t]{0.16\linewidth}
			\centering
			\includegraphics[width=\linewidth]{vis/6x6/sc4.0/10.pdf}
			\large \textbf{6839.4}
		\end{subfigure}
		\begin{subfigure}[t]{0.16\linewidth}
			\centering
			\includegraphics[width=\linewidth]{vis/6x6/sc4.0/14.pdf}
			\large \textbf{7182.4}
		\end{subfigure}
		\begin{subfigure}[t]{0.16\linewidth}
			\centering
			\includegraphics[width=\linewidth]{vis/6x6/sc4.0/8.pdf}
			\large \textbf{7456.1}
		\end{subfigure}
		\phantomsubcaption
		\label{fig:resppropb}
	\end{subfigure}
	\Large \textbf{(c)} \\
	\begin{subfigure}[t]{\linewidth}
		\centering
		\begin{subfigure}[t]{0.16\linewidth}
			\centering
			\includegraphics[width=\linewidth]{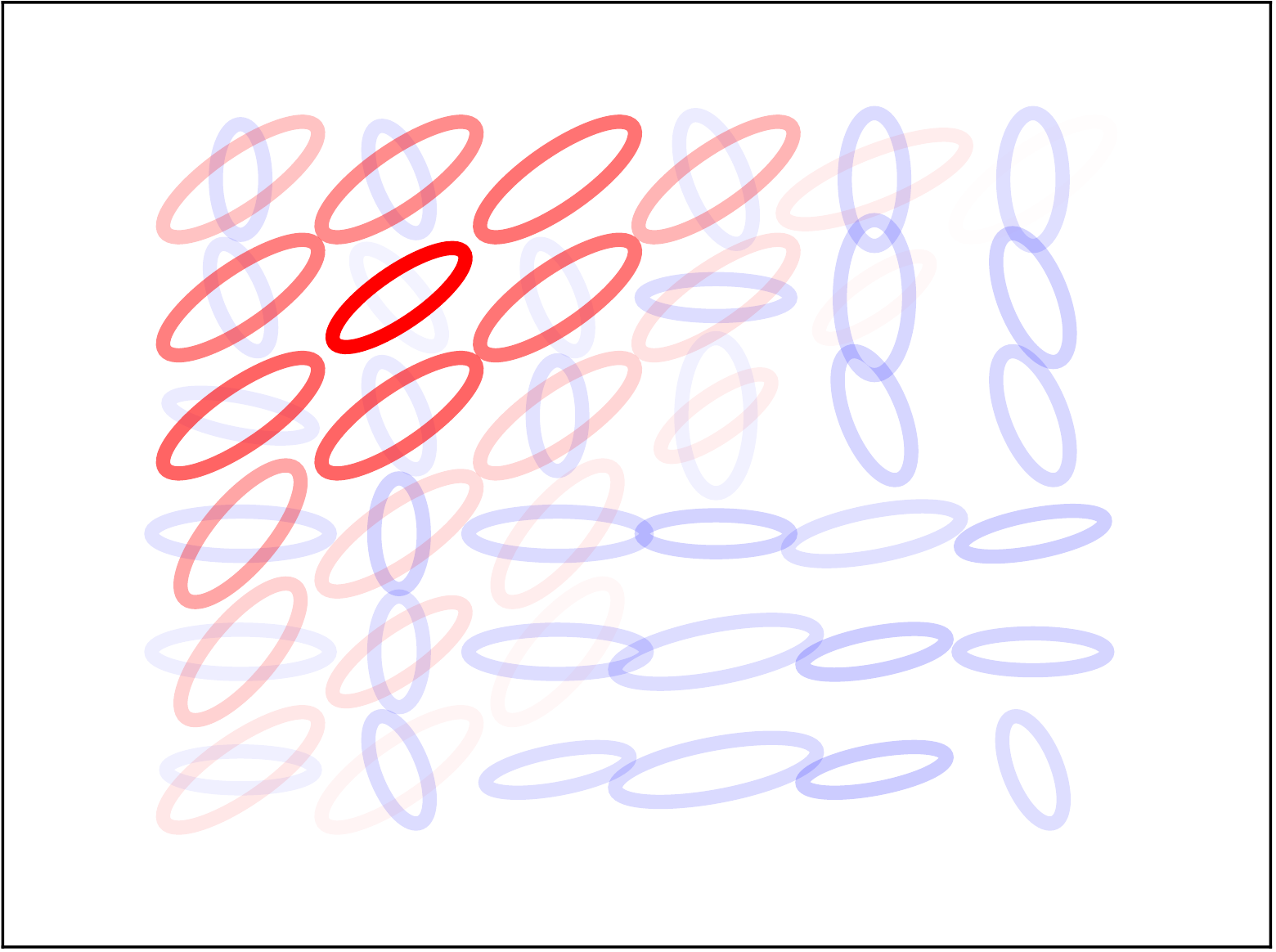}
			\large \textbf{26.2}
		\end{subfigure}
		\begin{subfigure}[t]{0.16\linewidth}
			\centering
			\includegraphics[width=\linewidth]{vis/6x6/ica/10.pdf}
			\large \textbf{16.9}
		\end{subfigure}
		\begin{subfigure}[t]{0.16\linewidth}
			\centering
			\includegraphics[width=\linewidth]{vis/6x6/ica/36.pdf}
			\large \textbf{15.4}
		\end{subfigure}
		\begin{subfigure}[t]{0.16\linewidth}
			\centering
			\includegraphics[width=\linewidth]{vis/6x6/ica/8.pdf}
			\large \textbf{15.5}
		\end{subfigure}
		\begin{subfigure}[t]{0.16\linewidth}
			\centering
			\includegraphics[width=\linewidth]{vis/6x6/ica/42.pdf}
			\large \textbf{19.0}
		\end{subfigure}
		\phantomsubcaption
		\label{fig:resppropc}
	\end{subfigure}
	\Large \textbf{(d)} \\
	\begin{subfigure}[t]{\linewidth}
		\centering
		\includegraphics[width=0.68\columnwidth]{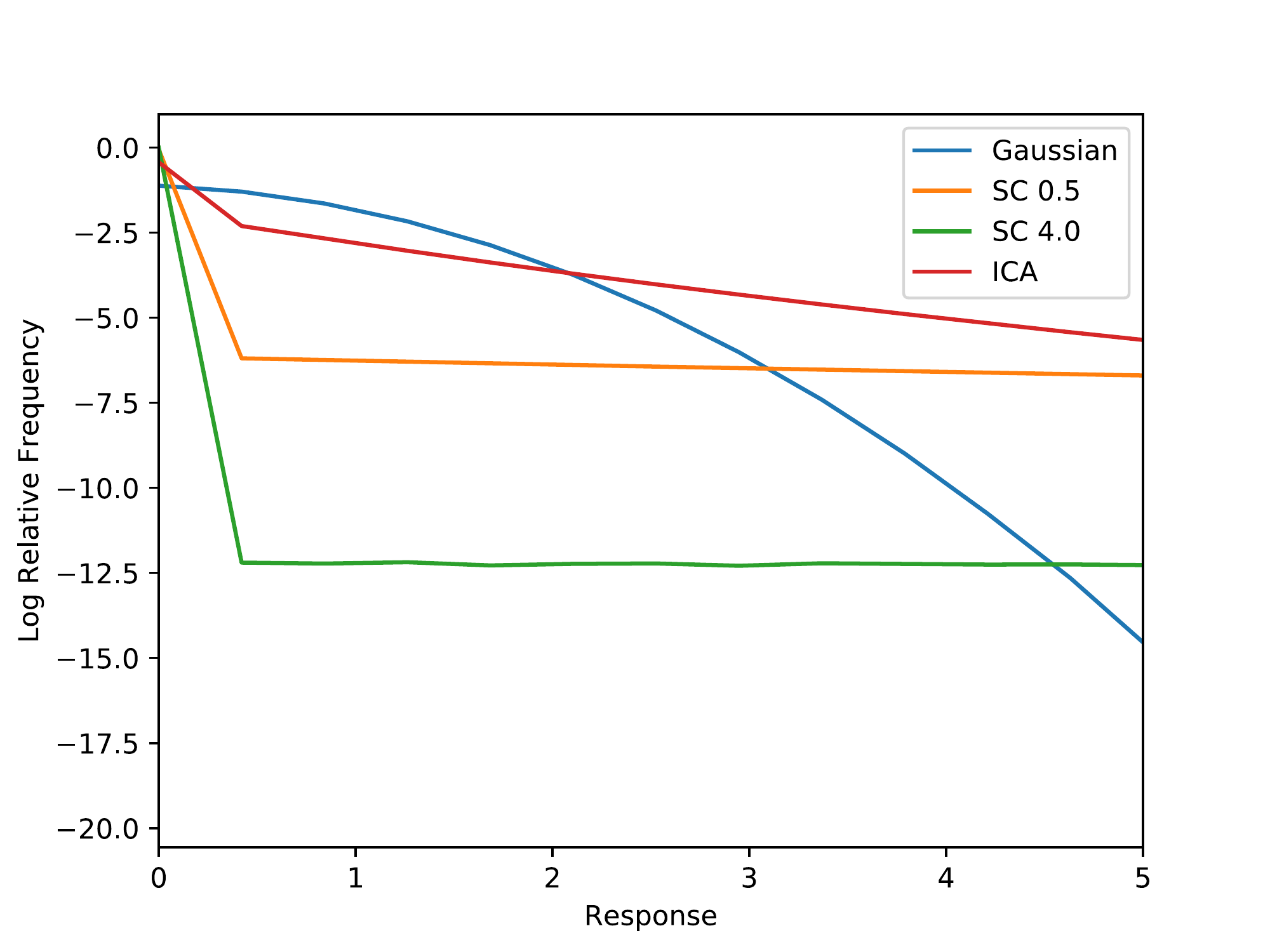}
		\phantomsubcaption
		\label{fig:resppropd}
	\end{subfigure}
	\caption{\textbf{Model Response Properties.} (a-c) Kurtosis of exemplary units of (a) sparse coding with a regularization coefficient of 0.5, (b) sparse coding with a regularization coefficient of 4.0, (c) ICA. High kurtosis indicates more involvement of a unit in reconstructing particular images. (d) Histogram in the Log domain of the responses to all 400,000 image patches for each of the three models. }
	\label{fig:respprop}
\end{figure}

The kurtosis values were much larger for non-negative sparse coding with a regularization coefficient of 4.0 than for a coefficient of 0.5 or overcomplete ICA (see Figure \ref{fig:boxplot}). Exemplary units for each model are shown in Figures \ref{fig:resppropa}, \ref{fig:resppropb}, and \ref{fig:resppropc}. Compared to other units, texture units had a large contribution to individual images (high kurtosis) in non-negative sparse coding, while overcomplete ICA relied often on iso-oriented excitation with broad inhibition units when assigning the largest coefficients. The distributions of the responses for each model (see Figure \ref{fig:resppropd}) were similar to a mixture between an exponential distribution and a delta at zero reflecting the rectification operation. The mixing proportion for the delta component is higher for the models with larger average kurtosis. Non-negative sparse coding distributions were more sparse and had higher kurtosis, with a regularization coefficient of 4.0 being the most sparse.

\subsection{Image Classification}

\begin{figure}
	\centering
	\includegraphics[width=0.65\columnwidth]{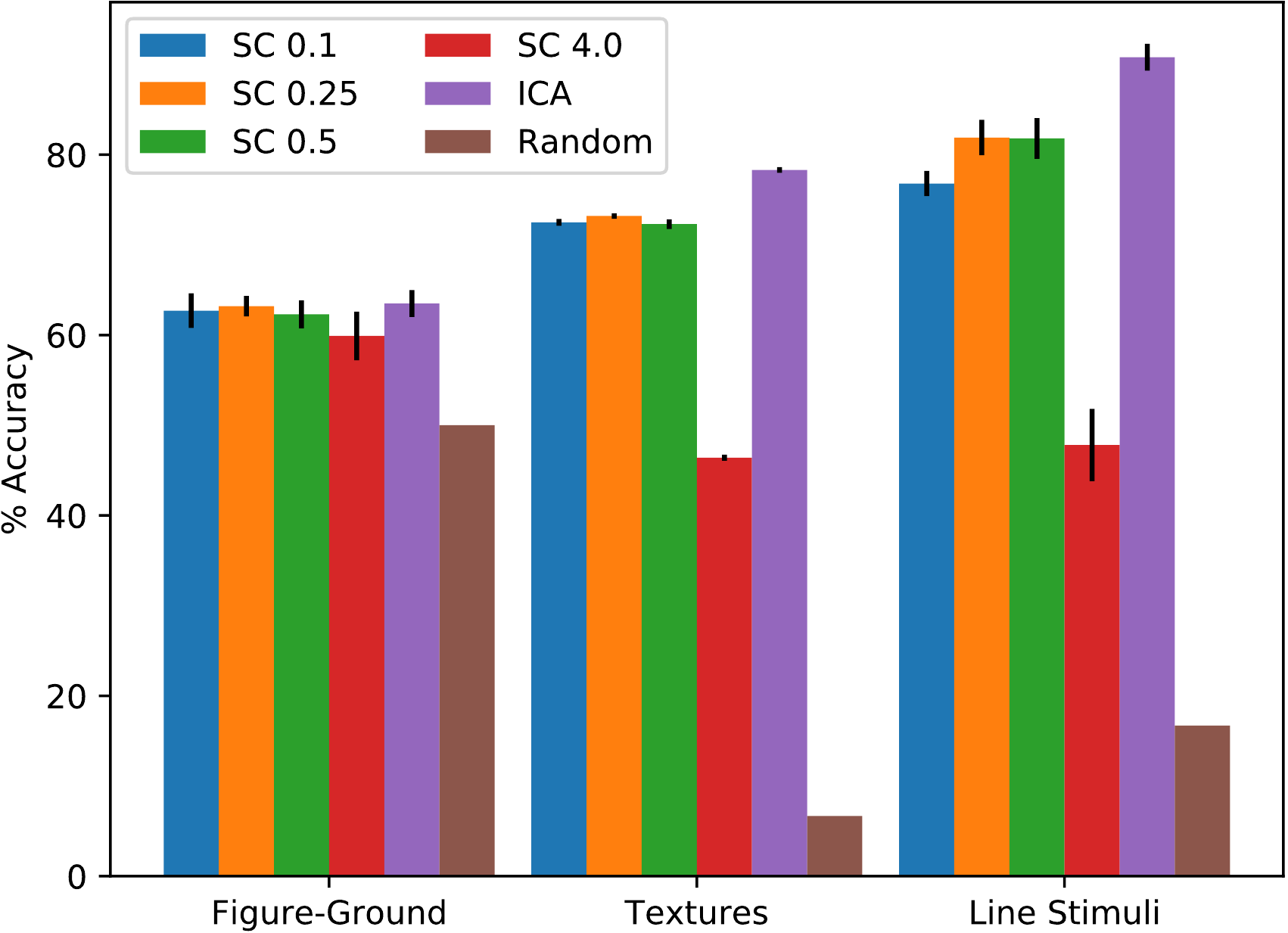}
	\caption{\textbf{Classification Accuracies.}  Average accuracy over 5-fold cross validation for non-negative sparse coding with regularization coefficients of 0.1, 0.25, 0.5 and 4.0 (listed as SC followed by the regularization coefficient) and overcomplete ICA (listed as ICA). Error bars reflect standard deviation over the 5 folds. Random denotes the result of guessing (expectation computed for the number of labels). }
	\label{fig:accbarchart}
\end{figure}

A common metric of vision models is performance on image classification tasks. A few classification tasks were explored here which test the ability to distinguish between figure and ground, multiple texture classes, and the angles between line segments connected at one point. The results for these experiments are shown in Figure \ref{fig:accbarchart}. Overcomplete ICA performed the best. Non-negative sparse coding with a regularization coefficient of 0.5 was competitive with overcomplete ICA on the figure-ground and texture classification tasks, but non-negative sparse coding with a regularization coefficient of 4.0 was only competitive on the figure-ground task. Non-negative sparse coding with a regularization coefficient of 4.0 performed the worst on all tasks. For the figure-ground, texture, and line stimuli tasks, non-negative sparse coding with a regularization coefficient of 0.5 had percent accuracies of 62.3\%, 72.3\%, 81.8\% respectively, non-negative sparse coding with a regularization coefficient of 4.0 had percent accuracies of 59.9\%, 46.4\%, and 47.8\% respectively, and overcomplete ICA had percent accuracies of 63.5\%, 78.3\%, and 90.8\% respectively (all shown to one decimal place). We also report the average number of support vectors used for each manipulation of each experiment in figure \ref{fig:svs}.

\begin{figure}
	\centering
	\begin{tabular}{ |l||c|c|c|c| } 
		\hline
		{} & \textbf{Figure-Ground} & \textbf{Texture} & \textbf{Line Stimuli} \\ 
		\hline
		\textbf{SC 0.1} & 5642.0 & 981.5 & 226.5 \\
		\textbf{SC 0.25} & 5594.0 & 969.9 & 224.8 \\
		\textbf{SC 0.5} & 5669.5 & 976.0 & 221.2 \\
		\textbf{SC 4.0} & 6039.5 & 1232.3 & 222.3 \\
		\textbf{ICA} & 5520.5 & 775.1 & 193.5 \\
		\hline
	\end{tabular}
	\caption{\textbf{Average Number of Support Vectors.}  Average number of support vectors for each experiment over all classes (2 for figure-ground, 15 for texture, and 6 for line stimuli). }
	\label{fig:svs}
\end{figure}

\subsection{Texture Sensitivity}

\begin{figure}
	\centering
	\includegraphics[width=0.65\columnwidth]{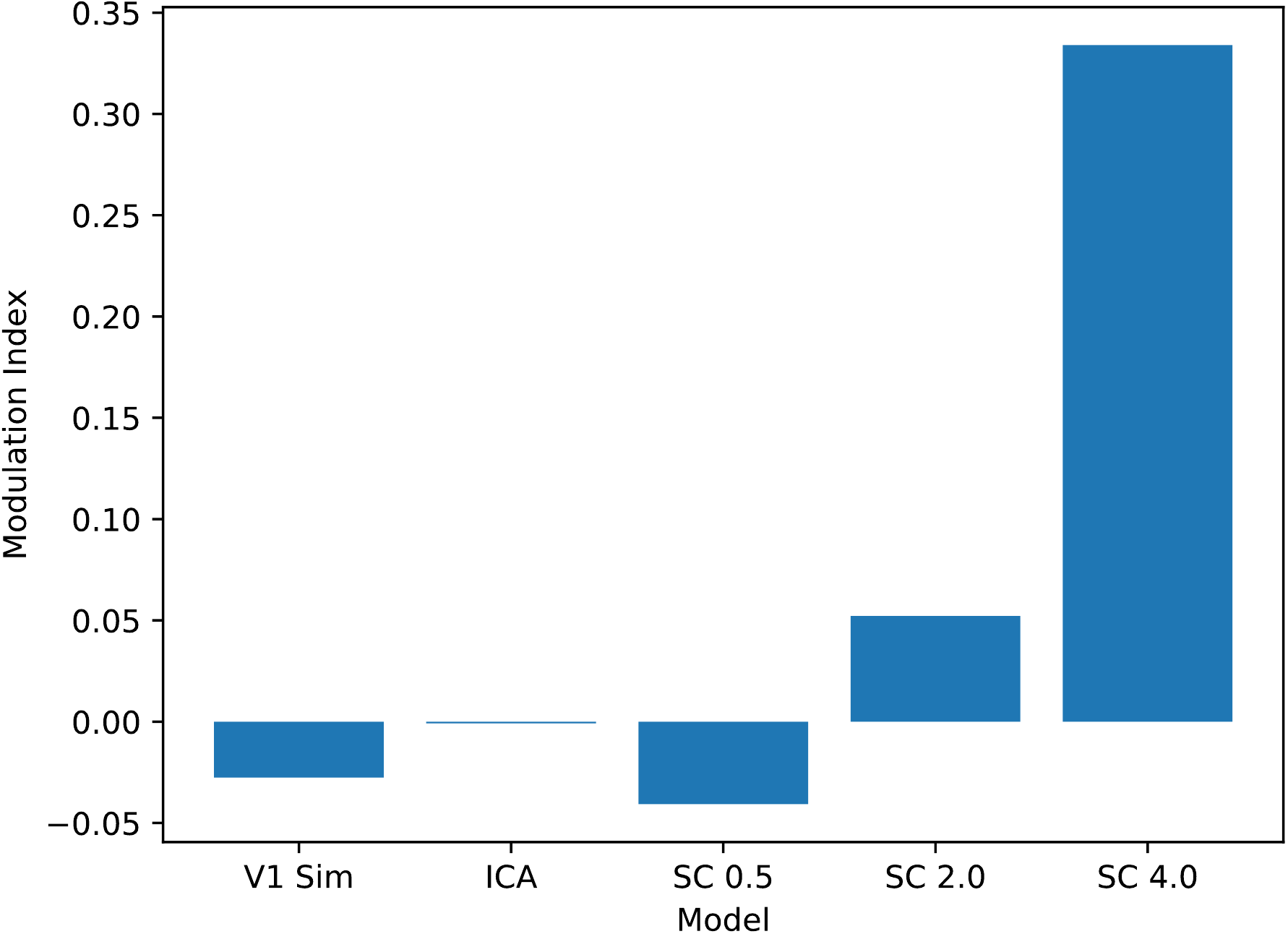}
	\caption{\textbf{Texture Modulation Indices.}  Modulation indices for vision models. The models include the initial simulated V1 stage via Gabor filters (V1 Sim), overcomplete ICA (ICA), and non-negative sparse coding with a regularization coefficient of 0.5 (SC 0.5), 2.0 (SC 2.0), and 4.0 (SC 4.0). }
	\label{fig:modbar1}
\end{figure}

A comparison to human vision can be made by analyzing the responses of these models to textures of varying classes, such as the textures of the second classification experiment and their spectrally-matched noise versions that preserve the amplitude spectrums of the original textures but have randomized phase. Secondary visual cortex shows sensitivity to texture that is absent in V1 \citep{freeman:natneuro13,ziemba:nas16,kohler:jneuro16,laskar:jov20}. For instance, in an fMRI experiment, the modulation index (see \nameref{sec:methods}) for textures versus noise was much larger in V2 than in V1 with an average modulation index of about 0.13 across subjects for V2 \citep{freeman:natneuro13}. We used the same texture and noise stimuli as in the \cite{freeman:natneuro13} studies. For the models studied here, a similar difference in modulation index (between the V1 stage and V2 stage) would suggest that the trend of texture sensitivity in primary and secondary visual cortex is also present in these models. The texture modulation indices for all models were computed by taking the responses to the 30,000 texture patches along with the responses to 30,000 spectrally-matched noise versions of the texture patches, taking the difference between each, and normalizing via the sum of each (see \nameref{sec:methods}) to yield 30,000 modulation indices. Modulation indices for texture-noise pairs that both yielded 0 response were discarded because they did not provide any response information. The modulation indices for overcomplete ICA and non-negative sparse coding with regularization parameters of 0.5, 2.0, and 4.0 are shown in Figure \ref{fig:modbar1}.

\begin{figure}
	\centering
	\includegraphics[width=0.65\columnwidth]{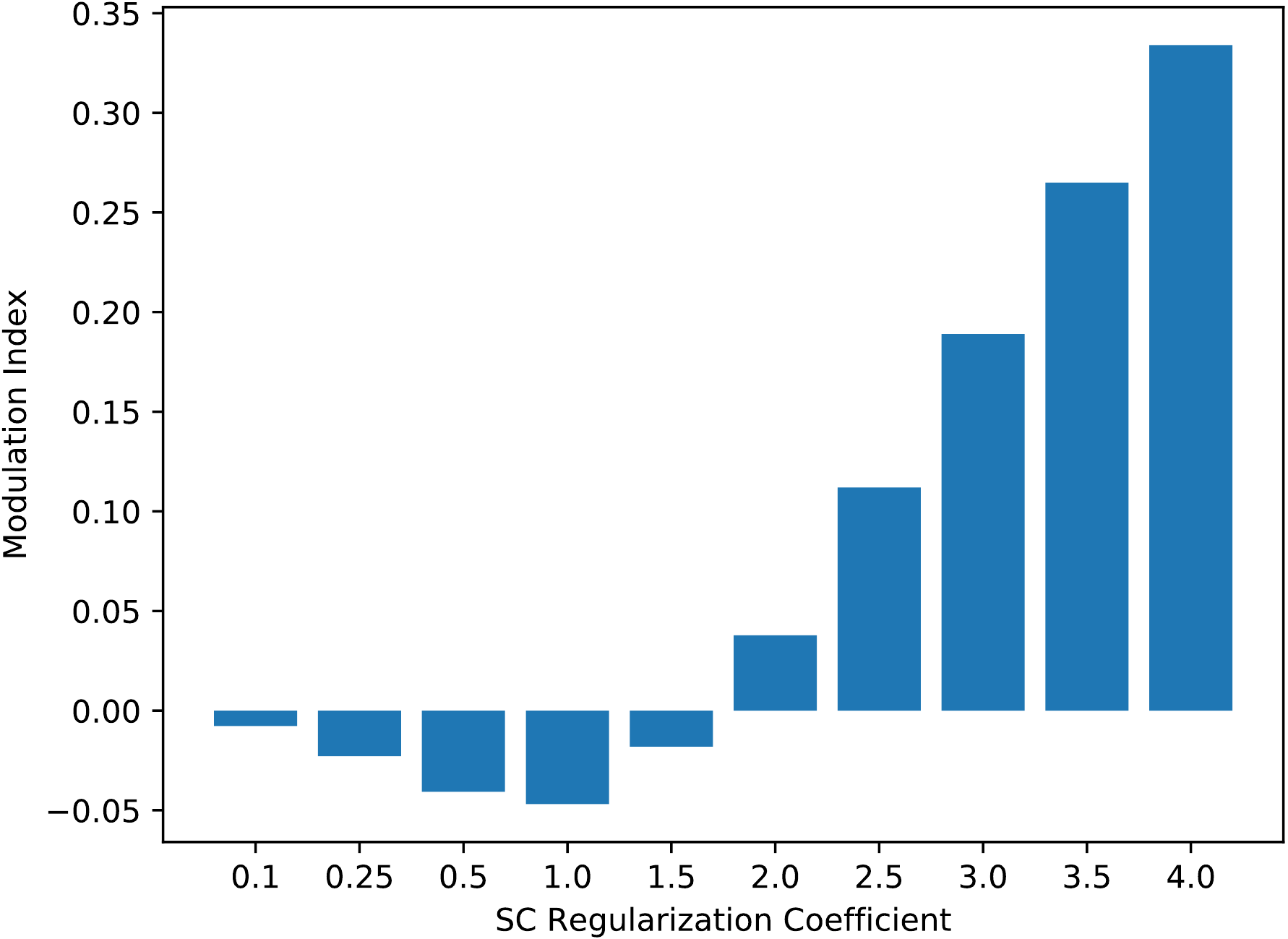}
	\caption{\textbf{Texture Modulation Indices for Non-Negative Sparse Coding with Various Values of the Regularization Coefficient.}  As the regularization coefficient increased, the texture modulation increased. A regularization coefficient of 2.5 yielded a modulation index of about 0.112 which roughly approximates the texture modulation index of V2 as measured by fMRI \citep{freeman:natneuro13}. }
	\label{fig:modbar2}
\end{figure}

The large kurtosis of non-negative sparse coding resulted in many more texture-noise pairs with no response (0 response to the texture and noise image) as the regularization coefficient increased. The percentage of texture-noise pairs with a response for overcomplete ICA was 73.9\% and for non-negative sparse coding with regularization parameters of 0.1, 0.25, 0.5, 1.0, 1.5, 2.0. 2.5, 3.0, 3.5, and 4.0 were 21.7\%, 18.9\%, 15.2\%, 9.73\%, 6.16\%, 3.84\%, 2.38\%, 1.46\%, 0.925\%, and 0.594\% respectively. However, overcomplete ICA does not have a sparsity control, so it could not yield representations with higher kurtosis. Interestingly, as the regularization coefficient of non-negative sparse coding increased, the modulation index increased. Over the range of values tested, a regularization coefficient of 2.5 most closely matched the modulation index of V2. The modulation indices for a few of the values tested are shown in Figure \ref{fig:modbar2}. The increase of the modulation index with sparsity is consistent with previous findings that considered deep neural networks \citep{zhuang:fcn17}.

\subsection{Patch Completion}

\begin{figure}
	\begin{flushleft}
		\hspace{1.5mm} \large \textbf{Image}
		\hspace{9.5mm} \large \textbf{V1}
		\hspace{4.5mm} \large \textbf{V1C Mod}
		\hspace{3mm} \large \textbf{PCA}
		\hspace{9mm} \large \textbf{ICA}
		\hspace{7mm} \large \textbf{SC 2.0}
		\hspace{4.5mm} \large \textbf{SC 4.0}
	\end{flushleft}
	\begin{subfigure}[t]{\linewidth}
		\centering
		\includegraphics[width=\columnwidth]{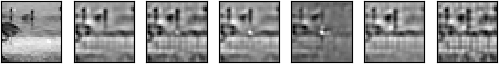}
		\includegraphics[width=\columnwidth]{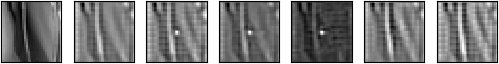}
		\includegraphics[width=\columnwidth]{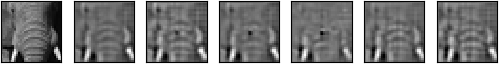}
		\includegraphics[width=\columnwidth]{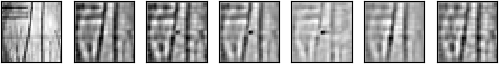}
		\includegraphics[width=\columnwidth]{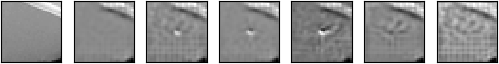}
		\includegraphics[width=\columnwidth]{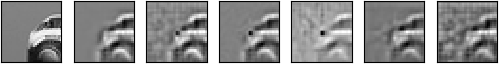}
		\includegraphics[width=\columnwidth]{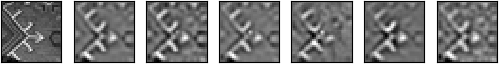}
	\end{subfigure}
	\caption{\textbf{Image Reconstructions of 32x32 Image Patches With a 1x1 Missing V1 Complex Region.} The first column shows the original image, the second the V1 reconstruction of the image, the third the V1 complex reconstruction with 1x1 missing region, the fourth the PCA reconstruction, the fifth the overcomplete ICA reconstruction, the sixth the non-negative sparse coding reconstruction with a regularization coefficient of 2.0, and the last non-negative sparse coding with a regularization coefficient of 4.0. }
	\label{fig:patcomp1}
\end{figure}

\begin{figure}
	\begin{flushleft}
		\hspace{1.5mm} \large \textbf{Image}
		\hspace{9.5mm} \large \textbf{V1}
		\hspace{4.5mm} \large \textbf{V1C Mod}
		\hspace{3mm} \large \textbf{PCA}
		\hspace{9mm} \large \textbf{ICA}
		\hspace{7mm} \large \textbf{SC 2.0}
		\hspace{4.5mm} \large \textbf{SC 4.0}
	\end{flushleft}
	\begin{subfigure}[t]{\linewidth}
		\centering
		\includegraphics[width=\columnwidth]{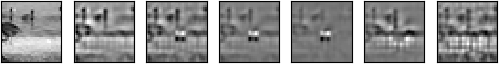}
		\includegraphics[width=\columnwidth]{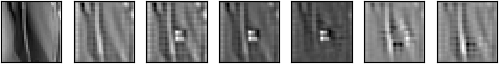}
		\includegraphics[width=\columnwidth]{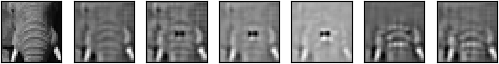}
		\includegraphics[width=\columnwidth]{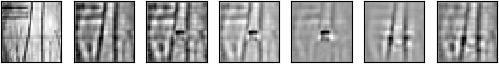}
		\includegraphics[width=\columnwidth]{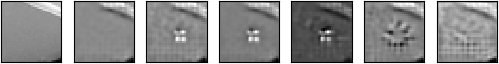}
		\includegraphics[width=\columnwidth]{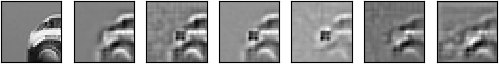}
		\includegraphics[width=\columnwidth]{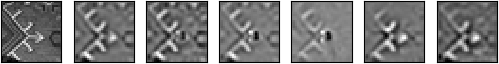}
	\end{subfigure}
	\caption{\textbf{Image Reconstructions of 32x32 Image Patches With 2x2 Missing V1 Complex Region.} The first column shows the original image, the second the V1 reconstruction of the image, the third the V1 complex reconstruction with a 2x2 missing region, the fourth the PCA reconstruction, the fifth the overcomplete ICA reconstruction, the sixth the non-negative sparse coding reconstruction with a regularization coefficient of 2.0, and the last non-negative sparse coding with a regularization coefficient of 4.0. }
	\label{fig:patcomp2}
\end{figure}

A less-studied, but important, metric for vision models is their ability to infer missing structure in images. In this experiment, 1x1 and 2x2 regions were deleted at the level of the V1 complex cell responses. Selected patch reconstructions are shown in Figures \ref{fig:patcomp1} and \ref{fig:patcomp2}. The columns in Figures \ref{fig:patcomp1} and \ref{fig:patcomp2} correspond to the different stages of visual processing in the model starting with the original image. The next two columns show the image reconstruction by the model after Gabor filtering (V1) and energy pooling along with the information removal (V1C Mod). Since the inverse transform of the V1 complex responses was undone exactly by saving the angles between quadrature pair responses during the forward transform, the V1 complex stage did not change the appearance of the reconstruction. For this reason the V1 complex reconstruction without the information removal was omitted (see the V1 representation instead). The next column shows the reconstruction after redundancy reduction with PCA. The final three columns show the model’s final stage of processing with either overcomplete ICA (ICA), non-negative sparse coding with a regularization coefficient of 2.0 (SC 2.0), and non-negative sparse coding with a regularization coefficient of 4.0 (SC 4.0). The mean-squared-error (MSE) for the patches with a 1x1 V1 complex region deleted for non-negative sparse coding with a regularization coefficient of 2.0 was 0.0129, with a regularization coefficient of 4.0 was 0.0218, and for overcomplete ICA was 0.786. The MSE for the patches with a 2x2 V1 complex region deleted for non-negative sparse coding with a regularization coefficient of 2.0 was 0.0355, with a regularization coefficient of 4.0 was 0.0304, and for overcomplete ICA was 2.67. Both non-negative sparse coding manipulations performed much better than overcomplete ICA. Student’s t-tests for independent samples showed that both were significant ($p < 0.01$). The differences between non-negative sparse coding with both values of the regularization coefficient were also significant ($p < 0.01$; t-test for independent samples) with non-negative sparse coding and a regularization coefficient of 2.0 performing better when only a 1x1 V1 complex region was deleted, and with a regularization coefficient of 4.0 performing better when a 2x2 V1 complex region was deleted. Overcomplete ICA reconstructions did not appear to attempt to complete missing information. Interestingly, non-negative sparse coding with different regularization coefficients inferred missing information with different plausible image reconstructions such as how it inferred the car’s bumper and the arrow in the last two rows in Figure \ref{fig:patcomp1}. Also, higher sparsity allowed for better inference when more information was missing in the line structure in the second row and the low spatial frequency region in the fifth row in Figure \ref{fig:patcomp2}.

\section{Discussion}

This work built upon the hierarchical unsupervised learning V2 model of \cite{hosoya:jneuro15}. Their work investigated overcomplete independent component analysis as a sparse coding model, but they did not investigate the original sparse coding model of \cite{olshausen:nat96}. In this paper, the different structures learned by incorporating ICA versus sparse coding in the V2 model were shown, a characterization of the marginal statistics of the filter responses was performed, and the implications for performing inference and classification tasks with these approaches were demonstrated.

This paper examined the tradeoffs of ovecomplete ICA versus non-negative sparse coding, and non-negative sparse coding with different sparsity levels, for vision. Good performance on a single task like image classification does not imply good performance on other tasks like image inference. Furthermore, image classification accuracy may suffer with a certain degree of kurtosis while texture sensitivity becomes more like V2. While seemingly worse for the model, such a change may be tolerated in the light of V2 being an early visual processing area. Perhaps further transformations are needed before classification accuracy increases to the degree observed by overcomplete ICA. However, it is not obvious that the strategy of sparse coding by \cite{olshausen:nat96} should be ascribed to V2, and other sparse coding algorithms may yield better classification accuracy with a representation that is more sparse. While the non-negative sparse coding model here was linked to V2 via the modulation index, other coding strategies may also yield similar values of the modulation index.

In terms of our choice of sparse coding implementation, a theoretical link exists between the sparse coding method of \cite{olshausen:nat96} and the neurally plausible locally competitive algorithm (LCA) \citep{rozell:nc08} approach to sparse coding based on the principles of thresholding and local competition. LCA is a dynamical systems approach to sparse coding that models a neural circuit via membrane potential-like quantities that govern the sparse coding response properties along with inhibitory signals from other sparse coding units (similar to neural inhibition). Interestingly, \cite{rozell:nc08} showed that their approach to deriving the sparse coding responses minimizes, under some constraints, the same loss function used here (LASSO) to perform sparse coding. Thus, results from the algorithm explored in this work can be connected to neurally plausible implementations. However, the LCA strategy mainly accounts for the forward response properties; the basis functions are still derived in a similar fashion to \cite{olshausen:nat96}. The degree of sparseness from the LASSO objective influences the learned basis functions, but it is still possible that other methods of deriving the basis functions may yield better classification results with a higher degree of sparseness, which is a limitation to this work. 

Future work can relate to biological data, by examining how well the model responses (with different sparsity levels) match measures of neural activity.  \cite{hosoya:jneuro15} examined V2 properties in their hierarchical overcomplete ICA model, and one can consider recent natural scenes data such as the large-scale natural scenes dataset (NSD) of \cite{allen:biorxiv21}. Their dataset provides V2 fMRI voxel data in response to viewing natural scenes. The non-negative sparse coding responses can be computed for the natural scenes in order to attempt fitting a linear classifier to the data with the model’s responses.

There is also interest in examining connections to deep convolutional neural networks, which have been shown to capture various cortical neural response properties \citep{kriegeskorte:arvs15,yamins:natneuro16,cadena:ploscb19,kindel:jov19,pospisil:elife18,laskar:jov20}. Such networks can perform a form of sparse coding by thresholding (setting to zero) responses with the ReLU activation function depending on the values of the bias weights (characterized by \cite{bowren:arxiv21}). A deep neural network can be trained with a constraint of several degrees of large negative bias weights to vary sparseness. If classification accuracy suffers and inference improves with larger kurtosis, then the result holds in another type of sparse coding model that has been popular for modeling cortical data. One could further test how the deep neural network models (modified for the different sparsity levels) capture neural data.

Non-negative sparse coding was a natural extension to make the original sparse coding model comparable with positive rectified overcomplete ICA, so it was explored here. Interestingly, the corresponding generative models of the two methods learned to represent images with different structures. Moreover, sparse coding found different structures depending on the degree of sparsity in the model which is fixed in overcomplete ICA. Regular positive and negative sparse coding was also found to yield different structures, but non-negative sparse coding was explored here because it performed better on the classification tasks and eliminated the need for rectification of model V2 units. In comparison to other vision models, the V1 complex cell stage is fixed rather than learned as in other image statistics models of visual cortical complex cells \citep{hyvarinen:vr01,karklin:nat09}.

The differences between overcomplete ICA and non-negative sparse coding may be attributed to the different computational strategies of the two models: overcomplete ICA has an explicit linear forward transform while non-negative sparse coding has an implicit nonlinear forward transform. The strategy of ICA is to find a set of filters whose responses to images are as independent as possible, assuming a linear transform (note that independence is not guaranteed, because of the existence of higher order couplings that cannot be removed by linear transformations, and by overcompleteness). Non-negative sparse coding by contrast does not learn filters, but a dictionary of basis functions that optimally reconstructs images with a linear combination of a few (depending on the regularization coefficient) of its basis functions. However, the different objectives of overcomplete ICA and non-negative sparse coding were not designed to classify images or infer unseen image information respectively. Roughly independent filter responses are not obviously better than sparse coding responses for image classification, and the advantage was not large compared to the best non-negative sparse coding configuration. For the image inference task, image reconstruction error grew as the regularization coefficient increased, but the reconstruction error for the original unmodified image decreased despite having the opposite effect on the input reconstruction error. In other words, the reconstruction error for the image without the deletion in the V1 complex stage was better for a regularization coefficient of 0.5 than 4.0, but when the deletion was present the opposite was true. The better reconstruction of the original image can be seen qualitatively in Figures \ref{fig:patcomp1} and \ref{fig:patcomp2} where 1x1 or 2x2 input regions were deleted. For example, consider the back-bumper of the car in row 6 of Figure \ref{fig:patcomp1}; if the model were simply performing reconstruction, the missing bumper region (blank space) of the car would have been reconstructed, but the model introduces new information into the image representation via its basis functions. The difference in performance on the image tasks was not an obvious result of the difference in loss functions, and previous patch completion (in-painting) results like that of \cite{mairal:cvpr09} only investigated inference in a single-layered model with smaller receptive fields and one level of sparsity. Here, the result of inference could be seen in single patches rather than reconstructing an entire image from its constituent image patches, and more importantly, the result across various levels of sparsity was also demonstrated. Also, another difference between sparse coding and ICA is that while ICA may be thought of as similar to sparse coding in the complete case, ICA tends to maximize coherence (redundancy) in its filter matrix when extended to the overcomplete case \citep{livezey:jmlr19}. In maximum-likelihood inspired ICA models, this is usually addressed by adding a coherence control to the loss function. Score matching ICA was incorporated in this model, similar to \cite{hosoya:jneuro15}, and while coherence control was not explicitly enforced, score matching provides an implicit, albeit data dependent, form of coherence control.

It was found that the resulting non-negative sparse coding units contained intuitively useful geometric primitives such as curves and corners (Figures \ref{fig:6x6gpsa} and \ref{fig:6x6gpsb}), unlike overcomplete ICA which found the units defined by \cite{hosoya:jneuro15} shown in Figure \ref{fig:6x6gpsc} mentioned in the \nameref{sec:methods}. \cite{hosoya:jneuro15} obtained orientation-convergent units, which they noted might be related to corner detection. Other hierarchical models have also resulted in structures such as curves and corners, including a two layer sparse deep belief net model \citep{lee:nips07}, a two layer model that included a statistically optimal divisive normalization at the V1-like stage \citep{coen:jov13}, and the second layer of particular deep convolutional neural networks \citep{zeiler:eccv14}. There is a need in future work to study differences in the resulting structure learned across different classes of models and computations, including intermediate layers of deep convolutional neural networks. In this work, the focus was on a fixed architecture and the influence of ICA versus sparse coding.

For non-negative sparse coding, as the regularization coefficient increases, the sizes of the geometric primitives increase because the model is constrained to represent entire images with only a few basis functions, so each basis function must contain more information to reduce reconstruction error. To see this more clearly, consider the extremes of the regularization coefficient: a coefficient of zero removes the L1 penalty term while increasing the coefficient excessively leads to fewer and fewer basis functions reconstructing the image until it is reconstructed as an image of all zeros with no basis functions. Reconstruction error increases with higher values of the regularization coefficient because the input representation is modified: more information has to be inferred. A similar effect was seen with sparse autoencoders trained on handwritten digits when the degree of sparsity was increased \citep{makhzani:nips15}. This inductive inference mechanism has potential use when sending information down a noisy pipe. When parts of the signal become corrupted, these parts can be safely interpolated by the contributions of the basis functions. When the regularization coefficient is low, these parts of the signal are more likely to be interpreted by the model as genuine parts of the signal. When the regularization coefficient is high, these parts of the signal are inferred over because the model does not have coefficients to spare on perturbations which are not well represented by the dictionary if the model is trained on uncorrupted signals. This inductive bias mechanism may be useful in the brain as well because of the stochasticity in the firing of neurons.

In addition to geometric primitives, non-negative sparse coding (with both choices of regularization coefficient shown in Figures \ref{fig:6x6gps} and \ref{fig:11x11gps}) also finds units that maximally respond to texture-like repeating patterns. \cite{hosoya:jneuro15} found some indication of more localized texture patterns with overcomplete ICA, but it was found here that the non-negative sparse coding units by contrast produced texture-like units covering the full extent of the receptive field. Texture patterns were also apparent in intermediate layers of deep convolutional neural networks \citep{zeiler:eccv14}. In practice, some of the unit types of non-negative sparse coding and overcomplete ICA were both maximally excited by similar images (Figure \ref{fig:maxpatches}). In non-negative sparse coding, lines that stop abruptly were maximally excited by images similar to those that maximally excite iso-oriented excitation with end inhibition units in overcomplete ICA: images where a line stops before reaching the end of the image. However, most of the non-negative sparse coding unit structures were different from those of overcomplete ICA. Non-negative sparse coding was also much more sparse than overcomplete ICA (Figures \ref{fig:boxplot} and \ref{fig:respprop}). The distribution of non-negative sparse coding units covers a different range of sparseness as measured by kurtosis (Figure \ref{fig:boxplot}) compared to overcomplete ICA. If non-negative sparse coding (with a certain regularization coefficient) is advantageous for vision tasks, this sparseness could be motivated by an efficient coding paradigm.

Non-negative sparse coding with the appropriate regularization coefficient better matched the level of texture sensitivity in V2 as measured with the texture modulation index. The modulation indices were computed for the Brodatz texture dataset \citep{brodatz:dover66} in order to determine if the texture sensitivity level would increase with an increasing regularization coefficient, but it is important to note that other texture datasets, like the one in \cite{freeman:natneuro13}, may yield different optimal coefficients, so the exact values should not be viewed as constants for the optimal V2 texture sensitivity match. Instead, the biological link is the increase in texture sensitivity with the increase in sparseness in the non-negative sparse coding model up to a point. The ability of sparse coding to derive bases with varying levels of sparsity allowed it to derive bases with varying levels of texture sensitivity as noted by \cite{zhuang:nas21} with other sparse models. It would also be interesting to see if the change in kurtosis of the spectrally-matched noise images due to the elimination of higher order statistics would elicit a model response pattern similar to that found in an fMRI study by \cite{puckett:neuroimage20} where humans viewed natural scenes degraded of higher order statistics.

Interestingly, while non-negative sparse coding had a high level of texture sensitivity (modulation index of 0.334), its performance on the classification tasks were poorer than that of non-negative sparse coding with a regularization coefficient of 0.5. The implication is that a model which spans the range of kurtosis of a low and high kurtosis sparse coding may better match V2. More on this approach is discussed later, however the main takeaway is that, within a sparse coding framework, texture sensitivity may be increased at the expense of classification accuracy (especially angle classification). This was likely due to the larger receptive fields of sparse coding with a higher regularization coefficient.

Non-negative sparse coding performed worse overall on image classification than overcomplete ICA followed by point-wise rectification (see Figure \ref{fig:accbarchart}). However, vision is a rich process that pertains to far more than distinguishing between classes of images. Vision systems must learn to perform inference when information is absent or lost due to error. Non-negative sparse coding seems to address this, but not rectified overcomplete ICA (see Figures \ref{fig:patcomp1} and \ref{fig:patcomp2}). Perturbations to the V1 complex representation were simply maintained by rectified overcomplete ICA while non-negative sparse coding derives a representation much closer to the original V1 representation before perturbations were introduced. It is likely non-negative sparse coding with a regularization coefficient of 4.0 suffered the most on the line classification task because the receptive fields of most units were too large to pick up on the angle between the two lines. However, the rectification step in overcomplete ICA seemed to remove structural information in the ICA representation, but this step was found to be necessary to maintain its high image classification accuracy; image classification accuracy fell significantly and performed worse than non-negative sparse coding when rectification was not incorporated. Non-negative sparse coding has an implicit built-in rectification mechanism, so it did not experience a similar effect; rather the learned representations were derived such that no negative responses were needed. Interestingly, non-negative sparse coding reconstructed images and completed missing regions, so while it suffered at image classification, it excelled at image inference and overcomplete ICA experienced the opposite. An interesting question, but beyond the scope of this work, is exploring the existence of any asymmetries in the activations of non-negative sparse coding, especially given the fact that changing the sign of an activation at the V2 level does not merely correspond to a contrast inversion.

Also, different reconstructions were formed with different values of the regularization coefficient. Larger values of the regularization coefficient led to representations that were more sparse and had more latent information introduced by the model. Smaller values of the regularization coefficient led to representations that were more faithful to the original V1 representation. Indeed, when only a 1x1 V1 complex region was deleted, a regularization coefficient of 2.0 yielded a smaller MSE, however, when a 2x2 V1 complex region was deleted a regularization coefficient of 4.0 yielded a smaller MSE. This is consistent with the idea that representations that are more sparse introduce more prior knowledge into the representation through the model’s basis functions. When the model was constrained to make due with fewer basis functions, more information had to be inferred. Depending on the amount of missing information, different degrees of sparsity were more useful in building the model representation that best explained the data.

One question that arises when attempting to add inference and content generation mechanisms to vision models is where in the brain do such mechanisms exist? Inference within the receptive field may occur throughout the visual cortex and is harder to localize, but complete image generation (imagining images) can be studied with fMRI. \cite{despoito:npsych97} asked subjects to imagine images while in a fMRI machine and found that the visual association cortex was activated, but not the primary visual cortex. It seems content generation arises in higher visual areas and should not be expected from low-level vision models like the one described in this work. Instead, only low-level local-region inference might be expected in the early visual system. The degree of sparsity expected might be related to the vast overcompleteness in the primary visual cortex \citep{olshausen:vn14}. See \cite{olshausen:wavelets09} for an application of very overcomplete and sparse coding.

One future direction may be to attempt to incorporate different degrees of sparsity into one overall model that is both able to reconstruct images with low error and perform inferences that best explain the data. Such a model would learn a representation that spans the range of kurtosis distributions in Figure \ref{fig:boxplot}. Another direction is applying the perturbations of Figures \ref{fig:patcomp1} and \ref{fig:patcomp2} to the original image. This way, missing information in lower level receptive fields may be inferred via higher level sparse coding, and future higher level models may be shown to complete larger regions of images. The original image was not modified in this work because of limitations in the underlying V1 complex energy model. The energy model pooled V1 responses by taking the magnitude of each quadrature pair of Gabor filters in polar coordinates, but discarding the phase. Because the model contains a large spatial stride, when reconstructing images with a randomized phase a large part of the image structure was lost and inference was unfeasible without attempting a method of phase recovery \citep{gerchberg:optik72}.

In the future, more Gabor filters could be applied at more spatial locations, orientations, and phases to better recover the original image structure. An appealing approach would be to learn a representation with sparse coding for the V1 representation coupled with some form of pooling. Besides pooling after V1, the sparse coding derived V2 responses can be enhanced. A natural improvement is to make the model convolutional \citep{szlam:arxiv10} to reduce the redundancy in the learned basis functions. One could also include a more sophisticated approach to sparse coding, such as approximating variance structure with a nonlinear model as in \cite{karklin:nc05}. This gives the model an understanding of the underlying distribution of the sparse coding responses. Yet another appealing sparse coding approach is to learn both of the bases at the same time as done by \citep{boutin:plos21,zeiler:citeseer10}. The role of nonlinear computations such as divisive normalization motivated by image statistics \citep[e.g.,][]{coen:jov13} can also be explored within such models. All these methods allow for perturbations to be introduced in the original image.

\section{Conclusion}

Non-negative sparse coding discovers a unique set of intuitively useful basis functions that with different degrees of sparsity may be advantageous for particular vision tasks. Overcomplete ICA performs better than non-negative sparse coding on image classification, but performs poorly on image inference. With a high degree of sparsity in a high-level visual model, non-negative sparse coding is able to infer small regions of missing information. The inference mechanism postulated here is feasible.

\section{Appendix A}\label{sec:appA}
\subsection{Classification Results With All Models and Manipulations}
\begin{figure}[H]
	\centering
	\captionsetup[sub]{font=Large,labelfont={bf,sf}}
	\begin{subfigure}[h]{\columnwidth}
		\includegraphics[width=0.95\columnwidth]{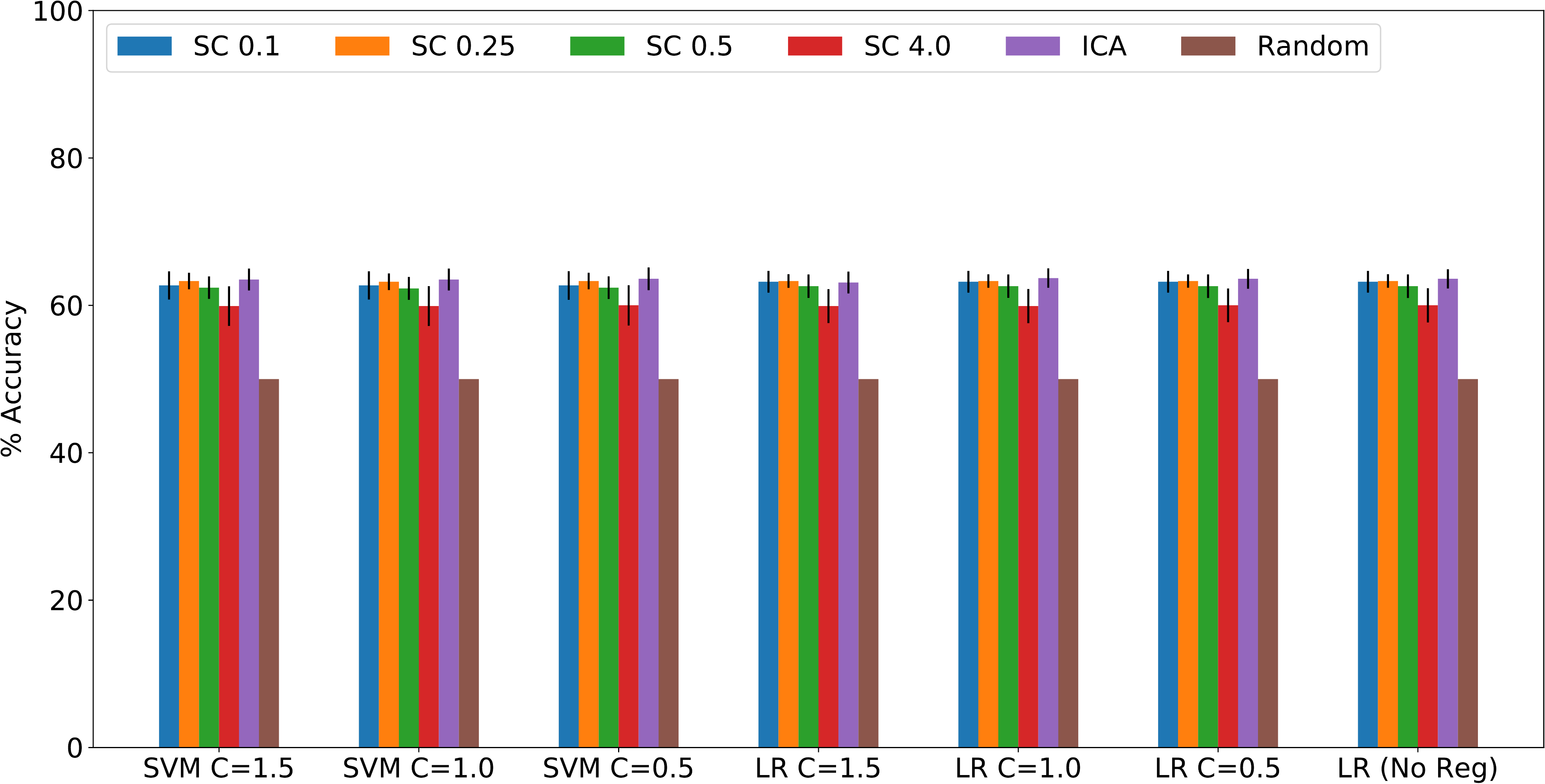}
		\caption{}
	\end{subfigure}
\end{figure}
\begin{figure}[H]
	\ContinuedFloat
	\centering
	\captionsetup[sub]{font=Large,labelfont={bf,sf}}
	\begin{subfigure}[h]{\columnwidth}
		\includegraphics[width=0.95\columnwidth]{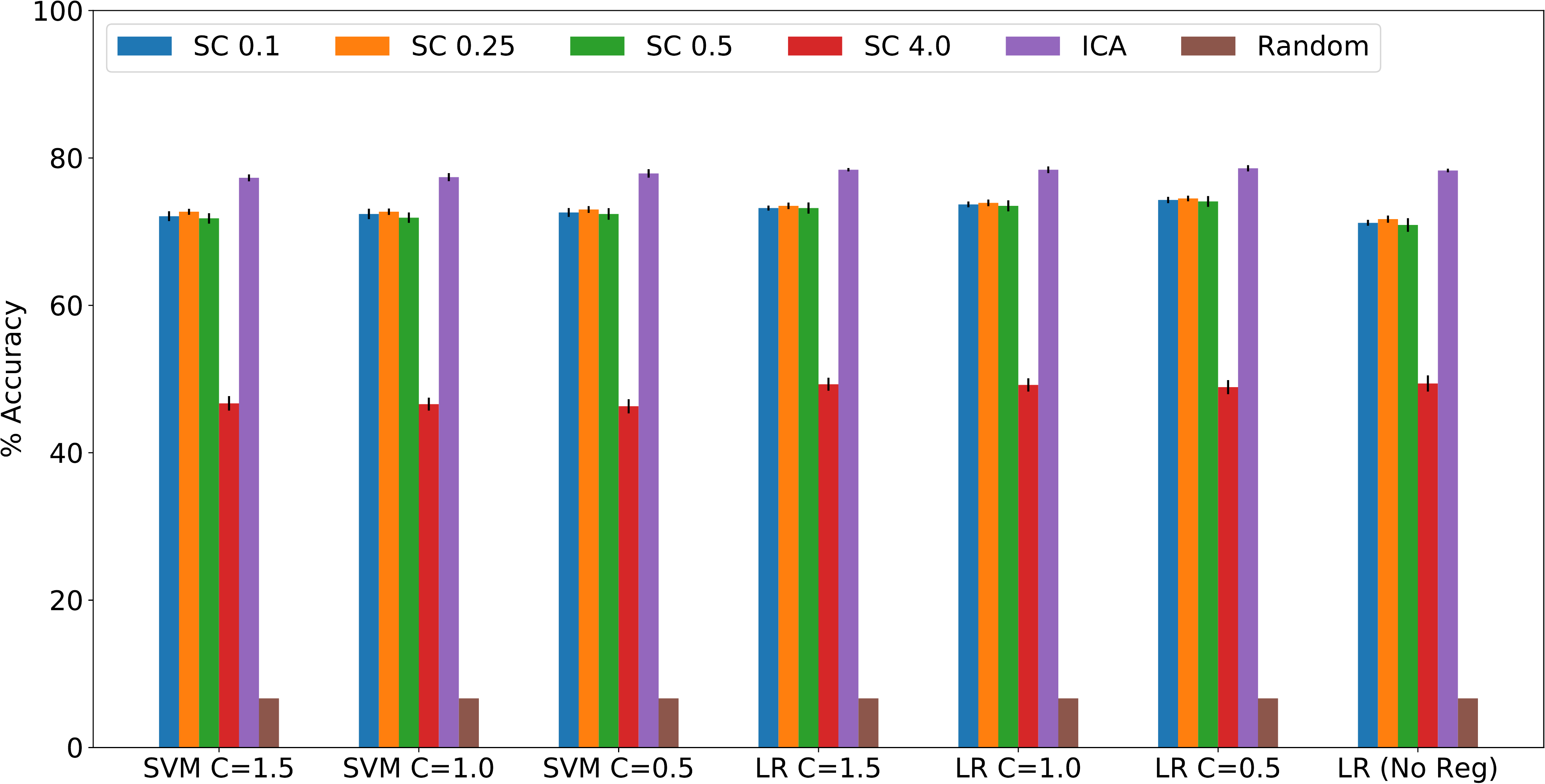}
		\caption{}
	\end{subfigure}
\end{figure}
\begin{figure}[H]
	\ContinuedFloat
	\centering
	\captionsetup[sub]{font=Large,labelfont={bf,sf}}
	\begin{subfigure}[h]{\columnwidth}
		\includegraphics[width=0.95\columnwidth]{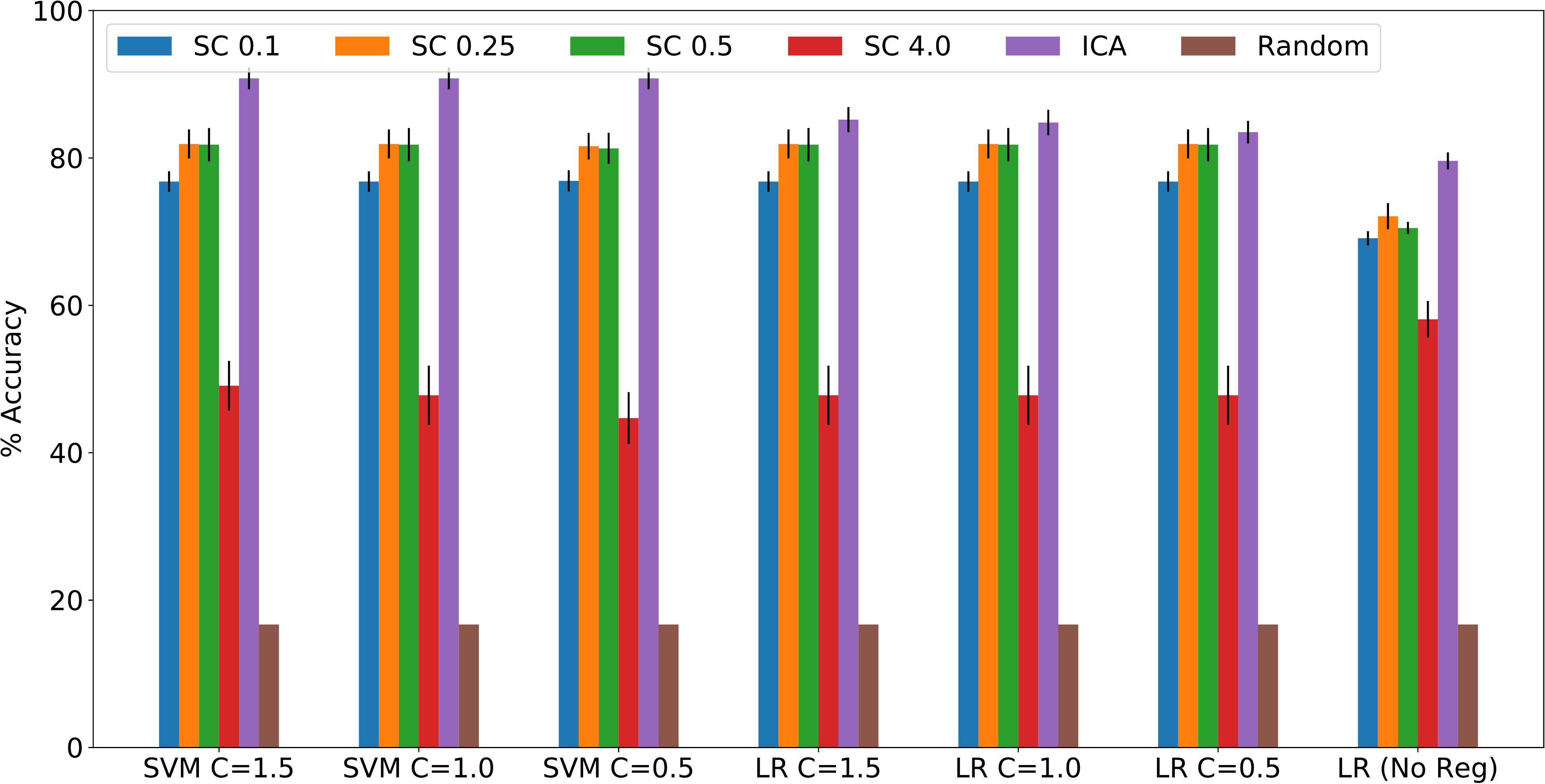}
		\caption{}
	\end{subfigure}
	\caption{\textbf{Classification Results For All Models and Manipulations.}  Average accuracy over 5-fold cross validation for non-negative sparse coding with regularization coefficients of 0.1, 0.25, 0.5 and 4.0 (listed as SC followed by the regularization coefficient) and overcomplete ICA (listed as ICA). Error bars reflect standard deviation over the 5 folds. Random denotes the result of guessing (expectation computed for the number of labels). }
	\label{fig:rbarchart1}
\end{figure}

\section{Appendix B}\label{sec:appB}
\subsection{Full Size Texture and Noise Images}
\begin{figure}[H]
	\centering
	\includegraphics[width=0.17\columnwidth]{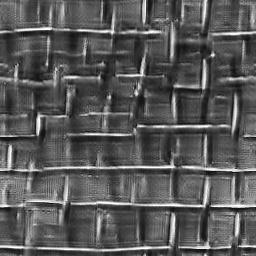}
	\includegraphics[width=0.17\columnwidth]{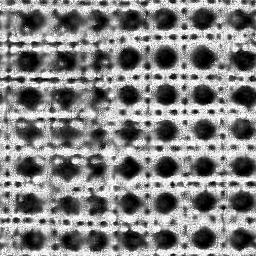}
	\includegraphics[width=0.17\columnwidth]{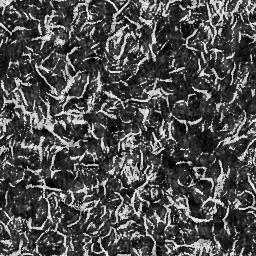}
	\includegraphics[width=0.17\columnwidth]{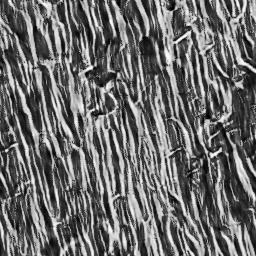}
	\includegraphics[width=0.17\columnwidth]{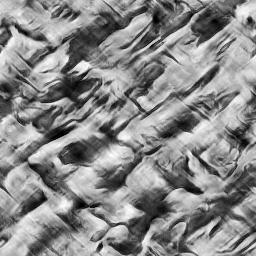}
	\includegraphics[width=0.17\columnwidth]{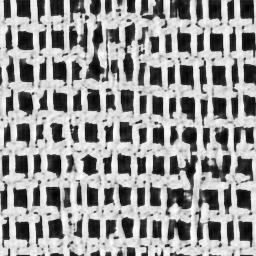}
	\includegraphics[width=0.17\columnwidth]{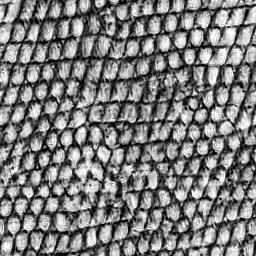}
	\includegraphics[width=0.17\columnwidth]{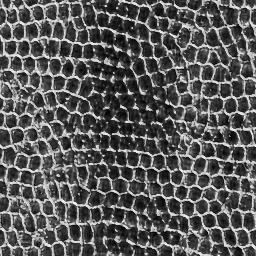}
	\includegraphics[width=0.17\columnwidth]{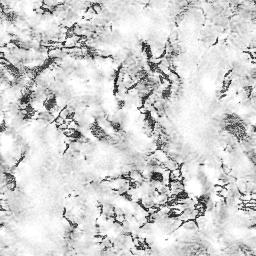}
	\includegraphics[width=0.17\columnwidth]{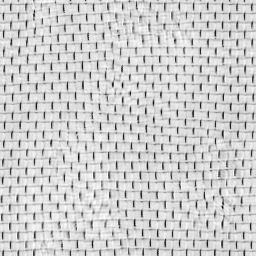}
	\includegraphics[width=0.17\columnwidth]{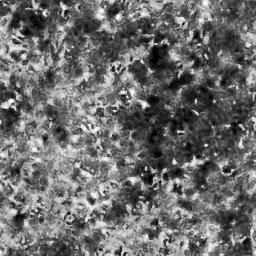}
	\includegraphics[width=0.17\columnwidth]{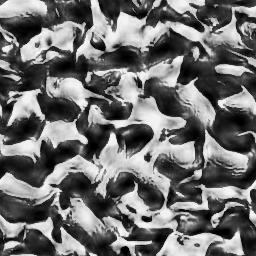}
	\includegraphics[width=0.17\columnwidth]{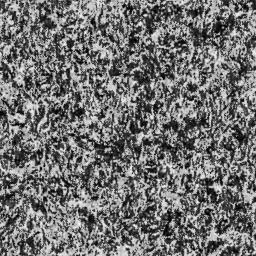}
	\includegraphics[width=0.17\columnwidth]{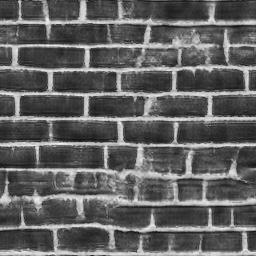}
	\includegraphics[width=0.17\columnwidth]{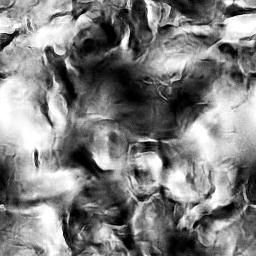}
	\caption{\textbf{Example 256x256 Texture Images Representing all 15 Classes.} For each texture class 50 images were generated (making a total of 750 images). The 15 images shown here each represent one texture class. The 32x32 patches in figure \ref{fig:tex} were randomly sampled from these images with an equal distribution across texture classes. }
	\label{fig:texbsd500}
\end{figure}
\begin{figure}[H]
	\centering
	\includegraphics[width=0.17\columnwidth]{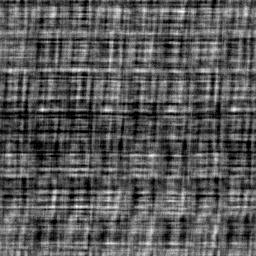}
	\includegraphics[width=0.17\columnwidth]{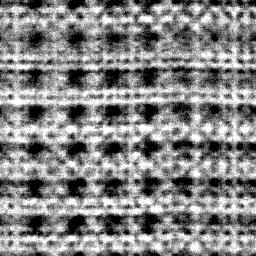}
	\includegraphics[width=0.17\columnwidth]{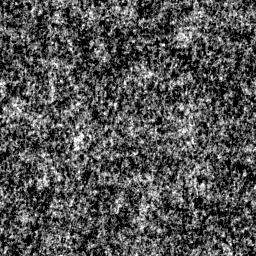}
	\includegraphics[width=0.17\columnwidth]{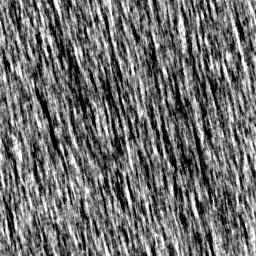}
	\includegraphics[width=0.17\columnwidth]{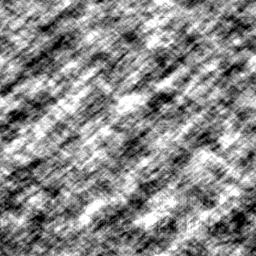}
	\includegraphics[width=0.17\columnwidth]{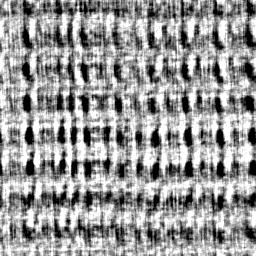}
	\includegraphics[width=0.17\columnwidth]{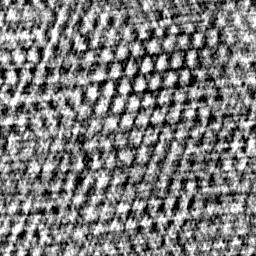}
	\includegraphics[width=0.17\columnwidth]{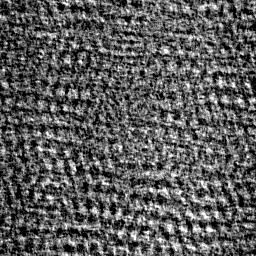}
	\includegraphics[width=0.17\columnwidth]{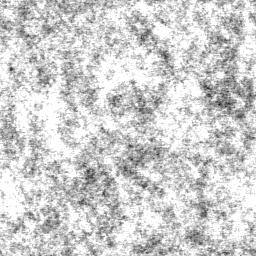}
	\includegraphics[width=0.17\columnwidth]{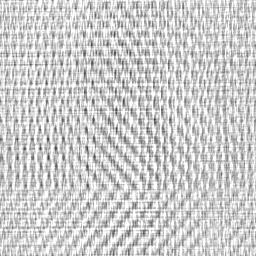}
	\includegraphics[width=0.17\columnwidth]{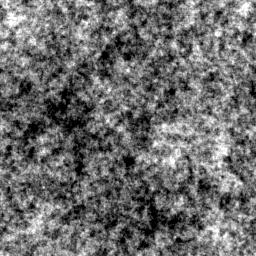}
	\includegraphics[width=0.17\columnwidth]{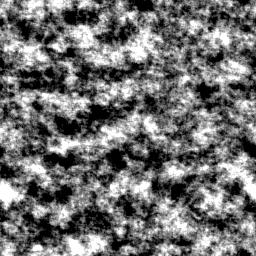}
	\includegraphics[width=0.17\columnwidth]{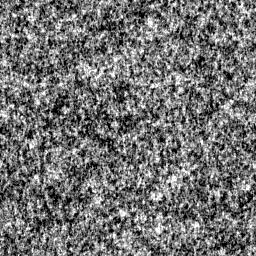}
	\includegraphics[width=0.17\columnwidth]{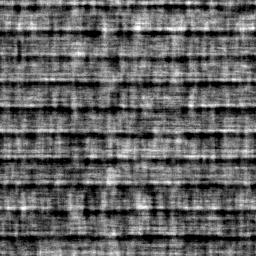}
	\includegraphics[width=0.17\columnwidth]{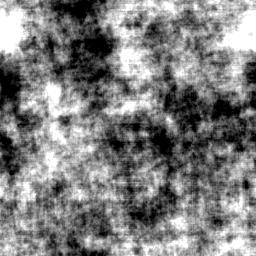}
	\caption{\textbf{Corresponding 256x256 Spectrally-Matched Noise Images.} Spectrally-matched noise versions of the textures in figure \ref{fig:texbsd500}. Each of the 750 full size texture images also has a spectrally-matched noise variant. }
	\label{fig:noisebsd500}
\end{figure}

\section{Acknowledgments}

We thank Hauro Hosoya for publicly providing his hierarchical overcomplete ICA model code \citep{hosoya:jneuro15}. We thank Bruno Olshausen for his helpful discussion regarding sparse coding in hierarchical models. We thank Nasir Laskar for the synthetic texture images in the classification experiment generated with the publicly available texture generation code associated with \cite{portilla:ijcv00}. We thank Ruben Coen-Cagli for providing us with his code for generating the figure-ground patches from the Berkeley segmentation dataset. This material is based upon work supported by the National Science Foundation Graduate Research Fellowship Program under Grant No. 1451511. Any opinions, findings, and conclusions or recommendations expressed in this material are those of the author(s) and do not necessarily reflect the views of the National Science Foundation.

\end{document}